\documentclass[Numbered,referee]{sn-jnl}

\usepackage[numbers,sort&compress]{natbib}
\usepackage[expansion=false]{microtype}
\usepackage{graphicx}
\usepackage{multirow}
\usepackage{amsmath,amssymb,amsfonts}
\usepackage{amsthm}
\usepackage{mathrsfs}
\usepackage{xcolor}
\usepackage{colortbl}
\usepackage{array}
\usepackage{textcomp}
\usepackage{booktabs}
\usepackage{algorithm}
\usepackage{algpseudocode}
\usepackage{listings}
\usepackage{comment}
\usepackage[font=small,labelfont=bf]{caption}
\usepackage{subcaption}
\usepackage{siunitx}
\usepackage{bm}
\usepackage{tikz}
\usepackage{pgfplots}
\usepackage{pgfplotstable}
\usepackage{float}
\usepackage{placeins}

\pgfplotsset{compat=1.18}
\usetikzlibrary{positioning,calc,matrix,patterns}
\usepgfplotslibrary{groupplots,colorbrewer,colormaps}

\definecolor{accent}{HTML}{1F4E79}
\hypersetup{colorlinks=true, linkcolor=accent, citecolor=accent, urlcolor=accent}

\newcommand{\R}{\mathbb{R}}
\newcommand{\E}{\mathbb{E}}
\newcommand{\N}{\mathcal{N}}
\newcommand{\U}{\mathcal{U}}
\newcommand{\loss}{\mathcal{L}}
\newcommand{\enc}{E_{\bm\phi}}
\newcommand{\dec}{D_{\bm\psi}}
\newcommand{\eps}{\bm\epsilon}
\newcommand{\epsth}{\bm\epsilon_{\bm\theta}}
\newcommand{\z}{\bm z}
\newcommand{\x}{\bm x}
\newcommand{\norm}[1]{\left\lVert #1 \right\rVert}
\newcommand{\abs}[1]{\left\lvert #1 \right\rvert}
\newcommand{\sg}{\operatorname{sg}}
\newcommand{\up}{$\uparrow$}
\newcommand{\dn}{$\downarrow$}
\newcommand{\zero}{$\!\to\!0$}

\begin{document}

\title[]{A continually expandable foundation model for brain imaging}

\author*[1,2]{\fnm{Michail} \sur{Mamalakis}}\email{mm2703@cam.ac.uk}
\author[5,10]{\fnm{Carmen} \sur{Jimenez-Mesa}}\email{cj473@cam.ac.uk}
\author[8]{\fnm{Yonghao} \sur{Li}}\email{yl860@cam.ac.uk}
\author[8]{\fnm{Hao} \sur{Chen}}\email{hc666@cam.ac.uk}
\author[6]{\fnm{Chao} \sur{Li}}\email{cl647@cam.ac.uk}
\author[9]{\fnm{Antonios} \sur{Mamalakis}}\email{npa4tg@virginia.edu}
\author[3]{\fnm{John} \sur{Suckling}}\email{js369@cam.ac.uk}
\author[4]{\fnm{Richard} \sur{Bethlehem}}\email{rb643@cam.ac.uk}
\author[7]{\fnm{Stephen J.} \sur{Price}}\email{sjp58@cam.ac.uk}
\author[1,7]{\fnm{Richard J.} \sur{Gilbertson}}\email{Richard.Gilbertson@cruk.cam.ac.uk}
\author[2]{\fnm{Pietro} \sur{Lio}}\email{pl219@cam.ac.uk}
\author[]{\textit{for the Alzheimer's Disease Neuroimaging Initiative}\textsuperscript{**}}

\affil*[1]{\orgdiv{Cancer Research UK Cambridge Institute}, \orgname{University of Cambridge Li Ka Shing Centre}, \orgaddress{\street{Robinson Way}, \city{Cambridge}, \postcode{CB2 0RE}, \state{Cambridgeshire}, \country{United Kingdom}}}
\affil[2]{\orgdiv{Department of Computer Science and Technology}, \orgname{University of Cambridge}, \orgaddress{\street{15 JJ Thomson Ave}, \city{Cambridge}, \postcode{CB3 0FD}, \state{Cambridgeshire}, \country{United Kingdom}}}
\affil[3]{\orgdiv{Department of Psychiatry}, \orgname{University of Cambridge}, \orgaddress{\street{Hills Road}, \city{Cambridge}, \postcode{CB2 2QQ}, \state{Cambridgeshire}, \country{United Kingdom}}}
\affil[4]{\orgdiv{Department of Psychology}, \orgname{University of Cambridge}, \orgaddress{\street{Downing Pl}, \city{Cambridge}, \postcode{CB2 3EB}, \state{Cambridgeshire}, \country{United Kingdom}}}
\affil[5]{\orgdiv{Department of Communication Engineering E.T.S. Ingenieria de Telecomunicacion}, \orgname{University of Malaga}, \orgaddress{\street{Blvd. Louis Pasteur 35}, \city{Malaga}, \postcode{29010}, \country{Spain}}}
\affil[6]{\orgdiv{Department of Applied Mathematics and Theoretical Physics}, \orgname{University of Cambridge}, \orgaddress{\street{Wilberforce Rd}, \city{Cambridge}, \postcode{CB3 0WA}, \state{Cambridgeshire}, \country{United Kingdom}}}
\affil[7]{\orgdiv{Department of Oncology}, \orgname{University of Cambridge}, \orgaddress{\street{Cambridge Biomedical Campus}, \city{Cambridge}, \postcode{CB2 0SP}, \state{Cambridgeshire}, \country{United Kingdom}}}
\affil[8]{\orgdiv{Department of Clinical Neuroscience}, \orgname{University of Cambridge}, \orgaddress{\street{Cambridge Biomedical Campus}, \city{Cambridge}, \postcode{CB2 0SP}, \state{Cambridgeshire}, \country{United Kingdom}}}
\affil[9]{\orgdiv{School of Data Science}, \orgname{University of Virginia}, \orgaddress{\city{Charlottesville}, \state{VA}, \country{United States of America}}}
\affil[10]{\orgdiv{DaSCI Andalusian Institute of Data Science and Computational Intelligence}, \orgname{University of Granada}, \orgaddress{\street{Av. del Conocimiento 37}, \city{Granada}, \postcode{18016}, \state{Granada}, \country{Spain}}}

\maketitle

\noindent\textbf{Author-affiliation mapping.}
M.M.\textsuperscript{1,2,*}, C.J.-M.\textsuperscript{5,10},
Y.L.\textsuperscript{8}, H.C.\textsuperscript{8}, C.L.\textsuperscript{6},
A.M.\textsuperscript{9}, J.S.\textsuperscript{3}, R.B.\textsuperscript{4},
S.J.P.\textsuperscript{7}, R.J.G.\textsuperscript{1,7} and
P.L.\textsuperscript{2}.

\noindent\textsuperscript{**}Data used in preparation of this article were obtained from the Alzheimer's Disease Neuroimaging Initiative (ADNI) database (\url{https://adni.loni.usc.edu}). As such, the investigators within ADNI contributed to the design and implementation of ADNI and/or provided data but did not participate in the analysis or writing of this report. A complete listing of ADNI investigators can be found at \url{https://adni.loni.usc.edu/wp-content/uploads/how_to_apply/ADNI_Acknowledgement_List.pdf}.

\vspace{2.6em}

Brain magnetic resonance imaging (MRI) is central to neuroscience and clinical
assessment, but models are commonly developed for individual diseases,
populations or imaging protocols \cite{wald2025openmind,kaczmarek2025neurosimclr,barba2025dune,tak2026brainiac,wu2026braindino}.
Foundation models promise more general representations, yet they are usually
pretrained once and can lose earlier capabilities when updated with new data
\cite{cata,kirkpatrick2017}. Here we show that Alcmaeon, a three-dimensional
brain MRI foundation model pretrained without manual labels on more than
425{,}000 volumes and derived imaging maps, can be expanded sequentially across
clinical domains. Alcmaeon combines volumetric encoding and latent diffusion
generation with Graph-Blueprint Pruning, which protects network modules
important to earlier domains while leaving the remaining capacity trainable
\cite{anthropic2025graphs,kang2022wsn,serra2018hat}. Across expansion from
healthy ageing and neurodegeneration to developmental, psychiatric and tumour
imaging, Graph-Blueprint Pruning showed less forgetting than sequential
adaptation and elastic weight consolidation across voxel-level reconstruction
measures, with its largest advantage after adaptation to tumour imaging. The
blueprints provided an inspectable record of how model capacity was protected
and reused. Representations from different model levels supported image
synthesis, disease classification, survival modelling and postoperative
prediction, although no single representation was optimal for every task.
These findings provide a route towards brain MRI foundation models that can
grow with emerging data while retaining earlier capabilities.

\vspace{2.6em}

\section*{Main}

Magnetic resonance imaging underpins much of modern neuroscience and
neurology, yet computational models for brain MRI remain highly fragmented.
Most are developed for a particular disease, cohort, imaging modality or
acquisition protocol and often transfer poorly when these conditions change.
Foundation models offer a possible route towards more general systems by
learning reusable representations from large and diverse datasets before
being adapted to specific applications. Developing such models for
three-dimensional neuroimaging remains challenging, however, because
volumetric data are computationally demanding, imaging protocols are
heterogeneous and clinically relevant datasets continue to expand.

Recent foundation and self-supervised models have shown that large-scale brain
MRI pretraining can support reconstruction, segmentation and clinical
prediction
\cite{wald2025openmind,kaczmarek2025neurosimclr,barba2025dune,tak2026brainiac,wu2026braindino}.
These approaches generally follow a static train--then--adapt design: the
model is trained once and subsequently probed or fine-tuned for individual
applications. In practice, new neuroimaging data arrive progressively as
additional populations, diseases, modalities and acquisition protocols become
available. Updating a pretrained model with these data can degrade previously
acquired capabilities, a problem known as catastrophic forgetting
\cite{cata}. Regularisation-based continual-learning methods limit changes to
important parameters \cite{reg,kirkpatrick2017}, while masking and subnetwork
approaches reserve selected components for earlier tasks
\cite{kang2022wsn,serra2018hat}. How to preserve earlier computational
pathways while continuing to expand a large three-dimensional MRI model
nevertheless remains unresolved \cite{chen2025pathwayprotection}.

Here we introduce Alcmaeon\footnote{The model is named after Alcmaeon of Croton, an early medical thinker associated with linking the brain to perception.}, a brain MRI foundation-model framework for
large-scale self-supervised pretraining, continual domain expansion and
downstream adaptation. Alcmaeon is trained without manual labels on more than
425{,}000 three-dimensional MRI volumes and derived imaging maps spanning
population imaging, healthy ageing, neurodegeneration, developmental and
psychiatric cohorts, and adult brain tumours. Rather than treating pretraining
as a fixed stage, we organise these data as a sequence of clinically and
biologically distinct domains and ask whether the model can acquire each new
domain while limiting deterioration in capabilities learned from earlier
cohorts.

To support continual expansion, we develop Graph-Blueprint Pruning (GBP), a
structural-memory mechanism that ranks domain-associated computational modules
using activation--gradient salience. Modules selected for earlier domains are
added to a cumulative blueprint and protected during subsequent adaptation,
whereas the remaining modules remain trainable. Unlike soft parameter
regularisation, GBP therefore places a hard structural constraint on selected
parts of the model. The resulting blueprints provide an inspectable,
stage-specific record of which modules are selected, protected or still
available for later learning. Across the evaluated sequence, GBP most
consistently reduced forgetting across voxel-wise reconstruction metrics, with
its largest advantage arising after adaptation to tumour imaging.

We first examine whether Alcmaeon can be trained stably at volumetric scale
and whether reconstruction depends more strongly on model size, latent design
or imaging modality. We then evaluate continual expansion across four
clinically distinct domains and test whether GBP limits deterioration when
both the clinical population and input format change. 

Finally, we evaluate representations extracted from different components and
depths of the model across cross-modal MRI synthesis,
neurodegenerative-disease classification, adult glioma survival modelling and
postoperative outcome prediction under limited supervision. These benchmarks
assess whether different model representations are suited to distinct
generative, diagnostic and prognostic objectives, and whether
blueprint-protected representations remain useful after continual expansion.
Together, our experiments examine whether a brain MRI foundation model can
learn sequentially from heterogeneous clinical domains while limiting the
deterioration of previously acquired capabilities. More broadly, Alcmaeon
provides a framework in which new neuroimaging domains can be incorporated as
data become available, without requiring a separate pretrained model for every
disease, cohort or imaging protocol.

\section*{Alcmaeon: a continually expandable foundation model for brain MRI}

\begin{figure}
\centering
\includegraphics[width=\textwidth,
trim={0.0cm 0.0cm 0.0cm 0.0cm},
clip]{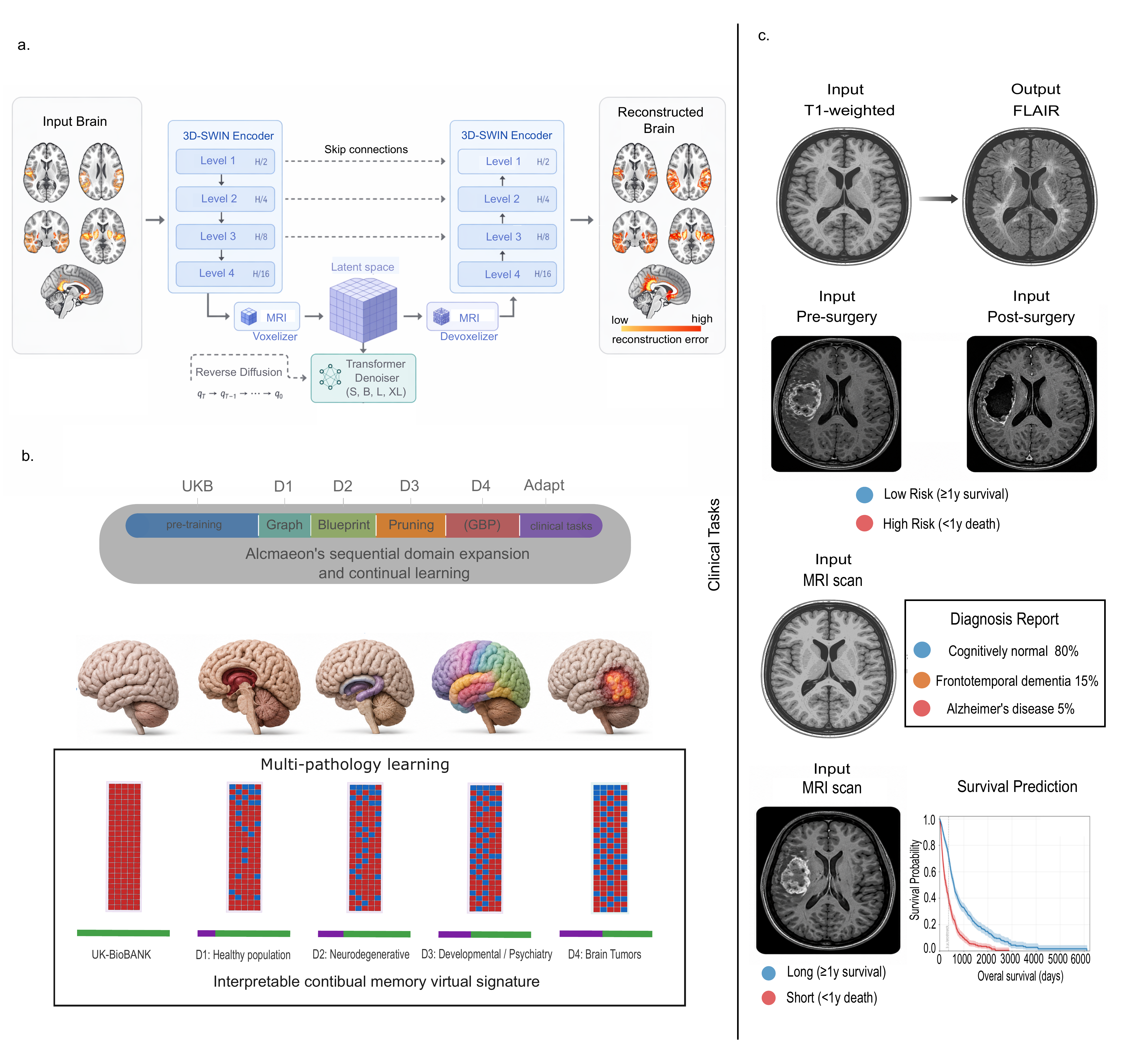}
\caption{
\textbf{Alcmaeon is a three-dimensional brain MRI foundation model with Graph-Blueprint continual learning.}
\textbf{a.} Alcmaeon is trained by reconstructing heterogeneous three-dimensional brain MRI inputs supplied as either paired multimodal channels or pooled single-channel scans. A Swin encoder maps the input volume to a compact latent representation, a transformer denoiser performs reverse diffusion in the latent space, and a devoxeliser and Swin decoder reconstruct the target volume.
\textbf{b.} Blueprint signatures track how trainable modules are allocated and protected across the sequential trajectory from UK Biobank pretraining to D1 healthy ageing, D2 neurodegeneration, D3 developmental and psychiatric imaging, and D4 adult glioma and brain-tumour imaging. Graph-Blueprint Pruning aggregates activation--gradient module salience, constructs a global blueprint using principal component analysis and protects high-salience modules through a cumulative update rule. Separate blueprints are maintained for the encoder--decoder and latent generator.
\textbf{c.} Alcmaeon's representations are adapted or evaluated for cross-modal brain-tumour MRI synthesis, neurodegenerative-disease classification, adult glioma survival modelling and postoperative brain-tumour outcome prediction, spanning generative, diagnostic and prognostic applications.
}
\label{fig:pipeline}
\end{figure}

Alcmaeon is a self-supervised foundation model trained without manual labels
on more than 425,000 three-dimensional brain MRI volumes and derived imaging
maps. It is designed to learn shared representations across heterogeneous MRI
data and to extend them sequentially as new clinical domains become available.
The architecture comprises four principal components
(Fig.~\ref{fig:pipeline}). A three-dimensional Swin Transformer
encoder--decoder, termed 3D-Swin
\cite{liu2021swin,tang2022self,cicek2016}, reconstructs structural MRI,
diffusion-derived images and quantitative imaging maps through a shared latent
representation. A volumetric Diffusion Transformer, termed 3D-DiT
\cite{peebles2023,dosovitskiy2021,ho2020}, models the distribution of these
latent representations and supports image synthesis and cross-modal
translation. Task-specific prediction heads connect the learned
representations to diagnostic and prognostic endpoints. Finally,
Graph-Blueprint Pruning (GBP) ranks domain-associated computational modules
and protects selected modules during subsequent adaptation, providing a
structural memory of previously learned domains.

The 3D-Swin reconstruction network adapts the hierarchical shifted-window
architecture of SwinUNETR and incorporates U-Net-style skip connections
\cite{liu2021swin,tang2022self,ronneberger2015}. The 3D-DiT generator operates
within the compact bottleneck representation rather than directly in the
full-resolution voxel space, reducing the computational cost of volumetric
generation and allowing the reconstruction and generative components to share
a latent space. GBP maintains separate blueprints for the encoder--decoder and
latent generator and can constrain these components independently or jointly.
After each domain, modules with high activation--gradient salience are added
to the cumulative blueprint and protected during subsequent adaptation,
whereas the remaining modules remain trainable. Architectural and
implementation details are provided in Methods.

\subsection*{Latent design influences reconstruction more strongly than model
scale}

Alcmaeon was initially trained through self-supervised reconstruction on UK
Biobank, which provided population-scale structural MRI, diffusion MRI and
neurite orientation dispersion and density imaging (NODDI)-derived maps from a
predominantly healthy population. The model was then expanded sequentially to
healthy ageing in D1, neurodegenerative disease in D2, developmental and
psychiatric imaging in D3, and adult glioma and brain-tumour imaging in D4.
This sequence introduced changes in age distribution, pathology, modality
composition and acquisition protocol. Tumour imaging was placed last to test
adaptation under a pronounced anatomical and pathological shift while
measuring retention of earlier-domain performance. We evaluated four model
scales: Small (approximately 33 million parameters), Base (131 million), Large
(580 million) and Extra-Large (676 million).

We first examined whether Alcmaeon could be trained stably at volumetric scale
and how reconstruction performance varied with optimization configuration,
latent design, imaging modality and model size (Extended Data
Figs.~\ref{fig:pretrain} and \ref{fig:pretrainb}). Among the evaluated
distributed-training configurations, those combining larger global batches
with greater hardware parallelism showed smoother optimization trajectories
and reached stable reconstruction losses earlier. We selected the
configuration with a global batch size of 512 for subsequent experiments.
Because batch size and hardware parallelism varied together, this comparison
characterizes the complete training configuration rather than an independent
effect of batch size. Full optimization analyses are provided in
Supplementary Information S9.1.1.

We next tested whether adding latent distribution modelling with 3D-DiT altered
the reconstruction fidelity of the deterministic encoder--decoder. Across the
evaluated training configurations, 3D-Swin and 3D-Swin-DiT differed by no more
than approximately \(0.1\,\mathrm{dB}\) in PSNR and \(0.001\) in SSIM.
Incorporating 3D-DiT was therefore associated with little measurable change in
reconstruction fidelity, while its generative performance was assessed
separately in the downstream synthesis experiments.

We then compared the deterministic latent representation used by 3D-Swin-DiT
with a variational autoencoder formulation
\cite{kingma2014}. Their relative reconstruction performance depended on the
imaging modality. For diffusion MRI, 3D-Swin-DiT achieved an SSIM of
approximately (0.872), compared with (0.796) for VAE-DiT, and produced
lower voxel-wise reconstruction errors. Structural MRI also favoured the
non-variational formulation. By contrast, for NODDI-derived microstructural
maps, models containing a variational bottleneck achieved an SSIM of
approximately (0.888) and reduced voxel-wise errors by more than (30\%)
relative to the corresponding non-variational model. Thus, deterministic
latent representations achieved higher reconstruction fidelity for structural
and diffusion MRI, whereas variational regularisation was advantageous for
some microstructural maps.

Finally, we examined whether increasing parameter count produced corresponding
improvements in reconstruction. Under the evaluated training conditions, gains
from the Small to Extra-Large models were modest, and their performance
intervals overlapped substantially (Extended Data
Fig.~\ref{fig:pretrainb}a--c). By contrast, comparisons between models using
the shallower level-2 and deeper level-4 representations showed larger
differences associated with latent depth and formulation, particularly for
NODDI-derived maps (Extended Data Fig.~\ref{fig:pretrain}d--f). These results
do not establish equivalence among model scales, but they indicate that
increasing parameter count alone provided limited gains under the evaluated
training conditions.

Together, these analyses show that reliable self-supervised learning across
diverse three-dimensional MRI data depended more on how the model encoded the
images than on model size alone. Because different types of MRI benefited from
different internal representations, we evaluated multiple feature levels and
adaptation strategies in the downstream tasks.

\section*{Graph-Blueprint Pruning limits forgetting when the
clinical domain or input format changes}

\subsection*{Adapting across clinical domains and input formats}

Continual neuroimaging models must accommodate variation arising from both the
clinical population and the structure of the input data. These sources of
variation are often confounded because new cohorts may differ simultaneously
in disease composition, acquisition protocol and available MRI modalities. We
therefore evaluated Alcmaeon under single-channel, two-channel and hybrid input
configurations designed to distinguish clinical-domain shift from changes in
channel semantics and input dimensionality.

In the single-channel configuration, each MRI volume was processed
independently as the model was sequentially adapted to healthy ageing (D1),
neurodegeneration (D2), developmental and psychiatric imaging (D3), and adult
brain tumours (D4). These domains represent distinct clinical and imaging
distributions rather than increasing levels of task complexity. D4 constituted
the most disruptive transition because tumour imaging introduced heterogeneous
lesions, oedema, mass effect and substantial anatomical distortion. In the
two-channel configuration, UK Biobank pretraining used two MRI modalities
acquired from the same participant. When consistently paired modalities were
unavailable in subsequent domains, the same image was duplicated across the
two channels to maintain the expected input dimensionality. The hybrid
configuration provided a more stringent input-format shift by combining
two-channel pretraining with single-channel adaptation. These experiments
therefore tested whether continual learning was affected by clinical-domain
shift, altered channel semantics or mismatched input dimensionality. Detailed
analyses are provided in Supplementary Information S9.2 and S9.5--S9.7.

Neither single-channel nor two-channel processing was uniformly superior during
adaptation from D1 to D4 (Extended Data Fig.~\ref{fig:pretrainb}d--f).
Single-channel models generally achieved comparable or higher SSIM in the
earlier domains and could incorporate datasets lacking complete multimodal
acquisitions. Two-channel models more often achieved higher PSNR and lower
voxel-wise errors in D3 and D4, with the clearest advantage in tumour imaging.
Thus, single-channel processing offered greater flexibility across
heterogeneous datasets, whereas genuinely paired inputs could exploit
within-participant anatomical correspondence. This advantage should not be
interpreted as a general benefit of multimodal learning in the later domains,
where duplicated images preserved dimensionality without providing additional
modality information.

We next evaluated the balance between plasticity and retention. Plasticity was
measured from performance on the domain immediately after it was learned,
whereas retention was assessed by re-evaluating earlier domains after each
subsequent adaptation stage. The largest deterioration in previous-domain
performance occurred after adaptation to D4 (Extended Data
Fig.~\ref{fig:forgetting}a). In the single-channel configuration, EWC retained
PSNR values of only \(10.1\), \(8.3\) and \(13.2\,\mathrm{dB}\) on D1--D3,
respectively. By comparison, GBP retained \(25.6\), \(18.5\) and
\(19.3\,\mathrm{dB}\) on these domains while achieving
\(21.4\,\mathrm{dB}\) on D4. SEQ showed an intermediate pattern that varied
across domains. GBP therefore preserved substantially more previous-domain
performance without preventing adaptation to the tumour domain, whereas EWC
did not retain earlier reconstruction performance after this pronounced
clinical shift.

This ordering was reproduced when the input structure changed and across the
evaluated model scales (Extended Data Fig.~\ref{fig:forgettingb}a). Following
D4 adaptation, two-channel GBP retained PSNR values of \(21.0\), \(17.1\) and
\(19.9\,\mathrm{dB}\) on D1--D3, compared with \(11.4\), \(9.3\) and
\(15.3\,\mathrm{dB}\) for EWC. In the hybrid configuration, GBP retained
\(20.4\), \(17.2\) and \(19.9\,\mathrm{dB}\), whereas EWC retained
\(11.7\), \(9.5\) and \(16.1\,\mathrm{dB}\). GBP therefore remained more
stable than EWC when adaptation involved either a clinical-domain shift alone
or a concurrent change in input format.

\begin{figure}[h!]
\centering
\includegraphics[
width=1.0\textwidth,
trim={0.0cm 0.0cm 0.0cm 0.0cm},
clip
]{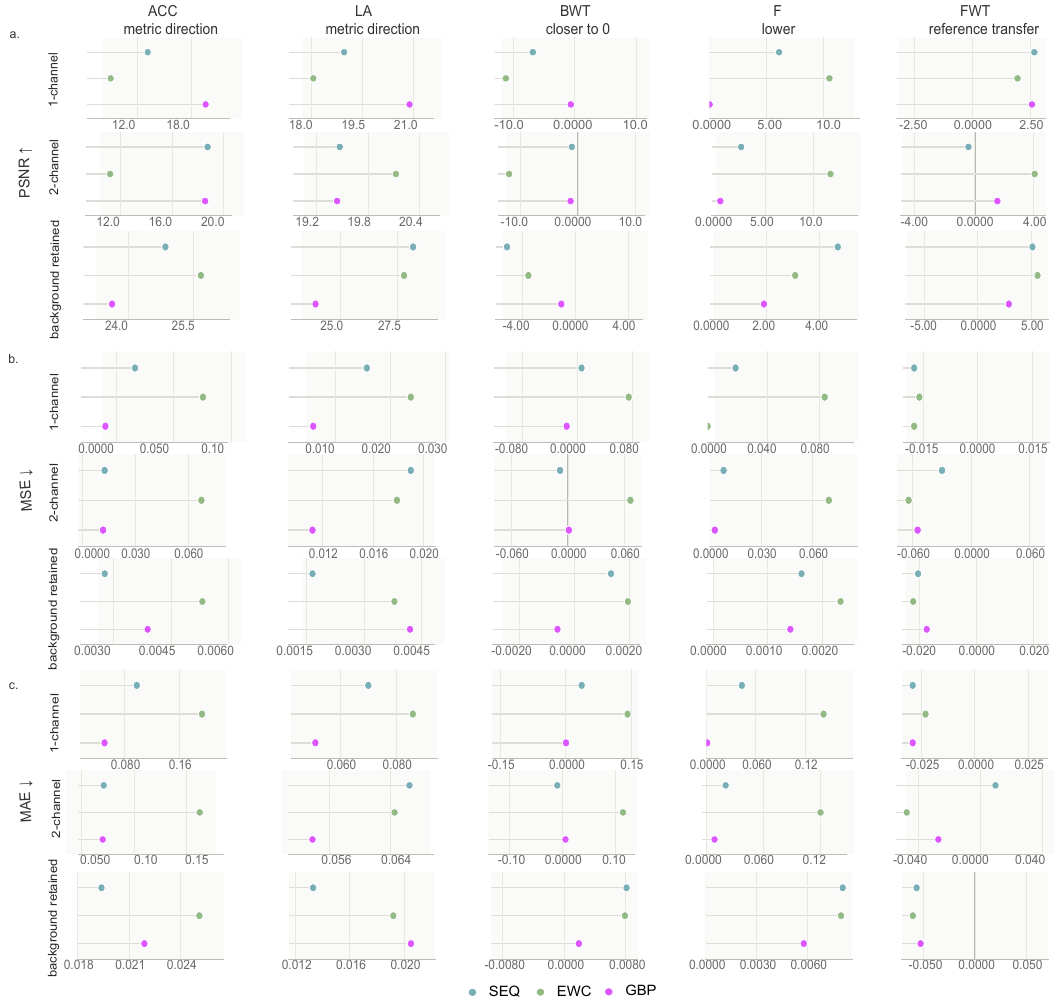}
\caption{\textbf{Continual-learning performance across channel
configurations and reconstruction metrics.}
\textbf{a--c}, Average accuracy after the final stage (ACC), learning
accuracy on the just-trained task (LA), backward transfer (BWT),
non-negative best-ever peak forgetting (\(F\)), and forward transfer
relative to the independent reference (FWT) for PSNR (\textbf{a}), MSE
(\textbf{b}) and MAE (\textbf{c}). Within each panel, rows show the
1-channel, 2-channel and background-retained 1-channel evaluations. Points
denote SEQ (blue), EWC (green) and GBP (magenta). Higher values indicate
better performance for PSNR, whereas lower values indicate better
performance for MSE and MAE; BWT values closer to zero and lower \(F\)
values indicate greater retention. FWT is interpreted relative to the
direction and scale of the corresponding image-quality metric. Axis limits
vary among metric--score combinations to show differences between methods
clearly.}
\label{new}
\end{figure}

Channel configuration also affected the unconstrained and regularized
baselines differently (Fig.~\ref{new}). For SEQ, peak PSNR forgetting decreased from
\(5.93\,\mathrm{dB}\) in the single-channel configuration to
\(3.00\,\mathrm{dB}\) in the two-channel configuration, indicating that
channel coupling was associated with greater stability . This association did
not extend to EWC, for which peak forgetting increased from
\(10.20\) to \(12.13\,\mathrm{dB}\). GBP showed consistently low peak
forgetting in the single- and two-channel configurations
\(0.03\) and \(0.87\,\mathrm{dB}\), respectively. In the hybrid
configuration, GBP similarly achieved \(F=0.87\,\mathrm{dB}\) and
\(\mathrm{BWT}=-0.14\,\mathrm{dB}\), compared with
\(F=11.56\,\mathrm{dB}\) for EWC. Collectively, these results indicate that
GBP was less sensitive than the comparator methods to changes in both clinical
domain and input structure.

Structural and voxel-wise metrics did not, however, always produce the same
ranking (Fig.~\ref{new}). In the single-channel configuration, SSIM-based ACC was similar for
EWC (\(0.835\)), SEQ (\(0.829\)) and GBP (\(0.810\)), but the difference
increased in the two-channel configuration, with values of \(0.802\) for SEQ,
\(0.782\) for EWC and \(0.688\) for GBP, and was greatest in the hybrid
configuration, where GBP achieved \(0.616\), compared with \(0.810\) for EWC
and \(0.802\) for SEQ . Because SSIM directly defined the autoencoder
reconstruction loss, the higher values achieved by SEQ and EWC may reflect
their greater freedom to optimize this loss-aligned structural criterion,
whereas blueprint protection imposed an adaptation cost on GBP. This cost was
amplified when the two-channel representation was collapsed to one channel,
although the present experiments do not isolate the contribution of the
channel-collapse adapter. Importantly, GBP showed the lowest forgetting across
PSNR, MSE, RMSE and MAE in all three configurations (see Fig.~\ref{new}). SSIM and the voxel-wise
metrics therefore capture complementary properties of reconstruction, and
their combined evaluation indicates that GBP most consistently preserved local
intensity fidelity across clinical and input-format shifts, while incurring a
structural-similarity cost under two-channel learning and channel collapse.
Further analyses are provided in Supplementary Information S9.10.1.
\subsection*{Robustness to background preprocessing}

Whole-volume MRI metrics are influenced by the large, relatively invariant
background surrounding the brain. Because this region is easier to reconstruct
than brain tissue, including it can inflate aggregate image-quality scores and
conceal errors within anatomically informative regions. We therefore repeated
the single-channel evaluation with background voxels retained to determine
whether the principal continual-learning result depended on background
preprocessing (Fig.~\ref{new}).

Retaining the background changed the absolute metric scale and reduced the
apparent differences among methods, but GBP continued to show the lowest peak
PSNR forgetting. Its \(F\) value was \(2.04\,\mathrm{dB}\), compared with
\(3.22\,\mathrm{dB}\) for EWC and \(4.83\,\mathrm{dB}\) for SEQ
(Fig.~\ref{new}). GBP also achieved the backward-transfer value closest to zero
\(\mathrm{BWT}=-0.57\,\mathrm{dB}\). These results indicate that the
relative retention advantage of GBP was present both when the evaluation was
restricted to the signal region and when the complete volume was assessed.
Additional results are provided in Supplementary Information S9.9.

Background retention nevertheless changed the failure modes visible in the
evaluation. With background voxels removed, EWC showed substantial forgetting
after D4 adaptation, with \(\mathrm{BWT}=-10.20\,\mathrm{dB}\) and
\(F=10.20\,\mathrm{dB}\). When the background was included, its performance
changed little after the final update, but it also learned D4 poorly: its D4
PSNR was \(19.26\,\mathrm{dB}\), compared with \(24.45\,\mathrm{dB}\) for
SEQ and \(22.75\,\mathrm{dB}\) for GBP. The apparent stability of EWC under
this evaluation therefore reflected limited plasticity rather than successful
retention combined with adaptation. GBP also incurred an adaptation cost,
showing the lowest background-inclusive learning accuracy
\(\mathrm{LA}=24.34\,\mathrm{dB}\); nevertheless, it learned the new domain
more effectively than EWC while retaining more performance on the earlier
domains (Supplementary Information S9.10.2).

Including the background also increased forward-transfer scores and caused
SSIM to approach saturation. Background-inclusive SSIM exceeded (0.997) on
D2 and D3 for all three methods and did not reveal either the voxel-wise
regressions or the limited D4 learning observed with EWC. Signal-region metrics
therefore provided the more stringent test of forgetting in brain tissue,
whereas background-inclusive metrics characterized reconstruction across the
complete image volume. Across both preprocessing regimes, GBP achieved the most
consistent balance between retaining earlier-domain performance and adapting
to D4. The comparison also reveals the cost of this constraint: protecting the
blueprint improved retention but reduced learning accuracy under the
background-inclusive evaluation. Further analyses are provided in
Supplementary Information S9.10.2.

\section*{Clinical adaptation reveals task-dependent representations}

Representations optimised for image reconstruction may not transfer equally to
generative, diagnostic and prognostic objectives. We therefore examined
whether Alcmaeon's pretrained representations retained information relevant to
four clinically motivated downstream tasks: cross-modal synthesis of
brain-tumour MRI, adult glioma survival modelling, neurodegenerative-disease
classification and postoperative outcome prediction under limited supervision
(Figs.~\ref{fig:finetune2c} and \ref{fig:finetune2d}; Extended Data
Figs.~\ref{fig:finetunea} and \ref{fig:finetuneb}).

For each task, Alcmaeon was compared with established task-specific baselines
and available foundation-model representations. We assessed performance using
the primary predictive or generative metrics for each endpoint and examined
latent-space organisation as a secondary, descriptive analysis. Features were
also extracted from the shallower S8-Lv2 and deeper S2-Lv4 configurations to
test whether the most informative representation depended on feature depth,
modality composition, adaptation strategy and downstream objective. Full
benchmarking protocols, statistical comparisons, encoder--generator analyses,
survival results and low-shot experiments are provided in Supplementary
Information S10.

\subsection*{Cross-modal tumour MRI synthesis}

\begin{figure}[h!]
\centering
\includegraphics[width=1.0\textwidth,
trim={0.0cm 0.0cm 0.0cm 0.0cm},
clip]{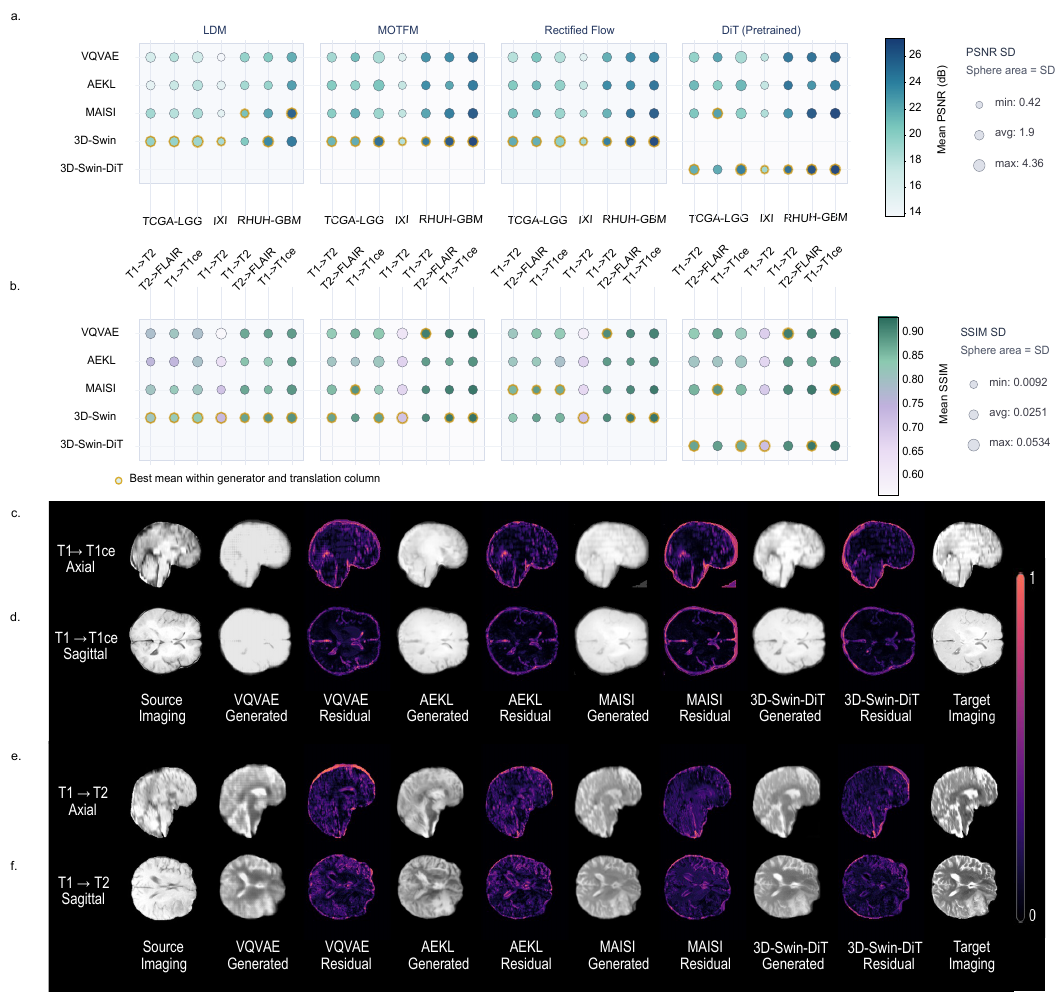}
\caption{
\textbf{Clinical adaptation of Alcmaeon across tasks.}
\textbf{a.} Brain-tumour data generation. Cross-modal MRI translation is evaluated across tumour and external cohorts using PSNR and SSIM matrices for multiple encoder--generator combinations. Qualitative examples and edge-filter visualisations illustrate preservation of anatomical structure and tumour-related image features during synthetic generation.
}
\label{fig:finetune2c}
\end{figure}

Cross-modal synthesis tests whether a learned representation preserves
anatomical correspondence while supporting translation between MRI contrasts.
We benchmarked VQVAE~\cite{van2017neural}, AEKL~\cite{rombach2022high},
MAISI~\cite{guo2025maisi}, 3D-Swin and 3D-Swin-DiT representations paired
with LDM~\cite{rombach2022high}, MOTFM~\cite{yazdani2025flow} and Rectified
Flow~\cite{liu2022flow}, each of which was adapted to the target translation
task (Fig.~\ref{fig:finetune2c}). The 3D-Swin representation achieved the
highest mean fidelity in most translation settings evaluated with these
task-adapted generators. When 3D-Swin-DiT was instead evaluated with its
pretrained 3D-DiT generator (the ``DiT (Pretrained)'' column in
Fig.~\ref{fig:finetune2c}a,b), it remained competitive with the alternative
latent-space generation strategies. The strongest configurations achieved
approximately \(28\,\mathrm{dB}\) PSNR and SSIM values of
\(0.93\)--\(0.94\) in the primary tumour cohorts. Performance was more
heterogeneous in the external IXI and RHUH-GBM cohorts, although the strongest
translations in RHUH-GBM retained PSNR values of approximately
\(25\)--\(27\,\mathrm{dB}\).

These results show that Alcmaeon's representations supported cross-modal
translation across several MRI contrasts, while the variation in external
cohorts indicates that translation fidelity remained sensitive to cohort
composition and acquisition characteristics.

\subsection*{Frozen representations support glioma survival modelling}
We tested whether frozen Alcmaeon representations contained prognostic
information in a multicentre cohort of 1,551 adults with glioma. Under a shared
evaluation protocol, the strongest EWC- and GBP-adapted representations showed
similar abilities to rank patients by mortality risk, achieving concordance
indices of $0.601\pm0.016$ and $0.601\pm0.020$, respectively. The corresponding
values were $0.574\pm0.013$ for ResNet-34, $0.581\pm0.022$ for DUNE U-AE and
$0.564\pm0.010$ for HLIP-ViT \cite{chen2019med3d,barba2025dune,zhao2025hlip}. The overlapping intervals do
not establish superiority between EWC and GBP, but show that GBP retained
prognostic information after continual expansion.

GBP and EWC also performed similarly when survival was evaluated at fixed
one-, two- and three-year intervals, with AUROCs ranging from
$0.719\pm0.036$ to $0.736\pm0.013$ for GBP and from $0.716\pm0.037$ to
$0.729\pm0.013$ for EWC. Their relative ordering changed across time points,
indicating that neither method was consistently superior for every survival
endpoint. Risk groups derived from the Alcmaeon representations showed partial
Kaplan--Meier separation, while the corresponding latent-space visualisations
provided descriptive evidence of outcome-related organisation (Extended Data
Fig.~\ref{fig:finetunea}).

The analysis also distinguished relative risk ranking from direct prediction
of survival time. Among patients with observed deaths, DUNE U-AE and U-VAE
achieved mean absolute errors of $510\pm32$ and $502\pm35$ days,
respectively, whereas the continual 3D-Swin-DiT configurations produced larger
errors. Alcmaeon therefore retained information relevant to ranking patients
by risk, but accurate estimation of survival time required different
representational properties.

\subsection*{Intermediate features support neurodegenerative-disease
classification}

We examined whether information relevant to Alzheimer's disease classification
was retained at different depths of Alcmaeon. Intermediate S8-Lv2 Swin
features, combined with a simple classifier, achieved an AUC of
$0.845\pm0.038$ and an accuracy of $0.757\pm0.023$. Performance was similar
to that of a fivefold CNN baseline (AUC, $0.846\pm0.033$; accuracy,
$0.748\pm0.026$) and higher in mean than that of PCA--SVM (AUC,
$0.812\pm0.011$; accuracy, $0.716\pm0.032$). With limited labelled data, the
AUC increased from $0.528$ without task-specific examples to
$0.825\pm0.015$ after five-shot adaptation.

We next compared matched continual-learning variants using one-channel S2-Lv4
encoder features. GBP achieved the highest mean accuracy
($0.743\pm0.025$) and balanced accuracy ($0.737\pm0.022$), compared with
accuracies of $0.724\pm0.035$ for SEQ and $0.724\pm0.036$ for EWC. AUCs were
similar across GBP ($0.820\pm0.012$), SEQ ($0.811\pm0.030$) and EWC
($0.809\pm0.022$). GBP also achieved the highest mean accuracy in the
three-class analysis ($0.613\pm0.034$). These differences were modest and do
not establish statistical superiority, but indicate that blueprint protection
retained disease-related information within the encoder features.

Direct classification using the complete 3D-Swin-DiT representation remained
close to chance for both GBP and EWC in the zero-, one- and three-shot settings
(Extended Data Fig.~\ref{fig:finetuneb}). Thus, the accessibility of diagnostic information
depended on selecting an appropriate representation depth: intermediate
encoder features were more informative for classification than the complete
generative representation.

\subsection*{GBP supports postoperative adaptation under limited supervision}

\begin{figure}[h!]
\centering
\includegraphics[width=1.0\textwidth,
trim={0.0cm 0.0cm 0.0cm 0.0cm},
clip]{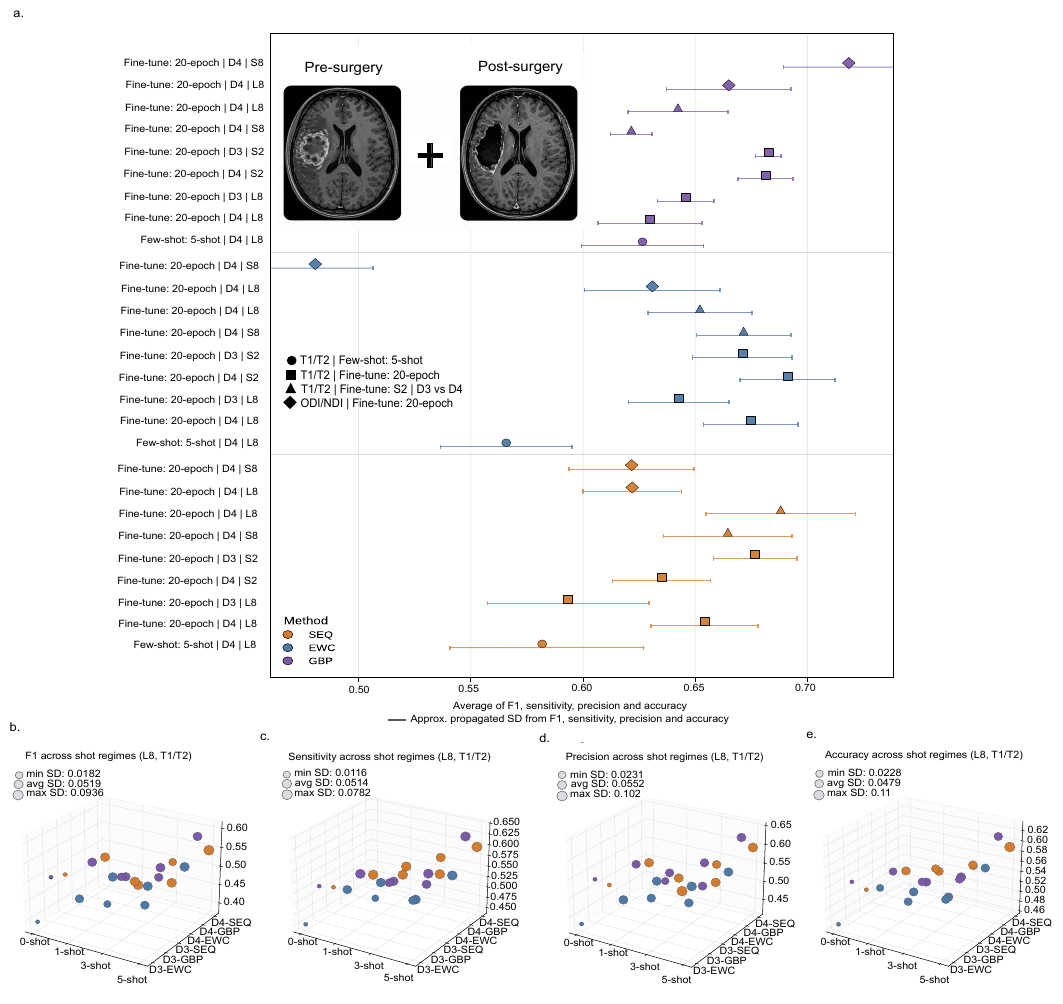}
\caption{
\textbf{Clinical adaptation of Alcmaeon across tasks.}
\textbf{a.} Postoperative brain-tumour outcome prediction. Before- and after-surgery imaging illustrates the anatomical shift from tumour mass to resection cavity, while surrounding performance summaries compare few-shot, chunk-maximum and ODI/NDI transfer settings across SEQ, EWC and GBP. Together, the four panels show that Alcmaeon supports task-specific clinical adaptation for prognostic modelling, disease classification, cross-modal generation and postoperative outcome estimation. The main symbol denotes the balanced mean, computed as the average of F1-score, sensitivity, precision and accuracy. Because all four metrics are derived from the same predictions, covariance terms were not estimated and the propagated s.d. should be interpreted as an approximate visual summary of variability rather than an independent statistical confidence interval.
}
\label{fig:finetune2d}
\end{figure}

Postoperative MRI introduces a substantial anatomical shift from tumour mass
to resection cavity, while labelled outcome datasets are often small. We
evaluated this setting using paired preoperative and postoperative MRI from 49
patients and compared SEQ, EWC and GBP under the same protocol
(Fig.~\ref{fig:finetune2d}). Using intermediate S8-Lv2 features derived from
ODI and NDI microstructural maps, GBP achieved an accuracy of
$0.7521\pm0.0553$, compared with $0.5938\pm0.0593$ for SEQ and
$0.4385\pm0.0738$ for EWC. GBP also achieved the highest F1 score
($0.6770\pm0.0591$), with the same ordering observed for sensitivity and
precision.

GBP retained this advantage in the low-shot experiments. Its F1 score reached
$0.548\pm0.085$ with three-shot adaptation in the S8 D4 configuration and
$0.601\pm0.050$ with five-shot adaptation in the L8 D4 configuration,
exceeding the corresponding SEQ and EWC results. However, GBP was not uniformly
superior, as EWC or SEQ performed better in some fully trained experiments
using conventional T1- and T2-weighted MRI. Blueprint protection therefore
provided its clearest downstream benefit when adaptation involved limited
labels, microstructural imaging or pronounced anatomical change. Given the
small cohort, these findings demonstrate comparative transfer performance
rather than clinical validity and require confirmation in larger independent
populations.

Together, these benchmarks clarify the complementary contributions of
Alcmaeon and GBP. Alcmaeon provides reusable representations from different
levels of the model, with each level supporting different applications.
Encoder representations, used alone or with the diffusion generator, supported
cross-modal image synthesis; intermediate features retained information
relevant to neurodegenerative-disease classification; and combined multimodal
representations supported glioma risk ranking. GBP helped preserve disease- and
prognosis-related information as the model learned new domains, with its
clearest downstream advantage observed during postoperative adaptation with
limited labelled data and substantial changes in anatomy or imaging modality.
When considered alongside GBP's lower forgetting across voxel-level
reconstruction measures, these downstream results support GBP as the most
suitable of the evaluated adaptation strategies when the main objective is to
learn from a substantially different domain while retaining earlier
capabilities. However, they do not establish GBP as a universally optimal
representation for every downstream task.

\section*{Graph-Blueprint Pruning maps structural memory across network modules}

\begin{figure}[h!]
  \centering
  \includegraphics[width=1.0\textwidth,
  trim={0.0cm 0.0cm 0.0cm 0.0cm},
  clip]{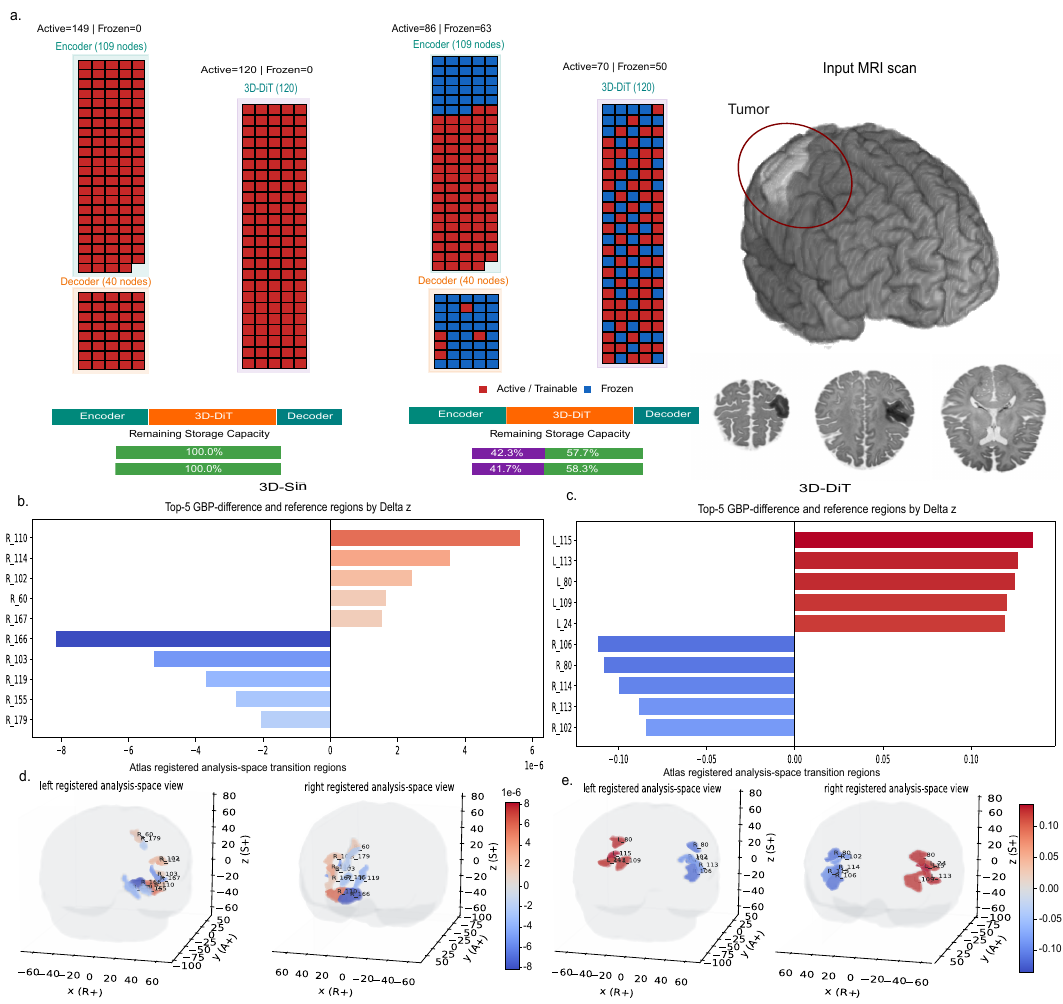}
  \caption{
  \textbf{Graph-Blueprint signatures of capacity allocation and anatomical
  interpretation during continual neuroimaging adaptation.}
  \textbf{a--e,} Blueprint signatures across the sequential adaptation
  trajectory. For each domain, GBP ranks task-associated computational modules
  in the 3D-DiT latent generator and 3D-Swin encoder--decoder, protects the
  selected modules as structural memory and leaves the remaining modules
  trainable for subsequent domains. Node grids distinguish protected and
  trainable modules, and capacity bars show the cumulative fraction of model
  capacity allocated during adaptation. Ranked blueprint scores illustrate
  that module selection is structured rather than uniformly distributed.
  Corresponding latent-space and anatomical projections provide illustrative
  examples of how the selected computational signatures differ across healthy
  ageing, neurodegeneration, developmental and psychiatric imaging, and brain
  tumours. The anatomical projections are based on representative participants
  and are intended for interpretation of model behaviour rather than
  population-level neuroanatomical inference.
  }
  \label{fig:blueprintd}
\end{figure}

Continual-learning methods are typically evaluated by how well they retain
performance, but aggregate retention scores do not reveal which parts of a
model have been constrained or remain available for subsequent learning.
Related masking and subnetwork approaches make parameter allocation explicit
\cite{kang2022wsn,serra2018hat}. GBP extends this principle by dividing
Alcmaeon into predefined computational modules---including attention,
feed-forward, normalisation and adaptive-modulation components---and ranking
their domain-associated contributions using dataset-aggregated
activation--gradient salience. The highest-ranked modules are incorporated
into a cumulative blueprint and protected during subsequent adaptation,
whereas the remaining modules remain trainable. Separate blueprints are
maintained for the 3D-Swin encoder--decoder and the 3D-DiT latent generator,
reflecting their distinct roles in anatomical representation and latent
generation (Fig.~\ref{fig:blueprintd}, Extended Data
Figs.~\ref{fig:blueprinta}--\ref{fig:blueprintc} and Methods).

GBP constrained a selective subset of modules rather than freezing the complete
architecture. Following sequential adaptation through D4, the number of
trainable 3D-DiT modules decreased from 120 to 70, leaving 50 protected and
(58.3\%) of the predefined modules available for further adaptation. In the
3D-Swin encoder--decoder, the number of trainable modules decreased from 149
to 86, leaving 63 protected and (57.7\%) available. These percentages refer
to the number of predefined modules rather than the corresponding proportion
of parameters or computational operations, because the modules differ in
size. Thus, GBP retained substantial trainable subspaces in both architectural
streams while protecting modules selected during earlier domains.

Descriptive comparison of the stage-specific blueprints showed that module
selection was distributed across the architecture and that successive domains
selected partly distinct combinations of attention, feed-forward,
normalisation and modulation components. The cumulative masks distinguish
modules that were newly selected, previously protected or still available for
adaptation at each stage. They therefore provide an inspectable record of how
the hard structural constraint accumulated as the model progressed through
the continual-learning sequence. This record complements the performance-based
analysis of the stability--plasticity trade-off by identifying where the
constraint was applied within the network.

To examine the anatomical distribution of model sensitivity associated with
the modules protected by GBP, we randomly selected one participant from each
domain after excluding cases that failed preprocessing or image-quality
checks. We then projected the corresponding attribution maps onto anatomical
atlases. In the healthy-ageing example, the map conditioned on GBP-protected
modules showed a more spatially restricted frontoparietal pattern than the
broader medial and posterior distribution obtained from the fully trainable
model. In the participant with Parkinson's disease, the protected-module map
placed greater emphasis on limbic, thalamic, temporal and midbrain regions. In the developmental example involving an adolescent, attribution was
concentrated within selected regions of the default-mode, attention and control
networks.
In the tumour example, the pattern was concentrated around the opercular,
insular and adjacent perisylvian regions. It partly overlapped the lesion and
extended into corresponding regions of the contralateral hemisphere that also
showed model-derived changes during tumour-image generation. This combined
lesion-associated and contralateral pattern was not evident in the attribution
map obtained from the fully trainable model.

Because each map was generated from only one randomly selected participant and
different anatomical atlases were used across domains, the projections should
be interpreted as exploratory visualisations of model sensitivity conditioned
on the GBP-protected modules. They do not establish reproducible
population-level anatomical patterns, biological circuitry or causal
associations. Full regional rankings, comparisons between model components and
analyses of the four cases are provided in Fig.~\ref{fig:blueprintd}, Extended
Data Figs.~\ref{fig:blueprinta}--\ref{fig:blueprintc} and Supplementary
Information S9.10.

Together, these analyses show that GBP protects modules identified as
important to previously learned domains while leaving the remaining modules
available for subsequent learning. The blueprints provide an inspectable
record of how this structural memory accumulates as new domains are introduced,
and the anatomical projections illustrate how the protected modules contribute
to model sensitivity across brain regions in individual examples.

\section*{Discussion}

Alcmaeon addresses a practical limitation of current brain MRI foundation
models. Large-scale pretraining can produce reusable representations, but the
resulting models are generally treated as fixed systems and adapted separately
to each application. Neuroimaging data instead accumulate across populations,
diseases, scanners and acquisition protocols. Our findings show that a
three-dimensional MRI model trained without manual labels on more than
425{,}000 volumes and derived imaging maps can be expanded across clinically
distinct domains while preserving much of its earlier reconstruction
performance. GBP provided the most consistent retention when the sequence
reached tumour imaging, which introduced the largest anatomical and
pathological shift evaluated here. These results support a broader view of
foundation models as systems that can continue to learn rather than being
pretrained only once.

The pretraining experiments further show that reconstruction quality was not
determined by model size alone. Increasing the number of parameters produced
modest gains, whereas representation design had a clearer effect.
Under the conditions evaluated, deterministic representations achieved higher
reconstruction fidelity for structural and diffusion MRI, while variational regularisation benefited some NODDI-derived microstructural
maps \cite{kingma2014}. Adding the latent 3D-DiT generator had little effect
on reconstruction fidelity. The imaging signal, learning objective and latent
representation should therefore be considered alongside scale when developing
volumetric foundation models.

The principal continual-learning challenge was to retain earlier knowledge
while remaining able to learn a new domain. Sequential adaptation and elastic
weight consolidation became less reliable after tumour imaging was introduced.
GBP instead protected selected modules and restricted subsequent learning to
the capacity that remained trainable. This structural constraint limited the
loss of earlier-domain performance while allowing adaptation to continue,
suggesting that module-level protection may be particularly useful when new
data differ substantially in pathology and anatomy. Input structure also
formed part of this shift: single-channel models accommodated incomplete
multimodal data, whereas paired channels could exploit anatomical
correspondence between MRI contrasts. Duplicating an image to maintain a
two-channel input preserved dimensionality but added no multimodal information.
Evaluation was similarly sensitive to metric choice. SSIM was aligned with the
reconstruction objective and approached saturation when background voxels were
included, whereas voxel-wise measures were more sensitive to changes within
brain tissue. Complementary metrics and anatomically relevant evaluation regions were
therefore required to distinguish genuine retention from apparent stability
caused by failure to learn the new domain or by the dominance of background
voxels.

The downstream benchmarks show that Alcmaeon's value lies not in a single
general-purpose representation, but in the complementary information available
at different levels of its encoder--diffusion architecture. Encoder features,
used alone or with the diffusion generator, supported cross-modal synthesis;
intermediate encoder features retained information relevant to
neurodegenerative-disease classification; and combined multimodal
3D-Swin-DiT representations supported glioma risk modelling. These findings
also show that reconstruction, generation, classification and prognosis require
different representational properties: accurate image synthesis did not
guarantee accessible diagnostic information, and effective risk ranking did
not ensure accurate prediction of survival time.

In the continual-learning comparisons, GBP was applied to the complete
3D-Swin-DiT architecture, including both the encoder--decoder and diffusion
components, even when downstream analyses subsequently extracted intermediate
encoder features. Representations from these GBP-adapted checkpoints retained
task-relevant information across diagnostic and prognostic applications after
sequential domain expansion. In glioma survival modelling, GBP retained
prognostic information and matched the strongest EWC configuration in risk
discrimination, although neither approach was consistently superior across all
survival endpoints. For Alzheimer's disease classification, intermediate
features extracted from the GBP-adapted model achieved the highest mean
accuracy among the matched continual-learning variants, although the
differences were modest. GBP showed its clearest mean performance advantage in postoperative
prediction, particularly when adaptation involved limited
labelled data, diffusion-derived microstructural maps and substantial
anatomical change. These results support GBP as a means of preserving
transferable information throughout the complete encoder--diffusion model,
especially when new domains differ markedly from earlier data. They do not,
however, imply that GBP or any single feature level is optimal for every
downstream objective.

These benchmarks also suggest directions for future biological and clinical
research. Generative models that preserve patient anatomy across MRI contrasts
could be developed to produce patient-specific counterfactual references,
estimating how the same brain might appear without disease-related changes
\cite{sanchez2022healthy,kumar2022counterfactual}. The present synthesis
experiments provide a technical basis for this possibility but do not validate
the generated images as true counterfactuals. Preoperative and postoperative
representations could similarly be used to study disease burden, surgical
change and prognosis \cite{palsson2022survival,yoon2020multiparametric}, while
intermediate representations could support normative modelling and
disease-stage inference in neurodegeneration
\cite{rutherford2022normative,young2018sustain}. All of these applications
require independent and prospective validation.

The blueprints complement performance evaluation by recording which modules
are protected and which remain available for later learning. Like
task-specific masking and subnetwork approaches
\cite{kang2022wsn,serra2018hat}, GBP makes this allocation inspectable. The
anatomical projections provide exploratory examples of how sensitivity
associated with the protected modules is distributed across brain regions.
However, each map was generated from only one randomly selected participant,
and different atlases were used across domains. The maps therefore cannot
establish population-level anatomical patterns, biological circuits or causal
associations. Larger studies are needed to determine whether these spatial
patterns are reproducible.

These findings demonstrate the potential of structural continual learning for
building expandable brain MRI models, while also identifying priorities for
further evaluation. We evaluated a single sequence in which tumour imaging was
introduced last. Different domain orders may produce different retention
patterns and blueprint allocations. Combining cohorts across sites and
populations increases diversity but may also allow the model to learn
scanner-, site- or population-specific shortcuts, requiring dedicated
demographic and site-bias audits. Because GBP protects modules cumulatively,
longer learning sequences may eventually leave insufficient capacity for
further adaptation and may require mechanisms for sharing, compressing or
safely releasing protected capacity. Finally, although the downstream cohorts
were independent of pretraining, the clinical analyses were retrospective, and
the postoperative evaluation included only 49 participants. Prospective
multi-institutional studies, expert clinical assessment and longitudinal
validation are therefore required before Alcmaeon can inform patient-level
decisions.

In summary, Alcmaeon combines large-scale volumetric pretraining, latent
generation and structural continual learning within one brain MRI framework.
Its central contribution is not a single representation for every application,
but an expandable set of task-relevant representations coupled to an
inspectable mechanism for protecting earlier knowledge. Among the evaluated
continual-learning strategies, GBP most consistently retained earlier
reconstruction capabilities under pronounced domain shift while preserving
features that remained useful for downstream diagnostic and prognostic tasks.
These findings provide a foundation for brain MRI models that can incorporate
new clinical domains while limiting the loss of previously acquired
capabilities.

\section*{Online content}
\noindent Any methods, additional references, source data, extended data,
Supplementary Information, acknowledgements, author contributions and competing
interests, and statements of data and code availability are available in the
online version of this paper.

\clearpage

\section*{Methods}

\subsection*{Study design}

Alcmaeon was developed as a three-dimensional neuroimaging foundation model for reconstruction, latent generative modelling, continual adaptation and downstream clinical prediction. All models operate on MRI volumes represented as
$x \in \mathbb{R}^{B \times C \times H \times W \times D}$, where $B$ is the batch size, $C$ is the number of MRI channels and $(H,W,D)$ are the spatial dimensions. Unless otherwise stated, inputs were trained in either a single-channel setting, in which scans from different modalities were pooled and treated independently, or a two-channel setting, in which paired modalities from the same participant were provided as separate channels. This design allowed the model to operate both on heterogeneous clinical datasets with incomplete imaging protocols and on paired multimodal cohorts.

The methodological framework consists of three components: a Swin-based volumetric autoencoder for self-supervised representation learning \cite{liu2021swin,tang2022self}, a latent Diffusion Transformer for generative modelling and cross-modal synthesis \cite{ho2020,peebles2023}, and a Graph-Blueprint Pruning (GBP) operator for continual learning. The frameworks and their mathematical foundations are described in detail in Supplementary Information S1--S3. After pretraining, the learned representations were evaluated on reconstruction, Alzheimer's progression classification, brain-tumour survival estimation, brain-tumour MRI synthesis and post-surgical few-shot outcome prediction. Architectures and fine-tuning strategies are described in detail in Supplementary Information S4.

\subsection*{Data sources and domain organisation}

Alcmaeon was developed using a large-scale collection of three-dimensional MRI data spanning population imaging, neurodegeneration, development, psychiatric variation and adult brain tumours. For self-supervised training, we assembled approximately 425{,}717 MRI scans from UK Biobank~\cite{ukbiobank,ukbiobank_paper}, the Parkinson's Progression Markers Initiative (PPMI)~\cite{ppmi}, the Alzheimer's Disease Neuroimaging Initiative (ADNI)~\cite{adni}, PREVENT-AD~\cite{preventad}, the Adolescent Brain Cognitive Development (ABCD) study~\cite{abcd}, Reproducible Brain Charts (RBC)~\cite{shafiei2025rbc}, the Harvard Aging Brain Study--Health Disparities (HABS-HD)~\cite{habs_hd}, and multi-institutional adult glioma cohorts. The training data included structural MRI, diffusion-derived maps, NODDI-derived maps, susceptibility-weighted images and standard multiparametric glioma MRI sequences.

The training corpus was organised into sequential domains to evaluate both representation learning and continual adaptation. UK Biobank was used to establish the initial population-scale representation and was subdivided into diffusion MRI, structural MRI and NODDI-derived subsets~\cite{ukbiobank,ukbiobank_paper}. Subsequent continual-learning domains comprised D1, healthy-population imaging from HABS-HD~\cite{habs_hd}; D2, neurodegenerative disease imaging from PPMI, ADNI and PREVENT-AD~\cite{ppmi,adni,preventad}; D3, developmental and lifespan imaging from ABCD and RBC~\cite{abcd,shafiei2025rbc}; and D4, adult brain-tumour imaging from UPenn-GBM, UCSF-PDGM and the Erasmus Glioma Database~\cite{upenn_gbm,ucsf_pdgm,egd}. The D4 tumour domain included T1-weighted, contrast-enhanced T1-weighted, T2-weighted and FLAIR MRI and was placed last in the continual-learning sequence to impose a strong pathological distribution shift.

Two complementary input settings were used. In the two-channel setting, paired modalities from the same participant were provided as separate input channels, preserving within-subject correspondence across contrasts. In the single-channel setting, scans from different modalities were pooled and treated independently, enabling training when paired acquisitions were unavailable. This design allowed the model to learn from both paired multimodal datasets and heterogeneous clinical cohorts with incomplete acquisition protocols.

All MRI volumes were processed using a harmonised three-dimensional pipeline. Images were loaded in channel-first format, reoriented to canonical RAS coordinates and resampled to isotropic 1-mm resolution. Structural MRI datasets were spatially normalised to Montreal Neurological Institute space, and skull stripping was performed where required using HD-BET~\cite{isensee2019hdbet}. Intensities were normalised using percentile-based clipping and scaled to the $[0,1]$ interval. Foreground cropping and resizing were applied to obtain a fixed network input size. For two-channel inputs, the second modality was resampled to the spatial grid of the first modality before concatenation. Stochastic augmentations included random flips along anatomical axes, random 90-degree rotations and intensity perturbations. Each sub-dataset was split into 70\% training and 30\% validation partitions. 

Fine-tuning and downstream evaluation used task-specific clinical cohorts. For Alzheimer's disease progression, we used the BrainLAT dataset~\cite{brainlat} and focused on T1-weighted MRI scans with diagnostic labels, defining binary Alzheimer's disease versus healthy-control classification and three-class Alzheimer's disease, frontotemporal dementia and healthy-control classification. The final BrainLAT subset contained 527 subjects. For synthetic MRI generation and cross-modal translation, we evaluated on public brain MRI datasets spanning adult glioma, paediatric high-grade glioma, brain metastases, post-operative glioblastoma and healthy controls, including BraTS-SSA~\cite{adewole2023brain}, BraTS-MET~\cite{moawad2024brain}, BraTS-PED~\cite{kazerooni2024brain}, TCGA-LGG~\cite{pedano2016tcgalgg}, IXI~\cite{ixi} and RHUH-GBM~\cite{cepeda2023rhuh}. These datasets provided heterogeneous acquisition settings and clinically relevant contrast combinations for testing generative transfer. The downstream test-participants do not overlap with the self-supervised training and adaptation datasets.

For adult brain-tumour survival estimation, we assembled a retrospective multicentre cohort of 1,551 patients with available overall-survival time, event status and multiparametric MRI. The cohort included UPenn-GBM~\cite{upenn_gbm}, UCSF-PDGM~\cite{ucsf_pdgm}, TCGA-LGG/GBM~\cite{pedano2016tcgalgg,tcga_gbm}, LUMIERE~\cite{lumiere}, REMBRANDT~\cite{rembrandt}, RHUH-GBM~\cite{cepeda2023rhuh}, IvyGAP-GBM~\cite{ivygap_gbm} and CPTAC-GBM~\cite{cptac_gbm}. All survival analyses used the same patient list across encoder families, ensuring that 3D-Swin-DiT and comparator models were evaluated on matched imaging and outcome data. For post-surgical outcome prediction, we used the institutional SIND dataset from Addenbrooke's Hospital, comprising paired pre-operative and post-operative T2-weighted MRI scans from 49 patients. Patients were stratified into shorter-term and longer-term survival groups using a 10-month post-operative survival threshold. Together, these downstream cohorts enabled evaluation of Alcmaeon across classification, survival modelling, cross-modal synthesis and few-shot post-surgical adaptation. The pretraining and fine-tuning datasets, strategies and associated demographic and clinical information are described in Supplementary Information S5.

\subsection*{Model architecture}

The self-supervised reconstruction module, termed 3D-Swin, is based on a hierarchical shifted-window Swin Transformer encoder--decoder adapted from SwinUNETR. The encoder produces multiscale features across four stages, with patch-merging operations reducing spatial resolution and increasing channel width. The deepest feature map defines the compact latent representation. The decoder reconstructs the volume through learned upsampling and U-Net-style skip connections, preserving multiscale anatomical information. Reconstruction was optimised using structural and voxel-wise fidelity criteria, including SSIM-based loss and standard error metrics.

The latent generative module is a volumetric Diffusion Transformer, termed 3D-DiT. Latent feature maps extracted from the 3D-Swin bottleneck are tokenised into a voxelised latent grid. A forward diffusion process corrupts the latent with Gaussian noise according to a variance schedule, and the transformer denoiser learns to predict the injected noise at each diffusion step. Diffusion time is encoded with sinusoidal embeddings and injected through adaptive normalisation. At generation, the reverse process iteratively denoises a latent sample, which is devoxelised and decoded by the 3D-Swin decoder to reconstruct or translate the target MRI volume.

The full 3D-Swin-DiT architecture couples the deterministic 3D-Swin with a latent generator operating on the bottleneck representation. The bottleneck generator was implemented either as a variational autoencoder or as the proposed diffusion-only transformer formulation (3D-DiT). Model scale was varied independently of token resolution using Small, Base, Large and Extra-Large backbones, with approximate parameter counts of 33M, 131M, 580M and 676M, respectively. Additional ablations evaluated patch levels, latent hierarchy depth and input-channel configuration.

\subsection*{Self-supervised pretraining}

Pretraining was performed by reconstructing the input MRI volume from its latent representation. The 3D-Swin established the latent anatomical metric, whereas the diffusion module learned the stochastic structure of the latent distribution. For the diffusion objective (3D-DiT), clean latents were noised over 500--1,000 diffusion steps and the denoiser was trained using a simplified denoising score-matching loss between the true and predicted noise. Batch size, spatial resolution and model scale were selected by controlled ablation. Distributed batches of 128--512 produced more stable optimisation than small single-GPU batches, and a batch size of 512 was used for subsequent experiments. Both $64^3$ and $128^3$ volumes preserved major anatomical and pathological structures, but $128^3$ did not provide sufficient downstream improvement to justify the increased cost. Therefore, $64^3$ was selected as the operating resolution. Training was implemented in PyTorch and MONAI with automatic mixed precision. The detailed results and hyperparameter analyses for self-supervised training are provided in Supplementary Information S8.

\subsection*{Continual-learning strategies}

Continual learning was evaluated across the sequential domains D1--D4. Three strategies were compared. The sequential baseline (SEQ) updated all trainable parameters without an explicit retention mechanism. Elastic Weight Consolidation (EWC; \cite{kirkpatrick2017}) used a diagonal Fisher approximation to penalise changes in parameters estimated to be important for previous tasks. Graph-Blueprint Pruning replaced soft anchoring with structural preservation of task-relevant computational circuits.

GBP partitions the network into computational units, including attention, feed-forward, normalisation and adaptive-modulation modules. For each probe sample, unit-level salience was estimated using the activation--gradient product, \(s_n(x)=\mathbb{E}_{j}\left[|a_{n,j}(x)\partial_{a_{n,j}}\mathcal{L}|\right]\), where \(a_{n,j}\) is the activation of unit \(n\) and \(\mathcal{L}\) is the task objective. Salience vectors were aggregated across probe samples and compressed by principal component analysis to define a task blueprint. A blueprint score combined mean salience with participation in the dominant salience subspace. At each continual-learning stage, an upper-quantile rule selected salient units among the currently trainable complement, and these units were frozen exactly by masking their optimiser updates. Frozen sets were monotone across tasks, so each new stage operated in a residual trainable subspace of the previous model. Blueprint scores from previous stages were fused by an element-wise extremal rule, ensuring that units salient for any earlier task remained eligible for protection. Separate blueprint streams were maintained for the 3D-Swin encoder/decoder and the 3D-DiT latent generator. The approach connects mechanistic circuit attribution~\cite{anthropic2025graphs} with subnetwork-preservation continual learning~\cite{kang2022wsn,serra2018hat}

\subsection*{Blueprint anatomical attribution}

To relate GBP capacity allocation to neuroanatomical structure, atlas-based attribution was performed for each continual-learning domain. To preserve realistic anatomical variability, the atlas was aligned to a randomly selected subject from the corresponding cohort. D1 used a HABS-HD subject and the HCP-MMP1.0--Glasser MNI cortical atlas; D2 used a PPMI subject and the AAL3 anatomical atlas; D3 used an ABCD subject and the Schaefer-400 cortical parcellation with 17-network Yeo ordering; and D4 used an EGD subject and the HCP-MMP1.0--Glasser MNI cortical atlas. Previously derived blueprint files defined all identified nodes, previously frozen nodes and remaining active nodes. For each blueprint scenario, individual modules were perturbed during inference, and the resulting objective was backpropagated to the atlas input to estimate voxel-wise sensitivity. Aggregated gradients produced spatial attribution maps identifying the anatomical or functional regions most dependent on each blueprint-defined circuit subset.

\subsection*{Clinical fine-tuning}

For Alzheimer’s progression classification, BrainLAT T1-weighted MRI scans were used to define binary and three-class diagnostic tasks involving Alzheimer’s disease, frontotemporal dementia and healthy controls. Features extracted from the Swin encoder were normalised at the sample level and standardised using training-set statistics. Classification was performed using support vector machines with radial basis-function kernels and multilayer perceptrons with three hidden layers, batch normalisation, ReLU activations and dropout. Five-fold cross-validation, 70/30 hold-out testing and zero-, one-, three- and five-shot settings were evaluated.

For brain-tumour MRI synthesis, cross-modal translation was evaluated across tumour and external cohorts, including BraTS-SSA, BraTS-MET, BraTS-PED, TCGA-LGG, IXI and RHUH-GBM. Translation tasks included T1$\rightarrow$T2, T2$\rightarrow$FLAIR and T1$\rightarrow$T1ce. Competing autoencoder backbones included VQVAE, AEKL, MAISI and the proposed Swin-based codec. Generative models included latent diffusion, rectified flow and medical optimal-transport flow matching. Models were trained using recommended repository settings where available, with a learning rate of $1\times10^{-4}$, 200 epochs and a 70/10/20 train/validation/test split.

For adult brain-tumour survival estimation, frozen MRI features were extracted from 3D-Swin-DiT and compared with MONAI 3D ResNet, DUNE and HLIP encoder families under a common downstream survival protocol. The validation cohort included patients with available overall survival time, event status and skull-stripped T1, T1c, T2 and FLAIR volumes registered to MNI152 space. Each encoder family used its native preprocessing pipeline, but downstream survival modelling was kept identical across feature sets. Survival tasks were evaluated using stratified five-fold cross-validation.

For post-surgical tumour adaptation, pretrained representations were evaluated under few-shot and transfer settings across pre- and post-operative imaging. SEQ, EWC and GBP variants were compared with direct tumour fine-tuning to assess robustness under limited supervision and D4 domain shift.

The detailed experiments and hyperparameter-selection strategies for the four fine-tuning tasks are described in Supplementary Information S7.

\subsection*{Evaluation metrics}
Reconstruction and cross-modal generation were evaluated using the structural similarity index measure (SSIM), root mean squared error (RMSE), mean squared error (MSE), mean absolute error (MAE) and peak signal-to-noise ratio (PSNR). Higher SSIM and PSNR indicate better image fidelity, whereas lower RMSE, MSE and MAE indicate lower reconstruction error. Continual learning was evaluated using cross-task matrices in which rows denote the training stage and columns denote the evaluation domain. Average final performance, backward transfer, forward transfer and average forgetting were computed across D1--D4, with sign conventions adjusted for higher-is-better and lower-is-better metrics. 

Classification tasks were evaluated using accuracy, balanced accuracy, precision, sensitivity, F1-score and AUROC. Model selection in imbalanced clinical settings was based primarily on F1-score. Survival models were evaluated using concordance for time-to-event discrimination, Kaplan--Meier out-of-fold risk stratification and fixed-horizon AUROC where applicable. For fixed-horizon binary survival tasks, patients censored before the horizon were excluded because their labels were undefined. 

The evaluation metrics used across all training and fine-tuning analyses are described in Supplementary Information S6.

The use of large language model tools during manuscript preparation is described in Supplementary Information S4.5.

\section*{Data availability}

\subsection*{Training cohorts}
The data that support the training analyses in this study were obtained from public, controlled-access and institutional neuroimaging resources, as detailed in Supplementary Information S5.1. UK Biobank data were accessed under UK Biobank Application Number 20904. Data from the Parkinson's Progression Markers Initiative (PPMI) were obtained on 25 August 2025 from the PPMI database (\url{https://www.ppmi-info.org/access-dataspecimens/download-data}; RRID:SCR\_006431). Access to PPMI requires registration and a data-use agreement through the LONI Image and Data Archive (\url{https://ida.loni.usc.edu/home/projectPage.jsp?project=PPMI}). Up-to-date information on PPMI is available at \url{https://www.ppmi-info.org}. PPMI is a public--private partnership funded by The Michael J. Fox Foundation for Parkinson's Research and its funding partners. ADNI data were accessed through the Alzheimer's Disease Neuroimaging Initiative/LONI portal (\url{https://adni.loni.usc.edu/}) and require registration and approval of the ADNI data-use agreement.

The ABCD diffusion-derived data used in this study were obtained from the Fiber Data Hub ABCD release (\url{https://brain.labsolver.org/abcd.html}). This resource provides derived ABCD FIB files, including baseline ($n=9,713$), 2-year follow-up ($n=7,199$) and 4-year follow-up ($n=2,736$) data, under the Creative Commons Attribution--ShareAlike 4.0 International licence. These files are derived diffusion reconstructions rather than raw MRI data; therefore, no additional NDA request was required for the derived FIB files used here. Raw ABCD MRI data, including T1-weighted, diffusion-weighted and SRC.GZ files, remain subject to the National Data Archive data-use agreement and are not redistributed in this study. Additional training resources included the Reproducible Brain Charts dataset (\url{https://reprobrainchart.github.io/docs/datasets/}), HABS-HD through LONI (\url{https://ida.loni.usc.edu/home/projectPage.jsp?project=HABS_HD}) and PREVENT-AD through the open PREVENT-AD/LORIS portal (\url{https://openpreventad.loris.ca/}).

The tumour-domain training data were assembled from multi-institutional glioma MRI cohorts with public, controlled-access or data-use-agreement provenance. These included UPenn-GBM (\url{https://doi.org/10.7937/TCIA.709X-DN49}) and UCSF-PDGM (\url{https://doi.org/10.7937/TCIA.BDGF-8V37}) from The Cancer Imaging Archive. The Erasmus Glioma Database was accessed through the BMIA XNAT repository (\url{https://xnat.bmia.nl/data/archive/projects/egd}) under a custom licence based on CC BY-NC-SA 4.0.

\subsection*{Fine-tuning cohorts}

The data that support the fine-tuning analyses were obtained from public, controlled-access and institutional neuroimaging datasets, as detailed in Supplementary Information S5.2. Public and repository-based glioma imaging resources were accessed under their original data-use terms and licences. The generation and translation tasks used BraTS 2025 through Synapse (\url{https://www.synapse.org/Synapse:syn64153130/wiki/630130}), TCGA-LGG (\url{https://www.cancerimagingarchive.net/collection/tcga-lgg/}) and RHUH-GBM (\url{https://www.cancerimagingarchive.net/collection/rhuh-gbm/}).

Public or repository access links for the adult brain-tumour survival cohorts are as follows: UPenn-GBM (\url{https://doi.org/10.7937/TCIA.709X-DN49}), TCGA-LGG (\url{https://doi.org/10.7937/K9/TCIA.2016.L4LTD3TK}), TCGA-GBM (\url{https://doi.org/10.7937/K9/TCIA.2016.RNYFUYE9}), LUMIERE (\url{https://doi.org/10.6084/m9.figshare.c.5904905.v1}), REMBRANDT (\url{https://doi.org/10.7937/K9/TCIA.2015.588OZUZB}), RHUH-GBM (\url{https://doi.org/10.7937/4545-C905}), IvyGAP-GBM (\url{https://doi.org/10.7937/K9/TCIA.2016.XLWAN6NL}) and CPTAC-GBM (\url{https://doi.org/10.7937/K9/TCIA.2018.3RJE41Q1}). LUMIERE was obtained from Figshare under CC BY 4.0. Across these resources, the available MRI sequences comprised the standard multiparametric glioma protocol: T1-weighted, contrast-enhanced T1-weighted, T2-weighted and FLAIR imaging. The BrainLAT dataset used for Alzheimer's disease and neurodegenerative-disease analyses is publicly available through Synapse (\url{https://www.synapse.org/Synapse:syn51549340/wiki/624187}).

The institutional SIND dataset cannot be deposited publicly or released through general request because of participant privacy, governance and data-use restrictions. Scientific collaborations may be considered by contacting S.J.P. (\href{mailto:sjp58@cam.ac.uk}{sjp58@cam.ac.uk}) or the corresponding author M.M. (\href{mailto:mm2703@cam.ac.uk}{mm2703@cam.ac.uk}); responses are expected within approximately two weeks. Any access would require independent institutional approvals and a formal inter-institutional data-use agreement. Associated Source Data files for submission should include machine-readable numerical values underlying all main figures, Extended Data figures, retained Supplementary Tables and plotted Supplementary Information results. Aggregated values that can be shared without breaching data-use agreements will be uploaded as Source Data files at submission. If any Source Data file cannot be uploaded at initial submission, the same numerical source data will be provided to editors and reviewers during assessment and deposited in a persistent repository before publication.

\section*{Code availability}
Code for reproducing the models and analyses will be made publicly available upon publication at \url{https://github.com/ece7048/Alcmaeon}, and the trained model weights will be released at \url{https://huggingface.co/ece7048/Alcmaeon}. Private access to the code and trained checkpoints will be provided to editors and reviewers during assessment. For publication, a versioned release of the code and trained checkpoints will be archived in a DOI-minting repository such as Zenodo, in addition to GitHub and Hugging Face.

\section*{Ethics approval and data governance}
This study analysed de-identified public, controlled-access and institutional neuroimaging data. For public and controlled-access datasets, participant consent, ethics approval and data governance were obtained by the original studies and repositories. No attempt was made to re-identify participants. Use of controlled-access resources, including UK Biobank, PPMI, ADNI and TCIA/GDC-linked datasets, complied with the access conditions and data-use agreements specified by the original data providers. This research was conducted using the UK Biobank Resource under Application Number 20904. Derived ABCD diffusion files were used from the publicly released Fiber Data Hub resource; raw ABCD MRI data were not redistributed and remain governed by National Data Archive access conditions.

The institutional SIND dataset was collected as part of the Assessing the impact of surgically induced deficits on patient functioning and quality of life study (SIND study; protocol SIND-2018). The study received a favourable ethical opinion from the West Midlands–Solihull Research Ethics Committee on 16 May 2019 (REC reference 19/WM/0152; IRAS project ID 256542). All participants provided written informed consent. S.J.P. was responsible for the ethical oversight and governance of the SIND dataset.

\subsection*{Data preparation of Alzheimer's Disease Neuroimaging Initiative (ADNI)}
Data used in the preparation of this article were obtained from the Alzheimer's Disease
Neuroimaging Initiative (ADNI) database (adni.loni.usc.edu). The ADNI was launched in
2003 as a public-private partnership, led by Principal Investigator Michael W. Weiner, MD.
The original goal of ADNI was to test whether serial magnetic resonance imaging (MRI),
positron emission tomography (PET), other biological markers, and clinical and
neuropsychological assessment can be combined to measure the progression of mild
cognitive impairment (MCI) and early Alzheimer's disease (AD). The current goals include
validating biomarkers for clinical trials, improving the generalizability of ADNI data by
increasing diversity in the participant cohort, and to provide data concerning the diagnosis
and progression of Alzheimer’s disease to the scientific community. For up-to-date
information, see adni.loni.usc.edu.

\section*{Acknowledgements}
Computations were performed in part on the Kelvin-2 High Performance Computing Facility, a UK Tier-2 National High Performance Computing facility funded by the Engineering and Physical Sciences Research Council and jointly managed by Queen's University Belfast and Ulster University. We acknowledge the facilities of the Research Computing Services of the University of Cambridge. This research was conducted using the UK Biobank Resource under Application Number 20904. We thank Rafael Romero-Garcia, Varun Warrier, Clara Pecci Terroba and Marcella Montagnese for their contributions to the curation and preprocessing of the UK Biobank cohort data.

Data used in the preparation of this article were obtained on 25 August 2025 from the Parkinson's Progression Markers Initiative (PPMI) database (\url{https://www.ppmi-info.org/access-dataspecimens/download-data}; RRID:SCR\_006431). For up-to-date information on the study, see \url{https://www.ppmi-info.org}. PPMI is sponsored and partially funded by The Michael J. Fox Foundation for Parkinson's Research, with support from the Aligning Science Across Parkinson's initiative. Additional support is provided by a consortium of industry partners, non-profit organisations and private individuals through financial and in-kind contributions. PPMI partners contribute to the study through the Partner Scientific Advisory Board, which provides feedback on study parameters and supports the evaluation of candidate precision-medicine measures for clinical testing.

Data collection and sharing for the Alzheimer's Disease Neuroimaging Initiative (ADNI) were funded by the National Institute on Aging, National Institutes of Health Grant U19AG024904. The grantee organization is the Northern California Institute for Research and Education. ADNI has also received funding from the National Institute of Biomedical Imaging and Bioengineering, the Canadian Institutes of Health Research and private-sector contributions through the Foundation for the National Institutes of Health (FNIH), including contributions from AbbVie; Alzheimer's Association; Alzheimer's Drug Discovery Foundation; Araclon Biotech; BioClinica, Inc.; Biogen; Bristol-Myers Squibb Company; CereSpir, Inc.; Cogstate; Eisai Inc.; Elan Pharmaceuticals, Inc.; Eli Lilly and Company; EuroImmun; F. Hoffmann-La Roche Ltd and its affiliated company Genentech, Inc.; Fujirebio; GE Healthcare; IXICO Ltd.; Janssen Alzheimer Immunotherapy Research \& Development, LLC; Johnson \& Johnson Pharmaceutical Research \& Development LLC; Lumosity; Lundbeck; Merck \& Co., Inc.; Meso Scale Diagnostics, LLC; NeuroRx Research; Neurotrack Technologies; Novartis Pharmaceuticals Corporation; Pfizer Inc.; Piramal Imaging; Servier; Takeda Pharmaceutical Company; and Transition Therapeutics.

The model name Alcmaeon acknowledges Alcmaeon of Croton, an ancient Greek figure in medical thought associated with linking the brain to perception and cognition.

\section*{Funding}
This work was supported by grants to R.J.G. from the Cancer Research UK Cambridge Centre and the Cancer Research UK Children's Brain Tumour Centre of Excellence. M.M. received high-performance-computing resources through UKRI Access to High Performance Computing facilities -- autumn 2025, under allocation APP94963, ``From MRI to Multi-Omics: An Explainable Foundation Model for Brain Cancer''. The work by C.J.M. is part of grant JDC2023-051807-I funded by MICIU/AEI/10.13039/501100011033 and by ESF+. S.J.P. was supported by the National Institute for Health and Care Research Career Development Fellowship CDF2018-11-ST2-003. Research at the Department of Psychiatry, University of Cambridge, is supported by the NIHR Cambridge Biomedical Research Centre (NIHR203312) and the NIHR Applied Research Collaboration East of England. The SIND-related work was supported by the NIHR HealthTech Research Centre in Brain Injury and the NIHR Cambridge Biomedical Research Centre (NIHR203312, S.J.P.). The views expressed are those of the authors and not necessarily those of the NHS, the NIHR or the Department of Health and Social Care.

\section*{Supplementary methodological details}

Further mathematical, methodological and implementation details are provided in the Supplementary Information. These include the formal problem setting and latent-variable notation; the Swin-Wrap encoder--decoder and hybrid 3D-Swin-DiT formulation; voxelisation and tokenisation, three-dimensional windowed self-attention, adaptive normalisation and de-voxelisation; the forward and reverse diffusion processes and denoising objective; and the derivations of elastic weight consolidation and Graph-Blueprint Pruning for continual learning. The Supplementary Information also describes dataset organisation, preprocessing and harmonisation, feature-extraction and fine-tuning protocols, optimisation and evaluation procedures, and ablations of model scale, spatial resolution, patch size, hierarchical depth, channel configuration and batch size. Additional sections provide clinical-task-specific methods and benchmarks for neurodegenerative-disease classification, brain-tumour reconstruction and cross-modal generation, and glioma survival modelling, together with few-shot evaluations, latent-space analyses and atlas-based attribution of blueprint-defined circuits. Collectively, these sections provide the mathematical background, step-by-step implementation details and supporting analyses required to reproduce the study while keeping the main Methods focused on the experimental design and core procedures.

\section*{Author information}
\textbf{University of Cambridge, Cambridge, UK}\\
Michail Mamalakis, Yonghao Li, Hao Chen, Chao Li, John Suckling, Richard Bethlehem, Stephen J. Price, Pietro Lio\\
\textbf{Cancer Research UK Cambridge Institute, University of Cambridge Li Ka Shing Centre, Cambridge, UK}\\
Richard J. Gilbertson, Michail Mamalakis\\
\textbf{University of Malaga, Malaga, Spain}\\
Carmen Jimenez-Mesa\\
\textbf{University of Virginia, Charlottesville, USA}\\
Antonios Mamalakis

\section*{Author contributions}
M.M. conceived the study, formulated the Alcmaeon framework, implemented the model architecture, developed the training strategy, designed and implemented Graph-Blueprint Pruning, performed the explainability and interpretability analyses, and supervised the experimental evaluation. M.M. also coordinated the overall analysis, interpreted the results and wrote the first draft of the manuscript.
R.B. curated and provided the UK Biobank data and contributed neuroscience feedback throughout the project. J.S. contributed neuroscience guidance and supported the coordination, organisation and storage of the D1--D4 datasets. Y.L. and S.J.P. provided resources and domain expertise for the D4 brain-tumour cohorts and the SIND dataset. Y.L. developed and evaluated the adult brain-tumour survival and fine-tuning pipelines under the supervision of M.M. and S.J.P. C.J. performed the neurodegenerative-disease fine-tuning analyses under the supervision of M.M. and J.S. M.M. and C.J. performed the post-surgical survival fine-tuning and analysis with clinical feedback from S.J.P. H.C. carried out the brain-tumour synthetic generation and cross-modal translation experiments under the supervision of M.M. and C.L.
P.L. provided high-performance computing and storage resources. A.M. contributed to the formulation and interpretation of the explainability and interpretability of the foundation model. R.J.G. and S.J.P. provided clinical guidance on brain-tumour tasks, feature importance and interpretation. R.J.G. and P.L. contributed project-management guidance, research-timeline planning and mentorship. All authors contributed to manuscript review and editing.

\section*{Competing interests}
The authors declare no competing interests.

\section*{Additional information}
Supplementary Information is available for this paper. Correspondence and requests for materials should be addressed to M.M.

\section*{Extended Figures}
\setcounter{figure}{0}
\renewcommand{\thefigure}{\arabic{figure}}
\renewcommand{\figurename}{Extended Fig}
\begin{figure}[h!]
  \centering
  \includegraphics[
    width=1.0\textwidth,
    trim={0.0cm 0.0cm 0.0cm 0.0cm},
    clip
  ]{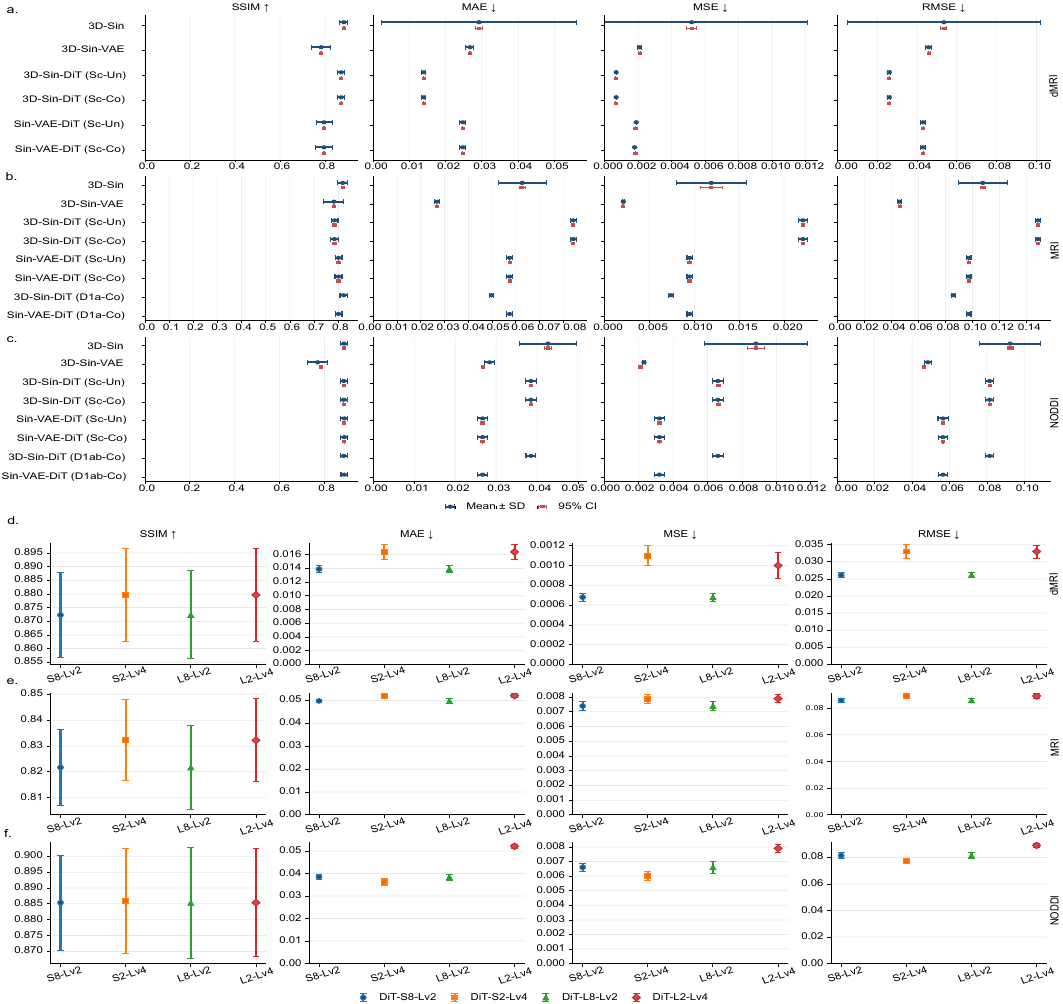}
  \caption{
  \textbf{Self-supervised reconstruction and scaling behaviour of
  3D-Swin-DiT.}
  \textbf{a--c,} Reconstruction performance of S8-Lv2 models on UK Biobank
  diffusion MRI (dMRI; \textbf{a}), structural MRI (\textbf{b}) and
  neurite orientation dispersion and density imaging (NODDI; \textbf{c}).
  Deterministic 3D-Swin and variational 3D-Swin-VAE models are compared with
  3D-Swin-DiT and Swin-VAE-DiT diffusion models under the indicated
  conditioning configurations. Performance is
  reported using structural similarity (SSIM), mean absolute error (MAE),
  mean squared error (MSE) and root mean squared error (RMSE). Higher SSIM and
  lower MAE, MSE and RMSE indicate better reconstruction. Blue intervals
  denote the mean~$\pm$~s.d., and red intervals denote the 95\% confidence
  interval where available.
  \textbf{d--f,} Effect of model scale and latent level on reconstruction of
  dMRI (\textbf{d}), structural MRI (\textbf{e}) and NODDI (\textbf{f}).
  DiT-S8-Lv2, DiT-S2-Lv4, DiT-L8-Lv2 and DiT-L2-Lv4 models are compared using
  the same four reconstruction metrics. 
  Sc: training from scratch; Sc-Un: diffusion without conditioning; Sc-Co: diffusion with conditioning; D1a-Co and D1ab-Co denote pretrained diffusion models.
  }
  \label{fig:pretrain}
\end{figure}

\begin{figure}[h!]
  \centering
  \includegraphics[
    width=1.0\textwidth,
    trim={0.0cm 0.0cm 0.0cm 0.0cm},
    clip
  ]{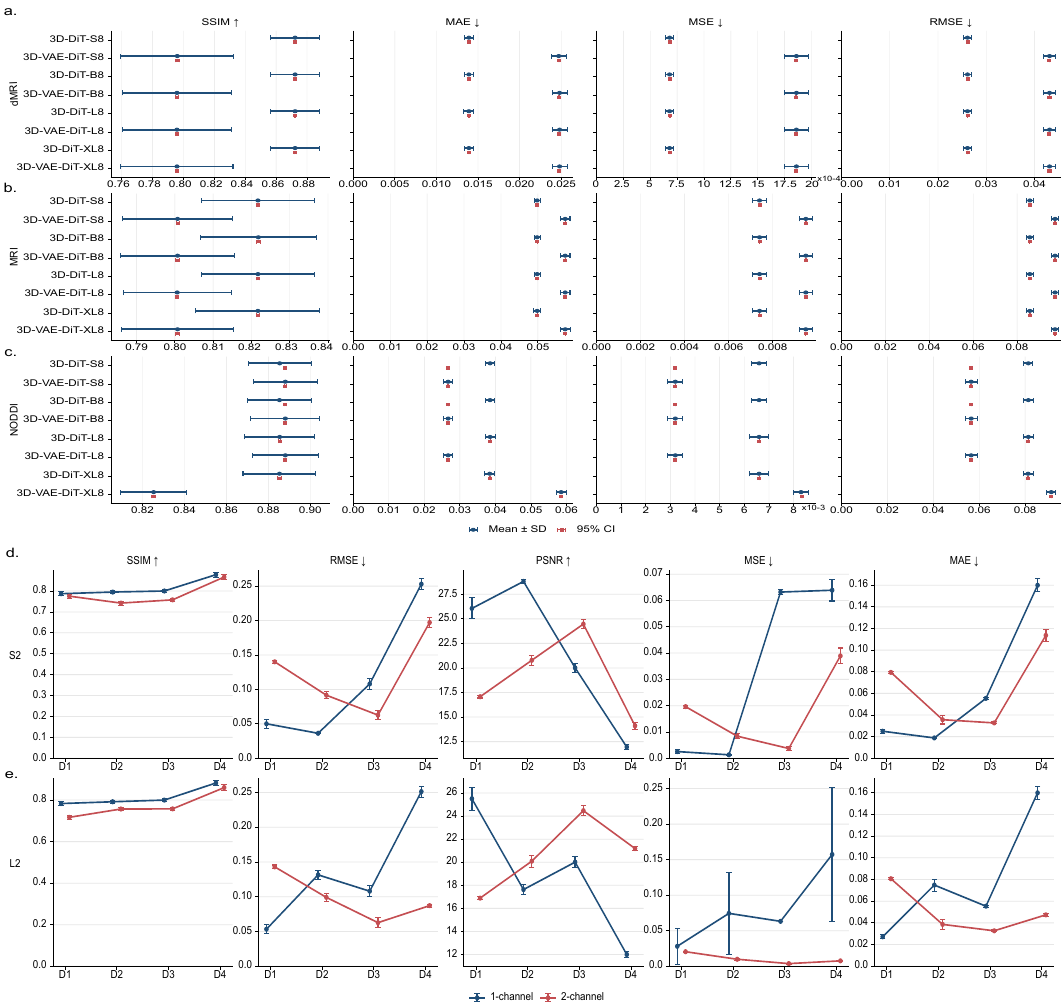}
  \caption{
  \textbf{Effects of architecture scale and channel configuration on
  self-supervised reconstruction.}
  \textbf{a--c,} Reconstruction performance of deterministic 3D-DiT and
  variational 3D-VAE-DiT models across Small (S8), Base (B8), Large (L8) and
  Extra-Large (XL8) scales, evaluated on UK Biobank diffusion MRI (dMRI;
  \textbf{a}), structural MRI (\textbf{b}) and neurite orientation dispersion
  and density imaging (NODDI; \textbf{c}). Performance is reported using
  structural similarity (SSIM), mean absolute error (MAE), mean squared error
  (MSE) and root mean squared error (RMSE). Higher SSIM and lower MAE, MSE and
  RMSE indicate better reconstruction. Blue intervals denote the mean~$\pm$~s.d., and red intervals
  denote the 95\% confidence interval.
  \textbf{d,e,} Comparison of one-channel (blue) and two-channel (red)
  reconstruction using S2 (\textbf{d}) and L2 (\textbf{e}) models across
  clinical domains D1--D4. Models are evaluated using SSIM, RMSE, peak
  signal-to-noise ratio (PSNR), MSE and MAE; higher SSIM and PSNR and lower
  RMSE, MSE and MAE indicate better performance. The relative advantage of
  each channel configuration varies across domains: one-channel models perform
  favourably on D1, whereas two-channel models generally yield lower
  voxel-wise errors on D3 and D4.
  }
  \label{fig:pretrainb}
\end{figure}

\begin{figure}[h!]
  \centering
  \includegraphics[width=1.0\textwidth,
trim={0.0cm 0.0cm 0.0cm 0.0cm},
clip]{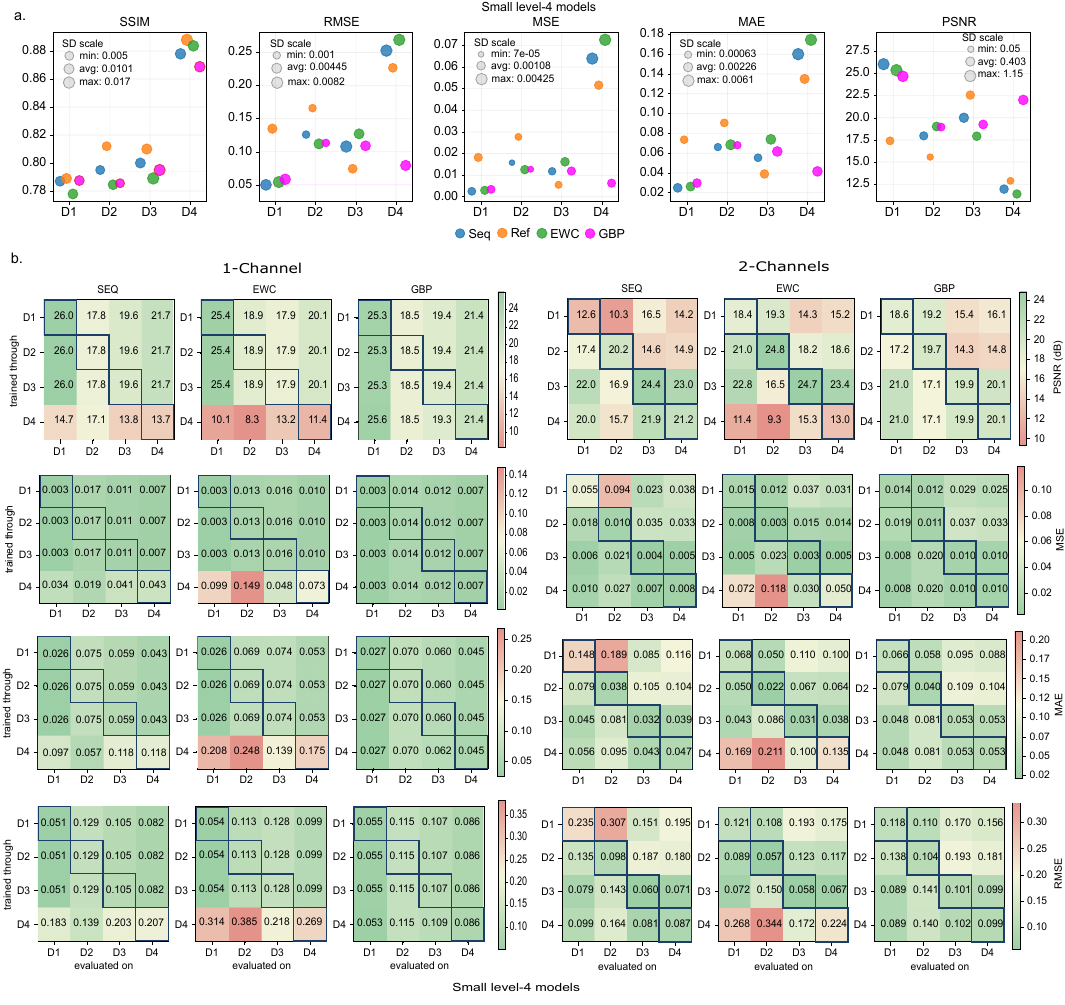}
\caption{
\textbf{Continual-learning performance across channel configurations.}
\textbf{a.} Bubble plots compare SEQ, EWC and GBP for single-channel small Lv4 model across SSIM, RMSE, MSE, MAE and PSNR. Bubble size is proportional to the standard deviation; higher SSIM and PSNR indicate better reconstruction, whereas lower RMSE, MSE and MAE indicate lower error.
\textbf{b.} Cross-task evaluation matrices for one-channel and two-channel models. Rows denote the evaluation dataset and columns denote the sequential training stage. Green shades indicate retained or improved performance, whereas red shades indicate degradation. GBP preserves reconstruction fidelity more consistently than SEQ or EWC, particularly in later domains and under stronger distribution shift.
}
  \label{fig:forgetting}
\end{figure}

\begin{figure}[h!]
  \centering
  \includegraphics[width=1.0\textwidth,
trim={0.0cm 0.0cm 0.0cm 0.0cm},
clip]{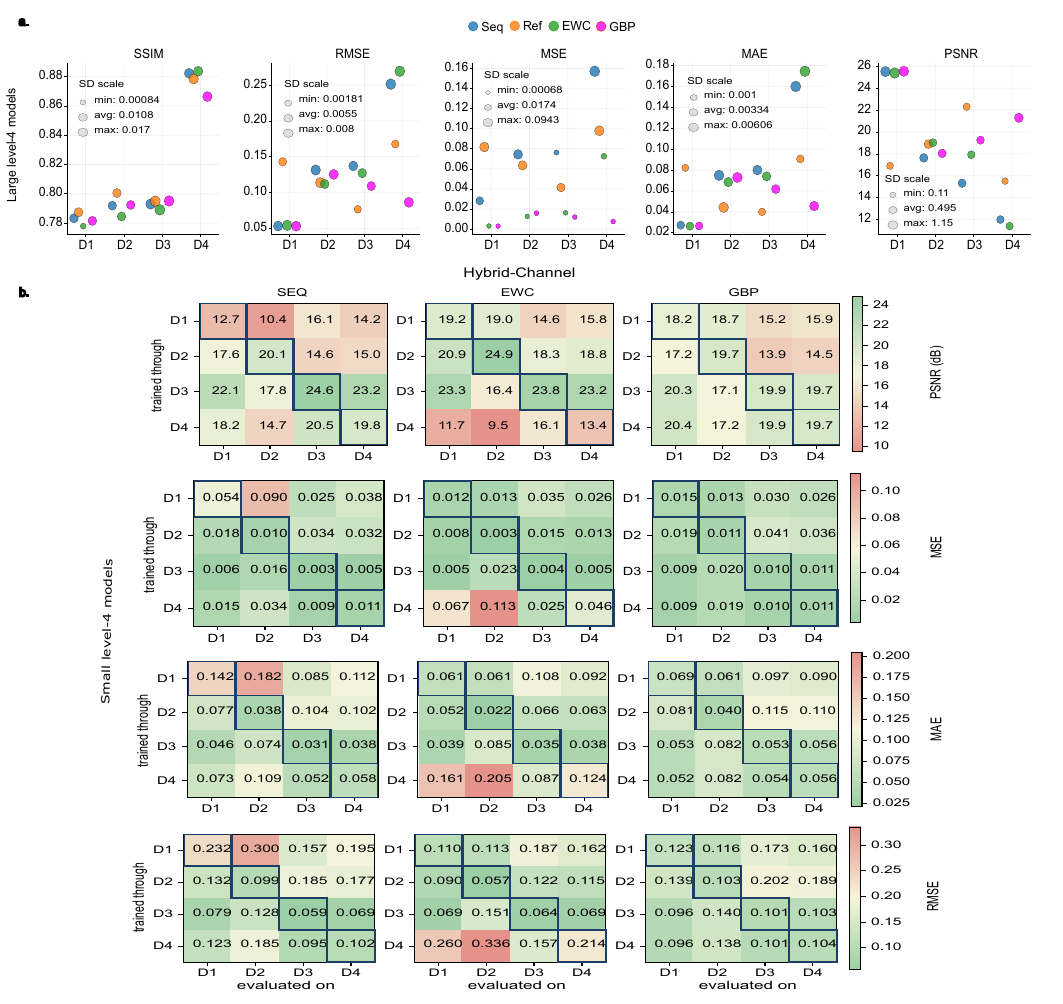}
\caption{
\textbf{Continual-learning performance across channel configurations.}
\textbf{a.} Bubble plots compare SEQ, EWC and GBP for single-channel large Lv4 models across SSIM, RMSE, MSE, MAE and PSNR. Bubble size is proportional to the standard deviation; higher SSIM and PSNR indicate better reconstruction, whereas lower RMSE, MSE and MAE indicate lower error.
\textbf{b.} Cross-task evaluation matrices for hybrid-channel model. Rows denote the evaluation dataset and columns denote the sequential training stage. Green shades indicate retained or improved performance, whereas red shades indicate degradation. GBP preserves reconstruction fidelity more consistently than SEQ or EWC, particularly in later domains and under stronger distribution shift.
}
  \label{fig:forgettingb}
\end{figure}

\begin{figure}[h!]
  \centering
  \includegraphics[
    width=1.0\textwidth,
    trim={0.0cm 0.0cm 0.0cm 0.0cm},
    clip
  ]{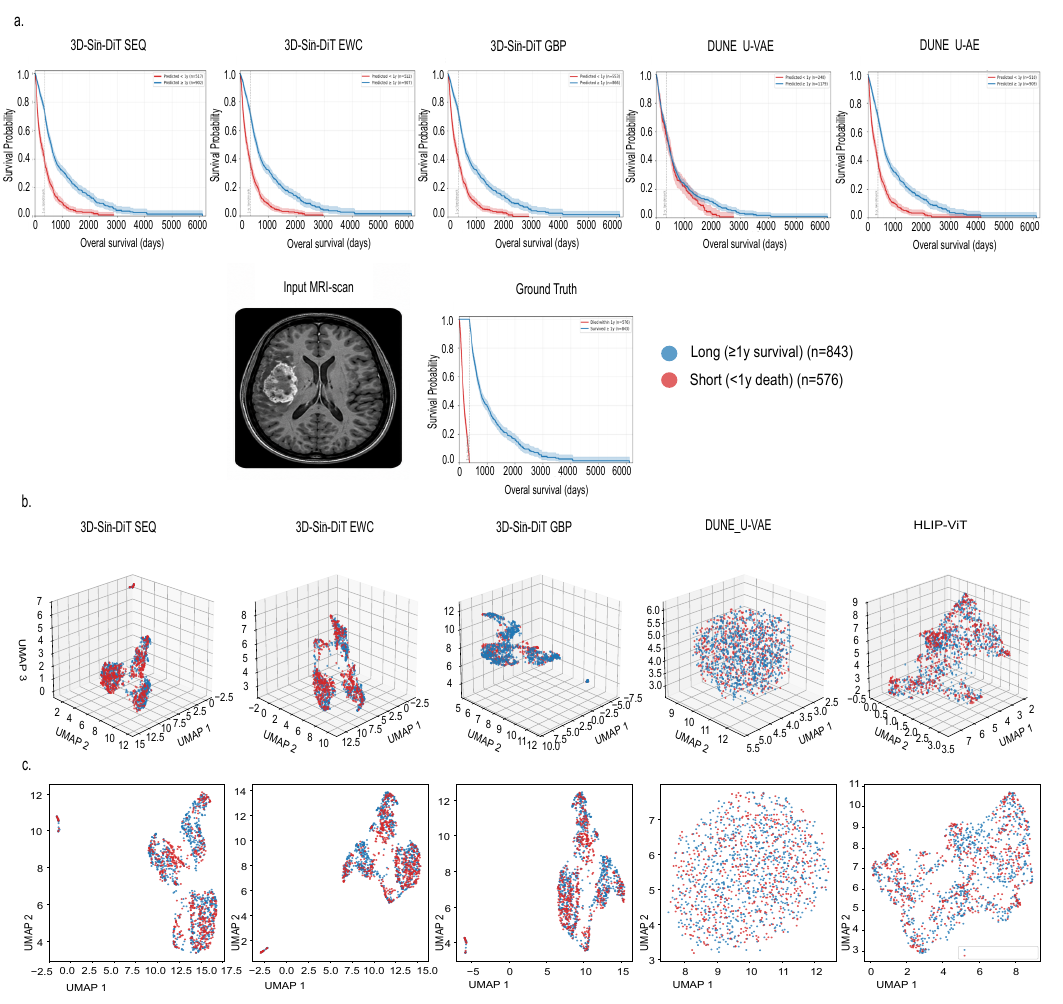}
  \caption{
  \textbf{Clinical adaptation of Alcmaeon for adult brain-tumour survival
  prediction.}
  \textbf{a,} Kaplan--Meier curves derived from out-of-fold predictions of
  3D-Swin-DiT adapted using unconstrained sequential fine-tuning (SEQ),
  elastic weight consolidation (EWC) or Graph-Blueprint Pruning (GBP), with
  DUNE U-VAE and DUNE U-AE included as comparison models. Predicted survival
  distributions are shown for patients with long survival
  (\(\geq 1\) year; \(n=843\), blue) and short survival
  (\(<1\) year; \(n=576\), red), together with the corresponding
  outcome-defined Kaplan--Meier reference and a representative input MRI
  scan. Shaded regions indicate uncertainty around the estimated survival
  curves.
  \textbf{b,} Three-dimensional UMAP projections of PCA-reduced latent
  features learned by 3D-Swin-DiT with SEQ, EWC or GBP, compared with
  DUNE U-VAE and HLIP-ViT. Points represent individual participants and are
  coloured according to the observed survival group defined in
  \textbf{a}.
  \textbf{c,} Corresponding two-dimensional UMAP projections of the same
  latent representations, showing the spatial organisation and overlap of
  short- and long-survival cases. The projections illustrate
  method-dependent differences in latent-space geometry and in the degree to
  which survival-related information is organised within the learned
  representation.
  }
  \label{fig:finetunea}
\end{figure}

\begin{figure}[h!]
  \centering
  \includegraphics[
    width=1.0\textwidth,
    trim={0.0cm 0.0cm 0.0cm 0.0cm},
    clip
  ]{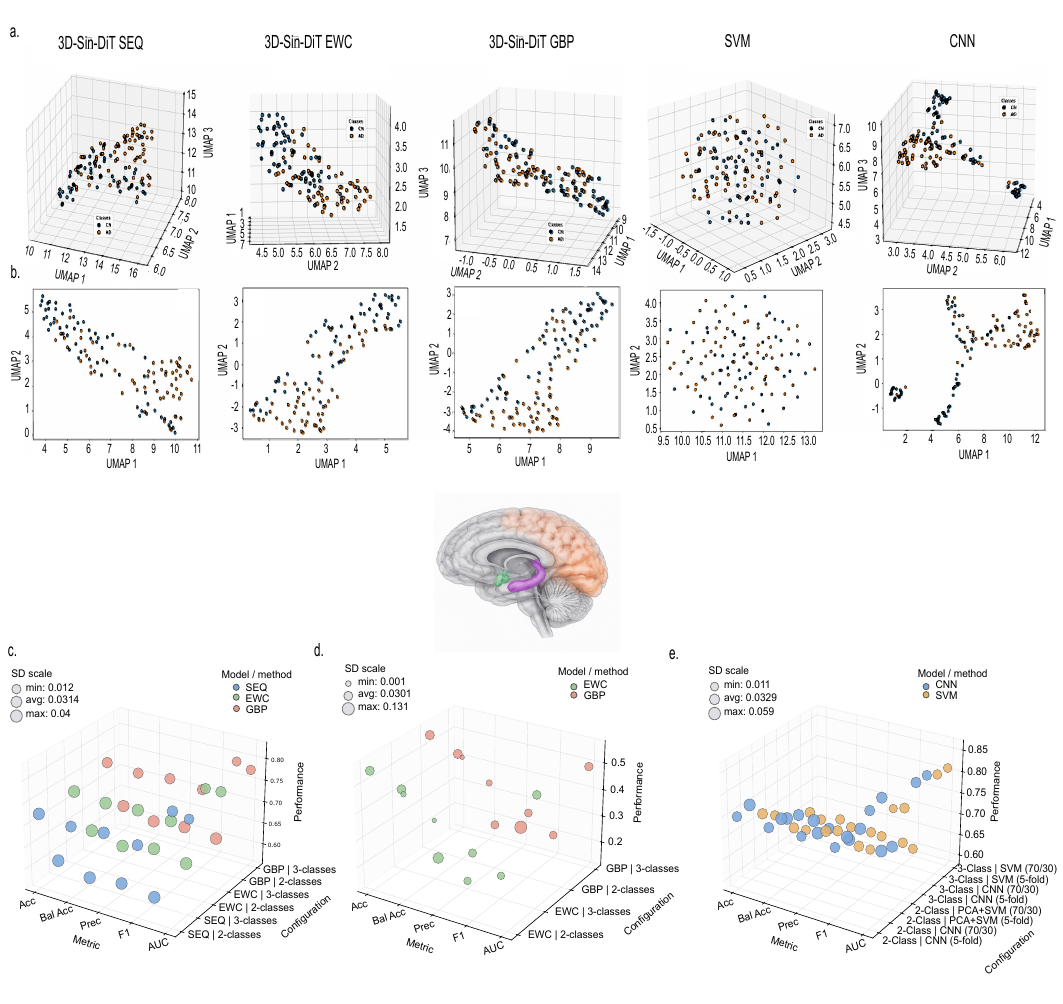}
  \caption{
  \textbf{Clinical adaptation of Alcmaeon for modelling Alzheimer's disease
  progression.}
  \textbf{a,} Three-dimensional UMAP projections of PCA-reduced latent
  features from one-channel 3D-Swin-DiT models adapted using unconstrained
  sequential fine-tuning (SEQ), elastic weight consolidation (EWC) or
  Graph-Blueprint Pruning (GBP). Conventional support vector machine (SVM)
  and convolutional neural network (CNN) feature spaces are shown for
  comparison. Points represent individual participants and are coloured by
  diagnostic group: cognitively normal (CN), mild cognitive impairment (MCI)
  or Alzheimer's disease (AD).
  \textbf{b,} Corresponding two-dimensional UMAP projections of the same
  representations, showing method-dependent differences in the organisation
  and overlap of the three diagnostic groups. The central brain rendering provides a most common anatomical representation of the regions associated in Alzheimer's disease.
  \textbf{c,} Downstream two-class and three-class classification performance
  obtained from the one-channel 3D-Swin-DiT representations learned using
  SEQ, EWC or GBP.
  \textbf{d,} Corresponding two-class and three-class classification results
  for the complete 3D-Swin-DiT configurations adapted using EWC or GBP.
  \textbf{e,} Classification baselines based on SVM and CNN models, including
  two-class and three-class evaluations using either a 70/30 train--test
  split or five-fold cross-validation. In \textbf{c--e}, performance is
  summarized using accuracy (Acc), balanced accuracy (Bal Acc), precision
  (Prec), F1 score and area under the receiver-operating-characteristic curve
  (AUC), together with the corresponding evaluation configuration. Point
  size represents the s.d. according to the scale shown in each panel.
  }
  \label{fig:finetuneb}
\end{figure}

\begin{figure}[h!]
  \centering
  \includegraphics[width=1.0\textwidth,
trim={0.0cm 0.0cm 0.0cm 0.0cm},
clip]{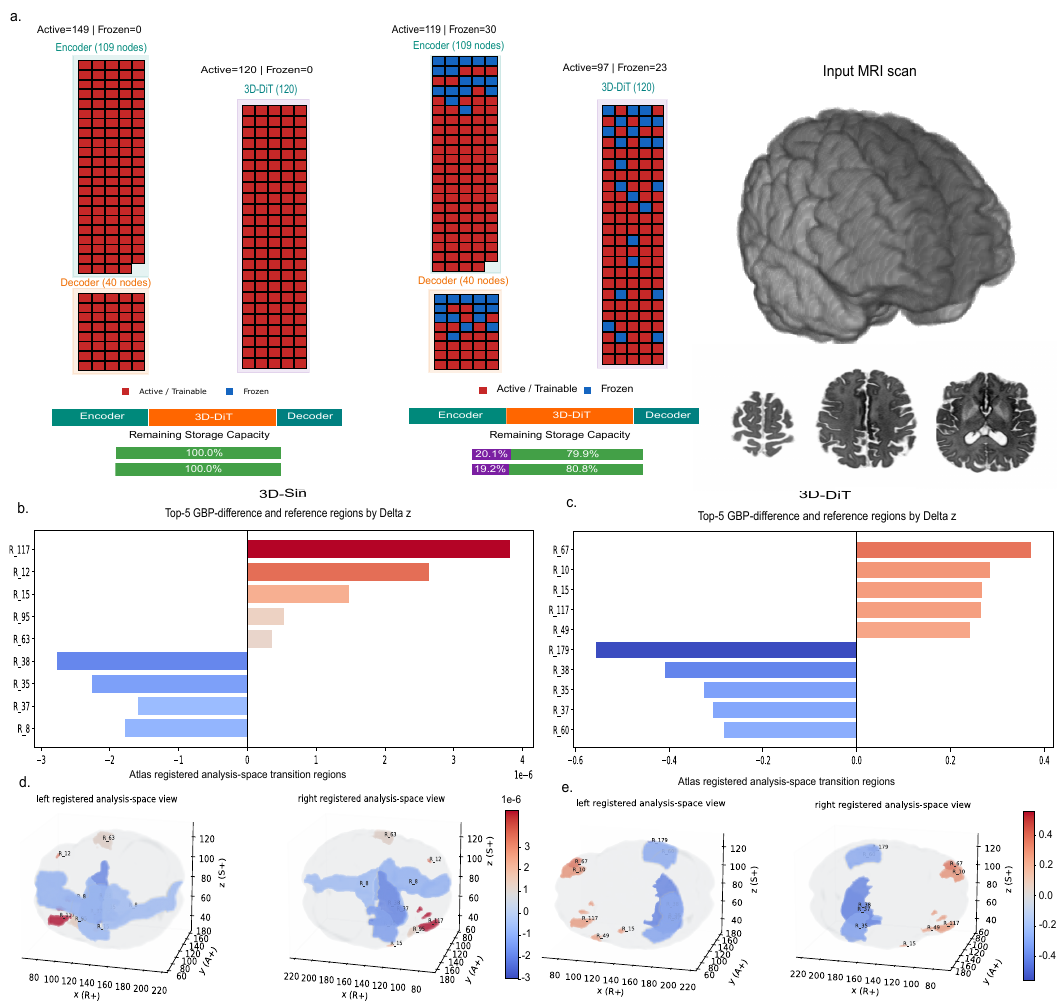}
\caption{
\textbf{Graph-Blueprint signatures of capacity allocation and anatomical interpretation during continual neuroimaging adaptation.}
\textbf{(a--e)} Blueprint maps are shown across the sequential domains D1 healthy population. GBP identifies task-relevant circuits in the 3D-DiT latent generator and 3D-Swin encoder--decoder, freezes the selected units as structural memory and leaves the remaining units trainable for subsequent adaptation. Active and frozen units are displayed as node grids, with accompanying capacity bars indicating the fraction of preserved and remaining trainable capacity.
Ranked blueprint scores identify the most influential units contributing to each domain-specific signature, illustrating that freezing is structured rather than uniformly distributed across the model. The corresponding latent-space and anatomical projections show that blueprint patterns become increasingly domain-specific across the continual-learning trajectory, from more diffuse healthy-population signatures to neurodegenerative, developmental/psychiatric and tumour-associated patterns. Together, these visualisations show that GBP provides both a consolidation mechanism for reducing forgetting and an interpretable structural readout of how Alcmaeon reallocates model capacity across heterogeneous neuroimaging domains.
}

  \label{fig:blueprinta}
\end{figure}

\begin{figure}[h!]
  \centering
  \includegraphics[width=1.0\textwidth,
trim={0.0cm 0.0cm 0.0cm 0.0cm},
clip]{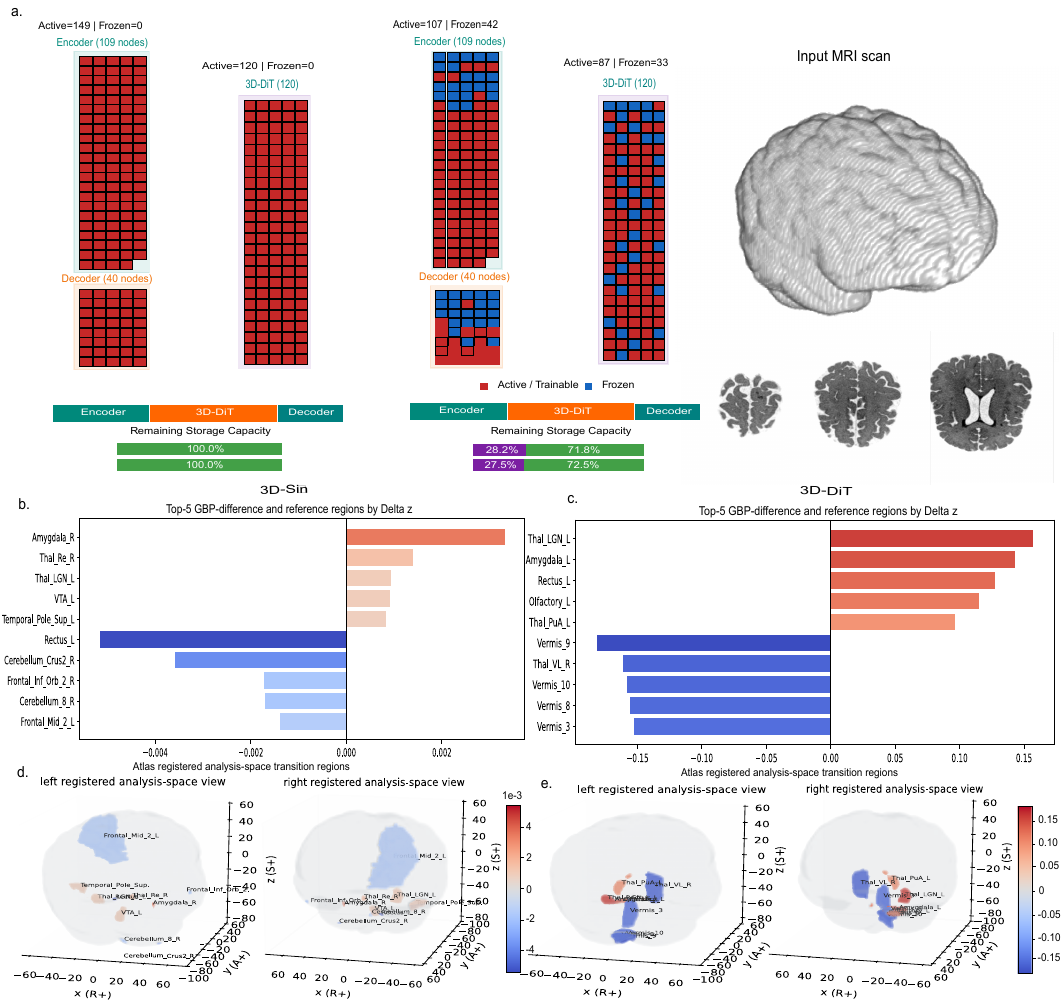}
\caption{
\textbf{Graph-Blueprint signatures of capacity allocation and anatomical interpretation during continual neuroimaging adaptation.}
\textbf{(a--e)} Blueprint maps are shown across the sequential domains D2 neurodegeneration. GBP identifies task-relevant circuits in the 3D-DiT latent generator and 3D-Swin encoder--decoder, freezes the selected units as structural memory and leaves the remaining units trainable for subsequent adaptation. Active and frozen units are displayed as node grids, with accompanying capacity bars indicating the fraction of preserved and remaining trainable capacity.
Ranked blueprint scores identify the most influential units contributing to each domain-specific signature, illustrating that freezing is structured rather than uniformly distributed across the model. The corresponding latent-space and anatomical projections show that blueprint patterns become increasingly domain-specific across the continual-learning trajectory, from more diffuse healthy-population signatures to neurodegenerative, developmental/psychiatric and tumour-associated patterns. Together, these visualisations show that GBP provides both a consolidation mechanism for reducing forgetting and an interpretable structural readout of how Alcmaeon reallocates model capacity across heterogeneous neuroimaging domains.
}

  \label{fig:blueprintb}
\end{figure}

\begin{figure}[h!]
  \centering
  \includegraphics[width=1.0\textwidth,
trim={0.0cm 0.0cm 0.0cm 0.0cm},
clip]{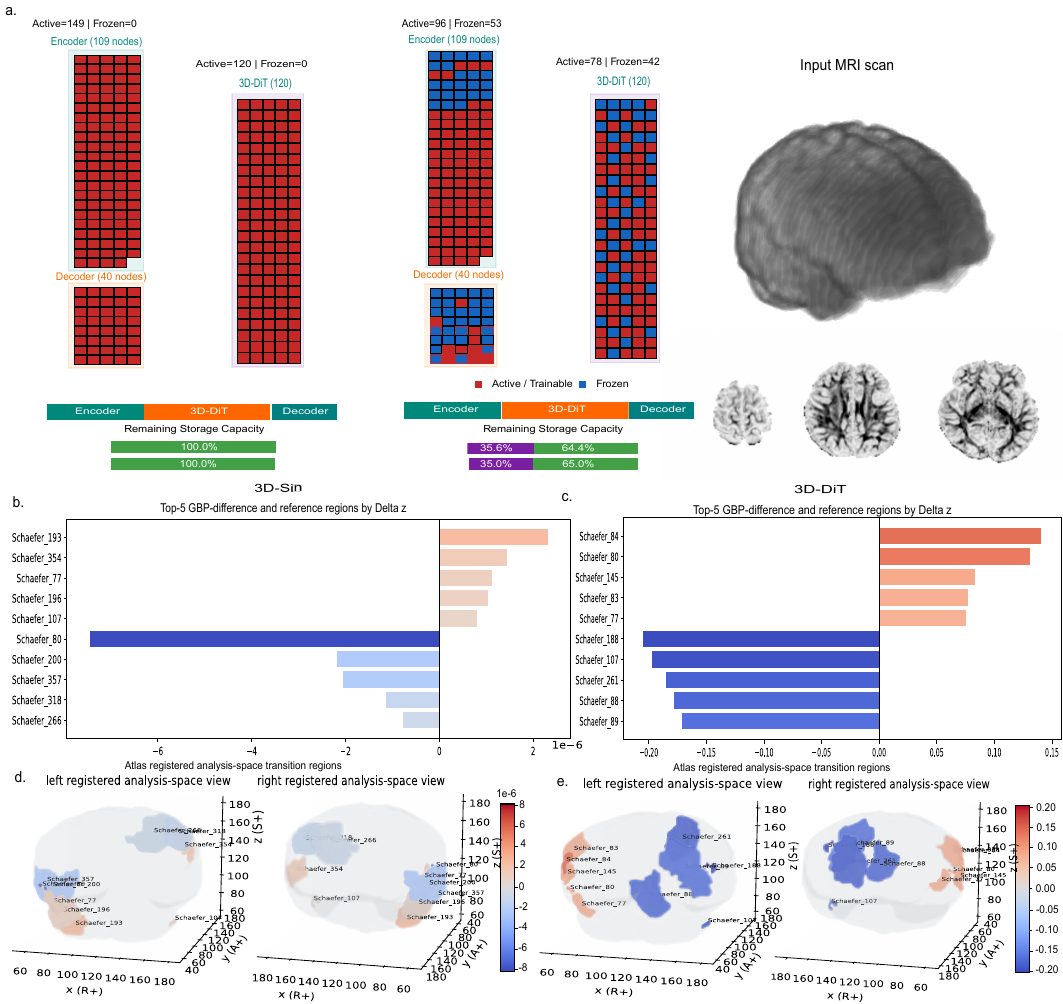}
\caption{
\textbf{Graph-Blueprint signatures of capacity allocation and anatomical interpretation during continual neuroimaging adaptation.}
\textbf{(a--e)} Blueprint maps are shown across the sequential domains D3 developmental/psychiatric cohorts. GBP identifies task-relevant circuits in the 3D-DiT latent generator and 3D-Swin encoder--decoder, freezes the selected units as structural memory and leaves the remaining units trainable for subsequent adaptation. Active and frozen units are displayed as node grids, with accompanying capacity bars indicating the fraction of preserved and remaining trainable capacity.
Ranked blueprint scores identify the most influential units contributing to each domain-specific signature, illustrating that freezing is structured rather than uniformly distributed across the model. The corresponding latent-space and anatomical projections show that blueprint patterns become increasingly domain-specific across the continual-learning trajectory, from more diffuse healthy-population signatures to neurodegenerative, developmental/psychiatric and tumour-associated patterns. Together, these visualisations show that GBP provides both a consolidation mechanism for reducing forgetting and an interpretable structural readout of how Alcmaeon reallocates model capacity across heterogeneous neuroimaging domains.
}

  \label{fig:blueprintc}
\end{figure}

\clearpage

\clearpage
\setcounter{section}{0}
\setcounter{subsection}{0}
\setcounter{subsubsection}{0}
\setcounter{figure}{0}
\setcounter{table}{0}
\setcounter{equation}{0}
\renewcommand{\thefigure}{S\arabic{figure}}
\renewcommand{\thetable}{S\arabic{table}}
\renewcommand{\theequation}{S\arabic{equation}}
\renewcommand{\thesection}{S\arabic{section}}
\renewcommand{\thesubsection}{S\arabic{section}.\arabic{subsection}}
\renewcommand{\thesubsubsection}{S\arabic{section}.\arabic{subsection}.\arabic{subsubsection}}
\renewcommand{\figurename}{Supplementary Fig.}
\renewcommand{\tablename}{Supplementary Table}
\setcounter{tocdepth}{2}
\renewcommand{\contentsname}{Supplementary Information Contents}

\begin{center}
{\LARGE Supplementary Information for ``A continually expandable foundation model for brain imaging''\par}
\vspace{1em}
{Michail Mamalakis, Carmen Jimenez-Mesa, Yonghao Li, Hao Chen, Chao Li, Antonios Mamalakis, John Suckling, Richard Bethlehem, Stephen J. Price, Richard J. Gilbertson and Pietro Lio\par}
\end{center}
\clearpage

\phantomsection
\section*{Content description and SI category map}
\addcontentsline{toc}{section}{Content description and SI category map}
This Supplementary Information PDF is organized as focused supporting material for the main Article. It combines the Nature ``flat'' Supplementary Information categories that are suitable for PDF presentation: Supplementary Methods, Supplementary Table(s), Supplementary Discussion, Supplementary Equation(s), and Supplementary Notes. Large numerical outputs, raw machine-readable tables, source data supporting plotted values, model checkpoints, and executable code are treated as Supplementary Data or repository material and should be supplied as separate editable files or repository records at submission.

\subsection*{Nature SI categories represented in this PDF}
\begin{description}
\item[\textbf{Supplementary Methods.}] Sections S1--S7 describe the model architectures, mathematical training framework, fine-tuning tasks, datasets, evaluation metrics, experimental settings, and hyperparameter-tuning procedures that support the Article.
\item[\textbf{Supplementary Equation(s).}] Equations embedded in Sections S2--S4 provide the formal notation for the Swin-Wrap autoencoder, 3D-DiT/3D-Swin-DiT diffusion transformer, diffusion objective, continual-learning penalties, Graph-Blueprint Pruning, and fine-tuning heads.
\item[\textbf{Supplementary Table(s).}] The tables retained in Sections S8--S10 summarize cohort characteristics, hyperparameter searches, ablation results, continual-learning outcomes, and task-specific performance where compact PDF presentation is needed for interpretation. Larger tabular source values should accompany the manuscript as editable Source Data or Supplementary Data files.
\item[\textbf{Supplementary Discussion.}] Interpretive discussion is limited to result-focused text in Sections S8--S10, including ablation interpretation, catastrophic-forgetting analysis, clinical-task comparisons, and cross-task conclusions.
\item[\textbf{Supplementary Notes.}] The data-availability, ethics and governance, and large-language-model disclosure notes clarify study governance, data access, and manuscript-preparation procedures.
\item[\textbf{Supplementary Data.}] Machine-readable source data, extended numerical outputs, trained checkpoints, and code are not best presented as fixed PDF pages and should be supplied separately through the Source Data files and permanent/versioned repositories described in the availability statements.
\end{description}

\subsection*{Section and subsection descriptions}
\begin{itemize}
\item \textbf{Supplementary Methods opening note.} Summarizes the foundation-model setup, MRI-derived inputs, and the relationship between pretraining, continual learning, and downstream clinical tasks.
\item \textbf{S1 Brief introduction of the proposed architectures.} Describes the three architecture families evaluated in the study and the main design choices for representation learning and generative modelling.
\item \textbf{S1.1 Swin-Wrap self-supervised reconstruction model.} Defines the reconstruction-based Swin-Wrap encoder--decoder and its role as a compact anatomical representation learner.
\item \textbf{S1.2 3D-DiT latent diffusion transformer.} Describes the 3D latent diffusion transformer used for generative modelling in volumetric neuroimaging data.
\item \textbf{S1.3 3D-Swin-DiT hybrid architecture.} Explains the hybrid design that combines Swin-style hierarchical representations with diffusion-transformer training.
\item \textbf{S1.4 Model scale, resolution, and batch-size variants.} Documents the tested architectural scales, voxel resolutions, input-channel settings, and batch-size trade-offs.
\item \textbf{S2 Mathematical background of training architectures and strategies.} Provides the formal notation and learning objectives for the pretraining models.
\item \textbf{S2.1 Problem setting and notation.} Establishes the symbols used for images, latent variables, time steps, conditioning variables, and model outputs.
\item \textbf{S2.2 Swin-Wrap latent autoencoder.} Details the shifted-window encoder, skip-coupled decoder, and reconstruction objective.
\item \textbf{S2.3 3D-DiT window-attention diffusion transformer.} Details voxelisation, tokenisation, windowed 3D attention, and adaptive-normalisation conditioning.
\item \textbf{S2.4 Diffusion process and learning objective.} Defines the forward and reverse diffusion dynamics, conditioning mechanism, and denoising training loss.
\item \textbf{S3 Mathematical background of catastrophic forgetting and continual training.} Defines how continual-learning behaviour is measured and how mitigation strategies are formulated.
\item \textbf{S3.1 Elastic Weight Consolidation.} Describes the regularized continual-learning baseline based on parameter importance.
\item \textbf{S3.2 Dataset-aggregated circuit and Graph-Blueprint Pruning.} Describes the proposed graph-guided pruning mechanism for preserving useful circuits across sequential datasets.
\item \textbf{S3.3 Training-formulation summary.} Consolidates the objectives and update rules used across continual-learning experiments.
\item \textbf{S4 Mathematical background of fine-tuning architectures and strategies.} Defines downstream heads, forward passes, mixed-precision execution, and task-specific objectives.
\item \textbf{S4.1 Proposed fine-tuned models.} Describes how Swin-Wrap and 3D-Swin-DiT representations are adapted for classification, generation, and survival tasks.
\item \textbf{S4.2 Alzheimer's disease progression classification.} Specifies the clinical classification task, output labels, and modelling formulation.
\item \textbf{S4.3 Synthetic data generation in brain-tumour cohorts.} Describes the generative fine-tuning task and evaluation context.
\item \textbf{S4.4 Brain-tumour survival estimation.} Specifies the survival-prediction task and associated modelling choices.
\item \textbf{S4.5 Use of large language models.} Documents the limited role of LLM tools in drafting and editing descriptive manuscript text.
\item \textbf{S5 Datasets for training and fine-tuning.} Summarizes the training and clinical fine-tuning cohorts used in the experiments.
\item \textbf{S5.1 Training setup.} Lists the pretraining datasets, modalities, preprocessing assumptions, and dataset sequencing.
\item \textbf{S5.2 Fine-tuning setup.} Lists the downstream cohorts and the task-specific data splits used for model evaluation.
\item \textbf{S6 Evaluation metrics.} Defines performance metrics for downstream tasks and forgetting metrics for sequential learning.
\item \textbf{S6.1 Fine-tuning task metrics.} Describes classification, generation, reconstruction, and survival-analysis metrics.
\item \textbf{S6.2 Continual-learning and catastrophic-forgetting protocol.} Defines the evaluation protocol for retention, forgetting, and cross-dataset transfer.
\item \textbf{S7 Experiments and hyperparameter tuning.} Describes the experimental procedures and search settings used to choose model configurations.
\item \textbf{S7.1 Brain-tumour surgery classification.} Summarizes the surgery-classification experiment and tuned settings.
\item \textbf{S7.2 Learning-rate scheduling.} Describes the tested learning-rate schedules and their use across tasks.
\item \textbf{S7.3 Brain-tumour survival estimation.} Summarizes the survival-estimation experiment and tuned settings.
\item \textbf{S7.4 Synthetic data generation in brain-tumour cohorts.} Summarizes the generation experiment and tuned settings.
\item \textbf{S8 Hyperparameter tuning ablation and state-of-the-art architecture comparison.} Presents compact ablation results and architecture comparisons supporting model selection.
\item \textbf{S8.1 Training.} Summarizes pretraining settings, hyperparameter combinations, and early architecture-comparison results.
\item \textbf{S9 Catastrophic forgetting of sequential learning in sub-groups D1--D4.} Presents sequential-training analyses across dataset groups and mitigation strategies.
\item \textbf{S9.1 Swin-Wrap forgetting in one-channel inputs.} Reports retention and forgetting behaviour for one-channel Swin-Wrap experiments.
\item \textbf{S9.2 Swin-Wrap and 3D-DiT forgetting in one-channel inputs.} Compares one-channel continual-learning behaviour between the two model families.
\item \textbf{S9.3 Model-scale comparison.} Summarizes how model size affects retention, transfer, and downstream behaviour.
\item \textbf{S9.4 Swin-Wrap and 3D-DiT forgetting in two-channel inputs.} Compares two-channel continual-learning behaviour across model families.
\item \textbf{S9.5 Previous-task forgetting and performance for one-channel inputs.} Reports forgetting scores and retained performance on earlier one-channel tasks.
\item \textbf{S9.6 Previous-task forgetting and performance for two-channel inputs.} Reports forgetting scores and retained performance on earlier two-channel tasks.
\item \textbf{S9.7 Previous-task forgetting and performance for hybrid-channel inputs.} Reports forgetting scores and retained performance for hybrid-channel configurations.
\item \textbf{S9.8 Best-performing max-two-channel configuration.} Compares channel selections to identify the strongest two-channel configuration.
\item \textbf{S9.9 Graph-Blueprint Pruning conclusions.} Summarizes evidence that graph-guided pruning can limit forgetting under specified experimental conditions.
\item \textbf{S9.10 Interpretability and memory-storage utilities of Graph-Blueprint Pruning.} Describes analyses linking retained model circuits to interpretable anatomical or task-relevant memory signatures.
\item \textbf{S10 Fine-tuning and clinical tasks.} Presents downstream task results and compact interpretation of clinical utility.
\item \textbf{S10.1 Alzheimer's disease progression.} Reports classification performance and interpretation for Alzheimer's disease progression labels.
\item \textbf{S10.2 Data generation in brain-tumour cohorts.} Reports synthetic-data generation outcomes and comparison metrics.
\item \textbf{S10.3 Reconstruction.} Presents reconstruction examples and quantitative reconstruction metrics.
\item \textbf{S10.4 Generation.} Presents generation examples and quantitative generation metrics.
\item \textbf{S10.5 Brain-tumour survival estimation.} Reports survival-estimation performance and clinical-task comparisons.
\item \textbf{S10.6 Survival analysis and UMAP embeddings.} Describes survival-stratification analyses and representation visualisations.
\item \textbf{S10.7 Brain-tumour survival after surgery.} Reports surgery-related survival analyses and model comparisons.
\item \textbf{S10.8 Cross-task analysis and conclusions.} Summarizes cross-task patterns while preserving the main Article's cautious conclusion that no method is uniformly superior.
\item \textbf{Data availability.} Explains access routes for training cohorts and fine-tuning cohorts and identifies restrictions for controlled human-participant data.
\item \textbf{Training-cohort availability.} Lists access conditions for datasets used in pretraining and continual-learning experiments.
\item \textbf{Fine-tuning-cohort availability.} Lists access conditions for downstream clinical cohorts and task-specific datasets.
\item \textbf{Ethics and data governance.} States ethics approval, governance, and data-use constraints for human-participant data.
\end{itemize}

\clearpage
\tableofcontents

\clearpage
\phantomsection
\section*{Supplementary Methods}
\addcontentsline{toc}{section}{Supplementary Methods}
\subsection*{Overview of Work}
\setcounter{figure}{0}
\renewcommand{\thefigure}{\arabic{figure}}
\renewcommand{\figurename}{Supplementary Fig}

All models operate on 3D neuroimaging volumes represented as
$x \in \mathbb{R}^{B \times C \times H \times W \times D}$,
where $B$ denotes the batch size, $C$ the number of channels, and $(H,W,D)$ the spatial dimensions. 
Unless otherwise stated, all experiments use paired anatomical inputs with $C = 1$ or $C = 2$ and volumes resampled to 
$64 \times 64 \times 64$. 
The methodological framework comprises three components:  
(i) a self-supervised reconstruction model for anatomical representation learning,  
(ii) a latent diffusion model for volumetric generative modelling, and  
(iii) supervised fine-tuning architectures for segmentation, classification, and clinical prediction.  
All models were implemented in PyTorch \cite{paszke2019pytorch} and MONAI (Medical Open Network for AI) \cite{cardoso2022monai}, with automatic mixed-precision training enabled throughout.

\section{Brief Introduction of the Different Proposed Architectures}

\subsection{Swin-Wrap: Self-supervised Reconstruction Model}
The Swin-Wrap model is based on the hierarchical Swin Transformer \cite{liu2021swin} encoder--decoder adapted from the 
SwinUNETR (Swin UNET Transformer) architecture \cite{tang2022self}. 
The encoder consists of four stages of shifted-window self-attention, generating multiscale feature maps 
$(h_0,h_1,h_2,h_3,h_4)$ that capture increasingly global anatomical structure across the 3D volume.  
Shallow convolutional blocks refine early-stage features, and a bottleneck module aggregates the deepest 
contextual representation.  
The decoder reconstructs the spatial hierarchy using learned upsampling and skip connections, following the design philosophy of U-Net \cite{ronneberger2015} and 3D U-Net \cite{cicek2016}.  
The final $1{\times}1{\times}1$ convolutional projection produces a two-channel volumetric reconstruction.  

This model provides high-resolution spatial fidelity and robust multiscale anatomical context, making it well-suited 
for self-supervised representation learning in neuroimaging.

\subsection{3D-DiT: Latent Diffusion Transformer}
To model the distribution of anatomical features, we employ a volumetric Diffusion Transformer (3D-DiT), inspired by 
Denoising Diffusion Probabilistic Models (DDPM) \cite{ho2020} and Diffusion Transformers (DiT) \cite{peebles2023}.  
Latent feature maps extracted from an intermediate Swin stage are reshaped into 
$q \in \mathbb{R}^{B \times C_{\mathrm{lat}} \times N}$, with $N = HWD$.

The forward diffusion process corrupts the latent using a variance schedule $\{\beta_t\}_{t=1}^T$, 
while the DiT denoiser---a volumetric Vision Transformer \cite{dosovitskiy2021} with time and optional 
class conditioning---learns the noise-prediction function needed for the reverse process.  
Iterative denoising reconstructs anatomically plausible latents from Gaussian noise, which are passed through the Swin 
decoder to yield 3D reconstructions.

This combination allows joint modelling of deterministic anatomical structure and stochastic variability in the latent space.

\subsection{3D-Swin-DiT: The Hybrid Architecture}
We develop \mbox{3D-Swin-DiT}, a hybrid generative architecture that
couples the Swin-Wrap volumetric autoencoder with a generative model
operating on its bottleneck representation. The bottleneck generator can
be instantiated either as a conventional three-dimensional variational
autoencoder \citep{kingma2014vae}, which provides a stochastic Gaussian
latent prior and an evidence lower-bound training criterion, or as the
3D~Diffusion Transformer (3D-DiT) introduced earlier in this work, which
replaces the variational prior with an iterative denoising process in the
same latent space. This factorisation realises three complementary
properties: (i)~anatomically informed latent representations inherited
from the hierarchical shifted-window encoder, whose deterministic latent
metric is stationary across continual stages; (ii)~tractable generative
modelling at high effective spatial resolution, since the diffusion or
variational dynamics act on the low-resolution latent rather than on the
full $64^3$ voxel grid; and (iii)~a modular substrate on which the
proposed continual-consolidation operator (Section~\ref{sec:gbp}) can be
applied to the encoder, the decoder, the bottleneck generator, or any
combination thereof.

\subsection{Different Scales, Resolutions, and Batch Sizes of the Proposed Foundation Model}
We follow the model scaling paradigm introduced in Diffusion Transformers (DiT)~\cite{dit} and subsequently extended to volumetric medical imaging in DiT-3D~\cite{3dit}. Consistent with the formulation proposed in~\cite{3dit}, our framework operates directly on voxelised three-dimensional MRI volumes rather than point-cloud or mesh-based representations. This design enables efficient spatial modelling while preserving compatibility with scalable transformer-based diffusion architectures across multiple computational regimes.

To systematically investigate scaling behaviour, the proposed framework supports multiple voxel resolutions (16, 32, and 64), patch tokenisation strategies ($p \in {2,4,8}$), and model capacities spanning the Small, Base, Large, and Extra-Large configurations originally defined in DiT~\cite{dit}. Such a hierarchical design allows the architecture to remain flexible across varying memory budgets, computational constraints, and representational capacities, while maintaining a unified transformer-based formulation.

In the present study, we primarily focus on the 3D-DiT-\texttt{*/8} family, where the symbol \texttt{*} denotes the Small, Base, Large, or Extra-Large backbone configurations, and the suffix \texttt{/8} indicates a patch size of $p = 8$ used for voxel tokenisation. In addition, controlled ablation and sensitivity analyses were conducted using smaller patch configurations ($p = 2$) to evaluate the influence of token granularity on representation learning, optimisation stability, and downstream generative performance.

The proposed architecture follows a scalable transformer design in which model variants differ in transformer depth $L$, hidden dimension $d$, and the number of attention heads. Representative configurations for the $p = 8$ setting include:
\begin{itemize}
\item \textbf{3D-DiT-S8}: $L=12$, $d=384$, $\sim$33M parameters;
\item \textbf{3D-DiT-B8}: $L=12$, $d=768$, $\sim$131M parameters;
\item \textbf{3D-DiT-L8}: $L=24$, $d=1152$, $\sim$580M parameters;
\item \textbf{3D-DiT-XL8}: $L=28$, $d=1152$, $\sim$676M parameters.
\end{itemize}

Analogously, the corresponding configurations using a finer patch granularity ($p = 2$) are:
\begin{itemize}
\item \textbf{3D-DiT-S2}: $L=12$, $d=384$, $\sim$33M parameters;
\item \textbf{3D-DiT-B2}: $L=12$, $d=768$, $\sim$131M parameters;
\item \textbf{3D-DiT-L2}: $L=24$, $d=1152$, $\sim$580M parameters;
\item \textbf{3D-DiT-XL2}: $L=28$, $d=1152$, $\sim$676M parameters.
\end{itemize}

This scalable formulation enables the model capacity and token resolution to be adapted according to dataset complexity and computational constraints, while preserving a unified architectural framework across all experimental settings.

Beyond transformer scaling, we further explored multiple architectural combinations integrating Swin-based autoencoding modules with diffusion-based generative backbones. In particular, we investigated both Swin-VAE and Swin-Diffusion hybrid formulations, including the proposed 3D-Swin-DiT framework, to assess the trade-offs between latent compression fidelity, reconstruction quality, and generative realism.

Extensive structural experiments were conducted across multiple spatial resolutions, including $64^3$ and $128^3$ volumetric inputs, to evaluate the relationship between anatomical detail preservation, computational complexity, and training scalability. Additional experiments examined single-channel and dual-channel MRI representations, alongside varying encoder depths and hierarchical latent levels within the Swin-based autoencoder framework, specifically considering level-2 and level-4 latent hierarchies.

Finally, we performed a broad exploration of optimisation dynamics under different mini-batch regimes, including batch sizes of 14, 128, 256, and 512. These experiments were designed to characterise the interaction between batch-scale training, diffusion stability, convergence behaviour, and large-scale distributed optimisation efficiency in high-dimensional volumetric generative modelling.

\section{Mathematical Background of Training Architectures and Strategies}

\subsection{Problem Setting and Notation}

Let $\x\in\mathcal{X}\subset\R^{C\times R^3}$ denote a brain MRI volume with
$C$ co-registered channels (modalities) on a cubic lattice
$\Lambda=\{0,\dots,R-1\}^3$, $R=64$, with intensities normalised to
$[0,1]$. We seek a generative model of the conditional law
$p(\x\mid \mathcal{C})$, where $\mathcal{C}$ is a conditioning context. Rather
than modelling $p$ directly in the $C R^3$-dimensional voxel space, we adopt a
two-stage \emph{latent} construction. A deterministic encoder
$\enc:\mathcal{X}\to\mathcal{Z}$ maps the volume to a compact latent
$\z=\enc(\x)\in\R^{C'\times r^3}$ with $r\ll R$, a diffusion process is learnt
on $\mathcal{Z}$, and a decoder $\dec:\mathcal{Z}\to\mathcal{X}$ returns the
synthesised volume. The composite model is
\begin{equation}
\hat\x \;=\; \dec\!\big(\,\Phi_{\bm\theta}(\z)\,\big),\qquad
\z=\enc(\x),
\end{equation}
where $\Phi_{\bm\theta}$ denotes the learnt latent generator (the reverse
diffusion map). The encoder and decoder are trained by a reconstruction
criterion; the generator is trained by a denoising score-matching criterion;
and the two objectives are optimised with respect to disjoint parameter
blocks, the latent being treated as a fixed target for the generator through a
stop-gradient operator $\sg(\cdot)$.

We study three nested objects. (i)~The \emph{baseline} diffusion transformer
$\Phi_{\bm\theta}$ acting on a voxelised representation, trained without any
continual-learning mechanism; this object exhibits catastrophic forgetting
under sequential training and defines the lower envelope of retained
performance. (ii)~The \emph{Swin autoencoder} $(\enc,\dec)$ furnishing the
latent metric. (iii)~The full \emph{latent diffusion} model
$\dec\circ\Phi_{\bm\theta}\circ\enc$, optionally equipped with a consolidation
operator $\mathsf{T}$ acting between successive tasks.

\subsection{The Swin-Wrap Latent Autoencoder}

\subsubsection{Hierarchical shifted-window encoder}
The encoder is a hierarchical Swin transformer producing a pyramid of
representations at geometrically decreasing resolutions. Writing
$\bm h^{(0)}=\x$ and indexing stages by $\ell\in\{0,\dots,L\}$ ($L=4$), each
stage applies a patch-merging contraction $M_\ell$ followed by a block of
windowed multi-head self-attention and a position-wise feed-forward map,
\begin{equation}
\bm h^{(\ell)} \;=\; \mathcal{S}_\ell\!\big(M_\ell\,\bm h^{(\ell-1)}\big),
\qquad
\bm h^{(\ell)}\in\R^{C_\ell\times (R/2^{\ell})^3},
\end{equation}
so that channel width grows as $C_\ell=2^{\ell}C_0$ while spatial extent
contracts by $2^\ell$. The latent supplied to the diffusion stage is the
bottleneck $\z=\bm h^{(L)}$. The encoder simultaneously retains the
intermediate \emph{skip} representations
$\mathcal{H}=\{\bm h^{(0)},\dots,\bm h^{(L)}\}$, which parameterise the decoder
below. Because no stochastic bottleneck is imposed, $\enc$ is a deterministic
map and the latent metric $d_{\mathcal{Z}}(\z_1,\z_2)=\norm{\z_1-\z_2}_2$ is
fixed throughout training; this stationarity of the latent geometry is
important when a single latent space must host a sequence of tasks.

\subsubsection{Skip-coupled decoder}
The decoder reconstructs through a symmetric expansion that fuses the upsampled
state with the corresponding encoder skip at each level. Denoting by
$U_\ell$ an upsampling (transpose-convolutional) operator and by $g_\ell$ a
residual fusion block,
\begin{equation}
\bm d^{(L)}=\z,\qquad
\bm d^{(\ell-1)} \;=\; g_\ell\!\big(\,U_\ell\,\bm d^{(\ell)}\,,\;\bm h^{(\ell-1)}\big),
\qquad
\hat\x \;=\; \Pi_{[0,1]}\!\big(W_{\mathrm{out}}\,\bm d^{(0)}\big),
\end{equation}
where $\Pi_{[0,1]}(\cdot)=\min(\max(\cdot,0),1)$ projects onto the admissible
intensity range. The autoencoder is trained to maximise structural fidelity
through a differentiable structural-similarity criterion. For two volumes with
local means $\mu$, variances $\sigma^2$ and covariance $\sigma_{xy}$ computed in
a sliding $8^3$ window $w$, with stabilisers $c_1,c_2$,
\begin{equation}
\mathrm{SSIM}_w(\hat\x,\x)=
\frac{(2\mu_{\hat x}\mu_x+c_1)(2\sigma_{\hat x x}+c_2)}
     {(\mu_{\hat x}^2+\mu_x^2+c_1)(\sigma_{\hat x}^2+\sigma_x^2+c_2)},
\qquad
\loss_{\mathrm{rec}}=1-\E_{w}\big[\mathrm{SSIM}_w(\hat\x,\x)\big].
\label{eq:rec}
\end{equation}

\subsection{The 3D-DiT Window-Attention Diffusion Transformer}

\subsubsection{Voxelisation and tokenisation}
The diffusion network treats its latent input as a set of $N=r^3$ feature
points on the lattice. Given features $\bm f\in\R^{C'\times N}$ at integer
coordinates $\bm c\in\R^{3\times N}$, a normalisation centres and rescales the
coordinates to the unit cube and maps them onto a resolution-$r$ grid,
\begin{equation}
\tilde{\bm c} \;=\; \Pi_{[0,r-1]}\!\left[\,
r\left(\frac{\bm c-\bar{\bm c}}{2\,\max_j\norm{\bm c_j-\bar{\bm c}}}+\tfrac12\right)\right],
\qquad \bar{\bm c}=\tfrac1N\textstyle\sum_j \bm c_j ,
\end{equation}
and an average-pooling voxelisation accumulates features into occupied cells,
\begin{equation}
V(\bm f,\bm c)_{:,v}
=\frac{1}{n_v}\sum_{j:\,\lfloor\tilde{\bm c}_j\rceil=v}\bm f_{:,j},
\qquad
n_v=\#\{j:\lfloor\tilde{\bm c}_j\rceil=v\},
\label{eq:vox}
\end{equation}
where $\lfloor\cdot\rceil$ is rounding to the nearest lattice node and the
adjoint of \eqref{eq:vox} (a count-normalised scatter) defines the backward
map for end-to-end differentiability. The voxel grid is partitioned into
non-overlapping cubic patches of edge $p$ and linearly embedded by a strided
volumetric convolution $W_{\mathrm{emb}}$, producing a token sequence
\begin{equation}
\bm u_0 \;=\; \mathrm{Patch}_p\big(V\big)\,W_{\mathrm{emb}} + \bm p_{\mathrm{pos}},
\qquad
\bm u_0\in\R^{T\times D},\quad T=(r/p)^3 ,
\end{equation}
with $D$ the model width and $\bm p_{\mathrm{pos}}$ a fixed three-dimensional
sinusoidal positional code: for an axis position $m$ and channel index $i$,
$\,p_{2i}=\sin(m\,\omega_i)$, $p_{2i+1}=\cos(m\,\omega_i)$,
$\omega_i=10^{-4\cdot 2i/(D/3)}$, concatenated over the three spatial axes.

\subsubsection{Windowed three-dimensional self-attention}
The defining computational primitive is self-attention restricted to local
cubic windows of the token lattice, which reduces the quadratic cost of global
attention to one that is linear in the token count. Reshaping the sequence to
its lattice form $\bm u\in\R^{(T_x\times T_y\times T_z)\times D}$ and tiling it
into windows of edge $w$, each window
$\Omega\in\R^{w^3\times D}$ is processed independently. With per-head query,
key and value projections $Q=\Omega W_Q$, $K=\Omega W_K$, $V=\Omega W_V$ and
head dimension $d_h=D/H$,
\begin{equation}
\mathrm{Attn}(\Omega)
=\mathrm{softmax}\!\left(\frac{QK^\top}{\sqrt{d_h}}+\mathcal{B}\right)V,
\label{eq:attn}
\end{equation}
where $\mathcal{B}$ is an optional decomposed relative-position bias that is
additively separable across the three axes,
$\mathcal{B}_{(a,b)}=R^x_{a_x-b_x}+R^y_{a_y-b_y}+R^z_{a_z-b_z}$, with learnable
per-axis tables $R^x,R^y,R^z$. The windowed outputs are reassembled into the
full lattice; cross-window information is propagated across depth by the
alternation of windowed blocks with non-windowed (global) blocks. The
asymptotic cost of \eqref{eq:attn} over the lattice is
$\mathcal{O}(T\,w^3 D)$ versus $\mathcal{O}(T^2 D)$ for global attention, the
saving being the factor $T/w^3$ that renders dense volumetric token grids
tractable.

\subsubsection{Adaptive-normalisation conditioning}
Diffusion time and conditioning are injected through adaptive layer
normalisation with zero-initialised residual gates. Let $\bm s$ be the
conditioning embedding (Section~\ref{sec:cond}). Each block produces six
modulation vectors $(\bm\gamma_1,\bm\beta_1,\bm\alpha_1,\bm\gamma_2,\bm\beta_2,
\bm\alpha_2)=\mathrm{MLP}(\bm s)$ and computes
\begin{align}
\bm u &\leftarrow \bm u + \bm\alpha_1\odot
\mathrm{Attn}\!\big((1+\bm\gamma_1)\odot\mathrm{LN}(\bm u)+\bm\beta_1\big),\\
\bm u &\leftarrow \bm u + \bm\alpha_2\odot
\mathrm{FFN}\!\big((1+\bm\gamma_2)\odot\mathrm{LN}(\bm u)+\bm\beta_2\big),
\end{align}
where the gates $\bm\alpha_1,\bm\alpha_2$ are initialised to zero, so that each
residual block is an identity map at initialisation and the network begins as
a well-conditioned near-isometry. After the final adaptive-norm layer the
tokens are de-patchified to the voxel grid and the field is returned to the
continuous sample locations by trilinear de-voxelisation: for a query point
with fractional offsets $(\delta_x,\delta_y,\delta_z)\in[0,1]^3$ within its
enclosing cell, the output is the convex combination over the eight corner
voxels $\{v_b\}_{b\in\{0,1\}^3}$,
\begin{equation}
\mathrm{DV}(\bm g)(\bm c)
=\sum_{b\in\{0,1\}^3}\Big[\textstyle\prod_{a\in\{x,y,z\}}
\big(b_a\delta_a+(1-b_a)(1-\delta_a)\big)\Big]\,\bm g_{:,v_b},
\label{eq:dv}
\end{equation}
whose corresponding adjoint scatters the upstream gradient onto the same eight
corners with identical weights.

\subsection{Diffusion Process and Learning Objective}
\label{sec:diff}

\subsubsection{Forward and reverse dynamics}
The generative model is a denoising diffusion probabilistic model defined on
the latent $\z_0=\sg\big(\enc(\x)\big)$. A variance-preserving forward Markov
chain progressively corrupts the latent over $T_{\!s}=500-1000$ steps with a noise
schedule $\{\beta_t\}_{t=1}^{T_{\!s}}$, $\alpha_t=1-\beta_t$,
$\bar\alpha_t=\prod_{s\le t}\alpha_s$:
\begin{equation}
q(\z_t\mid \z_{t-1})=\N\!\big(\sqrt{\alpha_t}\,\z_{t-1},\,\beta_t\bm I\big),
\qquad
q(\z_t\mid \z_0)=\N\!\big(\sqrt{\bar\alpha_t}\,\z_0,\,(1-\bar\alpha_t)\bm I\big).
\label{eq:fwd}
\end{equation}
The schedule is a warm-up--constant (``warm-$0.2$'') profile: $\beta_t$ rises
linearly from $\beta_{\min}=10^{-5}$ over the first fifth of the trajectory and
is thereafter held at $\beta_{\max}=5\times10^{-3}$. The comparatively small
terminal noise is matched to the low dynamic range of the encoder latent. The
exact Gaussian posterior of the forward chain,
\begin{equation}
q(\z_{t-1}\mid \z_t,\z_0)=\N\big(\tilde{\bm\mu}_t(\z_t,\z_0),\,\tilde\beta_t\bm I\big),\quad
\tilde{\bm\mu}_t=\frac{\sqrt{\bar\alpha_{t-1}}\beta_t}{1-\bar\alpha_t}\z_0
+\frac{\sqrt{\alpha_t}(1-\bar\alpha_{t-1})}{1-\bar\alpha_t}\z_t,\quad
\tilde\beta_t=\frac{1-\bar\alpha_{t-1}}{1-\bar\alpha_t}\beta_t,
\end{equation}
is the target that the reverse model approximates. The learnt reverse kernel
is parameterised in the noise-prediction form with a fixed (untrained) variance
$\sigma_t^2$,
\begin{equation}
p_{\bm\theta}(\z_{t-1}\mid \z_t)=\N\!\big(\bm\mu_{\bm\theta}(\z_t,t),\,\sigma_t^2\bm I\big),
\qquad
\bm\mu_{\bm\theta}(\z_t,t)=\frac{1}{\sqrt{\alpha_t}}
\Big(\z_t-\frac{\beta_t}{\sqrt{1-\bar\alpha_t}}\,\epsth(\z_t,t,\bm s)\Big).
\label{eq:rev}
\end{equation}

\subsubsection{Conditioning mechanism}
\label{sec:cond}
The model operates in a self-conditioned regime in which the clean latent
serves as the conditioning signal. The score network receives the
channel-wise concatenation of the noised and clean latents,
\begin{equation}
\epsth(\z_t,t,\bm s)\;\equiv\;\epsth\!\big([\,\z_t\,;\,\z_0\,],\,t\big),
\qquad
\bm s=\mathrm{MLP}_t(\bm\tau_t),
\end{equation}
where $\bm\tau_t$ is a sinusoidal embedding of the scalar timestep $t$,
analogous to the spatial code but over the temporal index. Consequently the
input channel dimension of the transformer is $2C'$ while its output dimension
is $C'$. This construction makes the denoiser an amortised conditional
estimator $\z_0\mapsto\,$``noise that would corrupt $\z_0$ into $\z_t$'',
i.e.\ a learnt posterior-mean residual.

\subsubsection{Training objective}
With $t\sim\U\{1,\dots,T_{\!s}\}$ and $\eps\sim\N(\bm0,\bm I)$, the forward
marginal \eqref{eq:fwd} gives
$\z_t=\sqrt{\bar\alpha_t}\,\z_0+\sqrt{1-\bar\alpha_t}\,\eps$, and the network is
trained by the simplified (unweighted) denoising score-matching loss
\begin{equation}
\loss_{\mathrm{diff}}(\bm\theta)
=\E_{\z_0,\,t,\,\eps}\;
\big\|\,\eps-\epsth\!\big([\z_t;\z_0],\,t\big)\,\big\|_2^2 ,
\label{eq:diff}
\end{equation}
which is a Monte-Carlo estimator of a re-weighted variational bound on
$-\log p_{\bm\theta}(\z_0)$. As a scale-invariant diagnostic of denoising
alignment we monitor the expected cosine similarity between the predicted and
true noise,
\begin{equation}
\rho(\bm\theta)=\E_{\z_0,t,\eps}\!\left[
\frac{\langle \eps,\,\epsth([\z_t;\z_0],t)\rangle}
{\norm{\eps}_2\,\norm{\epsth([\z_t;\z_0],t)}_2}\right].
\end{equation}
At generation time the reverse kernel \eqref{eq:rev} is iterated ancestrally
from $\z_{T_{\!s}}\sim\N(\bm0,\bm I)$ to $\z_0$, the recovered latent is
variance-normalised, and the decoder maps it to image space using the encoder
skips $\mathcal{H}$ of the conditioning volume.

\section{Mathematical Background of Catastrophic Forgetting and Continual Training Approaches in Foundation Models}
\subsection{Elastic Weight Consolidation}

Elastic Weight Consolidation (EWC) is a parameter-regularization approach for continual learning that mitigates catastrophic forgetting by penalizing changes to parameters that are important for previously learned tasks \cite{kirkpatrick2017}.

\subsubsection{Fisher Information Matrix.}
Let $\boldsymbol{\theta}$ denote the parameters of a model with likelihood $p(y \mid x, \boldsymbol{\theta})$, and let $\mathcal{D}$ denote the data distribution.
The Fisher Information Matrix (FIM) is defined as
\begin{equation}
\mathbf{F}(\boldsymbol{\theta})
=
\mathbb{E}_{(x,y)\sim\mathcal{D}}
\left[
\nabla_{\boldsymbol{\theta}} \log p(y \mid x, \boldsymbol{\theta})
\;
\nabla_{\boldsymbol{\theta}} \log p(y \mid x, \boldsymbol{\theta})^{\top}
\right].
\end{equation}
The Fisher matrix measures the sensitivity of the model likelihood to perturbations of each parameter.

\subsubsection{Diagonal Fisher approximation.}
In practice, EWC employs a diagonal approximation of the Fisher matrix, assigning an importance value to each parameter $\theta_i$:
\begin{equation}
F_i
=
\mathbb{E}_{(x,y)\sim\mathcal{D}}
\left[
\left(
\frac{\partial}{\partial \theta_i}
\log p(y \mid x, \boldsymbol{\theta})
\right)^2
\right].
\end{equation}

\subsubsection{Empirical Fisher estimation.}
Since the true data distribution is unknown, the Fisher information is estimated empirically using the training loss $\mathcal{L}(x,y;\boldsymbol{\theta})$, typically the negative log-likelihood or a reconstruction loss:
\begin{equation}
F_i
\approx
\frac{1}{|\mathcal{D}|}
\sum_{(x,y)\in\mathcal{D}}
\left(
\frac{\partial \mathcal{L}(x,y;\boldsymbol{\theta})}
{\partial \theta_i}
\right)^2.
\label{eq:fisher}
\end{equation}

\subsubsection{EWC regularization.}
When learning a new task $B$, EWC augments the task loss with a quadratic penalty that discourages deviation from parameters learned on a previous task $A$:
\begin{equation}
\mathcal{L}_{\mathrm{EWC}}
=
\frac{\lambda}{2}
\sum_i
F_i
\left(
\theta_i - \theta_i^{*}
\right)^2,
\label{eq:ewc}
\end{equation}
where $\theta_i^{*}$ denotes the value of parameter $\theta_i$ after training on task $A$, $F_i$ is the corresponding Fisher importance, and $\lambda$ controls the strength of the regularization.

\subsubsection{Total objective.}
The total loss optimized during training on task $B$ is therefore
\begin{equation}
\mathcal{L}_{\mathrm{total}}
=
\mathcal{L}_{\mathrm{Task\;B}}
+
\mathcal{L}_{\mathrm{EWC}}.
\end{equation}

\subsubsection{Interpretation.}
Parameters with large Fisher values are crucial for performance on previous tasks and are therefore constrained to change slowly, whereas parameters with small Fisher values remain more plastic. In this way, EWC transforms forgetting into a quadratic energy barrier in parameter space.

\subsubsection{Generative and reconstruction-based settings.}
In generative models where the likelihood is implicit, such as diffusion models for medical image reconstruction, the log-likelihood is replaced by a reconstruction loss $\mathcal{L}_{\mathrm{recon}}(x,y;\boldsymbol{\theta})$:
\begin{equation}
\log p(y \mid x, \boldsymbol{\theta})
\;\longrightarrow\;
-\,\mathcal{L}_{\mathrm{recon}}(x,y;\boldsymbol{\theta}).
\end{equation}
This yields the Fisher approximation
\begin{equation}
F_i
=
\mathbb{E}
\left[
\left(
\frac{\partial \mathcal{L}_{\mathrm{recon}}}
{\partial \theta_i}
\right)^2
\right],
\end{equation}
which is mathematically consistent with the original EWC formulation and commonly used in generative continual learning settings \cite{nguyen2018variational,farquhar2019towards}.

\subsection{Proposed Dataset-Aggregated Circuit: Graph-Blueprint Pruning}
\label{sec:gbp}

\begin{figure*}
\centering
\includegraphics[
  width=\textwidth,
  trim={1.cm 9cm 0.5cm 8cm},
  clip
]{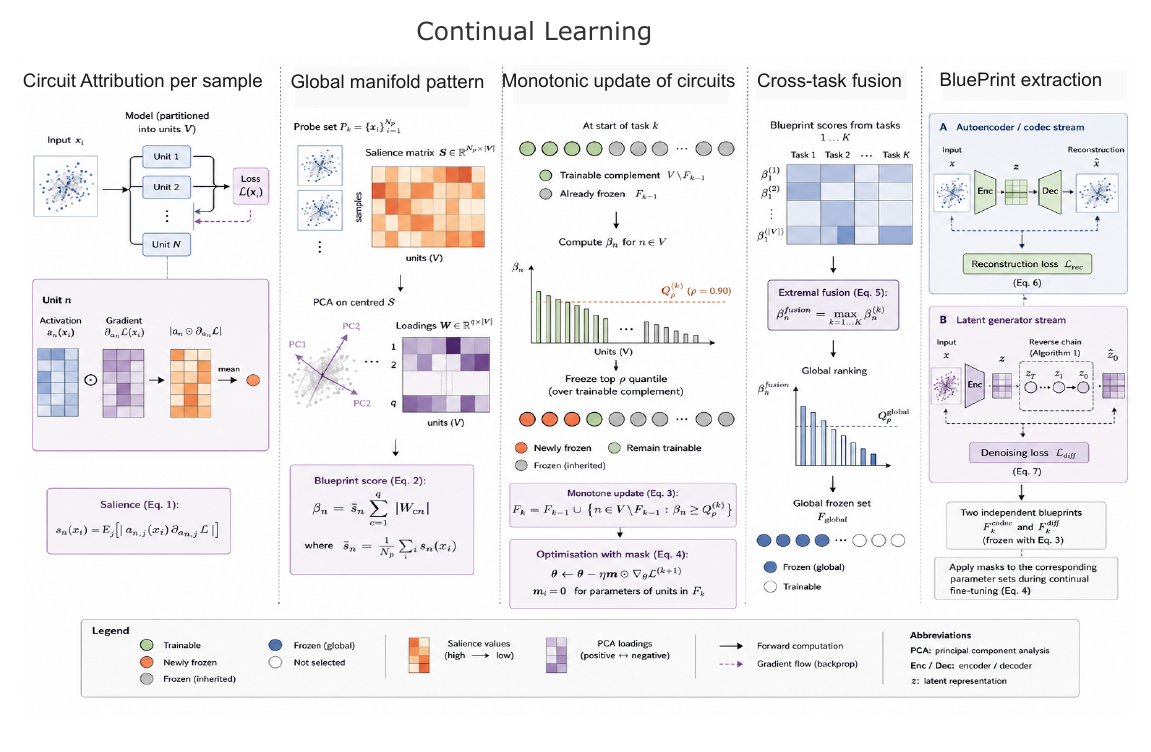}

\caption{
Graph-Blueprint Pruning (GBP) methodology. The proposed GBP operator identifies task-relevant computational circuits and preserves them during continual learning.
\textbf{(i)} For each probe sample, the model is partitioned into computational units, and unit-level mechanistic salience is computed from the activation--gradient product,
$s_n(\mathbf{x})=\mathbb{E}_j\!\left[|a_{n,j}(\mathbf{x})\partial_{a_{n,j}}\mathcal{L}|\right]$.
\textbf{(ii)} Salience vectors from the probe set are stacked into a task matrix and compressed by PCA. The blueprint score
$\beta_n=\bar{s}_n\sum_{c=1}^{q}|W_{cn}|$
combines mean attribution with participation in the dominant salience subspace.
\textbf{(iii)} A monotone hard projection freezes the upper-quantile units among the currently trainable complement, producing a cumulative frozen set
$\mathcal{F}_k=\mathcal{F}_{k-1}\cup\{n\in\mathcal{V}\setminus\mathcal{F}_{k-1}:\beta_n\ge Q^{(k)}_\rho\}$.
Frozen units are removed from subsequent optimiser updates through a binary mask.
\textbf{(iv)} Blueprints from previous tasks are fused by an elementwise extremal rule, ensuring that units salient for any previous task remain protected.
\textbf{(v)} Separate blueprint streams are maintained for the 3D-Swin and the latent generator, allowing structural preservation to be matched to the distinct reconstruction and denoising objectives.
}

\label{fig:gbp_methodology}
\end{figure*}

We replace soft parameter anchoring with a \emph{structural} operator that
identifies, and exactly preserves, the sub-circuits most responsible for a
task's behaviour, while leaving the remainder of the network free. The
construction proceeds in five elements: (i) per-sample mechanistic salience,
(ii) principal-subspace aggregation into a task blueprint, (iii) a monotone
hard projection of the parameter set, (iv) cross-task fusion of blueprints
through an extremal rule, and (v) two independent blueprint streams covering
the autoencoder and the latent generator.
 
\paragraph{(i) Per-sample salience of a computational unit.}
Partition the network into a set $\mathcal{V}$ of computational units (the
attention, feed-forward, normalisation and adaptive-modulation maps of each
transformer block) and let $\bm a_n(\x)$ denote the activation produced by
unit~$n\in\mathcal{V}$ on input~$\x$. To first order, the change in the
scalar objective $\loss$ induced by a perturbation of the activation is
$\langle\partial_{\bm a_n}\loss,\,\delta\bm a_n\rangle$, so the absolute
elementwise product $|\bm a_n\odot\partial_{\bm a_n}\loss|$ is a coordinate-wise
attribution map. We collapse it to a scalar by mean reduction,
\begin{equation}
s_n(\x)\;=\;\E_{j}\!\Big[\,\big|\,a_{n,j}(\x)\;\partial_{a_{n,j}}\loss\,\big|\,\Big],
\label{eq:gbp-salience}
\end{equation}
which equals to first order the average absolute contribution of unit $n$ to
the objective. The scalar functional $\loss$ used in \eqref{eq:gbp-salience}
is the task functional that the unit will subsequently be trained against
(reconstruction \eqref{eq:rec} for 3D-Swin units; the denoising loss
\eqref{eq:diff} for latent-generator units). For the latent-generator
blueprint, the gradient is taken \emph{through the full conditional
generative map}: the encoder produces $\z=\enc(\x)$, the reverse chain
\eqref{eq:rev} is unrolled to a sample $\hat\z_0$, and the loss is evaluated
against the supervised target. This causes \eqref{eq:gbp-salience} to attribute
importance with respect to the same computation that will be optimised, rather
than against a one-step proxy.
 
\paragraph{(ii) Principal-subspace blueprint of a task.}
Let $\mathcal{P}_k\subset\mathcal{D}_k$ be a probe subset of the current task,
of size $N_\mathrm{p}=\abs{\mathcal{P}_k}\ll\abs{\mathcal{D}_k}$. Stacking the
per-sample salience vectors over the probe set yields a node-score data matrix
\begin{equation}
\bm S\in\R^{N_\mathrm{p}\times\abs{\mathcal{V}}},\qquad
S_{ij}=s_{v_j}(\x_i),\quad \x_i\in\mathcal{P}_k.
\end{equation}
Let $\bar{\bm s}=\tfrac1{N_\mathrm{p}}\sum_i\bm S_{i,:}$ be the mean salience
profile and let the rows of $\bm W\in\R^{q\times\abs{\mathcal{V}}}$ be the top
$q=8$ right-singular vectors (principal loadings) of the centred matrix
$\bm S-\bm 1\bar{\bm s}^{\!\top}$. The \emph{blueprint score} of unit $n$
couples its mean salience with its aggregate participation in the principal
salience subspace,
\begin{equation}
\beta_n\;=\;\bar s_n \,\cdot\,\sum_{c=1}^{q}\big|W_{cn}\big|.
\label{eq:gbp-blueprint}
\end{equation}
The first factor measures the average attribution to unit $n$ across the
probe; the second up-weights units that participate \emph{coherently} across
the task distribution—those whose salience covaries with the dominant modes of
variation, rather than spiking on isolated samples. Equation
\eqref{eq:gbp-blueprint} therefore prefers stable, distributed circuitry over
sample-specific outliers, which is the property one wishes to protect against
forgetting. The use of a probe subset rather than the full task makes the
salience estimator a finite-sample empirical mean, with a corresponding
$\mathcal{O}(N_\mathrm{p}^{-1/2})$ standard error, and bounds the cost of
\eqref{eq:gbp-salience} to one forward and one backward pass per probe sample.
 
\paragraph{(iii) Monotone hard projection with quantile selection on the
trainable complement.}
Let $\mathcal{F}_{k-1}\subseteq\mathcal{V}$ be the cumulative frozen set
inherited from preceding stages, with $\mathcal{F}_0=\varnothing$.
\emph{Importantly, the quantile that drives selection is computed over the
trainable complement, not over all of $\mathcal{V}$:} otherwise units already
frozen at earlier stages would dominate the upper tail and block further
consolidation. Writing
\begin{equation}
Q_\rho^{(k)}\;=\;\rho\text{-quantile of }
\big\{\beta_n:n\in\mathcal{V}\setminus\mathcal{F}_{k-1}\big\},
\qquad \rho=0.90,
\end{equation}
the update of the frozen set is the monotone union
\begin{equation}
\mathcal{F}_k\;=\;\mathcal{F}_{k-1}\;\cup\;
\big\{n\in\mathcal{V}\setminus\mathcal{F}_{k-1}:\beta_n\ge Q_\rho^{(k)}\big\},
\qquad
\mathcal{F}_1\subseteq\mathcal{F}_2\subseteq\cdots\subseteq\mathcal{F}_K.
\label{eq:gbp-freeze}
\end{equation}
Freezing is realised by a binary parameter mask
$\bm m\in\{0,1\}^{\dim\bm\theta}$, with $m_i=0$ for every parameter
$\theta_i$ belonging to a unit in $\mathcal{F}_k$, applied multiplicatively in
the optimiser update:
\begin{equation}
\bm\theta\;\leftarrow\;\bm\theta\;-\;\eta\,\bm m\odot\nabla_{\bm\theta}\loss^{(k+1)}.
\label{eq:gbp-mask-step}
\end{equation}
Geometrically, \eqref{eq:gbp-mask-step} restricts the feasible set of the
stage-$k{+}1$ optimisation from $\R^{\dim\bm\theta}$ to the affine subspace
\begin{equation}
\Theta_k\;=\;\big\{\bm\theta:\bm m\odot(\bm\theta-\bm\theta^{(k)})=\bm 0\big\},
\end{equation}
i.e.\ the protected sub-circuits are held \emph{exactly} fixed rather than
penalised. The monotone property
$\mathcal{F}_1\subseteq\dots\subseteq\mathcal{F}_K$ implies a corresponding
chain of nested feasible sets $\Theta_1\supseteq\dots\supseteq\Theta_K$: each
stage operates in a residual subspace strictly contained in the previous one,
trading capacity for guaranteed retention.
 
\paragraph{(iv) Cross-task fusion of blueprints.}
Per-task blueprints are not used in isolation. At the start of stage $k$ the
current task blueprint $\beta^{(k)}$ is fused with the persistent blueprints of
all earlier stages by the element-wise extremal rule
\begin{equation}
\beta^{\mathrm{joint}}_n
\;=\;\max\!\big(\beta_n^{(k)},\,\beta_n^{(k-1)},\,\dots,\,\beta_n^{(1)}\big),
\label{eq:gbp-merge}
\end{equation}
and the quantile \eqref{eq:gbp-freeze} is applied to $\beta^{\mathrm{joint}}$
restricted to $\mathcal{V}\setminus\mathcal{F}_{k-1}$. Two design choices in
\eqref{eq:gbp-merge}--\eqref{eq:gbp-freeze} act in concert. The
\emph{exclude-frozen} rule removes already-protected units from the candidate
pool, so each stage contributes new structural protection; the \emph{max}
fusion ensures that a unit which was highly salient in any past task retains a
chance of being selected at later stages, even if its current-task salience
has decayed. Together they implement the principle ``protect what has been or
is now most coherently important, without re-deciding what is already
protected.''
 
\paragraph{(v) Parallel blueprints over 3D-Swin and 3D-DiT.}
The model contains two trainable subsystems with distinct functional roles:
the deterministic autoencoder $(\enc,\dec)$, which fixes the latent geometry,
and the conditional latent denoiser $\epsth$, which carries the generative
content. We maintain two independent blueprint streams, $\beta^{\mathrm{AE},(k)}$
and $\beta^{\mathrm{LG},(k)}$, with their own probe passes, fusion histories
and frozen sets $\mathcal{F}^{\mathrm{AE}}_k,\mathcal{F}^{\mathrm{LG}}_k$. The
salience functional differs between the two: for the autoencoder the loss in
\eqref{eq:gbp-salience} is the structural-similarity reconstruction
\eqref{eq:rec}; for the latent generator it is the denoising loss
\eqref{eq:diff} evaluated through the unrolled reverse chain. This
separation matches the two-objective training protocol and prevents the
attribution of a unit's importance from being confounded across roles.
 
\paragraph{Interpretation.}
The two consolidation operators—Fisher elastic anchoring \eqref{eq:ewc} and
graph-blueprint pruning \eqref{eq:gbp-freeze}--\eqref{eq:gbp-mask-step}—differ
in both granularity and mechanism. Elastic anchoring acts per coordinate and
is soft: it adds curvature $\lambda F_i$ to the loss while leaving the feasible
set equal to $\R^{\dim\bm\theta}$, trading a controllable bias against
residual drift. Graph-blueprint pruning acts per computational unit and is
hard: it removes a salience-selected sub-circuit from the optimisation by
restricting the feasible set to $\Theta_k$, guaranteeing zero forgetting on
the protected pathways at the cost of a monotone reduction of available
capacity. In the limit $\lambda\to\infty$ on a coordinate subset, elastic
anchoring approaches hard freezing of that subset; the proposed method
differs in selecting that subset by a graph-structured, principal-subspace
salience criterion \eqref{eq:gbp-blueprint} rather than by the diagonal
Fisher \eqref{eq:fisher}, and in compounding the selection across tasks
through the extremal fusion \eqref{eq:gbp-merge}.
 
\paragraph{Hyperparameters.}
The blueprint construction is governed by four scalars, all of which are
held fixed across stages: the probe size $N_\mathrm{p}$ (the number of
probe batches drawn from $\mathcal{D}_k$ for the salience expectation in
\eqref{eq:gbp-salience}--\eqref{eq:gbp-blueprint}); the principal-subspace
rank $q=8$ used in \eqref{eq:gbp-blueprint}; the freezing quantile
$\rho=0.90$ in \eqref{eq:gbp-freeze}; and the fusion mode in
\eqref{eq:gbp-merge}, fixed to the elementwise maximum. The mean reduction
in \eqref{eq:gbp-salience} and the absolute-value summation of loadings in
\eqref{eq:gbp-blueprint} ensure that $\beta_n\ge 0$, so that the quantile in
\eqref{eq:gbp-freeze} is well-posed without further sign conventions.

\subsubsection{Optimisation and stage procedure}

The encoder--decoder and the diffusion transformer are optimised by AdamW with
base learning rate $\eta=5\times10^{-4}$ and decoupled weight decay
$10^{-5}$, under automatic mixed precision with dynamic loss scaling. Micro-batches
of size $b$ are accumulated over $G=8$ steps to realise an effective batch
$b_{\mathrm{eff}}=b\,G\,n_{\mathrm{dev}}$ across $n_{\mathrm{dev}}$ data-parallel
devices; the latent is detached and variance-normalised,
$\z\leftarrow\sg(\enc(\x))/\mathrm{std}(\sg(\enc(\x)))$, before entering the
diffusion loss \eqref{eq:diff}, which stabilises the gradient magnitude when the
encoder produces low-variance codes. The per-stage continual procedure is given
in Algorithm~\ref{alg:cl}; the elastic variant replaces lines
\ref{ln:bp1}--\ref{ln:bp2} by Fisher estimation \eqref{eq:fisher} and adds the
penalty \eqref{eq:ewc} to the stage loss.
 
\begin{algorithm}[h]
\caption{Continual consolidation at stage $k$ (graph-blueprint variant).}
\label{alg:cl}
\begin{algorithmic}[1]
\Require parameters $\bm\theta^{(k-1)}$, task law $\pi_k$, prior blueprints
$\{\beta^{(j)}\}_{j<k}$, frozen set $\mathcal{F}_{<k}$
\State initialise $\bm\theta\leftarrow\bm\theta^{(k-1)}$
\For{each probe sample $\x\sim\pi_k$}\Comment{salience estimation} \label{ln:bp1}
  \State compute unit saliences $\bm s(\x)$ by one forward and one backward pass \eqref{eq:gbp-salience}
\EndFor
\State assemble $\bm X$, extract principal loadings $\bm W$, form blueprint $\beta$ \eqref{eq:gbp-blueprint}
\State fuse with prior tasks: $\beta\leftarrow\max\big(\beta,\,\beta^{(1)},\dots,\beta^{(k-1)}\big)$ \eqref{eq:gbp-merge}
\State update frozen set $\mathcal{F}_k$ by upper-quantile selection \eqref{eq:gbp-freeze}; build mask $\bm m$ \label{ln:bp2}
\For{optimisation iteration}
  \For{micro-batch $(\x,\x^\star)\sim\pi_k$}
     \State $\z\leftarrow \sg(\enc(\x))$; \; $\z\leftarrow\z/\mathrm{std}(\z)$
     \State draw $t\sim\U\{1,\dots,T_{\!s}\}$, $\eps\sim\N(\bm0,\bm I)$; \;
            $\z_t\leftarrow\sqrt{\bar\alpha_t}\,\z+\sqrt{1-\bar\alpha_t}\,\eps$
     \State $\loss\leftarrow \tfrac1G\,\norm{\eps-\epsth([\z_t;\z],t)}_2^2$
     \State accumulate $\bm m\odot\nabla_{\bm\theta}\loss$; apply AdamW step every $G$ micro-batches \eqref{eq:gbp-mask-step}
  \EndFor
\EndFor
\State evaluate $\{a_{k,j}\}_{j\le k}$ (MSE, RMSE, MAE, SSIM, PSNR, Dice); persist $\bm\theta^{(k)},\,\beta^{(k)},\,\mathcal{F}_k$
\end{algorithmic}
\end{algorithm}
 
\paragraph{Evaluation functionals.}
Generative fidelity is assessed in image space after decoding sampled latents.
For a reconstruction $\hat\x$ and reference $\x^\star$ on $\Omega=C R^3$ voxels,
\begin{equation}
\mathrm{MSE}=\tfrac{1}{\abs{\Omega}}\norm{\hat\x-\x^\star}_2^2,\quad
\mathrm{PSNR}=10\log_{10}\!\frac{1}{\mathrm{MSE}},\quad
\mathrm{Dice}_c=\frac{2\langle\hat\x_c,\x^\star_c\rangle}
{\norm{\hat\x_c}_1+\norm{\x^\star_c}_1},
\end{equation}
together with MAE, RMSE and the structural similarity of \eqref{eq:rec};
the Dice functional is reported per channel. Because the autoencoder is held
fixed during diffusion training, these functionals isolate the fidelity of the
conditional latent generator.

\subsubsection{Summary of the formulation}

\subsection{Summary of the Training Formulation}

\begin{table}[h]
\centering
\small
\renewcommand{\arraystretch}{1.18}
\begin{tabular}{@{}p{3.4cm}p{4.6cm}p{6.0cm}@{}}
\toprule
\textbf{Element} & \textbf{Mathematical form} & \textbf{Principle} \\
\midrule
Latent code & $\z=\enc(\x)$, deterministic & stationary latent metric across tasks \\
Tokenisation & average voxelisation \eqref{eq:vox} $+$ strided patch embedding & dense lattice to token sequence \\
Attention & windowed softmax \eqref{eq:attn}, cost $\mathcal{O}(Tw^3D)$ & local mixing, linear in token count \\
Conditioning & $\epsth([\z_t;\z_0],t)$, adaLN-zero gates & identity-initialised self-conditioning \\
Forward law & VP chain \eqref{eq:fwd}, warm-$0.2$, $T_{\!s}{=}500$ & low-noise regime for latents \\
Objective & noise-matching \eqref{eq:diff} & bound on $-\log p_{\bm\theta}(\z_0)$ \\
Reference CL & Fisher elastic penalty \eqref{eq:ewc} & soft curvature anchoring \\
Proposed CL & salience graph $\to$ PCA blueprint \eqref{eq:gbp-blueprint} $\to$ hard mask \eqref{eq:gbp-freeze} & exact structural preservation of coherent circuits \\
\bottomrule
\end{tabular}
\caption{Compact summary of the model and its two continual-consolidation
operators.}
\label{tab:summary}
\end{table}

The non-consolidated diffusion transformer defines the baseline whose
sequential training yields $\mathrm{BWT}<0$. Elastic Fisher anchoring is the
established reference operator. The proposed graph-blueprint operator is
distinguished by (a)~unit-level rather than coordinate-level granularity,
(b)~a hard feasible-set restriction rather than a soft penalty, and (c)~a
selection criterion \eqref{eq:gbp-blueprint} that combines first-order salience
with principal-subspace coherence and is fused across tasks by the extremal
rule \eqref{eq:gbp-merge}.

\section{Mathematical Background of Fine-tuning Architectures and Strategies}

Following the self-supervised training of the Swin-based reconstruction model and the unsupervised latent modelling achieved through the 3D diffusion backbone, we develop a set of fine-tuning frameworks designed to adapt the learned representations to downstream supervised tasks. These architectures leverage the strong anatomical priors and multiscale contextual features acquired during training, enabling efficient adaptation to segmentation, classification, and clinical decision-making.

We consider two complementary fine-tuning paradigms. First, \texttt{Swin-Wrap} extends the SwinUNETR backbone with task-specific heads and supports both voxelwise segmentation and global classification, allowing the network to operate either as a full encoder–decoder architecture or as a compact encoder-only predictor. Second, the \texttt{3D-Swin-DiT} fine-tuning model integrates frozen Swin-derived latent features with class-conditioned diffusion sampling and a discriminative classifier, forming a hybrid generative–discriminative approach optimised for clinical prediction.

Together, these architectures illustrate distinct but synergistic fine-tuning strategies: one rooted in direct feature decoding (\texttt{Swin-Wrap}) and one grounded in generative latent manipulation (\texttt{3D-Swin-DiT}). The following subsections detail the computational structure, optimisation procedures, and evaluation metrics associated with each fine-tuning model.

\subsection{Overview of The Proposed Models and How They Fine-Tuned}

\subsubsection{Swin-Wrap: fine-tuning classification model}

The class \texttt{Swin-Wrap} wraps the same \texttt{SwinUNETR} backbone with a configurable head.
In classification mode, the model bypasses the decoder and instead applies a linear projection to the bottleneck representation.
\\
\textbf{(i) Encoder features:}
\begin{equation}
  (\bm{h}_0, \bm{h}_1, \bm{h}_2, \bm{h}_3, \bm{h}_4)
  = \mathcal{T}(\bm{x}),
\end{equation}
\\
\textbf{(ii) Backbone features:}
\begin{equation}
  \bm{d}_4 = \mathcal{E}_{\text{bottleneck}}(\bm{h}_4).
\end{equation}
\\
\textbf{(iii) Flattening and linear projection:} Let the bottleneck output have shape
\[
  \bm{d}_4 \in \mathbb{R}^{B \times C_{\text{bottleneck}} \times H_b \times W_b \times D_b}.
\]
In the current implementation, the classification head receives a vector of dimension
$16 \cdot \texttt{feature\_size}$, implying an effective flattening of the form
\[
  \bm{v} \in \mathbb{R}^{B \times (16F)},
\]
where $F = \texttt{feature\_size}$ (here, $F = 48$). Formally,
\begin{equation}
  \bm{v} = \mathrm{flatten}(\bm{d}_4)
  \in \mathbb{R}^{B \times (16F)}.
\end{equation}

The classification head is a linear map
\[
  \mathcal{L} : \mathbb{R}^{16F} \rightarrow \mathbb{R}^{C_{\text{cls}}},
\]
parameterised by weights
$W \in \mathbb{R}^{C_{\text{cls}} \times (16F)}$
and bias
$\bm{b} \in \mathbb{R}^{C_{\text{cls}}}$,
where $C_{\text{cls}} = \texttt{class\_dim}$ (default: $1024$).
For each sample, the logits are:
\begin{equation}
  \bm{o} = \mathcal{L}(\bm{v}) = W \bm{v} + \bm{b},
  \qquad \bm{o} \in \mathbb{R}^{C_{\text{cls}}}.
\end{equation}

For a batch of size $B$, this becomes:
\begin{equation}
  \bm{O} = \mathcal{L}(\bm{V})
  \in \mathbb{R}^{B \times C_{\text{cls}}},
\end{equation}
where $\bm{V}$ stacks the individual feature vectors.

Thus, the overall classification mapping is
\begin{equation}
  f_{\text{swinunet-pure-cls}} : \bm{x} \mapsto \bm{O}.
\end{equation}

\subsubsection{3D-Swin-DiT: fine-tuning model}

Following self-supervised training of the Swin-Wrap encoder and unsupervised latent modelling with the diffusion backbone, we construct a fine-tuning model for binary clinical prediction. The fine-tuning module combines frozen anatomical features, class-conditioned generative sampling, and a discriminative classifier. Below, we describe (i) the architecture and (ii) the supervised learning scheme.

Let the input be a 3D volume
\begin{equation}
\bm{x} \in \mathbb{R}^{B \times 2/1 \times 64 \times 64 \times 64},
\end{equation}
and let $y \in \{0,1\}^B$ denote the corresponding binary class labels.

\paragraph{Latent extraction via the frozen Swin backbone.}
A frozen Swin-Wrap encoder produces a latent feature representation:
\begin{equation}
\bm{f} = \mathcal{E}_{\mathrm{swin}}(\bm{x}), \qquad 
\bm{f} \in \mathbb{R}^{B \times 2/1 \times F},
\end{equation}
where $F$ is the flattened spatial dimension of the selected encoder level.  
The Swin parameters remain frozen, i.e.
\[
\nabla_{\theta_{\mathrm{swin}}} \bm{f} = 0.
\]

\paragraph{Class-conditional diffusion sampling.}
For each label $y$, the DiT model generates a class-conditioned latent sample:
\begin{equation}
\tilde{\bm{q}} = \mathcal{G}_{\mathrm{DiT}}(\bm{f}, y),
\end{equation}
corresponding to the reverse diffusion trajectory applied to a noise-initialised latent:
\[
\tilde{\bm{q}} = f_{\mathrm{reverse}}(\bm{q}_T, y), \qquad 
\bm{q}_T \sim \mathcal{N}(0,I).
\]
This produces a generative embedding tailored to the diagnostic class.

\paragraph{Flattening and feature reduction.}
The generated latent is flattened and pooled:
\begin{align}
\bm{u} &= \mathrm{flatten}(\tilde{\bm{q}})
\quad \in \mathbb{R}^{B \times U}, \\
\bm{v} &= \mathrm{MaxPool}_k(\bm{u})
\quad \in \mathbb{R}^{B \times V},
\end{align}
where max-pooling reduces dimensionality while preserving discriminative structure.

\paragraph{Classification head.}
A multi-layer perceptron maps pooled latent features to prediction logits:
\begin{equation}
\hat{\bm{y}}
=
\mathcal{C}_{\mathrm{MLP}}(\bm{v})
\in \mathbb{R}^{B \times 2},
\end{equation}
where
\[
\mathcal{C}_{\mathrm{MLP}}(\bm{v})
=
W_3 \sigma(W_2 \sigma(W_1 \bm{v} + \bm{b}_1) + \bm{b}_2) + \bm{b}_3.
\]
Only the classifier and DiT components are trainable:
\[
\nabla_{\theta_{\mathrm{MLP}}} \neq 0, \qquad
\nabla_{\theta_{\mathrm{DiT}}} \neq 0.
\]

\paragraph{Overall mapping.}
The full architecture thus implements
\begin{equation}
f_{\mathrm{FT}} : 
\bm{x}
\;\xrightarrow{\;\mathcal{E}_{\mathrm{swin}}\;}
\bm{f}
\;\xrightarrow{\;\mathcal{G}_{\mathrm{DiT}}\;}
\tilde{\bm{q}}
\;\xrightarrow{\;\text{Pool}\;}
\bm{v}
\;\xrightarrow{\;\mathcal{C}_{\mathrm{MLP}}\;}
\hat{\bm{y}}.
\end{equation}

\subsubsection{Forward pass and automatic mixed precision}

In each training iteration, a mini-batch is sampled from the training loader and forwarded through the network
\begin{equation}
    \bm{z} = f_{\theta}(\bm{x}),
\end{equation}
where $f_{\theta}$ denotes the neural network parameterized by weights $\theta$.
We employ \emph{automatic mixed precision} (AMP) during training using 
\texttt{torch.cuda.amp.autocast()} to reduce GPU memory usage 
and accelerate training while maintaining numerical stability.
A gradient scaling mechanism (\texttt{GradScaler}) is applied before each backward pass, following NVIDIA's recommendations.

\subsubsection{Post-processing}

The raw network predictions are mapped to probability-like outputs through a pointwise sigmoid activation:
\begin{equation}
    \hat{\bm{y}} = \sigma(\bm{z}),
\end{equation}
where $\sigma$ denotes the logistic sigmoid applied channel-wise.
A normalization transform over the prediction domain $\hat{\bm{y}}$ is additionally used in accordance with MONAI's recommended practices for volumetric segmentation pipelines \cite{cardoso2022monai}.

\subsection{Clinical Task Adult's Alzheimer Progression Classification}
\subsubsection{Baseline models}
Two baseline approaches were considered for comparison with 3D-Swin-DiT: a classical machine learning pipeline based on dimensionality reduction and a deep learning model.

The first approach combines Principal Component Analysis (PCA) with a Support Vector Machine (SVM) classifier. PCA was applied to the input data retaining 90\% of the explained variance, in order to reduce dimensionality while preserving most of the relevant information. The resulting components were then used to train an SVM with a radial basis function (RBF) kernel. The model was trained using class-balanced weights to account for class imbalance, and probability estimates were enabled to allow probabilistic outputs. 

The second baseline is a Convolutional Neural Network (CNN) designed to learn hierarchical representations directly from the input images and was previously applied in \cite{arco2025explainable}. The architecture consists of successive convolutional layers with non-linear activation functions, interleaved with pooling operations to progressively reduce spatial dimensionality. These layers are followed by fully connected layers that perform the final classification. The CNN was trained using an input image size of $64 \times 64 \times 64$, with a training and test batch size of 4. The Adam optimizer was employed with a learning rate of $1 \times 10^{-4}$. The model was optimised using the CrossEntropyLoss function, suitable for multi-class classification problems. Training was performed for a maximum of 100 epochs but an early stopping criterion was applied, such that training was terminated if the average loss fell below 0.005, preventing overfitting and reducing unnecessary computation time.

\subsubsection{Swin-Wrap and 3D-Swin-DiT models}

We used and evaluated the Swin-Wrap autoencoder and 3D-Swin-DiT foundation models across two-channel and one-channel architectures at Levels 2 and 4 depth.
Features extracted from the Swin-UNet were first normalised at the sample level by dividing each embedding by its standard deviation. The resulting representations were then standardised to zero mean and unit variance using statistics computed from the training set. These normalised feature representations were then used as input to two classification models: SVM with a RBF kernel, and a three-layer Multi-Layer Perceptron (MLP). This MLP architecture comprised three hidden layers with 1024, 512, and 256 neurons, respectively, each followed by Batch Normalisation and ReLU activation functions to improve training stability and introduce non-linearity. Dropout regularisation with a rate of 0.3 was applied after each hidden layer to reduce overfitting. The final layer consisted of a fully connected output layer with a number of neurons equal to the number of classes. The hyperparameters used in this framework include the AdamW optimiser with a learning rate of $1 \times 10^{-3}$, applied only to trainable parameters. The input image size was set to $64 \times 64 \times 64$. The model was optimised using the CrossEntropyLoss function. A CosineAnnealing learning rate scheduler was employed, with $T_{max}$ set to the maximum number of training epochs, in order to gradually adjust the learning rate during training and improve convergence.

The model was assessed under multiple validation strategies to ensure a comprehensive evaluation. First, a 5-fold cross-validation scheme was applied to provide robust performance estimates across different data splits. In addition, a hold-out validation approach was used, consisting of a 70\% training and 30\% testing split. Both approaches used 20 epochs when applied with MLP. To further analyse the model’s behaviour under limited data conditions, several few-shot learning scenarios were explored using MLP as classifier. Specifically, zero-shot, one-shot, three-shot, and five-shot settings were evaluated. For these experiments, six different random seeds were used, and the reported results correspond to the average performance across these runs, ensuring stability and reproducibility.

\subsection{Clinical Task Synthetic Data Generation in Brain Tumor Cohorts}
\subsubsection{Reconstruction}
We benchmark four autoencoders that span the dominant design choices for latent representation of medical images: VQVAE~\cite{van2017neural},  AEKL~\cite{rombach2022high}, MAISI~\cite{guo2025maisi}, and our proposed OurAE. All baselines are evaluated using their publicly released code and checkpoints pre-trained on brain MRI.

{VQVAE} (Vector-Quantised Variational Autoencoder)~\cite{van2017neural} replaces the continuous latent with a discrete codebook, providing a fundamentally different latent geometry from AEKL and MAISI, and serves as the canonical discrete-latent baseline. {AEKL} (Autoencoder with Kullback--Leibler regularisation)~\cite{rombach2022high} is the autoencoder with a KL-regularised continuous latent space, introduced as the first stage of Latent Diffusion Models~\cite{rombach2022high}. It produces a smooth, low-dimensional latent and has become the de facto encoder--decoder backbone for latent generative modelling, included here as the canonical continuous-latent baseline. {MAISI} (Medical AI for Synthetic Imaging)~\cite{guo2025maisi} is a 3D foundation VAE for medical image compression~\cite{guo2025maisi}, pre-trained on tens of thousands of 3D CT and MRI volumes with support for variable volume size and voxel spacing; it represents the recent large-scale foundation-style medical autoencoders and allows us to assess whether such training translates into stronger reconstruction on the tested cohorts.  
\subsubsection{Generation}
We adopt three generative models that span the dominant paradigms for latent-space generation: {LDM} (Latent Diffusion Model)~\cite{rombach2022high}, {RF} (Rectified Flow)~\cite{liu2022flow}, and {MOTFM} (Medical Optimal Transport Flow Matching)~\cite{yazdani2025flow}. All three are well-suited to operate directly on the autoencoder's latent space, which makes them natural baselines for our setting. {LDM}~\cite{rombach2022high} is the canonical latent-space diffusion model and represents the score-based diffusion paradigm, providing a strong reference against which faster alternatives can be compared. {RF}~\cite{liu2022flow} represents the flow-matching paradigm, learning a straightened transport between source and target distributions and enabling high-quality generation in far fewer steps than diffusion. {MOTFM}~\cite{yazdani2025flow} is a recent optimal-transport flow-matching model designed specifically for medical image synthesis, included to assess whether a domain-specialised flow model offers advantages over the general-purpose formulations above. Together, these three baselines cover diffusion versus flow matching, and general-purpose versus medically specialised, allowing us to isolate the contribution of each design choice.

All models are based on publicly available implementations for medical 3D volumes. For RF, we use the implementation from FlowMI~\cite{chen2026contrastxmultimodalcontrastimage}. For all models, we follow the recommended settings from each repository and train for 200 epochs with a learning rate of $1\times10^{-4}$. We use a batch size of 16 for all encoders, except MAISI, where the batch size is set to 4 due to its larger latent shape.
We consider three cross-modal translation tasks: T1$\to$T2, T2$\to$FLAIR, and T1$\to$T1ce, where the arrow $\to$ denotes the translation direction from source to target modality. For each task, all datasets containing the required modality pair are pooled for joint training, and performance is reported per dataset to expose cohort-specific behaviour. For instance, in the T1$\to$T2 task, all available T1--T2 paired volumes across datasets are aggregated for training, while evaluation is conducted independently on each dataset's held-out test set. We adopt a 70\%/10\%/20\% train/validation/test split for each dataset.

\subsection{Clinical Task Adult's Brain Tumor Survival Estimation}
\textbf{Comparator families.} We evaluated 25 encoder configurations in total: the MONAI 3D ResNet family (ResNet-10, -18, -34, -50, -101, -152, and -200 with MedicalNet pre-training~\cite{chen2019med3d}), the DUNE family (U-AE, U-NET, and U-VAE~\cite{barba2025dune}), HLIP ViT~\cite{zhao2025hlip}, and multiple 3D-Swin-DiT checkpoints (single-modality and T1C+FLAIR concatenation, with baseline, EWC, EWC v2, and GBP joint continual-learning weights). Each family used its own preprocessing pipeline (Table~\ref{tab:preprocessing_by_model} and \S\ref{sec:feature_extraction}); downstream survival modelling was identical across families. The study flow is summarised in Figure~\ref{fig:workflow_methods}.

\begin{figure}[htbp]
\centering
\fbox{\begin{minipage}{0.92\linewidth}
\small
Single validation cohort ($n=1{,}551$): OS labels plus T1/T1C/T2/FLAIR (skull-stripped, MNI152).\\
Frozen feature extraction per encoder family (native preprocessing), same patients throughout.\\
Shared downstream survival tasks (Tasks 1--4) with five-fold stratified cross-validation.\\
Primary comparison: 3D-Swin-DiT vs.\ ResNet, DUNE, and HLIP families.
\end{minipage}}
\caption{Overview of the survival validation study design.}
\label{fig:workflow_methods}
\end{figure}

\subsubsection{MONAI 3D-ResNet family}
\label{sec:feat_resnet}

We evaluated seven depths (ResNet-10 through ResNet-200) implemented in MONAI with MedicalNet pre-training~\cite{chen2019med3d,cardoso2022monai,he2016resnet}. For each patient in the validation cohort, T1C and FLAIR were loaded as separate 3D volumes. When a brain mask was available, the volume was cropped to the mask bounding box with a five-voxel margin; otherwise the full field of view was used. Voxels inside the mask were z-scored (background set to zero), and the volume was resampled to $64 \times 64 \times 64$ by trilinear interpolation. Each modality was passed through the 3D ResNet with the classification head removed, yielding a 512-dimensional embedding. T1C and FLAIR embeddings were concatenated to form a \textbf{1,024-dimensional} patient feature vector.

\subsubsection{DUNE family}
\label{sec:feat_dune}

We used the three published DUNE architectures: U-AE, U-NET, and U-VAE~\cite{barba2025dune}. For each patient in the validation cohort, T1C and FLAIR were loaded separately, z-scored over the full volume, and cropped to $182 \times 218 \times 160$ voxels to match the spatial field of view used in DUNE training. Tensors were formed as single-channel inputs with layout $(1,1,D,H,W)$. Each architecture produced one embedding per sequence; T1C and FLAIR embeddings were concatenated, giving a \textbf{6,144-dimensional} feature vector per patient for each DUNE variant.

\subsubsection{HLIP ViT}
\label{sec:feat_hlip}

HLIP is a hierarchical vision--language pre-trained model for multiscan brain MRI~\cite{zhao2025hlip,dosovitskiy2021}. For each patient in the validation cohort, T1, T1C, T2, and FLAIR were reoriented to slice-first order, scaled to $[0,1]$ using 1st--99th percentile clipping, padded to a square axial field of view, resized to $256 \times 256$, and centre-cropped to $224 \times 224$. The through-plane axis was resampled to 72 slices per sequence (the setting used with our released checkpoint; HLIP pre-training used 48 slices in the original publication~\cite{zhao2025hlip}). The four sequences were encoded jointly with ImageNet-style normalisation applied slice-wise. The publicly released HLIP ViT backbone produced a \textbf{512-dimensional} CLS-pooled image embedding per patient.

\subsubsection{3D-Swin-DiT}
\label{sec:feat_dit}

Features were derived from frozen checkpoints of our model 3D-Swin-DiT: a Swin Transformer encoder~\cite{liu2021swin} paired with a diffusion transformer~\cite{ho2020}. We tested continual-learning weights (baseline, EWC~\cite{kirkpatrick2017}, EWC v2, and GBP joint) in addition to the default pre-trained checkpoint. T1C and FLAIR were processed in separate runs. Each volume was percentile-normalised (1st--99th) to $[0,1]$, resized to $64^3$, and encoded at diffusion timestep $t=0$ and spatial level 4. For each run we saved (i) the flattened Swin latent and (ii) the flattened model prediction $\hat{x}_0$, each \textbf{6,144-dimensional}. Reported concatenated models joined T1C and FLAIR features at the patient level, yielding \textbf{12,288 dimensions}. Fourteen DiT feature sets were evaluated in total (Swin-only, diffusion-only, and T1C+FLAIR concatenations across continual-learning strategies).

\subsection{Use of large language models}
This Supplementary Information section provides the disclosure of large language model use for the main Article.
Large language model tools (Claude, Anthropic; ChatGPT, OpenAI) were used to draft initial descriptions of results based on figures and Excel files generated by the authors. All AI-drafted text was subsequently reviewed, verified against the underlying data, and edited by the authors. These tools were not used for data analysis, data generation, or interpretation of findings. The authors take full responsibility for the accuracy and integrity of the final text.

\section{Datasets for Training and Fine-Tuning}

\subsection{Training Setup}
We pre-train \textbf{3D-Swin-DiT} on a collection of 425{,}717 magnetic resonance imaging (MRI) scans aggregated from multiple large-scale neuroimaging cohorts. These include: 
the UK Biobank (UKB; \textgreater 42{,}000 participants; ages 45--83, mean $\sim 64.5$; structural and diffusion MRI \cite{ukbiobank, ukbiobank_paper}), 
the Parkinson’s Progression Markers Initiative (PPMI; $\sim$1{,}500 participants; mean age $60.7 \pm 9.2$ years; structural MRI acquisitions used here \cite{ppmi}), 
the Alzheimer's Disease Neuroimaging Initiative (ADNI; $\sim$1{,}700 participants; ages 55--90, mean $\sim 73$ \cite{adni}), the PREVENT-AD cohort ($\sim$387 participants; cognitively normal older adults at elevated familial risk for Alzheimer's disease; ages 55--84; multimodal structural MRI \cite{preventad}),
the Adolescent Brain Cognitive Development study (ABCD; $\sim$11{,}500 participants; ages 9--10; diffusion MRI \cite{abcd}), 
the Reproducible Brain Charts (RBC; 6{,}346 participants; ages 5--85, predominantly 5--22 \cite{shafiei2025rbc}), 
the Health and Aging Brain Study–Health Disparities extension (HABS--HD; $\sim$4{,}000 participants; ages 30+ \cite{habs_hd}).

Lastly, we used some brain tumor patients from the brain-tumour cohort comprised the University of Pennsylvania Glioblastoma cohort (UPenn-GBM; \cite{upenn_gbm}; $n=611$; T1w, T1c, T2w and FLAIR), the University of California San Francisco Preoperative Diffuse Glioma MRI dataset (UCSF-PDGM; \cite{ucsf_pdgm}; $n=494$; T1w, T1c, T2w and FLAIR), (EGD; \cite{egd}; $n=774$; T1w, T1c, T2w and FLAIR). Together, these cohorts provided 1879 patients with standard multiparametric glioma MRI sequences for evaluating reconstruction, cross-modal synthesis, survival modeling and tumour-domain continual adaptation.

Structural MRI datasets (T1-weighted, T2-weighted, FLAIR, and related contrasts) were preprocessed using standard neuroimaging pipelines. All images were spatially normalised to Montreal Neurological Institute (MNI) space using \texttt{SPM12} (\url{https://www.fil.ion.ucl.ac.uk/spm/}) and resampled to an isotropic resolution of 1$\times$1$\times$1~mm$^3$, resulting in final dimensions of 157$\times$189$\times$156~mm. Skull-stripping was performed using HD-BET \cite{isensee2019hdbet}.

\subsubsection{Dataset grouping and Training Strategy}

To investigate the domain specificity of large-scale MRI training and to evaluate the transferability of the resulting foundation model across heterogeneous neuroimaging applications, all datasets were organised into four major domains (D1--D4), each representing a distinct biological, demographic, or clinical context. The domains were subsequently used as sequential tasks in the continual learning experiments.

Two complementary training strategies were explored. In the \textbf{two-channel setting}, paired imaging modalities acquired from the same participant were provided as separate input channels, preserving the correspondence between modalities. In the \textbf{single-channel setting}, all images were treated independently and pooled into a unified training corpus irrespective of modality, allowing the model to learn modality-agnostic representations. This design enables the framework to operate both in settings where paired multimodal acquisitions are available and in scenarios where only individual imaging modalities are present.

\paragraph{D1a--c: UK Biobank}

The first domain comprised large-scale population imaging data from UK Biobank and was subdivided into three modality-specific collections. Each subset was used to train a dedicated foundation model, enabling assessment of modality-specific representation learning prior to continual adaptation.

\begin{itemize}
\item \textbf{D1a: Diffusion MRI.} This subset consisted of diffusion-derived scalar maps, including fractional anisotropy (FA) and mean diffusivity (MD). In the two-channel configuration, FA and MD were paired, yielding 57,694 aligned scans per channel. In the single-channel configuration, all scans were pooled, resulting in 115,388 images.

\item \textbf{D1b: Structural MRI.} This subset contained T1-weighted and fluid-attenuated inversion recovery (FLAIR) images. The two-channel configuration included 59,570 paired scans per modality, whereas the single-channel configuration consisted of 119,140 images.

\item \textbf{D1c: NODDI-derived Imaging.} This subset comprised neurite orientation dispersion and density imaging (NODDI) derivatives, specifically orientation dispersion index (ODI) and isotropic volume fraction (ISOVF) maps. A total of 57,666 paired scans per channel were available in the two-channel setting, corresponding to 115,332 images in the single-channel configuration.

\end{itemize}

\paragraph{D1: Healthy Population Cohorts}

To evaluate generalisation beyond the UK Biobank population, a second healthy-control domain was assembled from the Healthy Aging Brain Study--Health Disparities (HABS-HD). This cohort comprised 9,851 MRI scans, including susceptibility-weighted imaging (SWI; 5,143 scans) and magnitude images from the second echo (MAG-IMAGE2; 4,708 scans). All images were spatially aligned and represented cognitively normal individuals across the adult lifespan.

\paragraph{D2: Neurodegenerative Disease}

The neurodegenerative disease domain consisted of imaging data from individuals with Parkinson's disease and Alzheimer's disease obtained from the Parkinson's Progression Markers Initiative (PPMI), Alzheimer's Disease Neuroimaging Initiative (ADNI), and PREVENT-AD cohorts. In total, this domain contained 13,196 MRI examinations. PPMI contributed 6,598 scans comprising T1-weighted, FLAIR, and MPRAGE acquisitions. PREVENT-AD provided 3,165 T1-weighted and FLAIR scans, while ADNI contributed 3,433 structural MRI examinations including T1-weighted and MPRAGE sequences.

\paragraph{D3: Developmental and Psychiatric Cohorts}

The developmental and lifespan domain combined data from the Reproducible Brain Charts (RBC) initiative and the Adolescent Brain Cognitive Development (ABCD) study, comprising 44,120 scans in total. RBC contributed 5,330 primarily T1-weighted structural MRI scans spanning a broad age range. ABCD contributed 38,790 diffusion MRI-derived measurements, including quality assurance and diffusion-derived metrics such as isotropic diffusion (ISO), fractional anisotropy (FA), and radial diffusivity (RD), representing one of the largest developmental neuroimaging resources currently available.

\paragraph{D4: Brain Tumour Cohorts}
The brain-tumour cohort included UPenn-GBM ($n=611$), UCSF-PDGM ($n=494$) and EGD ($n=774$). In total, the assembled brain-tumour benchmark included 1879 patients and combined open-access, controlled-access and data-use-agreement cohorts.

Across all cohorts, the domain contained 8690 scans and included the standard clinical imaging modalities commonly used for tumour characterisation: T1-weighted, contrast-enhanced T1-weighted (T1c), T2-weighted, and FLAIR MRI. This domain was selected to represent a substantial distributional shift from healthy, developmental, psychiatry and neurodegenerative populations, thereby providing a stringent benchmark for evaluating continual adaptation under severe pathological variability.

This grouping provides a principled framework for disentangling domain-specific effects during training and enables systematic evaluation of cross-domain transfer in downstream tasks.

We preprocessed all images using a standardised pipeline applied to two and one input channels (channel~1 and channel~2). Volumes were safely loaded, converted to channel-first format, re-oriented to the canonical RAS coordinate system, and resampled to an isotropic resolution of 1~mm using bilinear interpolation to ensure geometric consistency across subjects. When multiple channels were present, ancillary channels were resampled to match the spatial grid of channel~1. 

Intensity values were normalised within a fixed range (from 1 to 99 \% of maximum and minimum value of intensity) and scaled to the [0,1] interval with clipping to reduce the influence of outliers. Foreground regions were automatically cropped to remove non-informative background, and volumes were resized to a fixed spatial resolution compatible with network input. Channel~1 and channel~2 were then concatenated to form a two-channel input representation, which was also duplicated to serve as the training label in the self-supervised setting.

To improve robustness and reduce overfitting, we applied stochastic spatial augmentations, including random flips along all three anatomical axes, random 90\textdegree{} rotations, and intensity perturbations. All data were finally converted to PyTorch tensors on the target device for efficient training and batched processing.

Each sub-dataset was randomly partitioned into 70\% for training and 30\% for validation.

\subsection{Fine-tuning Setup}

\subsubsection{Clinical task adult Alzheimer progression classification}

We used BrainLAT dataset \cite{brainlat}. It comprises neuroimaging scans from patients with a range of neurological conditions, collected across several Latin American countries, including Colombia (CO), Mexico (MX), Chile (CL), Peru (PE), and Argentina (AR). The available modalities include T1-weighted (n$=$629), T2-weighted (n$=$57), and FLAIR (n$=$100) scans. Due to the substantially larger number of T1-weighted scans, the present study focuses exclusively on this modality. 

The dataset includes subjects diagnosed with Alzheimer’s disease (AD), behavioural variant frontotemporal dementia (bvFTD), multiple sclerosis (MS), Parkinson’s disease (PD), as well as healthy controls (HCs). For the purposes of this work, we selected AD, FTD, and HC groups, defining two classification scenarios: (i) a binary classification task (AD vs. HC), and (ii) a three-class classification task (AD vs. FTD vs. HC).

Only scans with available diagnostic labels were included in the analysis. The final cohort consists of 527 subjects distributed as follows: HC (n$=$226), AD (n$=$178), FTD (n$=$123). Table~\ref{tab:demo} includes the demographic information of these subjects.

\begin{table}
\centering
\caption{Demographic characteristics of the BrainLat cohort used in this study.}
\label{tab:demo}

\small

\begin{tabular}{ccccc}
\toprule
Group & $n$ & Age (years) & Sex (M/F) & Years of Education \\
\midrule

HC
& 226
& 67.88 $\pm$ 8.86
& 63/163
& 14.05 $\pm$ 4.26 \\

AD
& 178
& 72.97 $\pm$ 8.24
& 94/84
& 12.11 $\pm$ 4.48 \\

FTD
& 123
& 64.71 $\pm$ 8.62
& 63/60
& 12.76 $\pm$ 5.22 \\

\bottomrule
\end{tabular}

\end{table}

All images were skull-stripped prior to analysis. Subsequently, spatial normalisation to the Montreal Neurological Institute (MNI) space was performed using Advanced Normalization Tools (ANTs).

\subsubsection{Clinical task synthetic data generation in different neuroscience diseases}

We evaluate on public brain MRI datasets covering adult and pediatric populations, multiple tumor types, post-operative imaging, and healthy controls. 
We employ three sub-challenges from the Brain Tumor Segmentation (BraTS) cluster, each providing skull-stripped, co-registered mpMRI volumes accompanied by expert annotations of tumor sub-regions.
{BraTS-SSA}~\cite{adewole2023brain} comprises 95 adult glioma cases acquired at Sub-Saharan African imaging centres on heterogeneous, predominantly lower-field scanners, introducing acquisition and demographic shift relative to conventional BraTS cohorts.
{BraTS-MET}~\cite{moawad2024brain} consists of 238 annotated pre-treatment brain metastasis cases aggregated from eight international sites, characterised by multiple small, focal lesions rather than a single dominant mass.
{BraTS-PED}~\cite{kazerooni2024brain} contains 228 pediatric high-grade glioma cases curated through the Children's Brain Tumor Network (CBTN), the CONNECT consortium, and adopts a sub-region taxonomy tailored to pediatric disease.

The Cancer Genome Atlas Lower-Grade Glioma collection (TCGA-LGG)~\cite{pedano2016tcgalgg} contains 108 pre-operative adult lower-grade glioma cases from five institutions, with the four standard mpMRI sequences. 
The Information eXtraction from Images (IXI) dataset~\cite{ixi} comprises nearly 600 scans of healthy adult subjects collected at three London hospitals on different scanners (Philips 3T, Philips 1.5T, and GE 1.5T). Each subject is imaged with T1-, T2-, and proton density (PD)-weighted sequences, magnetic resonance angiography (MRA), and 15-direction diffusion tensor imaging (DTI); T1ce and FLAIR are not acquired. IXI serves as our tumor-free reference cohort and contributes additional scanner and field-strength diversity.
The R\'io Hortega University Hospital Glioblastoma (RHUH-GBM) dataset~\cite{cepeda2023rhuh} contains 40 adult glioblastoma patients (WHO grade 4), totalling 600 MRI series across three time points: pre-operative, early post-operative (within 72\,h), and follow-up at recurrence. Each study includes the four standard mpMRI sequences plus diffusion-derived ADC maps, with expert-corrected segmentations, enabling evaluation on post-resection cavities and recurrent disease.

\subsubsection{Clinical task adult brain tumor survival estimation}

The Imaging data we used was from the brain-tumour cohort comprised The Cancer Genome Atlas Glioblastoma Multiforme collection (TCGA-GBM; \cite{tcga_gbm}; $n=135$; T1w, T1c, T2w and FLAIR), The Cancer Genome Atlas Low Grade Glioma collection (TCGA-LGG; \cite{pedano2016tcgalgg}; $n=107$; T1w, T1c, T2w and FLAIR), the Clinical Proteomic Tumor Analysis Consortium Glioblastoma Multiforme collection (CPTAC-GBM; \cite{cptac_gbm}; $n=29$; T1w, T1c, T2w and FLAIR), the Ivy Glioblastoma Atlas Project collection (IvyGAP-GBM; \cite{ivygap_gbm}; $n=34$; T1w, T1c, T2w and FLAIR), the Repository of Molecular Brain Neoplasia Data collection (REMBRANDT; \cite{rembrandt}; $n=63$; T1w, T1c, T2w and FLAIR), the University of Pennsylvania Glioblastoma cohort (UPenn-GBM; \cite{upenn_gbm}; $n=611$; T1w, T1c, T2w and FLAIR), the University of California San Francisco Preoperative Diffuse Glioma MRI dataset (UCSF-PDGM; \cite{ucsf_pdgm}; $n=494$; T1w, T1c, T2w and FLAIR), the R\'io Hortega University Hospital Glioblastoma dataset (RHUH-GBM; \cite{cepeda2023rhuh}; $n=40$; T1w, T1c, T2w and FLAIR), the Longitudinal Glioblastoma MRI with Expert RANO Evaluation dataset (LUMIERE; \cite{lumiere}; $n=74$; T1w, T1c, T2w and FLAIR), and the RSNA--ASNR--MICCAI Brain Tumor Segmentation Challenge 2021 dataset (BraTS 2021; \cite{brats2021}; $n=40$; T1w, T1c, T2w and FLAIR). Together, these cohorts provided 1627 patients with standard multiparametric glioma MRI sequences for survival modelling and tumour-domain continual adaptation.

Based on these, the final adult brain-tumour survival cohort comprised 1,551 patients from eight glioma MRI resources: the University of Pennsylvania Glioblastoma cohort (UPenn-GBM; \cite{upenn_gbm}; $n=585$, 585 events, 0 censored), the University of California San Francisco Preoperative Diffuse Glioma MRI dataset (UCSF-PDGM; \cite{ucsf_pdgm}; $n=493$, 247 events, 246 censored), The Cancer Genome Atlas low-grade glioma and glioblastoma collections (TCGA-LGG/GBM; \cite{pedano2016tcgalgg,tcga_gbm}; $n=241$, 153 events, 88 censored), the Longitudinal Glioblastoma MRI with Expert RANO Evaluation dataset (LUMIERE; \cite{lumiere}; $n=70$, 70 events, 0 censored), the Repository of Molecular Brain Neoplasia Data collection (REMBRANDT; \cite{rembrandt}; $n=59$, 59 events, 0 censored), the R\'io Hortega University Hospital Glioblastoma dataset (RHUH-GBM; \cite{cepeda2023rhuh}; $n=40$, 31 events, 9 censored), the Ivy Glioblastoma Atlas Project collection (IvyGAP-GBM; \cite{ivygap_gbm}; $n=34$, 25 events, 9 censored), and the Clinical Proteomic Tumor Analysis Consortium Glioblastoma Multiforme collection (CPTAC-GBM; \cite{cptac_gbm}; $n=29$, 22 events, 7 censored). Overall, the cohort included 1,192 observed survival events and 359 censored cases, with multiparametric glioma MRI sequences available across cohorts.

This was a retrospective multicentre study designed to compare OS prediction from frozen MRI features extracted by 3D-Swin-DiT and by three published encoder families under a common downstream survival protocol. Imaging was drawn from publicly available, harmonised open-source glioma cohorts. For each patient, volumes were skull-stripped and registered to MNI152 standard space before any encoder was run. Overall survival time and event status were taken from curated clinical records accompanying the released imaging. Imaging, features, and outcomes were linked using a unique patient identifier that combined data source and case ID.

\textbf{Validation cohort ($n=1{,}551$).} One patient list was defined for the entire benchmark. Every encoder family (3D-Swin-DiT, MONAI ResNet, DUNE, and HLIP) and every downstream task used this same cohort. Patients were included if they had (i) non-missing OS time and event status; (ii) skull-stripped T1, T1C, T2, and FLAIR volumes in MNI152 (the four sequences required by HLIP and available for all reported comparisons); and (iii) successfully extracted feature vectors for each reported model configuration, matched to OS labels. After this alignment, \textbf{$n=1{,}551$} patients remained. In-house data were not used. No model was evaluated on a different patient set. The only additional exclusions were task-specific: for fixed-horizon binary survival (Task~4), patients censored before the horizon were omitted for that horizon because the label is undefined (\S\ref{sec:task4_labels}).

\subsubsection{Shared upstream imaging harmonisation}
\label{sec:imaging}

All encoders used skull-stripped volumes already registered to MNI152 at the cohort level. We did not repeat skull stripping or registration in the survival analysis code. Beyond this shared starting point, each family required additional steps (resampling, intensity normalisation, cropping, channel layout) consistent with its training setup. Table~\ref{tab:preprocessing_by_model} lists these differences; \S\ref{sec:feature_extraction} provides step-by-step detail for each family.

\begin{table}[htbp]
\centering
\caption{Imaging inputs and preprocessing for each encoder family before frozen feature extraction. All inputs were skull-stripped and registered to MNI152.}
\label{tab:preprocessing_by_model}
\footnotesize
\setlength{\tabcolsep}{4pt}
\renewcommand{\arraystretch}{1.2}
\begin{tabular}{p{2.1cm}p{2.0cm}p{1.8cm}p{2.1cm}p{1.6cm}}
\toprule
\textbf{Encoder family} & \textbf{MRI sequences used} & \textbf{Resampling / crop} & \textbf{Intensity normalisation} & \textbf{Output dimension} \\
\midrule
MONAI 3D ResNet (7 depths) & T1C + FLAIR & $64 \times 64 \times 64$ & Z-score inside brain mask & 1,024 (512 per modality) \\
DUNE (U-AE, U-NET, U-VAE) & T1C + FLAIR (separate passes) & Crop $182 \times 218 \times 160$ & Global z-score per volume & 6,144 \\
HLIP ViT & T1, T1C, T2, FLAIR & 72 axial slices at $224 \times 224$ & Percentile to [0,1], then ImageNet on slices & 512 \\
3D-Swin-DiT (14 checkpoints) & T1C and/or FLAIR & $64 \times 64 \times 64$ & Percentile (1st--99th) to [0,1] & 6,144 per channel; 12,288 when T1C+FLAIR concatenated \\
\bottomrule
\end{tabular}
\end{table}

\subsubsection{Feature extraction}
\label{sec:feature_extraction}

For 3D-Swin-DiT and each comparator family, we extracted patient-level feature vectors offline before any survival modelling. Backbone weights were frozen; no encoder was fine-tuned on OS labels. Performance therefore reflects the prognostic information already present in each representation rather than end-to-end survival training.

\subsubsection{Clinical task adult brain tumor survival surgery}

The main fine-tuning dataset (SIND) originates from Addenbrooke's Hospital (Cambridge, UK) and consists of 49 T2-weighted MRI scans acquired both before and after surgical tumour resection. As with the training data, scans were normalized to MNI space using \texttt{SPM12} and resampled to a 1$\times$1$\times$1~mm$^3$ resolution (157$\times$189$\times$156~mm). Skull-stripping was again performed using HD-BET \cite{isensee2019hdbet}.

Patients were categorized into two outcome groups: longer-term ((n = 32)) and shorter-term ((n = 17)) survival. Most individuals (42, 85\%) had glioblastoma, with additional cases of astrocytoma (1), gliosarcoma (3), and other tumour types (3). The shorter-term group comprised patients who died within 10 months of the postoperative scan, whereas the longer-term group survived for more than 10 months. All participants provided written informed consent, and the use of their data for research was approved by the Research Ethics Committee (REC reference: 19/WM/0152).

The model was trained to predict the survival class using either postoperative or preoperative imaging-derived scores as input. Performance was evaluated across both settings, and the mean and standard deviation of the evaluation metrics were computed.

\section{Evaluation Metrics}

\subsection{Evaluation Metrics for Fine-Tuning Tasks}

\subsubsection{Metrics for classification tasks}
Predicted labels are obtained via
\[
\tilde{y} = \arg\max \hat{\bm{y}}.
\]
Performance is assessed using:
\begin{align}
\mathrm{Sensitivity} &= 
\frac{\mathrm{TP}}{\mathrm{TP} + \mathrm{FN}},  \\
\mathrm{Accuracy} &= 
\frac{\mathrm{TP} + \mathrm{TN}}{\mathrm{TP}+\mathrm{TN}+\mathrm{FP}+\mathrm{FN}}, \\
\mathrm{Specificity} &= 
\frac{\mathrm{TN}}{\mathrm{TN} + \mathrm{FP}}, \\
\mathrm{Precision} &= 
\frac{\mathrm{TP}}{\mathrm{TP} + \mathrm{FP}}, \\
\mathrm{F1\text{-}score} &= 
2 \cdot \frac{\mathrm{Precision}\cdot\mathrm{Sensitivity}}
{\mathrm{Precision} + \mathrm{Sensitivity}}.
\end{align}

Model selection is performed by maximising the F1-score, which balances false positives and false negatives and is well suited for imbalanced clinical datasets.

\subsubsection{Metrics reconstruction and cross-modal generation}
We evaluate reconstruction quality using two standard image-fidelity metrics: peak signal-to-noise ratio (PSNR) and structural similarity index measure (SSIM). PSNR is reported in decibels (dB), and SSIM \cite{wang2004image} is reported on a $[0, 100]$ scale (i.e., scaled by $10^2$ from its native $[0, 1]$ range). For both metrics, higher values indicate better reconstruction.

\subsection{Evaluation Protocol for Continue Learning and Catastrophic Forgetting}
\label{sec:protocol}

 We evaluated continual learning across four sequential datasets (D1--D4) using three continual-learning strategies: sequential fine-tuning (SEQ), elastic weight consolidation (EWC), and Graph-Blueprint Pruning (GBP). Performance was evaluated using Structural Similarity Index Measure (SSIM), Root Mean Squared Error (RMSE), Mean Squared Error (MSE), Mean Absolute Error (MAE), and Peak Signal-to-Noise Ratio (PSNR).
 
Let $T=4$ denote the number of tasks. For each method
$m \in \{\text{SEQ}, \text{EWC}, \text{GBP}\}$ we form the performance matrix
$\mathbf{R}^{(m)} \in \mathbb{R}^{T \times T}$ where $R^{(m)}_{i,j}$ is the
model's score on task $j$ after sequential training on tasks $1,\dots,i$.
We additionally use the per-method single-task ceilings $R^{\star,m}_j$
(model $m$ trained only on $D_j$) and a method-agnostic random-init /
pre-CL reference $\bar b_j$.
 
\subsubsection{Average accuracy (ACC)}
\begin{equation}
\mathrm{ACC}^{(m)} = \tfrac{1}{T}\sum_{j=1}^{T} R^{(m)}_{T,j}.
\end{equation}
Headline number; performance on every task once all tasks have been seen \citep{lopezpaz2017gem,vandeven2020brain}.
 
\subsubsection{Learning accuracy (LA)}
$\mathrm{LA}^{(m)} = \tfrac{1}{T}\sum_{j=1}^{T} R^{(m)}_{j,j}.$
Mean of the diagonal: how well each task is learned at its training moment.
 
\subsubsection{Backward transfer (BWT) }
\begin{equation}
\mathrm{BWT}^{(m)} = \tfrac{1}{T-1}\sum_{j=1}^{T-1}
   \bigl(R^{(m)}_{T,j} - R^{(m)}_{j,j}\bigr).
\end{equation}
For higher-better metrics, negative BWT denotes forgetting \citep{lopezpaz2017gem}.
 
\subsubsection{Average forgetting (F) }
\begin{equation}
F^{(m)} = \tfrac{1}{T-1}\sum_{j=1}^{T-1}
   \Bigl(\max_{k<T} R^{(m)}_{k,j} - R^{(m)}_{T,j}\Bigr).
\end{equation}
Peak-minus-final, averaged over past tasks. Larger $F$ = stronger forgetting \citep{chaudhry2018riemannian}.

\subsubsection{Relative forgetting}

Relative forgetting was additionally computed to normalize for task-scale differences:
\begin{equation}
F^{rel}_j =
\frac{A_{j,j} - A_{T,j}}{A_{j,j}},
\end{equation}
for higher-is-better metrics, and:
\begin{equation}
F^{rel}_j =
\frac{A_{T,j} - A_{j,j}}{A_{j,j}},
\end{equation}
for lower-is-better metrics.

\subsubsection{Intransigence (I) }
$I^{(m)} = \tfrac{1}{T}\sum_{j=1}^{T}(R^{\star,m}_j - R^{(m)}_{j,j}).$
Inability to learn new tasks. Pairs naturally with BWT/F.
 \citep{chaudhry2018riemannian}
 
\subsubsection{Forward transfer (FWT) }
$\mathrm{FWT}^{(m)} = \tfrac{1}{T-1}\sum_{j=2}^{T}(R^{(m)}_{j-1,j} - \bar b_j).$
 \citep{lopezpaz2017gem}
\bigskip
 
The two stability metrics (BWT, F) are derived from past columns of $\mathbf{R}^{(m)}$;
the two plasticity metrics (LA, I) from the diagonal; FWT from the strictly
upper triangular part; and ACC integrates everything in the bottom row.
Together they fully characterise the stability--plasticity trade-off
\citep{parisi2019continual,wang2024comprehensive}.

\section{Experiments and Hyperparameter Tuning}

\subsection{Clinical Task Adult's Brain Tumor Survival Surgery Classification}

\subsubsection{Loss function}
The model is trained to predict clinical labels using binary cross-entropy (BCE) with logits.  
Let $\hat{\bm{y}} \in \mathbb{R}^{B \times 2}$ be the classifier logits and $y \in \{0,1\}^B$ the ground-truth labels.  
The supervised objective is
\begin{equation}
\mathcal{L}_{\mathrm{FT}}
=
\mathrm{BCE}\bigl(\hat{\bm{y}},\, y\bigr).
\end{equation}

The parameters are updated via AdamW optimisation:
\begin{equation}
\theta \leftarrow \theta 
- \eta \, \nabla_{\theta} \mathcal{L}_{\mathrm{FT}},
\end{equation}
with cosine-annealing learning rate schedule
\begin{equation}
\eta_t 
=
\eta_{\min}
+
\frac{1}{2}\bigl(\eta_0 - \eta_{\min}\bigr)
\left(1 + \cos\!\left(\frac{\pi t}{T_{\max}}\right)\right).
\end{equation}

\subsection{Learning Rate Scheduling}
A cosine annealing scheduler \cite{loshchilov2016sgdr} is applied with period $T_{\max}$:
\begin{equation}
    \eta_t = \eta_{\min} + \tfrac{1}{2}(\eta_0 - \eta_{\min})
            \left(1 + \cos \left( \frac{t}{T_{\max}}\pi \right)\right),
\end{equation}
which progressively reduces the learning rate, empirically improving stability on 3D medical segmentation tasks.

\subsubsection{Model checkpointing and early stopping}
We adopt a validation-based early stopping strategy: a checkpoint is saved whenever the mean validation score improves:
\begin{equation}
    \ell_{\text{val}} < \ell_{\text{best}},
\end{equation}
where $\ell_{\text{best}}$ stores the lowest validation loss observed so far.
This closely follows conventional supervised medical segmentation pipelines 
(\emph{e.g.}, U-Net and its derivatives) where the best model is selected based on held-out validation performance 
\cite{ronneberger2015,cicek2016,tang2022self}.

\subsubsection{Training dynamics}
The fine-tuning scheme therefore integrates  
(a) frozen anatomical priors from the Swin backbone,  
(b) generative class-conditioned diffusion features, and  
(c) a supervised classifier,  
forming a hybrid discriminative–generative model tailored for clinical decision-making.

\subsection{Clinical Task Adult's Brain Tumor Survival Estimation}

\subsubsection{Downstream survival tasks and labels}
\label{sec:downstream}

Let $T_i$ denote OS time in days and $\delta_i \in \{0,1\}$ the event indicator (1 = death, 0 = censored) for patient $i$. All encoders were evaluated on four downstream tasks using the same protocol (Table~\ref{tab:hyperparams}).

\textbf{Tasks 1 and 2 (time-to-event).} We used the full censored OS pair $(T_i,\delta_i)$ without horizon filtering. Features were linearly projected to 512 dimensions. Task~1 applied a three-layer multilayer perceptron (512-256-128-1) with ReLU activations and dropout $p=0.3$, trained with the Cox partial likelihood~\cite{cox1972}. Task~2 used a DeepSurv-style head~\cite{katzman2018deepsurv} (512-32-32-1 with batch normalisation) trained with the same Cox loss and mini-batches of size 256. Discrimination was summarised by Harrell's C-index~\cite{harrell1982cindex} on each validation fold.

\textbf{Task 3 (survival time regression).} We regressed $y_i=\log(1+T_i)$ with an MLP head matching Task~1. Training loss (L1) was computed on deceased patients only ($\delta_i=1$); censored patients were excluded from the regression loss but included when computing validation ranking. We report mean absolute error (MAE) and root mean squared error (RMSE) of predicted survival time in days among deceased patients, and a validation C-index based on $-\hat{T}_i$.

\textbf{Task 4 (fixed-horizon binary survival).}
\label{sec:task4_labels}

For horizon $h \in \{1,2,3\}$ years we set $\tau_h = 365h$ days and defined
\begin{equation}
y_i^{(h)} =
\begin{cases}
1, & T_i \ge \tau_h,\\
0, & T_i < \tau_h \;\wedge\; \delta_i = 1,\\
\text{undefined}, & T_i < \tau_h \;\wedge\; \delta_i = 0.
\end{cases}
\end{equation}
Patients censored before $\tau_h$ were excluded for that horizon. The same MLP head as Task~3 was trained with binary cross-entropy; classification used a 0.5 probability threshold. We report the area under the receiver operating characteristic curve (AUROC)~\cite{hanley1982roc}, precision--recall AUC, sensitivity, specificity, balanced accuracy, and F1. Effective sample size decreased at longer horizons (1-year prevalence $\approx 0.59$; 2-year $\approx 0.29$; 3-year $\approx 0.17$).

\subsubsection{Training, cross-validation, and metrics}
\label{sec:validation}

Models were trained with Adam ($\mathrm{lr}=10^{-3}$) for up to 150 epochs per fold. Random seeds were fixed (split seed 42; training seed $42 + \mathrm{fold\_id}$). Gradient norms were clipped at 5.0 when needed. We used five-fold stratified cross-validation with shuffling; stratification was based on the event indicator $\delta_i$ for all tasks. Reported values are the mean $\pm$ standard deviation of fold-wise validation scores. No separate locked test set was held out.

For each task, the training epoch that achieved the best validation score on the validation fold was used when reporting all metrics for that task (C-index for Tasks~1--3; AUROC for Task~4). Backbone encoders were never updated using survival labels. Hyperparameters were fixed from code defaults without nested cross-validation.

\begin{table}[htbp]
\centering
\caption{Downstream survival training settings (identical across encoder families except input dimension $d$).}
\label{tab:hyperparams}
\footnotesize
\begin{tabular}{ll}
\toprule
\textbf{Setting} & \textbf{Value} \\
\midrule
Cross-validation & 5-fold stratified \\
Stratification & Event indicator $\delta$ \\
Random seed (splits) & 42 \\
Per-fold training seed & $42 + \mathrm{fold\_id}$ \\
Optimiser & Adam, learning rate $10^{-3}$ \\
Max epochs & 150 \\
Projection dimension & 512 \\
MLP hidden layers (Cox-MLP, regression, classification) & 256, 128 \\
Dropout (MLP tasks) & 0.3 \\
Task 2 batch size & 256 \\
Gradient clipping & Max norm 5.0 \\
Classification threshold & 0.5 \\
\bottomrule
\end{tabular}
\end{table}

\subsubsection{Software and reproducibility}
\label{sec:software}

Feature extraction and survival training were implemented in Python using PyTorch. Analysis code and pretrained checkpoints will be released with the parent 3D-Swin-DiT manuscript to support reproduction of the benchmark.

\subsection{Clinical Task Synthetic Data Generation in Brain Tumor Cohorts}

In this experiment, our model acts as an autoencoder that serves as a foundation autoencoder for downstream generative tasks. An autoencoder is a neural architecture comprising an encoder--decoder pair trained to reconstruct its input from a low-dimensional latent representation, thereby learning a compressed code that retains the information necessary for recovery. Formally, given a 3D volume $x \in \mathbb{R}^{H \times W \times D}$, the encoder ${E}$ maps $x$ to a latent code $z = {E}(x) \in \mathbb{R}^{h \times w \times d \times c}$, and the decoder ${D}$ produces a reconstruction $\hat{x} = {D}(z)$. 

We evaluate it along two axes: \textbf{reconstruction} and \textbf{generation}. Reconstruction probes the compression-and-recovery capacity of the autoencoder, establishing how faithfully a 3D volume can be encoded into the latent space and decoded back. Generation probes whether cross-modality translation can be carried out directly in this latent space; success on this axis indicates that the latent representation is sufficiently structured for cross-modal mapping to be tractable.

\subsubsection{Reconstruction}
The reconstruction task assesses the model's ability to compress an input volume $x$ into a latent code $z = {E}(x)$ and recover it as $\hat{x} = {D}(z)$, thereby measuring how faithfully the encoder--decoder preserves the input through the latent bottleneck.

\subsubsection{Cross-modal generation}
Whereas reconstruction evaluates how well the autoencoder recovers $x$ from its own latent code, cross-modal generation evaluates the structure of the latent space itself: specifically, whether different modalities are organised such that the mapping between them can be learned directly in latent space. To this end, we train a translation model ${T}$ that maps a source-modality latent $z_s = {E}(x_s)$ to a target-modality latent $\hat{z}_t = {T}(z_s)$, which is then decoded as $\hat{x}_t = {D}(\hat{z}_t)$. For example, a T1 latent is mapped to a T2 latent and decoded into a T2 volume. The quality of $\hat{x}_t$ thus reflects both the geometric regularity of the latent space across modalities and the capacity of ${T}$ to exploit it.

Since $\hat{x}_t$ is decoded by the same ${D}$, reconstruction quality upper-bounds generation quality. Strong reconstruction with weak generation therefore points to the latent geometry, rather than the autoencoder, as the bottleneck. Conversely, when generation quality approaches the reconstruction ceiling, the latent space has captured a high-quality cross-modal structure that the translation model can effectively exploit.
 
\clearpage
\phantomsection
\section*{Supplementary Results and Tables}
\addcontentsline{toc}{section}{Supplementary Results and Tables}

\section{Hyperparameter Tuning Ablation and State-of-the-art Architecture Comparison}

\subsection{Training}
\subsubsection{The Importance of Batch Size for Latent Space Diffusion with Two Channels}

Batch size plays a critical role in determining the optimization dynamics, convergence behavior, and final generative quality of latent diffusion models. In our experiments, the configuration using a batch size of 14 on a single GPU exhibited significantly noisier gradient estimates and slower convergence compared to the larger distributed settings using global batch sizes of 128, 256, and 512 on two GPUs. According to the large-batch training theory of McCandlish et al.~\cite{mccandlish2018empirical} and Smith et al.~\cite{smith2018dont}, increasing the batch size reduces the variance of stochastic gradients approximately proportional to $1/\sqrt{B}$, leading to smoother optimization trajectories and more stable training. This behavior is particularly important in diffusion models, where the denoising objective must be optimized across a wide range of timesteps and noise levels.

Recent studies on diffusion model scalability confirm these observations. Li et al.~\cite{li2024scalability} demonstrated that increasing the batch size from 2048 to 4096 during the training of SDXL-based latent diffusion models consistently improved image quality metrics, including FID, CLIP score, ImageReward, and TIFA, while also producing substantially smoother convergence curves. Similarly, Xu et al.~\cite{xu2024faster} showed that diffusion training benefits from improved optimization stability, reporting up to $2\times$ faster convergence on CIFAR-10 and $2.6\times$ faster convergence on ImageNet through better exploitation of the loss landscape. Furthermore, Hang et al.~\cite{hang2023minsnr} demonstrated that reducing gradient conflicts across diffusion timesteps significantly accelerates convergence and improves sample quality, highlighting the importance of stable gradient estimation during training.

For the specific case of a two-channel latent representation, larger batch sizes are expected to be even more beneficial because the compressed latent space contains less redundancy than conventional four-channel latent representations. While a batch size of 14 may introduce beneficial stochastic regularization during early training, it often results in unstable updates and increased sensitivity to initialization. In contrast, batch sizes of 128 and 256 provide a better balance between optimization stability and generalization, yielding faster convergence and more consistent latent representations. A batch size higher than 512 further reduces gradient noise and maximizes hardware utilization, the literature suggests that diminishing returns may emerge when optimization noise is no longer the dominant limitation and the representational capacity of the latent bottleneck becomes the primary constraint. This observation is consistent with the findings of Rombach et al.~\cite{rombach2022high}, who showed that stronger latent compression improves computational efficiency but can eventually reduce reconstruction fidelity. Therefore, for two- and one- channel latent diffusion models, batch sizes in the range of 128--512 appear to provide the most favorable trade-off between training stability, convergence speed, computational efficiency, and final generative quality.In this study, a batch size of 512 was adopted for all training experiments.

\begin{table}[t]
\centering
\caption{Comparison of 3D-Swin-DiT-S2-Lv4 models on D1a, S2 level-4. Values are reported as mean $\pm$ standard deviation.}
\label{tab:dit_results}

\begin{tabular}{lcccc}
\toprule
Model & SSIM & RMSE & MSE & MAE \\
\midrule
3D-Swin-DiT-S2-Lv4 (5k) 
& $0.8723 \pm 0.0157$
& $0.0139 \pm 0.0005$
& $0.00068 \pm 0.00004$
& $0.0261 \pm 0.0008$ \\

3D-Swin-DiT-S2-Lv4 (10k) &
$0.882 \pm 0.004$ &
$0.0320 \pm 0.0007$ &
$0.00107 \pm 0.00003$ &
$0.01526 \pm 0.00030$ \\

3D-Swin-DiT-S2-Lv4 (20k) &
$0.882 \pm 0.007$ &
$0.0320 \pm 0.0008$ &
$0.00108 \pm 0.00005$ &
$0.01531 \pm 0.00040$ \\

3D-Swin-DiT-S2-Lv4 (76k) &
$0.880 \pm 0.017$ &
$0.0332 \pm 0.0018$ &
$0.00110 \pm 0.00012$ &
$0.01645 \pm 0.00100$ \\
\bottomrule
\end{tabular}

\vspace{0.5cm}

\begin{tabular}{lccc}
\toprule
Model & Training Time & Steps & Batch Size \\
\midrule
3D-Swin-DiT-S2-Lv4 (5k)  & 4 h  & 5k  & 512 \\
3D-Swin-DiT-S2-Lv4 (10k) & 6 h  & 10k & 256 \\
3D-Swin-DiT-S2-Lv4 (20k) & 8 h  & 20k & 128 \\
3D-Swin-DiT-S2-Lv4 (76k) & 70 h & 76k & 14 \\
\bottomrule
\end{tabular}

\end{table}

Based on previous studies in computer vision and neuroimaging, we expect the reconstruction quality achieved after latent diffusion training to approach, but not surpass, that of the pretrained SwinUNETR autoencoder. This behavior is consistent with the latent diffusion paradigm, where the autoencoder establishes the reconstruction fidelity ceiling and the diffusion model focuses on learning the latent data distribution rather than enhancing reconstruction accuracy~\cite{rombach2022high}. Therefore, any degradation in SSIM or PSNR is expected to be modest and primarily attributable to the stochastic latent generation process and information loss introduced during latent-space modeling. As a result, the ultimate reconstruction performance remains bounded by the quality of the latent representations learned by the autoencoder.

Similar observations have been reported in brain imaging applications \cite{pinaya2022brain} and diffusion autoencoder architectures \cite{preechakul2022diffusion}, where deterministic autoencoders maintain superior reconstruction metrics. The diffusion model' s value lies in generation and sampling capabilities, not in improved reconstruction accuracy.
\begin{table}[t]
\centering
\caption{Comparison of 3D-Swin and 3D-Swin-DiT models on D1a, S2 level-4. Values are reported as mean $\pm$ standard deviation.}
\label{tab:model_comparison}

\begin{tabular}{llccc}
\toprule
Model & Batch & SSIM & RMSE & PSNR \\
\midrule
3D-Swin-S2-Lv4     & 512 & $0.889 \pm 0.006$ & $0.0144 \pm 0.0006$ & $36.10 \pm 0.14$ \\
3D-Swin-DiT-S2-Lv4 & 512 & $0.888 \pm 0.007$ & $0.0147 \pm 0.0007$ & $36.06 \pm 0.15$ \\

3D-Swin-S2-Lv4     & 256 & $0.889 \pm 0.010$ & $0.0205 \pm 0.0011$ & $33.73 \pm 0.24$ \\
3D-Swin-DiT-S2-Lv4 & 256 & $0.888 \pm 0.011$ & $0.0210 \pm 0.0011$ & $32.79 \pm 0.23$ \\

3D-Swin-S2-Lv4     & 128 & $0.8884 \pm 0.0099$ & $0.0156 \pm 0.0008$ & $36.17 \pm 0.48$ \\
3D-Swin-DiT-S2-Lv4 & 128 & $0.889 \pm 0.004$ & $0.0156 \pm 0.0005$ & $36.20 \pm 0.30$ \\
\bottomrule
\end{tabular}

\vspace{0.3cm}

\begin{tabular}{llccc}
\toprule
Model & Batch & MSE & MAE & Steps \\
\midrule
3D-Swin-S2-Lv4     & 512 & $0.00020 \pm 0.00005$ & $0.0070 \pm 0.0004$ & 2k \\
3D-Swin-DiT-S2-Lv4 & 512 & $0.00021 \pm 0.00005$ & $0.0072 \pm 0.0004$ & 14k \\

3D-Swin-S2-Lv4     & 256 & $0.00043 \pm 0.00009$ & $0.010 \pm 0.0007$ & 4k \\
3D-Swin-DiT-S2-Lv4 & 256 & $0.00044 \pm 0.00009$ & $0.011 \pm 0.0007$ & 28k \\

3D-Swin-S2-Lv4     & 128 & $(2.43 \times 10^{-4}) \pm (2.79 \times 10^{-5})$ & $0.0079 \pm 0.0004$ & 6k \\
3D-Swin-DiT-S2-Lv4 & 128 & $0.00024 \pm 0.00004$ & $0.0079 \pm 0.0003$ & 56k \\
\bottomrule
\end{tabular}
\end{table}

\subsubsection{Determine the spatial resolution of the input images in 1 and 2 channels}

As the model operates on \textbf{3D volumetric data}, using high spatial resolutions such as $512^3$ or $256^3$ voxels across all dimensions was computationally infeasible. A practical alternative is to downsample the volumes to $128^3$ or $64^3$ resolutions.

\begin{figure}[t]
\centering
\begin{subfigure}{0.8\textwidth}
    \centering
    \includegraphics[width=\linewidth]{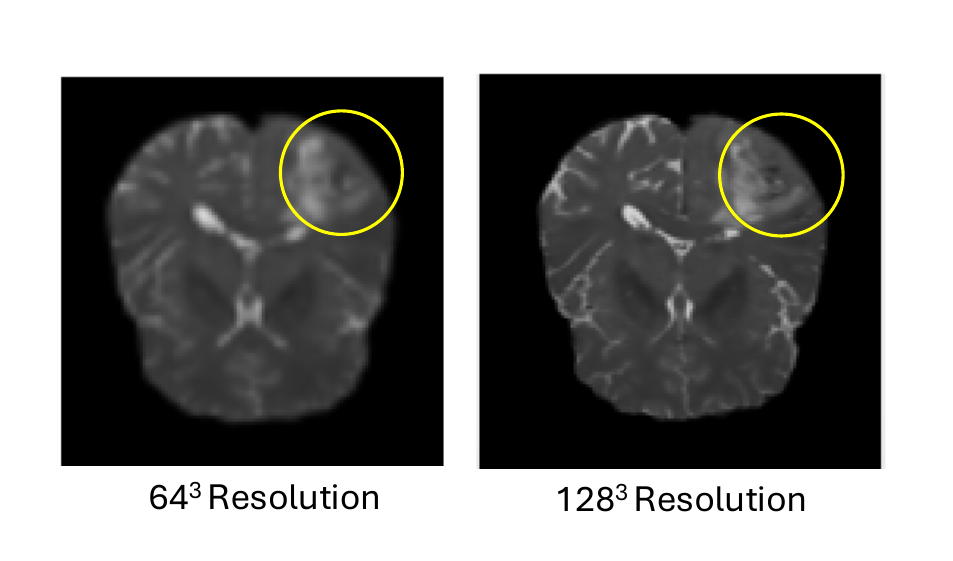}
\end{subfigure}
\hfill
\caption{
\textbf{Comparison of volumetric resolutions used for 3D modeling.}
Axial brain slices reconstructed at $64\times64\times64$ and $128\times128\times128$ voxel resolutions. 
The highlighted region (yellow circle) corresponds to a tumor located in the upper-right cortical area. 
Despite the lower spatial resolution, the key anatomical and pathological structures remain clearly identifiable, 
demonstrating that clinically relevant features are preserved after downsampling. 
This supports the use of $64^3$ volumes as a computationally efficient representation while maintaining 
the critical visual characteristics required for downstream learning tasks.
}
\label{fig:resolution_comparison}
\end{figure}

Figure~\ref{fig:resolution_comparison} illustrates the spatial representation of the data at $64 \times 64 \times 64$ and $128 \times 128 \times 128$ resolutions. Importantly, the main anatomical structures and salient image characteristics (e.g., tumor regions) remain clearly preserved at both resolutions.

In preliminary experiments on the D1b dataset, training with a resolution of $64^3$ and a batch size of 512 samples for 10k steps in a 2-channel task required approximately 8 hours, whereas $128^3$ required around 32 hours for the small model (3D-Swin-DiT-S2-Lv4). However, the downstream prediction performance for the SIND fine-tuning task did not show significant improvement when using the higher resolution (see Tab. \ref{tab:chunk_max_results_pm_L8_S8_D3}).

Therefore, we adopted the $64^3$ resolution as it provides the most computationally efficient and balanced trade-off between training cost and predictive performance.

\begin{table}[t]
\centering
\caption{Mean $\pm$ Standard Deviation of 20-epochs fine-tuning ($3D-Swin-DiT-S2-Lv4-128$ and $3D-Swin-DiT-S2-Lv4-64$) at D1b UK-BioBANK data point}
\label{tab:chunk_max_results_pm_L8_S8_D3}
\begin{tabular}{lll}
\toprule
Model & Metric & Mean $\pm$ SD \\
\midrule
3D-Swin-DiT-S2-Lv4-64 & f1\_score & $0.6422 \pm 0.0446$ \\
3D-Swin-DiT-S2-Lv4-128 & f1\_score & $0.5954 \pm 0.0118$ \\
3D-Swin-DiT-S2-Lv4-64 & Sens & $0.6944 \pm 0.0627$ \\
3D-Swin-DiT-S2-Lv4-128 & Sens & $0.6345 \pm 0.0021$ \\
3D-Swin-DiT-S2-Lv4-64 & Prec\_metric & $0.6884 \pm 0.0552$ \\
3D-Swin-DiT-S2-Lv4-128 & Prec\_metric & $0.6432 \pm 0.0103$ \\
3D-Swin-DiT-S2-Lv4-64 & Acc & $0.6979 \pm 0.0470$ \\
3D-Swin-DiT-S2-Lv4-128 & Acc & $0.6563 \pm 0.0395$ \\
\bottomrule
\end{tabular}
\end{table}

\subsubsection{Ablation and Architecture Comparison of Two-Channel Self-supervised training on D1a–c UK-Biobank sub-groups}

Across the three UK-Biobank reconstruction targets, two architectures
consistently outperform the rest: the deterministic
3D-Swin-S8-Lv2 trained from scratch, and the
3D-Swin-DiT-S8-Lv2 with conditioning, with the relative ranking
between the two depending on the imaging modality (Table~\ref{tab:ukbb_results_mean_sd}). On the diffusion-MRI target (\textbf{D1a}), 3D-Swin-DiT-S8-Lv2 attains
the smallest voxel-wise error of any architecture by a substantial
margin ($\mathrm{MAE} = 0.0139 \pm 0.0005$,
$\mathrm{MSE} = 0.0007 \pm 0.0000$,
$\mathrm{RMSE} = 0.0261 \pm 0.0008$), with reconstruction errors
roughly half those of the deterministic Swin baseline
($\mathrm{RMSE} = 0.0535 \pm 0.0483$) and an order of magnitude smaller
than the variational alternatives. Its SSIM
($0.8722 \pm 0.0161$) is within $0.01$ of the highest value reported on
this dataset by 3D-Swin-S8-Lv2 ($0.8837 \pm 0.0183$), but with
substantially tighter confidence intervals on every voxel-wise score.
On the NODDI target (\textbf{D1c}), 3D-Swin-DiT-S8-Lv2 again attains
the second-highest SSIM ($0.8853 \pm 0.0156$) with voxel-wise errors
(MAE $0.0385$, MSE $0.0066$, RMSE $0.0814$) comparable to the
3D-Swin-S8-Lv2 baseline on perceptual quality but with tighter
confidence intervals. On the T1-weighted target (\textbf{D1b}),
pretraining on D1a recovers most of the gap to the deterministic Swin
baseline: 3D-Swin-DiT-S8-Lv2 (D1a-Co) reaches
$\mathrm{SSIM} = 0.8217 \pm 0.0148$ and
$\mathrm{RMSE} = 0.0858 \pm 0.0015$, improving on the from-scratch
variant by approximately $40\%$ in MAE, MSE and RMSE and confirming
that the diffusion-MRI pretraining transfers usefully across modalities.

\begin{table*}[t]
\centering
\caption{\textbf{Mean $\pm$ standard deviation on UK-Biobank datasets for the S8-Lv2 architecture.}
SSIM ($\uparrow$) indicates better performance, whereas MAE, MSE, and RMSE ($\downarrow$) indicate lower reconstruction error. Sc: training from scratch; Sc-Un: diffusion without conditioning; Sc-Co: diffusion with conditioning; D1a-Co and D1ab-Co denote pretrained diffusion models.}
\label{tab:ukbb_results_mean_sd}
\scriptsize
\resizebox{\textwidth}{!}{
\begin{tabular}{llcccc}
\toprule
\multirow{2}{*}{Dataset} &
\multirow{2}{*}{Model} &
\multicolumn{1}{c}{Quality} &
\multicolumn{3}{c}{Error Metrics} \\
\cmidrule(lr){3-3}\cmidrule(lr){4-6}
& & SSIM $\uparrow$ & MAE $\downarrow$ & MSE $\downarrow$ & RMSE $\downarrow$ \\
\midrule

\multirow{6}{*}{dMRI}
& 3D-Swin                 & $0.8837\pm0.0183$ & $0.0292\pm0.0270$ & $0.0052\pm0.0069$ & $0.0535\pm0.0483$ \\
& 3D-Swin-VAE             & $0.7831\pm0.0408$ & $0.0267\pm0.0011$ & $0.0021\pm0.0001$ & $0.0458\pm0.0015$ \\
& 3D-Swin-DiT (Sc-Un)     & $0.8722\pm0.0161$ & $0.0139\pm0.0005$ & $0.0007\pm0.0000$ & $0.0261\pm0.0008$ \\
& 3D-Swin-DiT (Sc-Co)     & $0.8723\pm0.0157$ & $0.0139\pm0.0005$ & $0.0007\pm0.0000$ & $0.0261\pm0.0008$ \\
& 3D-Swin-VAE-DiT (Sc-Un) & $0.7960\pm0.0354$ & $0.0247\pm0.0009$ & $0.0019\pm0.0000$ & $0.0431\pm0.0013$ \\
& 3D-Swin-VAE-DiT (Sc-Co) & $0.7960\pm0.0365$ & $0.0247\pm0.0009$ & $0.0018\pm0.0000$ & $0.0431\pm0.0013$ \\

\midrule

\multirow{8}{*}{MRI}
& 3D-Swin                 & $0.8183\pm0.0213$ & $0.0627\pm0.0101$ & $0.0119\pm0.0039$ & $0.1077\pm0.0183$ \\
& 3D-Swin-VAE             & $0.7816\pm0.0412$ & $0.0269\pm0.0011$ & $0.0021\pm0.0001$ & $0.0462\pm0.0015$ \\
& 3D-Swin-DiT (Sc-Un)     & $0.7843\pm0.0144$ & $0.0842\pm0.0013$ & $0.0221\pm0.0005$ & $0.1486\pm0.0016$ \\
& 3D-Swin-DiT (Sc-Co)     & $0.7841\pm0.0148$ & $0.0842\pm0.0013$ & $0.0221\pm0.0005$ & $0.1486\pm0.0016$ \\
& 3D-Swin-VAE-DiT (Sc-Un) & $0.8008\pm0.0143$ & $0.0574\pm0.0013$ & $0.0095\pm0.0003$ & $0.0973\pm0.0017$ \\
& 3D-Swin-VAE-DiT (Sc-Co) & $0.8006\pm0.0145$ & $0.0574\pm0.0013$ & $0.0095\pm0.0003$ & $0.0973\pm0.0017$ \\
& 3D-Swin-DiT (D1a-Co)    & $0.8217\pm0.0148$ & $0.0498\pm0.0008$ & $0.0074\pm0.0003$ & $0.0858\pm0.0015$ \\
& 3D-Swin-VAE-DiT (D1a-Co)& $0.8007\pm0.0143$ & $0.0574\pm0.0013$ & $0.0095\pm0.0003$ & $0.0973\pm0.0017$ \\

\midrule

\multirow{8}{*}{NODDI}
& 3D-Swin                 & $0.8869\pm0.0150$ & $0.0427\pm0.0069$ & $0.0088\pm0.0030$ & $0.0923\pm0.0161$ \\
& 3D-Swin-VAE             & $0.7689\pm0.0435$ & $0.0284\pm0.0013$ & $0.0023\pm0.0001$ & $0.0484\pm0.0017$ \\
& 3D-Swin-DiT (Sc-Un)     & $0.8853\pm0.0156$ & $0.0385\pm0.0013$ & $0.0066\pm0.0003$ & $0.0814\pm0.0020$ \\
& 3D-Swin-DiT (Sc-Co)     & $0.8853\pm0.0159$ & $0.0385\pm0.0013$ & $0.0066\pm0.0003$ & $0.0814\pm0.0020$ \\
& 3D-Swin-VAE-DiT (Sc-Un) & $0.8881\pm0.0154$ & $0.0267\pm0.0012$ & $0.0032\pm0.0003$ & $0.0565\pm0.0027$ \\
& 3D-Swin-VAE-DiT (Sc-Co) & $0.8881\pm0.0158$ & $0.0267\pm0.0012$ & $0.0032\pm0.0003$ & $0.0565\pm0.0026$ \\
& 3D-Swin-DiT (D1ab-Co)   & $0.8854\pm0.0150$ & $0.0385\pm0.0012$ & $0.0066\pm0.0003$ & $0.0814\pm0.0020$ \\
& 3D-Swin-VAE-DiT (D1ab-Co)
                            & $0.8881\pm0.0152$ & $0.0267\pm0.0012$ & $0.0032\pm0.0003$ & $0.0565\pm0.0026$ \\

\bottomrule
\end{tabular}}
\end{table*}

\begin{table*}[t]
\centering
\caption{\textbf{95\% confidence intervals on UK-Biobank datasets ($N=4000$) for the S8-Lv2 architecture. Co: conditioning in the diffusion latent space; Un: no conditioning in the diffusion latent space.}}
\label{tab:ukbb_results_ci_s8}
\scriptsize
\resizebox{\textwidth}{!}{
\begin{tabular}{llcccc}
\toprule
\multirow{2}{*}{Dataset} &
\multirow{2}{*}{Model} &
\multicolumn{1}{c}{Quality} &
\multicolumn{3}{c}{Error Metrics} \\
\cmidrule(lr){3-3}\cmidrule(lr){4-6}
& & SSIM & MAE & MSE & RMSE \\
\midrule

\multirow{6}{*}{dMRI}
& 3D-Swin                 & $[0.8826,0.8848]$ & $[0.0283,0.0301]$ & $[0.0049,0.0055]$ & $[0.0520,0.0550]$ \\
& 3D-Swin-VAE             & $[0.7820,0.7850]$ & $[0.0266,0.0268]$ & $[0.00205,0.00215]$ & $[0.0456,0.0468]$ \\
& 3D-Swin-DiT (Un)        & $[0.8717,0.8727]$ & $[0.01388,0.01392]$ & $[0.00068,0.00069]$ & $[0.02610,0.02615]$ \\
& 3D-Swin-DiT (Co)        & $[0.8717,0.8728]$ & $[0.01388,0.01392]$ & $[0.00068,0.00069]$ & $[0.02611,0.02616]$ \\
& 3D-Swin-VAE-DiT (Un)    & $[0.7956,0.7964]$ & $[0.02472,0.02477]$ & $[0.00185,0.00186]$ & $[0.04303,0.04311]$ \\
& 3D-Swin-VAE-DiT (Co)    & $[0.7956,0.7965]$ & $[0.02472,0.02477]$ & $[0.00185,0.00186]$ & $[0.04303,0.04311]$ \\
\midrule

\multirow{6}{*}{MRI}
& 3D-Swin                 & $[0.8176,0.8190]$ & $[0.0614,0.0640]$ & $[0.0107,0.0131]$ & $[0.1057,0.1097]$ \\
& 3D-Swin-VAE             & $[0.7820,0.7850]$ & $[0.0266,0.0268]$ & $[0.00205,0.00215]$ & $[0.0456,0.0468]$ \\
& 3D-Swin-DiT (Un)        & $[0.7838,0.7848]$ & $[0.08415,0.08423]$ & $[0.02207,0.02210]$ & $[0.14855,0.14864]$ \\
& 3D-Swin-DiT (Co)        & $[0.7837,0.7849]$ & $[0.08415,0.08423]$ & $[0.02207,0.02210]$ & $[0.14855,0.14864]$ \\
& 3D-Swin-VAE-DiT (Un)    & $[0.8004,0.8012]$ & $[0.05740,0.05748]$ & $[0.00946,0.00948]$ & $[0.09725,0.09736]$ \\
& 3D-Swin-VAE-DiT (Co)    & $[0.8002,0.8010]$ & $[0.05740,0.05748]$ & $[0.00946,0.00948]$ & $[0.09725,0.09736]$ \\
\midrule

\multirow{6}{*}{NODDI}
& 3D-Swin                 & $[0.8856,0.8882]$ & $[0.0419,0.0435]$ & $[0.0083,0.0093]$ & $[0.0910,0.0941]$ \\
& 3D-Swin-VAE             & $[0.7820,0.7850]$ & $[0.0266,0.0268]$ & $[0.00205,0.00215]$ & $[0.0456,0.0468]$ \\
& 3D-Swin-DiT (Un)        & $[0.8850,0.8857]$ & $[0.03848,0.03858]$ & $[0.00662,0.00664]$ & $[0.0813,0.0815]$ \\
& 3D-Swin-DiT (Co)        & $[0.8850,0.8857]$ & $[0.03848,0.03858]$ & $[0.00662,0.00664]$ & $[0.0813,0.0815]$ \\
& 3D-Swin-VAE-DiT (Un)    & $[0.8876,0.8886]$ & $[0.02664,0.02672]$ & $[0.00318,0.00322]$ & $[0.0564,0.0566]$ \\
& 3D-Swin-VAE-DiT (Co)    & $[0.8876,0.8886]$ & $[0.02664,0.02672]$ & $[0.00318,0.00322]$ & $[0.0564,0.0566]$ \\

\bottomrule
\end{tabular}}
\end{table*}

Although the deterministic 3D-Swin-S8-Lv2 baseline attains the highest
SSIM on two of three datasets, it is an autoencoder, not a generative
model: it can reconstruct existing inputs but cannot sample new MRI
volumes from a prior. Of the architectures that can generate
volumes (the two diffusion-transformer variants), the deterministic-latent
3D-Swin-DiT-S8-Lv2 outperforms the variational
3D-Swin-VAE-DiT-S8-Lv2 on both perceptual and voxel-wise metrics on
D1a (SSIM $0.8722$ vs $0.7960$;
RMSE $0.0261$ vs $0.0431$) and on D1b (SSIM $0.8217$ for the
D1a-pretrained variant vs $0.8007$; RMSE $0.0858$ vs $0.0973$). The
variational latent reduces reconstruction fidelity uniformly across
datasets, with SSIM losses of $0.05$--$0.08$ relative to the
deterministic latent on D1a and D1b. The latent regularisation that the
VAE introduces, while attractive in principle for downstream generative
sampling, costs more in reconstruction quality than it is worth at the
present model scale: the deterministic Swin produces
substantially sharper reconstructions and the diffusion transformer on
top of it already provides a generative prior over those latents. For
this reason we adopt \textbf{3D-Swin-DiT-S8-Lv2} as the
generation-capable architecture in the rest of the work. The
conditioned and unconditioned variants of each diffusion-transformer
pair are statistically indistinguishable in this evaluation
(overlapping 95\% confidence intervals on every metric across all three
datasets, (Table~\ref{tab:ukbb_results_ci_s8})); we adopt the conditioned variant going forward because the
conditioning channel is required for the downstream continual-learning
experiments, even though it contributes negligibly under the present
reconstruction-only regime.

\subsubsection{Best results Architecture of different scale and the whole grouping of the dataset in the Self-supervised training in 2 channels task}

The results presented in Table~\ref{tab:ukbb_results_ci_only2} demonstrate several consistent trends across the UK-Biobank datasets. First, scaling the DiT backbone from Small (S) to XL produces negligible differences in performance, as evidenced by the nearly identical mean values and highly overlapping 95\% confidence intervals across all model scales Table~\ref{tab:ukbb_results_ci_only22}. This suggests that increasing the transformer capacity does not provide additional benefits once the latent diffusion model has sufficient representational power for the considered reconstruction tasks. Second, the impact of the latent representation differs across modalities. For both dMRI and structural MRI, the standard 3D-Swin-DiT models consistently outperform their VAE-based counterparts, achieving higher SSIM values and substantially lower MAE, MSE, and RMSE. For example, on the dMRI dataset, the standard DiT models achieve an SSIM of approximately 0.872 compared with 0.796 for the VAE variants, while reducing the reconstruction error by nearly a factor of three. In contrast, the NODDI dataset exhibits the opposite behavior, where the VAE-based models achieve the best overall performance, reaching an SSIM of approximately 0.888 and reducing MAE, MSE, and RMSE by more than 30\% relative to the corresponding standard DiT models. This observation indicates that the latent VAE representation is particularly effective for capturing the microstructural characteristics encoded in NODDI maps, whereas direct latent diffusion conditioning is more suitable for dMRI and structural MRI reconstruction. Finally, the narrow confidence intervals obtained from the large evaluation cohort ($N=4000$) indicate high statistical stability and reproducibility of the reported results, confirming that the observed performance differences arise from the model architecture and latent representation rather than random variation.

\begin{table*}[t]
\centering
\caption{\textbf{Mean $\pm$ SD across UK-Biobank datasets and foundation model scales. Co: training from scratch with conditioning in the diffusion latent space. All models follow the 3D-Swin-\textit{Name}-Lv2 architectural configuration.}}
\label{tab:ukbb_results_ci_only2}
\scriptsize
\resizebox{\textwidth}{!}{
\begin{tabular}{llcccc}
\toprule
\multirow{2}{*}{Dataset} &
\multirow{2}{*}{Model} &
\multicolumn{1}{c}{Quality} &
\multicolumn{3}{c}{Error Metrics} \\
\cmidrule(lr){3-3}\cmidrule(lr){4-6}
& & SSIM & MAE & MSE & RMSE \\
\midrule

\multirow{8}{*}{dMRI}
& DiT-S8      & $0.8723\pm0.0157$ & $0.0139\pm0.0005$ & $0.00068\pm0.00004$ & $0.0261\pm0.0008$ \\
& VAE-DiT-S8  & $0.7960\pm0.0365$ & $0.0247\pm0.0009$ & $0.00186\pm0.00011$ & $0.0431\pm0.0013$ \\
& DiT-B8      & $0.8723\pm0.0161$ & $0.0139\pm0.0005$ & $0.00068\pm0.00004$ & $0.0261\pm0.0008$ \\
& VAE-DiT-B8  & $0.7958\pm0.0351$ & $0.0248\pm0.0009$ & $0.00186\pm0.00011$ & $0.0431\pm0.0013$ \\
& DiT-L8      & $0.8723\pm0.0158$ & $0.0139\pm0.0006$ & $0.00068\pm0.00004$ & $0.0261\pm0.0008$ \\
& VAE-DiT-L8  & $0.7958\pm0.0352$ & $0.0248\pm0.0009$ & $0.00186\pm0.00011$ & $0.0431\pm0.0013$ \\
& DiT-XL8     & $0.8723\pm0.0161$ & $0.0139\pm0.0005$ & $0.00068\pm0.00004$ & $0.0261\pm0.0008$ \\
& VAE-DiT-XL8 & $0.7958\pm0.0363$ & $0.0248\pm0.0009$ & $0.00186\pm0.00011$ & $0.0431\pm0.0013$ \\
\midrule

\multirow{8}{*}{MRI}
& DiT-S8      & $0.8217\pm0.0148$ & $0.0498\pm0.0008$ & $0.0074\pm0.0003$ & $0.0858\pm0.0015$ \\
& VAE-DiT-S8  & $0.8007\pm0.0143$ & $0.0574\pm0.0013$ & $0.0095\pm0.0003$ & $0.0973\pm0.0017$ \\
& DiT-B8      & $0.8218\pm0.0151$ & $0.0498\pm0.0008$ & $0.0074\pm0.0003$ & $0.0858\pm0.0015$ \\
& VAE-DiT-B8  & $0.8007\pm0.0148$ & $0.0574\pm0.0013$ & $0.0095\pm0.0003$ & $0.0973\pm0.0017$ \\
& DiT-L8      & $0.8217\pm0.0148$ & $0.0498\pm0.0008$ & $0.0074\pm0.0003$ & $0.0858\pm0.0015$ \\
& VAE-DiT-L8  & $0.8006\pm0.0141$ & $0.0574\pm0.0013$ & $0.0095\pm0.0003$ & $0.0973\pm0.0017$ \\
& DiT-XL8     & $0.8217\pm0.0162$ & $0.0498\pm0.0009$ & $0.0074\pm0.0003$ & $0.0858\pm0.0016$ \\
& VAE-DiT-XL8 & $0.8007\pm0.0146$ & $0.0574\pm0.0014$ & $0.0095\pm0.0003$ & $0.0973\pm0.0017$ \\
\midrule

\multirow{8}{*}{NODDI}
& DiT-S8      & $0.8854\pm0.0150$ & $0.0385\pm0.0012$ & $0.0066\pm0.0003$ & $0.0814\pm0.0020$ \\
& VAE-DiT-S8  & $0.8881\pm0.0152$ & $0.0267\pm0.0012$ & $0.0032\pm0.0003$ & $0.0565\pm0.0026$ \\
& DiT-B8      & $0.8853\pm0.0153$ & $0.0385\pm0.0013$ & $0.0066\pm0.0003$ & $0.0814\pm0.0021$ \\
& VAE-DiT-B8  & $0.8881\pm0.0165$ & $0.0267\pm0.0012$ & $0.0032\pm0.0003$ & $0.0565\pm0.0026$ \\
& DiT-L8      & $0.8854\pm0.0167$ & $0.0385\pm0.0014$ & $0.0066\pm0.0004$ & $0.0814\pm0.0022$ \\
& VAE-DiT-L8  & $0.8881\pm0.0157$ & $0.0267\pm0.0012$ & $0.0032\pm0.0003$ & $0.0565\pm0.0026$ \\
& DiT-XL8     & $0.8853\pm0.0174$ & $0.0384\pm0.0014$  & $0.0066\pm0.0004$ & $0.0813\pm0.0022$\\
& VAE-DiT-XL8 & $0.8253\pm0.0156$ & $0.0585\pm0.0014$ & $0.0083\pm0.0003$ & $0.0914\pm0.0020$ \\
\bottomrule
\end{tabular}}
\end{table*}

From a foundation-model perspective, the standard 3D-Swin-DiT architecture appears to be the most suitable candidate for large-scale training across diverse cohorts. Although the VAE-based variants achieve superior performance on the NODDI dataset, the standard DiT models consistently outperform the VAE counterparts on both dMRI and structural MRI, while maintaining remarkably stable performance across all model scales. This robustness suggests that the latent representations learned by the DiT architecture generalize more effectively across heterogeneous imaging modalities and acquisition settings. Furthermore, the negligible performance differences observed between the Small, Base, Large, and XL variants indicate that scaling model capacity alone does not substantially improve reconstruction quality, implying that the learned latent representation, rather than model size, is the dominant factor governing performance. Therefore, for large-scale foundation model training on multiple cohorts, the standard DiT architecture offers the most favorable balance between accuracy, stability, and cross-dataset generalization, whereas VAE-based models may be advantageous for specialized microstructural imaging tasks such as NODDI reconstruction.

\begin{table*}[t]
\centering
\caption{\textbf{95\% confidence intervals on UK-Biobank datasets ($N=4000$). Co: training from scratch with conditioning in the diffusion latent space. All models follow the 3D-Swin-\textit{Name}-Lv2 architectural configuration.}}
\label{tab:ukbb_results_ci_only22}
\scriptsize
\resizebox{\textwidth}{!}{
\begin{tabular}{llcccc}
\toprule
\multirow{2}{*}{Dataset} &
\multirow{2}{*}{Model} &
\multicolumn{1}{c}{Quality} &
\multicolumn{3}{c}{Error Metrics} \\
\cmidrule(lr){3-3}\cmidrule(lr){4-6}
& & SSIM & MAE & MSE & RMSE \\
\midrule

\multirow{8}{*}{dMRI}
& DiT-S8      & $[0.8718,0.8728]$ & $[0.01390,0.01393]$ & $[0.000682,0.000686]$ & $[0.02612,0.02615]$ \\
& VAE-DiT-S8  & $[0.7949,0.7972]$ & $[0.02472,0.02478]$ & $[0.001854,0.001860]$ & $[0.04303,0.04311]$ \\
& DiT-B8      & $[0.8718,0.8728]$ & $[0.01390,0.01393]$ & $[0.000682,0.000686]$ & $[0.02612,0.02615]$ \\
& VAE-DiT-B8  & $[0.7947,0.7969]$ & $[0.02473,0.02479]$ & $[0.001855,0.001861]$ & $[0.04305,0.04313]$ \\
& DiT-L8      & $[0.8719,0.8728]$ & $[0.01389,0.01393]$ & $[0.000682,0.000686]$ & $[0.02611,0.02615]$ \\
& VAE-DiT-L8  & $[0.7947,0.7970]$ & $[0.02472,0.02479]$ & $[0.001855,0.001861]$ & $[0.04304,0.04313]$ \\
& DiT-XL8     & $[0.8718,0.8729]$ & $[0.01390,0.01394]$ & $[0.000683,0.000686]$ & $[0.02612,0.02615]$ \\
& VAE-DiT-XL8 & $[0.7946,0.7970]$ & $[0.02472,0.02479]$ & $[0.001855,0.001861]$ & $[0.04305,0.04313]$ \\

\midrule

\multirow{8}{*}{MRI}
& DiT-S8      & $[0.8212,0.8222]$ & $[0.04977,0.04983]$ & $[0.00739,0.00741]$ & $[0.08575,0.08585]$ \\
& VAE-DiT-S8  & $[0.8003,0.8011]$ & $[0.05736,0.05744]$ & $[0.00949,0.00951]$ & $[0.09727,0.09733]$ \\
& DiT-B8      & $[0.8213,0.8223]$ & $[0.04977,0.04983]$ & $[0.00739,0.00741]$ & $[0.08575,0.08585]$ \\
& VAE-DiT-B8  & $[0.8002,0.8012]$ & $[0.05736,0.05744]$ & $[0.00949,0.00951]$ & $[0.09727,0.09733]$ \\
& DiT-L8      & $[0.8212,0.8222]$ & $[0.04977,0.04983]$ & $[0.00739,0.00741]$ & $[0.08575,0.08585]$ \\
& VAE-DiT-L8  & $[0.8002,0.8010]$ & $[0.05736,0.05744]$ & $[0.00949,0.00951]$ & $[0.09727,0.09733]$ \\
& DiT-XL8     & $[0.8212,0.8222]$ & $[0.0498,0.0498]$ & $[0.0074,0.0074]$ & $[0.0858,0.0859]$ \\
& VAE-DiT-XL8 & $[0.8002,0.8011]$ & $[0.0574,0.0575]$ & $[0.0095,0.0095]$ & $[0.0972,0.0973]$ \\

\midrule

\multirow{8}{*}{NODDI}
& DiT-S8      & $[0.8876,0.8886]$ & $[0.02666,0.02674]$ & $[0.00319,0.00321]$ & $[0.05642,0.05658]$ \\
& VAE-DiT-S8  & $[0.8876,0.8886]$ & $[0.02666,0.02674]$ & $[0.00319,0.00321]$ & $[0.05642,0.05658]$ \\
& DiT-B8      & $[0.8876,0.8886]$ & $[0.02666,0.02674]$ & $[0.00319,0.00321]$ & $[0.05642,0.05658]$ \\
& VAE-DiT-B8  & $[0.8876,0.8886]$ & $[0.02666,0.02674]$ & $[0.00319,0.00321]$ & $[0.05642,0.05658]$ \\
& DiT-L8      & $[0.8848,0.8859]$ & $[0.0385,0.0386]$ & $[0.0066,0.0066]$ & $[0.0813,0.0815]$ \\
& VAE-DiT-L8  & $[0.8876,0.8886]$ & $[0.0266,0.0267]$ & $[0.0032,0.0032]$ & $[0.0564,0.0566]$ \\
& DiT-XL8     & $[0.8844,0.8862]$ & $[0.0811,0.0814]$ & $[0.0066,0.0066]$ & $[0.0384,0.0385]$ \\
& VAE-DiT-XL8 & $[0.8244,0.8261]$ & $[0.05845,0.05860]$ & $[0.00834,0.00837]$ & $[0.0913,0.0915]$ \\

\bottomrule
\end{tabular}}
\end{table*}

\subsubsection{Choice of 3D-Swin-DiT for downstream generative training.}
The multi-scale evaluation in Tab.~\ref{tab:ukbb_results_mean_sd} and
Tab.~\ref{tab:ukbb_results_ci_only2} supports adopting the
deterministic-latent \textbf{3D-Swin-DiT} as the foundation model for the
generative tasks that follow, in preference to its variational counterpart
3D-Swin-VAE-DiT. Two observations drive this choice. First, across the
four parameter scales (S8\,$\to$\,B8\,$\to$\,L8\,$\to$\,XL8), reconstruction
quality saturates within each architectural family: the metrics on D1 dMRI
and D1 MRI are identical to four decimal places across scales, with
overlapping 95\% confidence intervals
($\mathrm{SSIM} \in [0.8718, 0.8729]$ on D1 dMRI;
$\mathrm{SSIM} \in [0.8212, 0.8223]$ on D1 MRI), indicating that the small
3D-Swin-DiT-S8-Lv2 already captures the reconstruction signal that the
larger scales offer. We therefore retain the S8 scale as the default, since
additional capacity provides no measurable benefit on the reconstruction
objective within the present training budget. Second, at every scale the
deterministic 3D-Swin-DiT outperforms 3D-Swin-VAE-DiT on the two largest
clinical modalities: on D1 dMRI by $0.076$ SSIM
($0.8723 \pm 0.0157$ vs $0.7960 \pm 0.0365$) and $39\%$ lower RMSE
($0.0261 \pm 0.0008$ vs $0.0431 \pm 0.0013$); on D1 MRI by $0.021$ SSIM
($0.8217 \pm 0.0148$ vs $0.8007 \pm 0.0143$) and $12\%$ lower RMSE. The
VAE variant achieves comparable or marginally better quality only on
D1 NODDI ($\mathrm{SSIM} = 0.8881$ vs $0.8854$), where the derived
microstructural maps have smoother spatial statistics; on the underlying
dMRI and structural-MRI signals, the prior-matching regularisation imposed
by the variational bottleneck costs more reconstruction fidelity than it
recovers. For a foundation model whose downstream role is generating
unseen MRI volumes from a learned prior, the latent must support sharp,
voxel-accurate decoding; the deterministic Swin meets this
requirement consistently across modalities while the variational variant
does not. We therefore adopt 3D-Swin-DiT-S8-Lv2 (deterministic latent,
conditioned) as the architecture for the continual-learning and
generative experiments that follow.

\subsubsection{Effect of Patch Size and Network Depth Across UK-Biobank Modalities}
Table~\ref{tab2} presents an ablation study investigating the effects of model scale and hierarchical depth across three UK-Biobank imaging modalities, namely diffusion MRI (dMRI), structural MRI, and NODDI. The analysis compares Small (S), Large (L), and Extra-Large (XL) architectures under different latent patch resolutions and hierarchical levels.

A consistent trend across all modalities is that increasing the hierarchical depth from Lv2 to Lv4 leads to improvements in structural similarity, as measured by SSIM. For dMRI, the Lv4 configuration improves SSIM from $0.8723$ to $0.8796$, while for structural MRI the increase is from $0.8217$ to $0.8322$. Similar improvements are observed for NODDI, where SSIM rises from $0.8854$ to $0.8860$. These findings suggest that deeper hierarchical representations capture richer anatomical and structural information, resulting in improved perceptual reconstruction quality.

However, the benefits of increased hierarchical depth are not always reflected in voxel-wise reconstruction metrics. For both dMRI and structural MRI, the Lv4 configurations exhibit higher MAE, MSE, and RMSE values compared with their Lv2 counterparts. This indicates a trade-off between perceptual fidelity and pixel-level reconstruction accuracy, where deeper latent hierarchies may prioritise the preservation of global anatomical structure over exact intensity reconstruction. In contrast, NODDI benefits from increased depth in both SSIM and reconstruction error metrics, achieving the best overall performance with the S2-Lv4 configuration. This behaviour suggests that the more complex microstructural information encoded within NODDI data may benefit from higher-resolution hierarchical representations.

The influence of model scale appears considerably less significant than the influence of hierarchical depth. Across all datasets, the S8-Lv2 and L8-Lv2 configurations produce nearly identical performance, with differences generally falling within the reported standard deviations. Likewise, the XL2-Lv4 models provide negligible improvements compared with the corresponding S2-Lv4 configurations. These results indicate that increasing the parameter count alone does not substantially improve reconstruction performance under the current self-supervised training regime.

From a modality-specific perspective, NODDI consistently achieves the highest SSIM values across all configurations, reaching approximately $0.886$, whereas structural MRI produces the lowest SSIM values, ranging between $0.822$ and $0.832$. This observation likely reflects differences in signal complexity and anatomical variability between modalities. NODDI-derived microstructural maps may contain more regular spatial patterns that are easier to reconstruct, whereas structural MRI presents greater anatomical diversity and higher-frequency details.

Overall, the results demonstrate that hierarchical depth constitutes the primary architectural factor influencing reconstruction quality, whereas increasing model scale beyond the Small configuration yields minimal benefit. These findings suggest that future development of foundation models for neuroimaging should prioritise improvements in hierarchical latent representations and multi-scale feature extraction rather than simply increasing model capacity. Furthermore, the modality-dependent improvements observed for NODDI indicate that optimal architectural choices may depend on the spatial and statistical characteristics of the underlying imaging modality.

\begin{table*}[t]
\centering
\caption{\textbf{Mean $\pm$ SD across UK-Biobank datasets and architectural configurations. Co: training from scratch with conditioning in the diffusion latent space. All models follow the 3D-Swin-\textit{Name} architectural configuration.}}
\label{tab2}

\scriptsize
\setlength{\tabcolsep}{3pt}

\resizebox{\textwidth}{!}{
\begin{tabular}{lcccc}
\toprule
\textbf{Model}
& \textbf{SSIM}
& \textbf{MAE}
& \textbf{MSE}
& \textbf{RMSE} \\
\midrule

\multicolumn{5}{l}{\textbf{D1 dMRI: UK-Biobank}} \\
\midrule

DiT-S8-Lv2
& $0.8723\pm0.0157$
& $0.0139\pm0.0005$
& $0.00068\pm0.00004$
& $0.0261\pm0.0008$ \\

DiT-S2-Lv4
& $0.8796\pm0.0170$
& $0.0164\pm0.0011$
& $0.0011\pm0.0001$
& $0.0330\pm0.0020$ \\

DiT-L8-Lv2
& $0.8723\pm0.0161$
& $0.0139\pm0.0005$
& $0.00068\pm0.00004$
& $0.0261\pm0.0008$ \\

DiT-XL2-Lv4
& $0.8796\pm0.0169$
& $0.0164\pm0.0011$
& $0.0010\pm0.00013$
& $0.0330\pm0.0019$ \\

\midrule

\multicolumn{5}{l}{\textbf{D1 MRI: UK-Biobank}} \\
\midrule

DiT-S8-Lv2
& $0.8217\pm0.0148$
& $0.0498\pm0.0008$
& $0.0074\pm0.0003$
& $0.0858\pm0.0015$ \\

DiT-S2-Lv4
& $0.8322\pm0.0157$
& $0.0521\pm0.0008$
& $0.0079\pm0.0003$
& $0.0891\pm0.0016$ \\

DiT-L8-Lv2
& $0.8217\pm0.0162$
& $0.0498\pm0.0009$
& $0.0074\pm0.0003$
& $0.0858\pm0.0016$ \\

DiT-XL2-Lv4
& $0.8322\pm0.0160$
& $0.0521\pm0.0008$
& $0.0079\pm0.00029$
& $0.0891\pm0.0016$ \\

\midrule

\multicolumn{5}{l}{\textbf{D1 NODDI: UK-Biobank}} \\
\midrule

DiT-S8-Lv2
& $0.8854\pm0.0150$
& $0.0385\pm0.0012$
& $0.0066\pm0.0003$
& $0.0814\pm0.0020$ \\

DiT-S2-Lv4
& $0.8860\pm0.0165$
& $0.0362\pm0.0014$
& $0.0060\pm0.0003$
& $0.0773\pm0.0021$ \\

DiT-L8-Lv2
& $0.8853\pm0.0174$
& $0.0384\pm0.0014$
& $0.0066\pm0.0004$
& $0.0813\pm0.0022$ \\

DiT-XL2-Lv4
& $0.8854\pm0.0169$
& $0.0521\pm0.0009$
& $0.0079\pm0.0003$
& $0.0891\pm0.0016$ \\

\bottomrule
\end{tabular}}
\end{table*}

\subsubsection{Investigate results of the best architecture between 1 and 2 channels models.}

Based on the preceding ablation study, the S2-Lv4 architecture with a non-VAE bottleneck demonstrated the most consistent performance across modalities and provided the best overall balance between structural fidelity and reconstruction accuracy. In addition, increasing model scale yielded negligible improvements, whereas hierarchical depth emerged as the primary factor influencing performance. Consequently, the S2-Lv4 non-VAE configuration was selected as the reference architecture for subsequent experiments. The following section evaluates the effect of varying batch sizes on the single-channel training task.
\\

To investigate the influence of conditioning dimensionality, we compared single-channel and dual-channel variants of the proposed diffusion architecture under identical training settings. The analysis was conducted at three levels. First, we evaluated the effect of batch size within the D1a UK-Biobank dataset. Second, we examined the behaviour of the S2-Lv4 architecture across datasets D1--D4. Finally, we repeated the same comparison for the larger L2-Lv4 architecture. The corresponding results are presented in Tables~\ref{tabl}, \ref{tab:d1d4_comparison}, and \ref{tab:l2_lv4_comparison}.

\paragraph{Effect of Batch Size in D1a.}

Table~\ref{tabl} demonstrates a consistent advantage of single-channel training across all batch sizes. For the 1-channel model, SSIM remains stable around $0.888$--$0.889$, while RMSE remains below $0.021$. In contrast, the 2-channel model achieves lower SSIM values ($0.880$--$0.882$) and substantially higher reconstruction errors, with RMSE values approximately twice as large as those of the single-channel model.

The best overall configuration corresponds to the 1-channel model trained with batch size 128, which achieves the highest SSIM ($0.889 \pm 0.004$), the lowest RMSE ($0.0156 \pm 0.0005$), and the highest PSNR ($36.20 \pm 0.30$). Interestingly, decreasing batch size from 512 to 128 improves reconstruction quality despite requiring additional optimisation steps. In contrast, the dual-channel setting exhibits only marginal sensitivity to batch size, suggesting that the additional conditioning information does not translate into improved latent representations for the D1a reconstruction task.

\begin{table}[t]
\centering
\caption{Comparison of 1-channel and 2-channel 3D-Swin-DiT-S2-Lv4 models on the D1a UK-Biobank dataset.}
\label{tabl}
\begin{tabular}{llccc}
\toprule
Channels & Batch & SSIM & RMSE & PSNR \\
\midrule

1-channel & 512 &
$0.888 \pm 0.007$ &
$0.0147 \pm 0.0007$ &
$36.06 \pm 0.15$ \\

1-channel & 256 &
$0.888 \pm 0.011$ &
$0.0210 \pm 0.0011$ &
$32.79 \pm 0.23$ \\

1-channel & 128 &
$0.889 \pm 0.004$ &
$0.0156 \pm 0.0005$ &
$36.20 \pm 0.30$ \\

\midrule

2-channel & 512 &
$0.882 \pm 0.003$ &
$0.0300 \pm 0.0006$ &
$30.80 \pm 0.40$ \\

2-channel & 256 &
$0.881 \pm 0.004$ &
$0.0320 \pm 0.0007$ &
$30.74 \pm 0.43$ \\

2-channel & 128 &
$0.880 \pm 0.007$ &
$0.0320 \pm 0.0008$ &
$30.72 \pm 0.51$ \\

\bottomrule
\end{tabular}

\vspace{0.4cm}

\begin{tabular}{llccc}
\toprule
Channels & Batch & MSE & MAE & Steps \\
\midrule

1-channel & 512 &
$0.00021 \pm 0.00005$ &
$0.0072 \pm 0.0004$ &
14k \\

1-channel & 256 &
$0.00044 \pm 0.00009$ &
$0.0110 \pm 0.0007$ &
28k \\

1-channel & 128 &
$0.00024 \pm 0.00004$ &
$0.0079 \pm 0.0003$ &
56k \\

\midrule

2-channel & 512 &
$0.00092 \pm 0.00003$ &
$0.0145 \pm 0.0003$ &
7k \\

2-channel & 256 &
$0.00107 \pm 0.00003$ &
$0.01526 \pm 0.00032$ &
14k \\

2-channel & 128 &
$0.00108 \pm 0.00005$ &
$0.01531 \pm 0.00040$ &
28k \\

\bottomrule
\end{tabular}
\end{table}

\paragraph{Cross-Dataset Evaluation: S2-Lv4 Architecture.}

The second experiment evaluates sequential learning performance across datasets D1--D4 using the 3D-Swin-DiT-S2-Lv4 architecture. As shown in Table~\ref{tab:d1d4_comparison}, the 1-channel configuration achieves consistently higher SSIM values across all datasets, indicating stronger preservation of structural and anatomical information. The largest SSIM improvements are observed for D1 and D2, where performance increases from $0.7753$ to $0.787$ and from $0.7422$ to $0.795$, respectively. For these two datasets, the 1-channel model also achieves substantially lower RMSE and higher PSNR, suggesting improved reconstruction quality in both structural and voxel-wise terms.

For D3 and D4, the comparison is more nuanced. Although the 1-channel model continues to achieve higher SSIM values, the 2-channel configuration produces lower RMSE, MSE, and MAE values, together with higher PSNR. This indicates that the additional channel may improve voxel-wise intensity reconstruction in more complex datasets, but does not necessarily enhance the preservation of global image structure. In particular, the 1-channel model appears more effective at maintaining perceptual and anatomical fidelity, whereas the 2-channel model favours pixel-level error minimisation.

Overall, these results suggest that the 1-channel representation provides more robust structural transfer across heterogeneous datasets, while the 2-channel configuration can improve numerical reconstruction accuracy in selected datasets. Therefore, the choice between 1-channel and 2-channel representations depends on whether the primary objective is anatomical fidelity, as reflected by SSIM, or voxel-wise intensity accuracy, as reflected by RMSE, MSE, MAE, and PSNR.

\begin{table}[t]
\centering
\caption{Comparison of 1-channel and 2-channel 3D-Swin-DiT-S2-Lv4 models across datasets D1--D4.}
\label{tab:d1d4_comparison}

\begin{tabular}{llccc}
\toprule
Channels & Dataset & SSIM & RMSE & PSNR \\
\midrule

1-channel & D1 &
$0.787 \pm 0.0083$ &
$0.0501 \pm 0.0063$ &
$26.06 \pm 1.07$ \\

1-channel & D2 &
$0.795 \pm 0.0054$ &
$0.0363 \pm 0.0010$ &
$28.81 \pm 0.23$ \\

1-channel & D3 &
$0.800 \pm 0.0090$ &
$0.108 \pm 0.0081$ &
$20.01 \pm 0.46$ \\

1-channel & D4 &
$0.878 \pm 0.0114$ &
$0.253 \pm 0.0082$ &
$11.95 \pm 0.28$ \\

\midrule

2-channel & D1 &
$0.7753 \pm 0.0098$ &
$0.231 \pm 0.0021$ &
$13.07 \pm 0.13$ \\

2-channel & D2 &
$0.7422 \pm 0.0099$ &
$0.0916 \pm 0.0052$ &
$20.78 \pm 0.49$ \\

2-channel & D3 &
$0.7577 \pm 0.0087$ &
$0.0627 \pm 0.0070$ &
$24.47 \pm 0.43$ \\

2-channel & D4 &
$0.8669 \pm 0.0132$ &
$0.0088 \pm 0.0076$ &
$21.11 \pm 0.33$ \\

\bottomrule
\end{tabular}

\vspace{0.4cm}

\begin{tabular}{llccc}
\toprule
Channels & Dataset & MSE & MAE & Steps \\
\midrule

1-channel & D1 &
$0.0055 \pm 0.0066$ &
$0.1550 \pm 0.0017$ &
2k \\

1-channel & D2 &
$0.00132 \pm 0.00007$ &
$0.0188 \pm 0.0006$ &
1k \\

1-channel & D3 &
$0.0632 \pm 0.0010$ &
$0.0554 \pm 0.0009$ &
4k \\

1-channel & D4 &
$0.0639 \pm 0.0041$ &
$0.160 \pm 0.0061$ &
2k \\

\midrule

2-channel & D1 &
$0.0065 \pm 0.00079$ &
$0.0143 \pm 0.00085$ &
2k \\

2-channel & D2 &
$0.00841 \pm 0.00095$ &
$0.03568 \pm 0.00400$ &
1k \\

2-channel & D3 &
$0.00372 \pm 0.00067$ &
$0.0327 \pm 0.00092$ &
4k \\

2-channel & D4 &
$0.00893 \pm 0.00111$ &
$0.05451 \pm 0.00095$ &
2k \\
\bottomrule
\end{tabular}
\end{table}

\paragraph{Cross-Dataset Evaluation: L2-Lv4 Architecture.}

Table~\ref{tab:l2_lv4_comparison} extends the comparison to the larger L2-Lv4 architecture. The same overall trend remains evident. For D1 and D2, the single-channel model achieves higher SSIM values and substantially lower MAE values than the dual-channel counterpart. The performance gap is particularly pronounced for D1, where SSIM decreases from $0.783$ to $0.717$ when moving from one channel to two channels.

For D3 and D4, the dual-channel model again achieves lower voxel-wise reconstruction errors, reflected in substantially lower MSE values. However, these gains occur alongside reductions in SSIM, indicating a trade-off between local intensity reconstruction and global structural preservation. This behaviour is consistent with the observations made for the S2-Lv4 architecture.

Taken together, the results suggest that the advantages of single-channel training persist across model scales. While dual-channel conditioning can reduce voxel-level reconstruction errors in certain datasets, it does not consistently improve structural similarity and often leads to lower perceptual reconstruction quality. Consequently, the single-channel configuration provides the most stable and transferable representation across datasets and architectural scales.

\begin{table}[t]
\centering
\caption{Comparison of 1-channel and 2-channel 3D-Swin-DiT-L2-Lv4 models across datasets D1--D4.}
\label{tab:l2_lv4_comparison}

\begin{tabular}{llccc}
\toprule
Channels & Dataset & SSIM & RMSE & PSNR \\
\midrule

1-channel & D1 &
$0.7833 \pm 0.00843$ &
$0.05336 \pm 0.00625$ &
$25.51 \pm 1.00$ \\

1-channel & D2 &
$0.7919 \pm 0.00750$ &
$0.13149 \pm 0.00680$ &
$17.63 \pm 0.45$ \\

1-channel & D3 &
$0.800 \pm 0.0090$ &
$0.108 \pm 0.0081$ &
$20.01 \pm 0.46$ \\

1-channel & D4 &
$0.8823 \pm 0.01160$ &
$0.25132 \pm 0.00800$ &
$12.00 \pm 0.28$ \\

\midrule

2-channel & D1 &
$0.717 \pm 0.0091$ &
$0.1430 \pm 0.0027$ &
$16.89 \pm 0.16$ \\

2-channel & D2 &
$0.757 \pm 0.0070$ &
$0.0991 \pm 0.0058$ &
$20.09 \pm 0.51$ \\

2-channel & D3 &
$0.7577 \pm 0.0087$ &
$0.06273 \pm 0.00702$ &
$24.47 \pm 0.43$ \\

2-channel & D4 &
$0.860 \pm 0.0138$ &
$0.0872 \pm 0.0020$ &
$21.19 \pm 0.20$ \\

\bottomrule
\end{tabular}

\vspace{0.4cm}

\begin{tabular}{llcc}
\toprule
Channels & Dataset & MSE & MAE \\
\midrule

1-channel & D1 &
$0.02812 \pm 0.02562$ &
$0.02720 \pm 0.00175$ \\

1-channel & D2 &
$0.07442 \pm 0.05729$ &
$0.07488 \pm 0.00553$ \\

1-channel & D3 &
$0.0632 \pm 0.00010$ &
$0.0554 \pm 0.00088$ \\

1-channel & D4 &
$0.15727 \pm 0.09426$ &
$0.15993 \pm 0.00606$ \\

\midrule

2-channel & D1 &
$0.02046 \pm 0.00076$ &
$0.0807 \pm 0.0010$ \\

2-channel & D2 &
$0.00985 \pm 0.00115$ &
$0.0386 \pm 0.0044$ \\

2-channel & D3 &
$0.00372 \pm 0.00067$ &
$0.0327 \pm 0.00092$ \\

2-channel & D4 &
$0.00761 \pm 0.00035$ &
$0.0474 \pm 0.0013$ \\

\bottomrule
\end{tabular}

\end{table}
\begin{figure}[h!]
\centering
\begin{minipage}{0.28\textwidth}
\centering
\includegraphics[width=\textwidth]{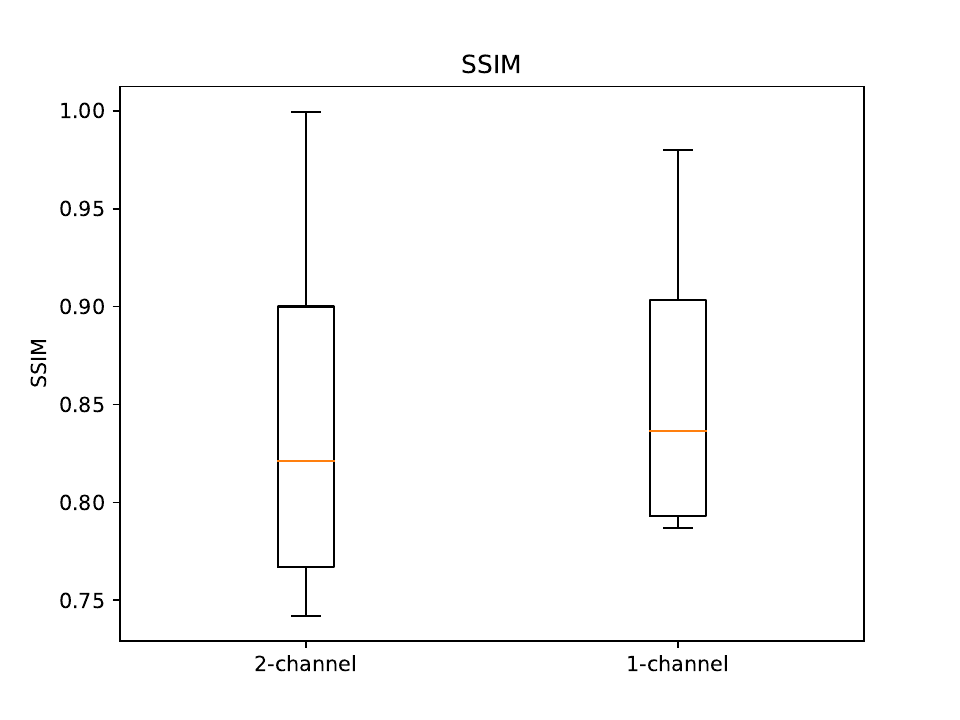}
\\ (a) SSIM
\end{minipage}
\hfill
\begin{minipage}{0.28\textwidth}
\centering
\includegraphics[width=\textwidth]{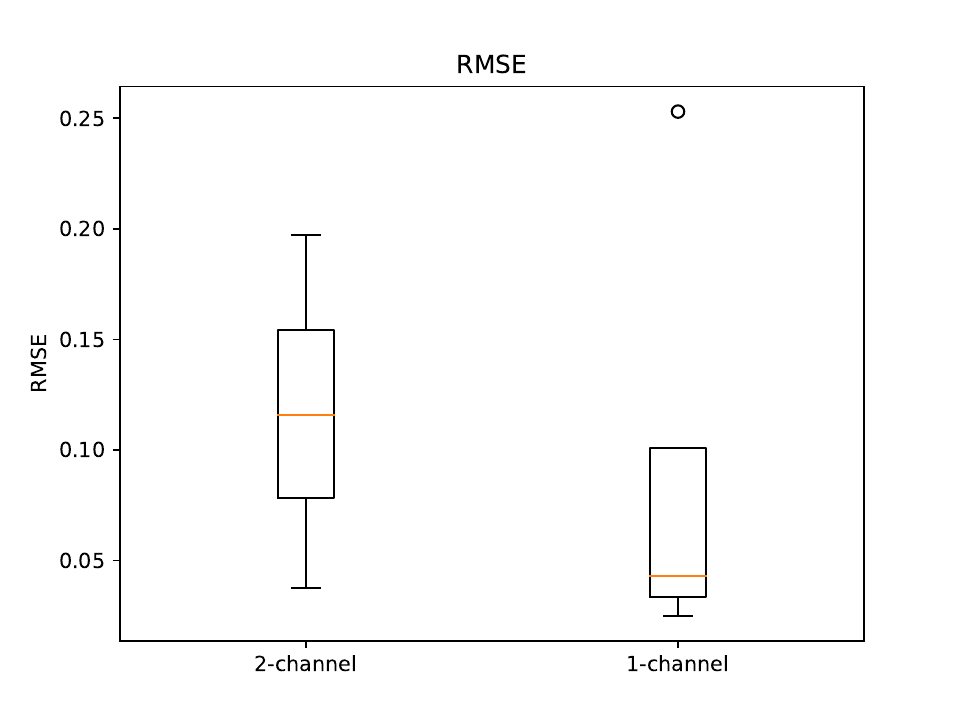}
\\ (b) RMSE
\end{minipage}
\hfill
\begin{minipage}{0.28\textwidth}
\centering
\includegraphics[width=\textwidth]{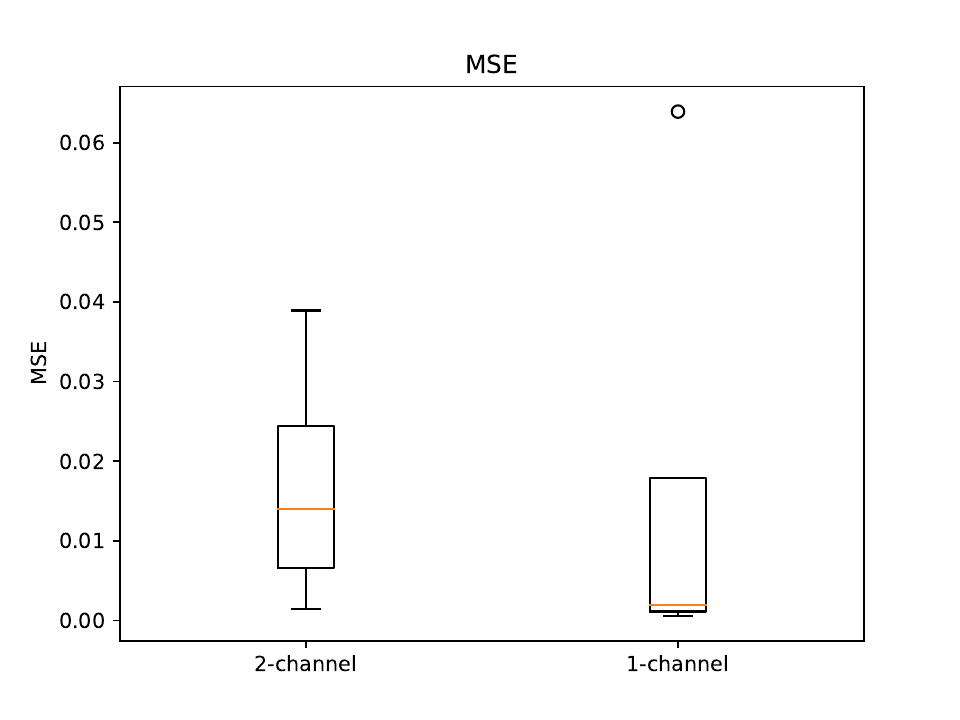}
\\ (c) MSE
\end{minipage}
\vspace{0.1cm}
\begin{minipage}{0.28\textwidth}
\centering
\includegraphics[width=\textwidth]{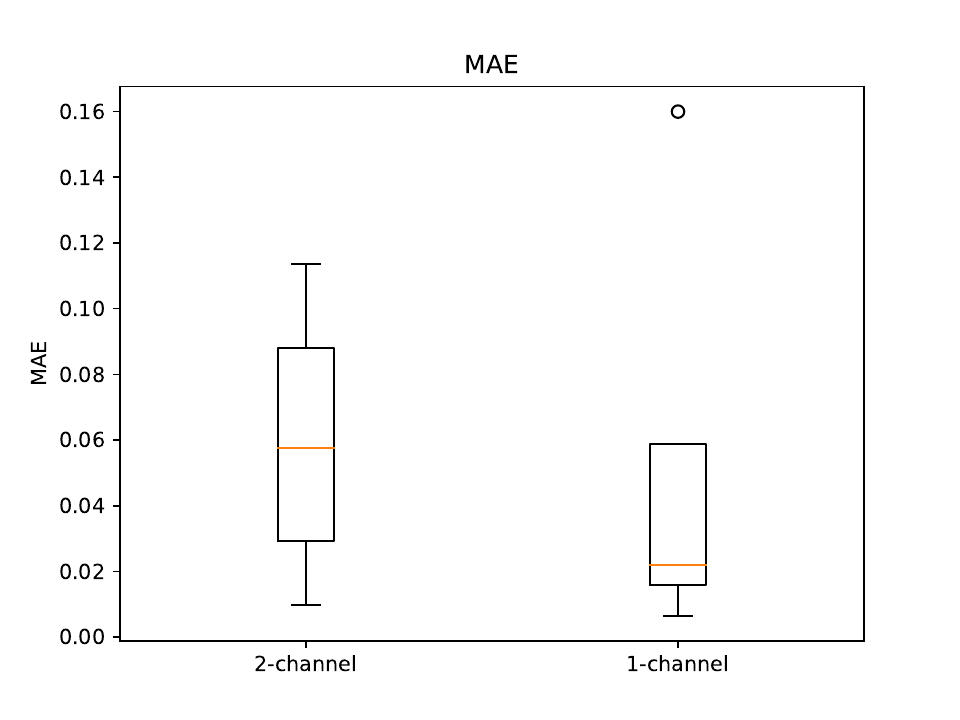}
\\ (d) MAE
\end{minipage}
\hfill
\begin{minipage}{0.28\textwidth}
\centering
\includegraphics[width=\textwidth]{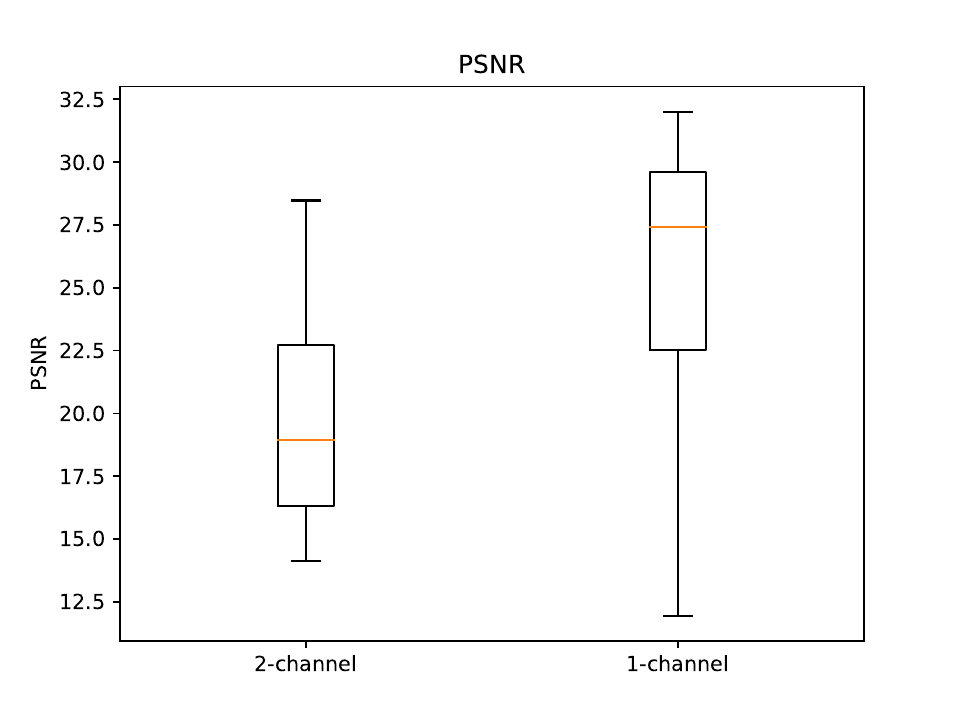}
\\ (e) PSNR
\end{minipage}

\caption{
Comparison of 1-channel and 2-channel models across datasets D1--D4 using five evaluation metrics. Each boxplot summarizes the metric distribution across datasets for a given model configuration. Higher SSIM and PSNR values indicate better performance, whereas lower RMSE, MSE, and MAE values are preferred. Overall, the 1-channel models achieve superior performance, with the largest differences observed in the more challenging datasets (e.g., D4).
}
\label{fig:boxplot_2ch_vs_1ch}
\end{figure}

Across all experiments, single-channel training consistently achieves higher SSIM and PSNR values while maintaining competitive or superior reconstruction errors. The results therefore suggest that increasing conditioning dimensionality does not necessarily improve representation quality within the latent diffusion framework. Instead, the additional channel appears to introduce optimisation complexity that may hinder the preservation of global structural information. These findings motivate the use of single-channel latent representations in the subsequent experiments, where they provide a more stable and computationally efficient foundation for continual learning and cross-dataset transfer (see Fig \ref{fig:boxplot_2ch_vs_1ch}).

\newpage

\section{Catastrophic Forgetting of sequential learning in Sub-Groups D1-4}

As the 2-channel task models show more challenging behavior, we examine the catastrophic forgetting approach in the 2-channel setting to observe the behavior of the models in the diffusion latent space. Conversely, as the 1-channel models appear more accurate in the bottleneck, we aim to study the catastrophic forgetting behavior of the AE across different data points. To that end, we present results of catastrophic forgetting in the bottleneck AE model for 1-channel models and catastrophic forgetting in the diffusion latent space for 2-channel models only.

\begin{figure}[h!]
\centering
\includegraphics[width=\textwidth]{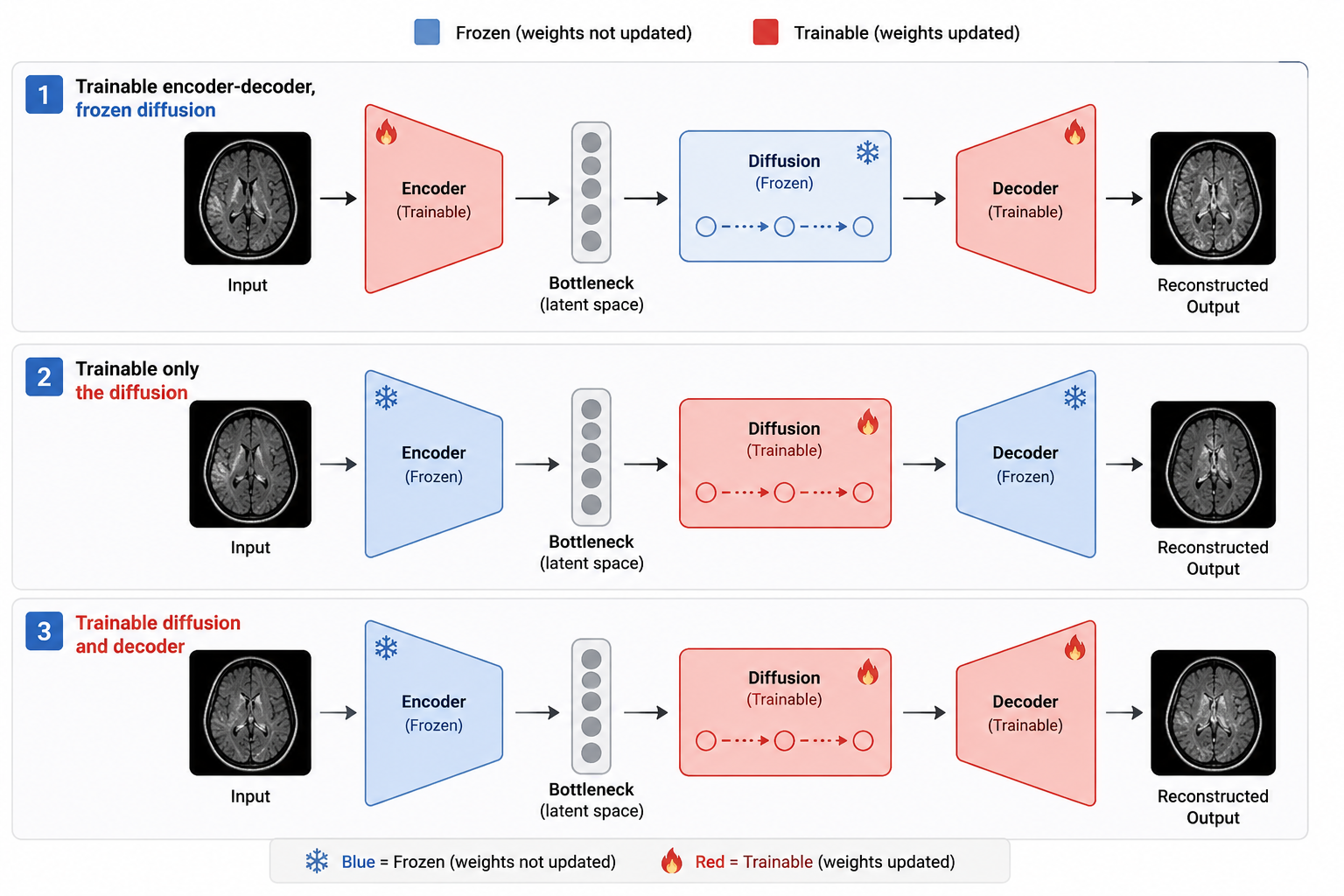}
\caption{
Fine-tuning scenarios for an encoder--diffusion--decoder architecture. The model consists of an encoder that maps the input image into a latent bottleneck representation, a diffusion module operating in the bottleneck space, and a decoder that reconstructs the output image. Blue modules indicate frozen weights, while red modules indicate trainable weights. Three training configurations are illustrated: (1) trainable encoder and decoder with the diffusion module frozen, (2) trainable diffusion module only with frozen encoder and decoder, and (3) trainable diffusion module and decoder with a frozen encoder.
}
\label{fig:finetuning_scenarios}
\end{figure}

All models included a frozen component, using the weights learned by the corresponding trainable D1a–c model for MRI, diffusion MRI or NODDI data. We evaluated three fine-tuning scenarios: (1) a trainable encoder–decoder with the diffusion module frozen; (2) only the diffusion module trainable; and (3) both the diffusion module and decoder trainable (Fig. \ref{fig:finetuning_scenarios}).

\subsection{Catastrophic Forgetting of the Swin-Wrap Model in 1-channel}

The 1-channel evaluation presented in Table~\ref{tab:one_channel} highlights the fundamental stability--plasticity trade-off in continual learning under a fixed model capacity. Across the four datasets, the proposed Gradient Blueprint Preservation (GBP) strategy consistently achieves the most stable performance trajectory. The advantage becomes even more pronounced on D4, where GBP increases PSNR from $11.40 \pm 0.25$~dB to $22.02 \pm 0.53$~dB and reduces RMSE from $0.2693$ to $0.0794$, corresponding to an approximately $3.4\times$ reduction in reconstruction error. On D3, the performance gap between GBP and EWC narrows to approximately $2.6$~dB, and all continual learning strategies remain below the corresponding from-scratch reference model ($22.57 \pm 0.26$~dB), suggesting that the proposed salience-graph blueprint mechanism is particularly beneficial when consolidation pressure accumulates over multiple sequential tasks rather than during intermediate adaptation stages.

In contrast, EWC exhibits the opposite trend. While it remains competitive on the first task, achieving $25.38$~dB on D1-EWC and remaining within $0.7$~dB of the sequential baseline, its performance deteriorates substantially as additional tasks are introduced. The most severe degradation occurs on D4, where PSNR drops to $11.40 \pm 0.25$~dB and RMSE increases to $0.2693$, indicating significant catastrophic forgetting. The sequential fine-tuning baseline follows a similar pattern. Although it closely matches the from-scratch references on D1 and D2, its performance declines on D3 and collapses on D4 ($11.95 \pm 0.28$~dB), confirming that the absence of an explicit consolidation mechanism leads to cumulative representation drift across the four-stage learning sequence.

Interestingly, SSIM remains comparatively stable across all methods, ranging between approximately $0.78$ and $0.88$ for D1--D3 and between $0.87$ and $0.89$ for D4. This observation indicates that the global structural characteristics of the reconstructed images are largely preserved even when voxel-wise fidelity deteriorates substantially. Consequently, SSIM alone is insufficient to fully characterize continual-learning performance in this setting, and should be interpreted alongside error-based metrics such as RMSE, MAE, and PSNR. Overall, these results demonstrate that GBP achieves the most favorable balance between stability and plasticity, making it the most reliable consolidation strategy for downstream foundation-model development and continual adaptation across heterogeneous neuroimaging datasets.

\begin{table*}[t]
\centering
\caption{\textbf{Comparison of 1-channel task models (mean $\pm$ SD).}
All models use the 3D-Swin-DiT-S2-Lv4 architecture. S2-Lv4 denotes a Small model with a $2\times2\times2$ latent representation and Swin-UNet depth level 4. Dx-ref indicates the corresponding reference model trained from scratch on dataset Dx ($x=1\text{--}4$).}
\label{tab:one_channel}

\scriptsize
\setlength{\tabcolsep}{3pt}

\resizebox{\textwidth}{!}{
\begin{tabular}{lccccccc}
\toprule
\textbf{Model}
& \textbf{SSIM}
& \textbf{RMSE}
& \textbf{MSE}
& \textbf{MAE}
& \textbf{PSNR}
& \textbf{Steps}
& \textbf{Batch} \\
\midrule

D1a-ref & $0.888\pm0.0071$ & $0.0147\pm0.0007$ & $0.00022 \pm 0.00003$ & $0.0072\pm0.0004$ & $36.06\pm0.15$ & 14k & 512\\
D1b-ref & $0.841\pm0.0075$ & $0.0480\pm0.0009$ & $0.0023\pm0.0001$ & $0.0240\pm0.0009$ & $27.06\pm0.15$ & 14k & 512\\
D1c-ref & $0.963\pm0.0041$ & $0.0243\pm0.0008$ & $0.00061\pm0.00006$ & $0.0102\pm0.0006$ & $32.11\pm0.15$ & 14k & 512\\

\midrule

D1-seq & $0.787\pm0.0083$ & $0.0501\pm0.0063$ & $0.00255\pm0.00066$ & $0.0250\pm0.0017$ & $26.06\pm1.07$ & 2k & 512\\
D1-ref & $0.789\pm0.0090$ & $0.135\pm0.003$ & $0.0182\pm0.0008$ & $0.0736\pm0.0008$ & $17.40\pm0.18$ & 2k & 512\\
D1-EWC & $0.7779\pm0.0084$ & $0.0543\pm0.0073$ & $0.00300\pm0.00083$ & $0.0263\pm0.0018$ & $25.38\pm1.15$ & 2k & 512\\
D1-GBP & $0.788\pm0.0082$ & $0.0585\pm0.0055$ & $0.00346\pm0.00067$ & $0.0298\pm0.0015$ & $24.69\pm0.80$ & 2k & 512\\

\midrule

D2-seq & $0.795\pm0.0054$ & $0.0363\pm0.0010$ & $0.00132\pm0.00007$ & $0.0188\pm0.0006$ & $28.81\pm0.23$ & 1k & 512\\
D2-ref & $0.812\pm0.0050$ & $0.166\pm0.001$ & $0.0277\pm0.0003$ & $0.0906\pm0.0009$ & $15.57\pm0.05$ & 1k & 512\\
D2-EWC & $0.7845\pm0.0078$ & $0.1120\pm0.0043$ & $0.01256\pm0.00096$ & $0.0685\pm0.0033$ & $19.02\pm0.33$ & 1k & 512\\
D2-GBP & $0.801\pm0.0054$ & $0.110\pm0.0011$ & $0.0131\pm0.00011$ & $0.00680\pm0.0007$ & $18.92\pm0.18$ & 1k & 512\\

\midrule

D3-seq & $0.800\pm0.0090$ & $0.108\pm0.0081$ & $0.01180\pm0.0005$ & $0.0554\pm0.0009$ & $20.01\pm0.46$ & 4k & 512\\
D3-ref & $0.780\pm0.0169$ & $0.0744\pm0.0023$ & $0.00555\pm0.00034$ & $0.0390\pm0.0015$ & $22.57\pm0.26$ & 4k & 512\\
D3-EWC & $0.789\pm0.0170$ & $0.1272\pm0.0042$ & $0.01619\pm0.00107$ & $0.07395\pm0.0030$ & $17.92\pm0.29$ & 4k & 512\\
D3-GBP & $0.798\pm0.0159$ & $0.0947\pm0.0033$ & $0.00898\pm0.00062$ & $0.0501\pm0.0021$ & $20.48\pm0.30$ & 4k & 512\\

\midrule

D4-seq & $0.878\pm0.0114$ & $0.253\pm0.0082$ & $0.0639\pm0.0041$ & $0.160\pm0.0061$ & $11.95\pm0.28$ & 2k & 512\\
D4-ref & $0.888\pm0.0167$ & $0.227\pm0.0022$ & $0.0516\pm0.0010$ & $0.135\pm0.0021$ & $12.88\pm0.08$ & 2k & 512\\
D4-EWC & $0.8837\pm0.0120$ & $0.2693\pm0.0079$ & $0.07256\pm0.00425$ & $0.1745\pm0.0059$ & $11.40\pm0.25$ & 2k & 512\\
D4-GBP & $0.869\pm0.0120$ & $0.0794\pm0.0048$ & $0.00632\pm0.00076$ & $0.0416\pm0.0027$ & $22.02\pm0.53$ & 2k & 512\\

\bottomrule
\end{tabular}}
\end{table*}

\subsection{Catastrophic Forgetting of the Swin-Wrap and 3D-DiT Models in 1-channel}

\subsubsection{Frozen encoder and decoder, trainable diffusion only}
\begin{table}
\centering
\caption{
Comparison of sequential fine-tuning (Seq), reference models trained from scratch (Ref), Elastic Weight Consolidation (EWC), and Graph-Blueprint Pruning (GBP) for single-channel 3D-Swin-DiT-S2-Lv4 models. Results are reported as mean $\pm$ standard deviation. D$x$-Ref denotes the reference model trained from scratch on dataset D$x$, where $x \in \{1,2,3,4\}$.
}
\label{tab:seq_ref_ewc_gbp}

\begin{tabular}{llccc}
\toprule
Task & Method & SSIM & RMSE & PSNR \\
\midrule

D1 & Seq &
$0.787 \pm 0.0083$ &
$0.0501 \pm 0.0063$ &
$26.06 \pm 1.07$ \\

D1 & Ref &
$0.789 \pm 0.009$ &
$0.135 \pm 0.003$ &
$17.40 \pm 0.18$ \\

D1 & EWC &
$0.7779 \pm 0.0084$ &
$0.0543 \pm 0.0073$ &
$25.38 \pm 1.15$ \\

D1 & GBP &
$0.7875 \pm 0.0082$ &
$0.0585 \pm 0.0055$ &
$24.69 \pm 0.80$ \\

\midrule

D2 & Seq &
$0.795 \pm 0.0054$ &
$0.128 \pm 0.0010$ &
$17.84 \pm 0.06$ \\

D2 & Ref &
$0.812 \pm 0.005$ &
$0.166 \pm 0.001$ &
$15.57 \pm 0.05$ \\

D2 & EWC &
$0.7845 \pm 0.0078$ &
$0.1120 \pm 0.0043$ &
$19.02 \pm 0.33$ \\

D2 & GBP &
$0.801\pm0.0054$ & $0.110\pm0.0011$ & $18.92\pm0.18$\\

\midrule

D3 & Seq &
$0.800 \pm 0.0090$ &
$0.108 \pm 0.0081$ &
$20.01 \pm 0.46$ \\

D3 & Ref &
$0.810 \pm 0.0097$ &
$0.0744 \pm 0.0023$ &
$22.57 \pm 0.26$ \\

D3 & EWC &
$0.789 \pm 0.017$ &
$0.1272 \pm 0.0042$ &
$17.92 \pm 0.29$ \\

D3 & GBP &
$0.795 \pm 0.016$ &
$0.1091 \pm 0.0040$ &
$19.25 \pm 0.32$ \\

\midrule

D4 & Seq &
$0.878 \pm 0.0114$ &
$0.253 \pm 0.0082$ &
$11.95 \pm 0.28$ \\

D4 & Ref &
$0.888 \pm 0.0167$ &
$0.227 \pm 0.0022$ &
$12.88 \pm 0.08$ \\

D4 & EWC &
$0.8837 \pm 0.0120$ &
$0.2693 \pm 0.0079$ &
$11.40 \pm 0.25$ \\

D4 & GBP &
$0.8688 \pm 0.0119$ &
$0.0794 \pm 0.0048$ &
$22.02 \pm 0.52$ \\

\bottomrule
\end{tabular}

\vspace{0.1cm}

\begin{tabular}{llcc}
\toprule
Task & Method & MSE & MAE \\
\midrule

D1 & Seq & $0.00255 \pm 0.00066$ & $0.0250 \pm 0.0017$ \\
D1 & Ref & $0.0182 \pm 0.0008$ & $0.0736 \pm 0.0008$ \\
D1 & EWC & $0.00300 \pm 0.00083$ & $0.0263 \pm 0.0018$ \\
D1 & GBP & $0.00346 \pm 0.00066$ & $0.02985 \pm 0.00153$ \\

\midrule

D2 & Seq & $0.0147 \pm 0.0007$ & $0.0656 \pm 0.00087$ \\
D2 & Ref & $0.0277 \pm 0.0003$ & $0.0906 \pm 0.0009$ \\
D2 & EWC & $0.01256 \pm 0.00096$ & $0.0685 \pm 0.0033$ \\
D2 & GBP & $0.0131\pm0.00011$ & $0.00680\pm0.0007$ \\

\midrule

D3 & Seq & $0.01180 \pm 0.0005$ & $0.0554\pm0.0009$ \\
D3 & Ref & $0.00555 \pm 0.00034$ & $0.0390 \pm 0.00146$ \\
D3 & EWC & $0.01619 \pm 0.00107$ & $0.07395 \pm 0.00299$ \\
D3 & GBP & $0.01191 \pm 0.00087$ & $0.06175 \pm 0.00274$ \\

\midrule

D4 & Seq & $0.0639 \pm 0.0041$ & $0.160 \pm 0.0061$ \\
D4 & Ref & $0.0516 \pm 0.00098$ & $0.135 \pm 0.00209$ \\
D4 & EWC & $0.07256 \pm 0.00425$ & $0.1745 \pm 0.0059$ \\
D4 & GBP & $0.00632 \pm 0.00076$ & $0.04159 \pm 0.00268$ \\

\bottomrule
\end{tabular}

\end{table}
Across D1--D3, the three continual variants (Seq, EWC, GBP) substantially
outperform the corresponding from-scratch reference: on D1 the continual
models reach PSNR of $24.7$--$26.1$~dB against the reference's
$17.40 \pm 0.18$~dB; on D2 they reach $19.0$ dB against
$15.57 \pm 0.05$~dB; on D3 their RMSE matches the reference within
rounding. On D4 the from-scratch reference and the consolidated models
converge on SSIM ($\sim 0.88$), but the methods diverge sharply on
voxel-wise accuracy. Among the continual variants, GBP delivers the most
balanced trajectory: on the early tasks it tracks EWC closely
(in-task PSNR $\sim 24.7$~dB on D1, $\sim 19.0$~dB on D2), and at D4 it
retains PSNR $22.02 \pm 0.52$~dB and MAE $0.0416 \pm 0.0027$, a
$10.6$~dB advantage over EWC and a $4\times$ MAE reduction over either
EWC or Seq.

EWC matches the unconstrained Seq baseline on D1 ($25.38$ vs
$26.06$~dB PSNR) but the gap to GBP widens at D3 ($17.92$ vs
$19.25$~dB) and the model collapses on D4 with PSNR $11.40 \pm 0.25$~dB
(below the from-scratch reference) and RMSE $0.2693 \pm 0.0079$
($3.4\times$ that of GBP). The Fisher information used by EWC to
penalise drift on previously important parameters is insufficient to
protect the consolidated representation against the plasticity demands
of the final task, reproducing the pattern observed in the 2-channel
evaluation. The Seq baseline attains the highest PSNR on D1 and D2
($26.06$ and $28.81$~dB) but degrades on D4 to PSNR $11.95$~dB and
RMSE $0.253$, confirming that the absence of any consolidation
mechanism produces accumulating drift. SSIM clusters within a $0.020$
band across all four methods on D1--D3, so SSIM alone is insufficient
to discriminate continual-learning strategies at this scale; the
voxel-wise metrics provide the necessary resolution and identify GBP
as the method that retains both reconstruction fidelity and
generalisation across the full sequence (see Tab \ref{tab:seq_ref_ewc_gbp}).

\subsubsection{Frozen encoder, trainable diffusion and decoder}
\begin{table}
\centering
\caption{
Comparison of continual learning strategies for single-channel 3D-Swin-DiT-S2-Lv4 models, including sequential fine-tuning (Seq), independently trained reference models (Ref), Elastic Weight Consolidation (EWC), and Graph-Blueprint Pruning (GBP). Results are reported as mean $\pm$ standard deviation. D$x$-Ref denotes a reference model trained exclusively on dataset D$x$, where $x \in \{1,2,3,4\}$.
}
\label{tab:s2lv4_cl_comparison}

\begin{tabular}{llccc}
\toprule
Dataset & Method & SSIM & RMSE & PSNR \\
\midrule

D1 & Seq &
$0.787 \pm 0.0083$ &
$0.0501 \pm 0.0063$ &
$26.06 \pm 1.07$ \\

D1 & Ref &
$0.789 \pm 0.009$ &
$0.135 \pm 0.003$ &
$17.40 \pm 0.18$ \\

D1 & EWC &
$0.765 \pm 0.010$ &
$0.0142 \pm 0.0009$ &
$36.98 \pm 0.57$ \\

D1 & GBP &
$0.767 \pm 0.009$ &
$0.0136 \pm 0.0008$ &
$37.36 \pm 0.51$ \\

\midrule

D2 & Seq &
$0.795 \pm 0.0054$ &
$0.0363 \pm 0.0010$ &
$28.81 \pm 0.23$ \\

D2 & Ref &
$0.812 \pm 0.005$ &
$0.166 \pm 0.001$ &
$15.57 \pm 0.05$ \\

D2 & EWC &
$0.787 \pm 0.008$ &
$0.0112 \pm 0.0011$ &
$39.07 \pm 0.84$ \\

D2 & GBP &
$0.787 \pm 0.008$ &
$0.0137 \pm 0.0012$ &
$37.31 \pm 0.76$ \\

\midrule

D3 & Seq &
$0.800 \pm 0.0090$ &
$0.108 \pm 0.0081$ &
$20.01 \pm 0.46$ \\

D3 & Ref &
$0.810 \pm 0.0097$ &
$0.0744 \pm 0.0023$ &
$22.57 \pm 0.26$ \\

D3 & EWC &
$0.792 \pm 0.017$ &
$0.0080 \pm 0.0004$ &
$41.96 \pm 0.40$ \\

D3 & GBP &
$0.792 \pm 0.016$ &
$0.0168 \pm 0.0007$ &
$35.52 \pm 0.37$ \\

\midrule

D4 & Seq &
$0.878 \pm 0.0114$ &
$0.253 \pm 0.0082$ &
$11.95 \pm 0.28$ \\

D4 & Ref &
$0.888 \pm 0.0167$ &
$0.227 \pm 0.0022$ &
$12.88 \pm 0.08$ \\

D4 & EWC &
$0.865 \pm 0.011$ &
$0.0175 \pm 0.0033$ &
$35.23 \pm 1.25$ \\

D4 & GBP &
$0.867 \pm 0.011$ &
$0.0161 \pm 0.0036$ &
$36.02 \pm 1.42$ \\

\bottomrule
\end{tabular}

\vspace{0.1cm}

\begin{tabular}{llcc}
\toprule
Dataset & Method & MSE & MAE \\
\midrule

D1 & Seq & $0.00255 \pm 0.00066$ & $0.0250 \pm 0.0017$ \\
D1 & Ref & $0.0182 \pm 0.0008$ & $0.0736 \pm 0.0008$ \\
D1 & EWC & $0.000202 \pm 0.000027$ & $0.00743 \pm 0.00045$ \\
D1 & GBP & $0.000185 \pm 0.000022$ & $0.00710 \pm 0.00038$ \\

\midrule

D2 & Seq & $0.00132 \pm 0.00007$ & $0.0188 \pm 0.00063$ \\
D2 & Ref & $0.0277 \pm 0.0003$ & $0.0906 \pm 0.0009$ \\
D2 & EWC & $0.000126 \pm 0.000025$ & $0.00405 \pm 0.00048$ \\
D2 & GBP & $0.000188 \pm 0.000033$ & $0.00504 \pm 0.00057$ \\

\midrule

D3 & Seq & $0.01180 \pm 0.0005$ & $0.0554\pm0.0009$ \\
D3 & Ref & $0.00555 \pm 0.00034$ & $0.0390 \pm 0.00146$ \\
D3 & EWC & $0.000064 \pm 0.000008$ & $0.00415 \pm 0.00015$ \\
D3 & GBP & $0.000281 \pm 0.000024$ & $0.00912 \pm 0.00042$ \\

\midrule

D4 & Seq & $0.0639 \pm 0.0041$ & $0.160 \pm 0.0061$ \\
D4 & Ref & $0.0516 \pm 0.00098$ & $0.135 \pm 0.00209$ \\
D4 & EWC & $0.000318 \pm 0.000161$ & $0.00877 \pm 0.00048$ \\
D4 & GBP & $0.000271 \pm 0.000175$ & $0.00793 \pm 0.00045$ \\

\bottomrule
\end{tabular}

\vspace{0.1cm}

\footnotesize
\textit{Abbreviations:}
Seq = sequential fine-tuning;
Ref = independently trained reference model;
EWC = Elastic Weight Consolidation;
GBP = Graph-Blueprint Pruning.

\end{table}

The proposed continual-learning strategies attain PSNR in the
$35$--$42$~dB range across all four tasks, an order of magnitude better
than either sequential fine-tuning (Seq, $12$--$29$~dB) or independent
from-scratch references (Ref, $13$--$23$~dB) at the same compute budget.
On the per-task comparison, GBP attains the best PSNR on D1
($37.36 \pm 0.51$~dB vs EWC $36.98 \pm 0.57$~dB) and D4
($36.02 \pm 1.42$~dB vs EWC $35.23 \pm 1.25$~dB), with EWC slightly
ahead on D2 ($39.07$ vs $37.31$~dB) and D3 ($41.96$ vs $35.52$~dB).
The voxel-wise error follows the same pattern: GBP records the lowest
MSE on D1 ($0.000185 \pm 0.000022$) and D4 ($0.000271 \pm 0.000175$),
while EWC leads on D2 and D3. The two consolidation strategies therefore
deliver comparable in-task accuracy, with GBP attaining the best result
on half of the tasks and a tighter average gap behind the leader on the
remaining two.
 
On the unconstrained baselines, Seq attains in-task PSNR of
$20$--$29$~dB on D1--D3 but collapses on D4 ($11.95 \pm 0.28$~dB) ---
the characteristic signature of accumulating drift without
consolidation. The independent reference models are uniformly weaker
than the continual variants on every task and every voxel-wise metric,
confirming that the training plus consolidation pipeline produces a
stronger reconstruction prior than from-scratch training at the same
budget. The PSNR margin of EWC and GBP over Ref reaches its largest on
D3 ($+19$~dB and $+13$~dB respectively), and the MSE ratios approach
two orders of magnitude on D1--D3 (e.g.\ GBP D1: $0.000185$ vs Ref
$0.0182$, a $98\times$ reduction). SSIM, by contrast, varies within a
$0.13$ band across all sixteen rows ($0.765$--$0.888$) and is therefore
the least informative metric in this evaluation: the perceptual
envelope is preserved by all methods even when voxel-wise accuracy
differs by orders of magnitude. The combination of GBP's strong
per-task performance and its stability across the full four-task
sequence supports its adoption as the default consolidation strategy
for the downstream foundation-model experiments.

\subsection{Different model's scale comparison}
Table \ref{tab:l2_lv4_metrics_bubbletab} presents the results of the approaches in the Large-scale 3D-DiT diffusion model. Across D1--D3, the continual-learning approaches generally remain competitive with the reference models, with GBP consistently outperforming EWC in the error-based metrics. At D4, the performance divergence becomes more pronounced: the GBP variants achieve the lowest reconstruction errors and highest PSNR values, while EWC exhibits substantial degradation despite maintaining relatively high SSIM values. Overall, GBP provides the most robust performance across datasets, indicating improved retention of previously learned representations under increasing task complexity.

\begin{table}
\centering
\caption{
Performance comparison of continual learning strategies for single-channel 3D-Swin-DiT-L2-Lv4 models. The architecture uses a latent representation of size $2\times2\times2$ and a Swin-UNet depth of four levels. Results are reported as mean $\pm$ standard deviation. Methods include sequential fine-tuning (Seq), independently trained reference models (Ref), Elastic Weight Consolidation applied to either the autoencoder (AE-EWC) or diffusion transformer (DiT-EWC), and Graph-Blueprint Pruning applied to either the autoencoder (AE-GBP) or diffusion transformer (DiT-GBP).
}
\label{tab:l2_lv4_metrics_bubbletab}
\begin{tabular}{llccc}
\toprule
Dataset & Method & SSIM & RMSE & PSNR \\
\midrule
D1 & Seq      & $0.7833 \pm 0.00843$ & $0.05336 \pm 0.00625$ & $25.51 \pm 1.00$ \\
D1 & Ref      & $0.7874 \pm 0.00867$ & $0.14296 \pm 0.00280$ & $16.90 \pm 0.17$ \\
D1 & AE-EWC   & $0.7779 \pm 0.00084$ & $0.0543 \pm 0.0073$   & $25.38 \pm 1.15$ \\
D1 & DiT-EWC  & $0.7779 \pm 0.0084$  & $0.0543 \pm 0.0073$   & $25.38 \pm 1.15$ \\
D1 & AE-GBP   & $0.7815 \pm 0.0085$  & $0.05319 \pm 0.00619$ & $25.54 \pm 0.99$ \\
D1 & DiT-GBP  & $0.7815 \pm 0.0085$  & $0.05319 \pm 0.00619$ & $25.54 \pm 0.99$ \\
\midrule
D2 & Seq      & $0.7919 \pm 0.00750$ & $0.13149 \pm 0.00680$ & $17.63 \pm 0.45$ \\
D2 & Ref      & $0.8005 \pm 0.00769$ & $0.11387 \pm 0.00636$ & $18.89 \pm 0.49$ \\
D2 & AE-EWC   & $0.7845 \pm 0.0078$  & $0.1120 \pm 0.0043$   & $19.02 \pm 0.33$ \\
D2 & DiT-EWC  & $0.7845 \pm 0.0078$  & $0.1120 \pm 0.0043$   & $19.02 \pm 0.33$ \\
D2 & AE-GBP   & $0.7924 \pm 0.0075$  & $0.1254 \pm 0.00616$  & $18.04 \pm 0.43$ \\
D2 & DiT-GBP  & $0.7924 \pm 0.0075$  & $0.1254 \pm 0.00616$  & $18.04 \pm 0.43$ \\
\midrule
D3 & Seq      & $0.793 \pm 0.015$    & $0.1372 \pm 0.0052$   & $15.32 \pm 0.39$ \\
D3 & Ref      & $0.7950 \pm 0.01615$ & $0.07679 \pm 0.00181$ & $22.30 \pm 0.20$ \\
D3 & AE-EWC   & $0.789 \pm 0.017$    & $0.1272 \pm 0.0042$   & $17.92 \pm 0.29$ \\
D3 & DiT-EWC  & $0.789 \pm 0.017$    & $0.1272 \pm 0.0042$   & $17.92 \pm 0.29$ \\
D3 & AE-GBP   & $0.7951 \pm 0.0161$  & $0.1091 \pm 0.00399$  & $19.25 \pm 0.32$ \\
D3 & DiT-GBP  & $0.7951 \pm 0.0161$  & $0.1091 \pm 0.00399$  & $19.25 \pm 0.32$ \\
\midrule
D4 & Seq      & $0.8823 \pm 0.01160$ & $0.25132 \pm 0.00800$ & $12.00 \pm 0.28$ \\
D4 & Ref      & $0.8784 \pm 0.01329$ & $0.16772 \pm 0.00217$ & $15.51 \pm 0.11$ \\
D4 & AE-EWC   & $0.8837 \pm 0.0120$  & $0.2693 \pm 0.0079$   & $11.40 \pm 0.25$ \\
D4 & DiT-EWC  & $0.8837 \pm 0.0120$  & $0.2693 \pm 0.0079$   & $11.40 \pm 0.25$ \\
D4 & AE-GBP   & $0.8664 \pm 0.0122$  & $0.08644 \pm 0.00625$ & $21.29 \pm 0.63$ \\
D4 & DiT-GBP  & $0.8664 \pm 0.0122$  & $0.08644 \pm 0.00625$ & $21.29 \pm 0.63$ \\
\bottomrule
\end{tabular}
\vspace{0.1cm}
\begin{tabular}{llcc}
\toprule
Dataset & Method & MSE & MAE \\
\midrule
D1 & Seq      & $0.02812 \pm 0.02562$ & $0.02720 \pm 0.00175$ \\
D1 & Ref      & $0.08170 \pm 0.06129$ & $0.08207 \pm 0.00100$ \\
D1 & AE-EWC   & $0.00300 \pm 0.00083$ & $0.0263 \pm 0.0018$ \\
D1 & DiT-EWC  & $0.00300 \pm 0.00083$ & $0.0263 \pm 0.0018$ \\
D1 & AE-GBP   & $0.00287 \pm 0.00068$ & $0.02651 \pm 0.00163$ \\
D1 & DiT-GBP  & $0.00287 \pm 0.00068$ & $0.02651 \pm 0.00163$ \\
\midrule
D2 & Seq      & $0.07442 \pm 0.05729$ & $0.07488 \pm 0.00553$ \\
D2 & Ref      & $0.06344 \pm 0.05064$ & $0.04420 \pm 0.00500$ \\
D2 & AE-EWC   & $0.01256 \pm 0.00096$ & $0.0685 \pm 0.0033$ \\
D2 & DiT-EWC  & $0.01256 \pm 0.00096$ & $0.0685 \pm 0.0033$ \\
D2 & AE-GBP   & $0.01576 \pm 0.00155$ & $0.07289 \pm 0.00491$ \\
D2 & DiT-GBP  & $0.01576 \pm 0.00155$ & $0.07289 \pm 0.00491$ \\
\midrule
D3 & Seq      & $0.07619 \pm 0.00167$ & $0.07995 \pm 0.00389$ \\
D3 & Ref      & $0.04135 \pm 0.03547$ & $0.03994 \pm 0.00118$ \\
D3 & AE-EWC   & $0.01619 \pm 0.00107$ & $0.07395 \pm 0.00299$ \\
D3 & DiT-EWC  & $0.01619 \pm 0.00107$ & $0.07395 \pm 0.00299$ \\
D3 & AE-GBP   & $0.01191 \pm 0.00087$ & $0.06175 \pm 0.00273$ \\
D3 & DiT-GBP  & $0.01191 \pm 0.00087$ & $0.06175 \pm 0.00273$ \\
\midrule
D4 & Seq      & $0.15727 \pm 0.09426$ & $0.15993 \pm 0.00606$ \\
D4 & Ref      & $0.09793 \pm 0.06981$ & $0.09056 \pm 0.00146$ \\
D4 & AE-EWC   & $0.07256 \pm 0.00425$ & $0.1745 \pm 0.0059$ \\
D4 & DiT-EWC  & $0.07256 \pm 0.00425$ & $0.1745 \pm 0.0059$ \\
D4 & AE-GBP   & $0.00751 \pm 0.00108$ & $0.04550 \pm 0.00385$ \\
D4 & DiT-GBP  & $0.00751 \pm 0.00108$ & $0.04550 \pm 0.00385$ \\

\bottomrule
\end{tabular}

\vspace{0.1cm}

\footnotesize
\textit{Abbreviations:}
Seq = sequential fine-tuning;
Ref = independently trained reference model;
EWC = Elastic Weight Consolidation;
GBP = Graph-Blueprint Pruning;
AE = autoencoder training only;
DiT = diffusion transformer training only.

\end{table}

\subsection{Catastrophic Forgetting of the Swin-Wrap and 3D-DiT Models in 2-channel}

Table \ref{tab:s2_lv4_metrics_bubble} presents the performance of the approaches in the 2 channels models in the small scale.  The two-channel S2-Lv4 configuration reveals a clear trade-off between average performance and robustness. EWC achieves the lowest reconstruction errors and highest PSNR values on D1--D3, demonstrating effective knowledge retention during continual learning. However, its performance deteriorates substantially on D4, where RMSE, MSE, and MAE increase sharply and PSNR decreases by more than 10 dB relative to D3. In contrast, GBP exhibits more stable behavior across datasets and maintains competitive reconstruction quality under the most challenging task conditions. These results suggest that EWC offers stronger average performance, whereas GBP provides greater robustness to distribution shifts and increasing task difficulty.

\begin{table}[h!]
\centering
\caption{
Comparison of continual learning strategies for dual-channel 3D-Swin-DiT-S2-Lv4 models. Results are reported as mean $\pm$ standard deviation. Methods include sequential fine-tuning (Seq), independently trained reference models (Ref), Elastic Weight Consolidation (EWC), and Graph-Blueprint Pruning (GBP).
}
\label{tab:s2_lv4_metrics_bubble}

\begin{tabular}{llccc}
\toprule
Dataset & Method & SSIM & RMSE & PSNR \\
\midrule

D1 & Seq &
$0.717 \pm 0.0091$ &
$0.2350 \pm 0.0027$ &
$12.59 \pm 0.16$ \\

D1 & Ref &
$0.719 \pm 0.009$ &
$0.385 \pm 0.008$ &
$8.23 \pm 0.20$ \\

D1 & EWC &
$0.7796 \pm 0.0081$ &
$0.1206 \pm 0.0027$ &
$18.38 \pm 0.19$ \\

D1 & GBP &
$0.773 \pm 0.009$ &
$0.1179 \pm 0.0024$ &
$18.57 \pm 0.17$ \\

\midrule

D2 & Seq &
$0.757 \pm 0.0070$ &
$0.0991 \pm 0.0058$ &
$20.09 \pm 0.51$ \\

D2 & Ref &
$0.774 \pm 0.007$ &
$0.453 \pm 0.027$ &
$6.85 \pm 0.17$ \\

D2 & EWC &
$0.7706 \pm 0.0078$ &
$0.0580 \pm 0.0034$ &
$24.75 \pm 0.52$ \\

D2 & GBP &
$0.784 \pm 0.009$ &
$0.1050 \pm 0.0067$ &
$19.59 \pm 0.57$ \\

\midrule

D3 & Seq &
$0.7577 \pm 0.0087$ &
$0.06273 \pm 0.00702$ &
$24.47 \pm 0.43$ \\

D3 & Ref &
$0.796 \pm 0.017$ &
$0.102 \pm 0.003$ &
$19.80 \pm 0.31$ \\

D3 & EWC &
$0.751 \pm 0.017$ &
$0.0581 \pm 0.0018$ &
$24.73 \pm 0.27$ \\

D3 & GBP &
$0.695 \pm 0.018$ &
$0.1011 \pm 0.0023$ &
$19.90 \pm 0.20$ \\

\midrule

D4 & Seq &
$0.860 \pm 0.0138$ &
$0.0872 \pm 0.0020$ &
$21.19 \pm 0.20$ \\

D4 & Ref &
$0.870 \pm 0.017$ &
$0.0782 \pm 0.0025$ &
$22.12 \pm 0.14$ \\

D4 & EWC &
$0.8358 \pm 0.0112$ &
$0.2238 \pm 0.0074$ &
$13.01 \pm 0.29$ \\

D4 & GBP &
$0.771 \pm 0.011$ &
$0.0993 \pm 0.0031$ &
$20.07 \pm 0.27$ \\

\bottomrule
\end{tabular}

\vspace{0.1cm}

\begin{tabular}{llcc}
\toprule
Dataset & Method & MSE & MAE \\
\midrule

D1 & Seq &
$0.0546 \pm 0.00076$ &
$0.1478 \pm 0.0010$ \\

D1 & Ref &
$0.146 \pm 0.006$ &
$0.238 \pm 0.003$ \\

D1 & EWC &
$0.01454 \pm 0.00065$ &
$0.0682 \pm 0.0010$ \\

D1 & GBP &
$0.01392 \pm 0.00055$ &
$0.0655 \pm 0.0009$ \\

\midrule

D2 & Seq &
$0.00985 \pm 0.00115$ &
$0.0386 \pm 0.0044$ \\

D2 & Ref &
$0.207 \pm 0.024$ &
$0.186 \pm 0.021$ \\

D2 & EWC &
$0.00338 \pm 0.00039$ &
$0.0222 \pm 0.0024$ \\

D2 & GBP &
$0.01108 \pm 0.00140$ &
$0.0407 \pm 0.0052$ \\

\midrule

D3 & Seq &
$0.00372 \pm 0.00067$ &
$0.0327 \pm 0.00092$ \\

D3 & Ref &
$0.0104 \pm 0.0007$ &
$0.0536 \pm 0.0019$ \\

D3 & EWC &
$0.00337 \pm 0.00021$ &
$0.03102 \pm 0.00115$ \\

D3 & GBP &
$0.01023 \pm 0.00046$ &
$0.05265 \pm 0.00128$ \\

\midrule

D4 & Seq &
$0.00761 \pm 0.00035$ &
$0.0474 \pm 0.0013$ \\

D4 & Ref &
$0.00614 \pm 0.00039$ &
$0.0400 \pm 0.0015$ \\

D4 & EWC &
$0.05014 \pm 0.00332$ &
$0.1343 \pm 0.0055$ \\

D4 & GBP &
$0.00987 \pm 0.00061$ &
$0.0530 \pm 0.0015$ \\

\bottomrule
\end{tabular}

\footnotesize
\textit{Abbreviations:}
Seq = sequential fine-tuning;
Ref = independently trained reference model;
EWC = Elastic Weight Consolidation;
GBP = Graph-Blueprint Pruning.

\end{table}
\newpage

\subsection{Forgetting Score and Performance in Previous Tasks (1-channel) }

\subsubsection{SSIM structural similarity}
\label{sec:ssim-1ch}

On SSIM the three methods sit within a narrow band
($\mathrm{LA}\in[0.810,0.814]$). EWC takes the highest average accuracy
($\mathrm{ACC}=0.835$) and, with SEQ, records zero peak forgetting
($F=0.000$), its stage-4 row sitting above the corresponding diagonal
cells. GBP records the smallest absolute backward transfer ($-0.003$) and the
only nonzero peak forgetting ($F=0.003$), reflecting that its
blueprint-constrained final update leaves SSIM marginally (at most $0.004$)
below the level first reached on each task. SSIM forward transfer is small and
negative across all methods ($-0.016$ to $-0.023$).

\begin{table}
\centering
\caption{CL metrics on SSIM (1-channel). \textbf{Bold} marks the best per
column; \textit{italics} the worst. For BWT and FWT, closer-to-zero is better;
for $F$, lower is always better.}
\label{tab:ssim-1ch}
\setlength{\tabcolsep}{8pt}\renewcommand{\arraystretch}{1.15}
\begin{tabular}{lS[round-mode=places,round-precision=3]S[round-mode=places,round-precision=3]S[round-mode=places,round-precision=3]S[round-mode=places,round-precision=3]S[round-mode=places,round-precision=3]}
\toprule
Method & {ACC \up} & {LA \up} & {BWT \zero} & {F \dn} & {FWT \zero} \\
\midrule
SEQ & 0.829           & \bfseries 0.814 & 0.020            & \bfseries 0.000 & -0.017           \\
EWC & \bfseries 0.835 & \itshape 0.810  & \itshape 0.033   & \bfseries 0.000 & \itshape -0.023  \\
GBP & \itshape 0.810  & 0.812           & \bfseries -0.003 & \itshape 0.003  & \bfseries -0.016 \\
\bottomrule
\end{tabular}
\end{table}

\begin{figure}
\centering
\includegraphics[width=0.85\textwidth]{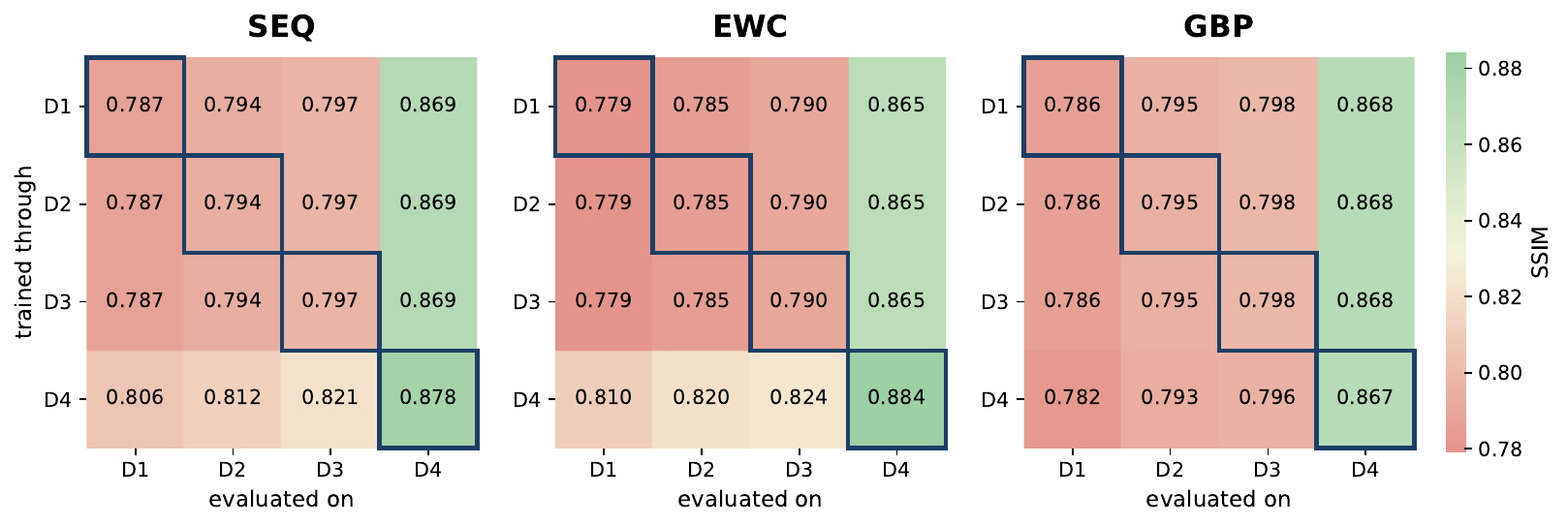}
\caption{Performance matrices $\mathbf{R}^{(m,\mathrm{SSIM})}$ for the
1-channel models. Rows = training stage, columns = evaluation task; diagonal
cells (outlined in dark blue) mark the just-learned task. Green = higher SSIM
(better); coral = lower SSIM (worse).}
\label{fig:R-ssim-1ch}
\end{figure}

\subsubsection{PSNR peak signal-to-noise ratio in dB}
\label{sec:psnr-1ch}

\begin{table}
\centering
\caption{CL metrics on PSNR (dB), 1-channel. $F$ is the non-negative best-ever
average forgetting, applied identically to all three methods.}
\label{tab:psnr-1ch}
\setlength{\tabcolsep}{8pt}\renewcommand{\arraystretch}{1.15}
\begin{tabular}{lS[round-mode=places,round-precision=2]S[round-mode=places,round-precision=2]S[round-mode=places,round-precision=2]S[round-mode=places,round-precision=2]S[round-mode=places,round-precision=2]}
\toprule
Method & {ACC \up} & {LA \up} & {BWT \zero} & {F \dn} & {FWT \zero} \\
\midrule
SEQ & 14.83           & 19.28           & -5.93           & 5.93            & \itshape 2.68  \\
EWC & \itshape 10.75  & \itshape 18.40  & \itshape -10.20 & \itshape 10.20  & \bfseries 1.97 \\
GBP & \bfseries 21.20 & \bfseries 21.15 & \bfseries 0.07  & \bfseries 0.03  & 2.58           \\
\bottomrule
\end{tabular}
\end{table}

\begin{figure}
\centering
\includegraphics[width=0.85\textwidth]{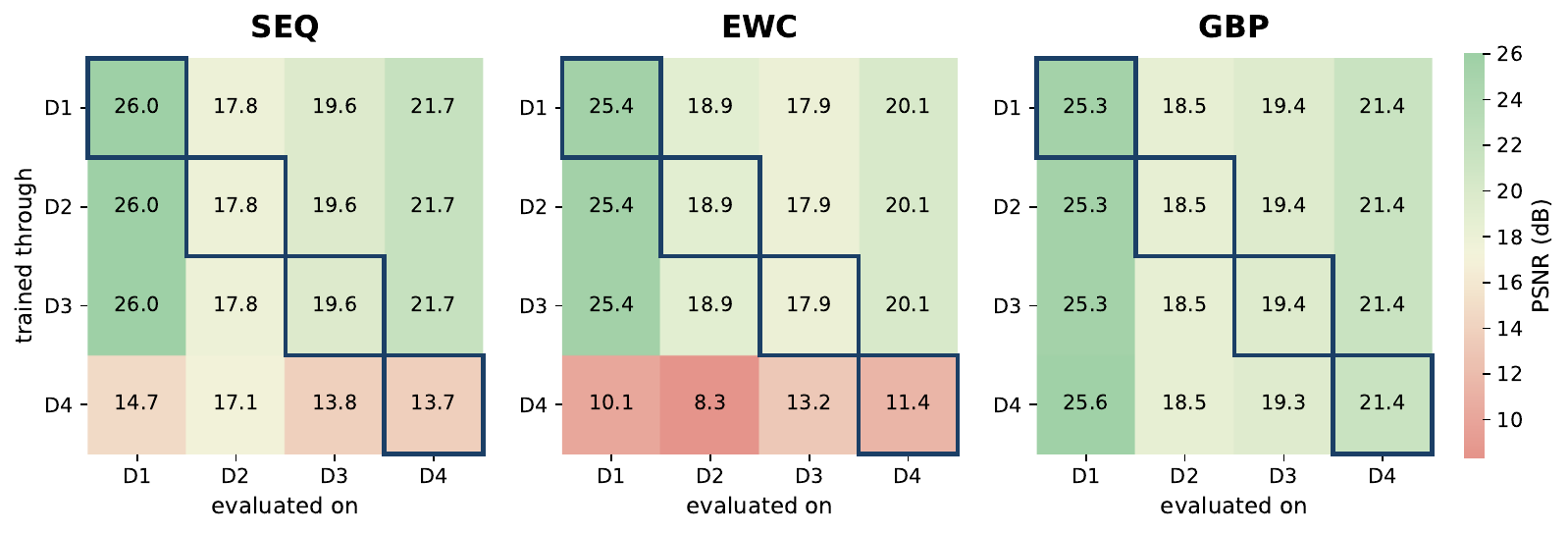}
\caption{Performance matrices $\mathbf{R}^{(m,\mathrm{PSNR})}$ for the
1-channel models, in dB. The EWC stage-4 collapse is visible across all four
columns.}
\label{fig:R-psnr-1ch}
\end{figure}

The PSNR table makes the central CL story explicit. GBP attains the highest
average accuracy ($\mathrm{ACC}=21.20$~dB), the highest learning accuracy
($\mathrm{LA}=21.15$~dB), the smallest peak forgetting ($F=0.03$~dB) and the
smallest absolute backward transfer ($\mathrm{BWT}=+0.07$~dB). Its backward
transfer is in fact slightly positive: after the final update GBP sits on
average $0.07$~dB \emph{above} the level at which each earlier task was first
learned, so the blueprint freeze eliminates degradation of past tasks at the
reported precision. EWC achieves comparable per-task learning
($\mathrm{LA}=18.40$~dB) but its stage-4 row collapses on every column
($10.1$, $8.3$, $13.2$, $11.4$~dB), yielding peak forgetting $F=10.20$~dB and
the worst average accuracy in the table. SEQ falls between the two on $F$
($5.93$~dB), with a less severe stage-4 degradation than EWC but well short of
GBP's retention. All three methods show positive PSNR forward transfer in this
configuration ($+1.97$ to $+2.68$~dB).

\subsubsection{MSE mean squared error}
\label{sec:mse-1ch}

\begin{table}
\centering
\caption{CL metrics on MSE (1-channel). Sign conventions inverted (lower is
better): BWT above zero means error grew on past tasks; FWT below zero means
the continually-trained model is more accurate than the independent reference.}
\label{tab:mse-1ch}
\setlength{\tabcolsep}{8pt}\renewcommand{\arraystretch}{1.15}
\begin{tabular}{lS[round-mode=places,round-precision=4]S[round-mode=places,round-precision=4]S[round-mode=places,round-precision=4]S[round-mode=places,round-precision=4]S[round-mode=places,round-precision=4]}
\toprule
Method & {ACC \dn} & {LA \dn} & {BWT \zero} & {F \dn} & {FWT \zero} \\
\midrule
SEQ & 0.0343           & 0.0185           & 0.0210           & 0.0210           & \bfseries -0.0167 \\
EWC & \itshape 0.0923  & \itshape 0.0263  & \itshape 0.0880  & \itshape 0.0880  & \itshape -0.0153  \\
GBP & \bfseries 0.0090 & \bfseries 0.0090 & \bfseries 0.0000 & \bfseries 0.0000 & \bfseries -0.0167 \\
\bottomrule
\end{tabular}
\end{table}

\begin{figure}
\centering
\includegraphics[width=0.85\textwidth]{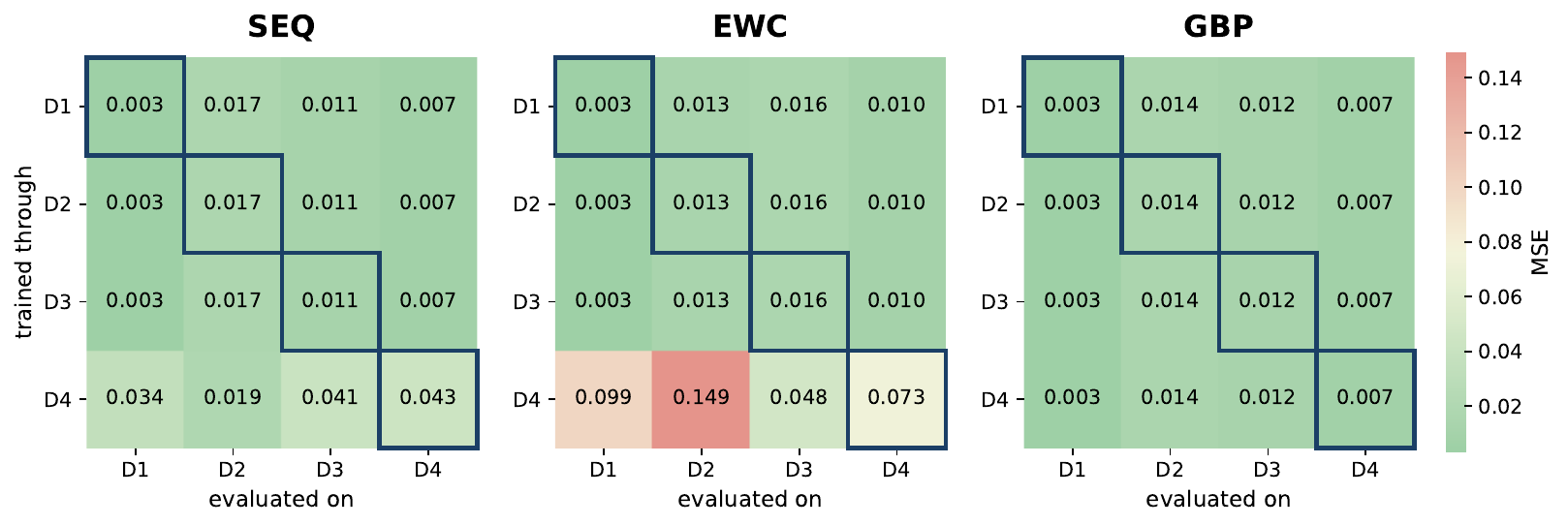}
\caption{Performance matrices $\mathbf{R}^{(m,\mathrm{MSE})}$ for the 1-channel
models. Green = low error (good), coral = high error (bad). EWC's stage-4 row
appears as a coral band; GBP's stage-4 row is the greenest of all final rows.}
\label{fig:R-mse-1ch}
\end{figure}

GBP records the smallest average accuracy ($\mathrm{ACC}=0.0090$), the smallest
learning accuracy ($\mathrm{LA}=0.0090$), zero backward transfer
($\mathrm{BWT}=0.0000$) and zero peak forgetting ($F=0.0000$): its stage-4 row
is identical to its earlier rows at the reported precision, so voxel-wise error
on past tasks does not grow at all across the final update. EWC suffers the
largest stage-4 collapse: MSE rises from a $0.003$--$0.073$ diagonal to
$0.048$--$0.149$ across the stage-4 row, giving $\mathrm{BWT}=+0.088$ and
$F=0.088$. SEQ is intermediate ($\mathrm{BWT}=F=0.021$). All methods show
comparable, favourable forward transfer ($-0.015$ to $-0.017$).

\subsubsection{RMSE root mean squared error}
\label{sec:rmse-1ch}

\begin{table}
\centering
\caption{CL metrics on RMSE (1-channel).}
\label{tab:rmse-1ch}
\setlength{\tabcolsep}{8pt}\renewcommand{\arraystretch}{1.15}
\begin{tabular}{lS[round-mode=places,round-precision=3]S[round-mode=places,round-precision=3]S[round-mode=places,round-precision=3]S[round-mode=places,round-precision=3]S[round-mode=places,round-precision=3]}
\toprule
Method & {ACC \dn} & {LA \dn} & {BWT \zero} & {F \dn} & {FWT \zero} \\
\midrule
SEQ & 0.183           & 0.123           & 0.080           & 0.080           & \bfseries -0.050 \\
EWC & \itshape 0.297  & \itshape 0.141  & \itshape 0.207  & \itshape 0.207  & \itshape -0.043  \\
GBP & \bfseries 0.091 & \bfseries 0.091 & \bfseries 0.000 & \bfseries 0.001 & -0.049           \\
\bottomrule
\end{tabular}
\end{table}

\begin{figure}
\centering
\includegraphics[width=0.85\textwidth]{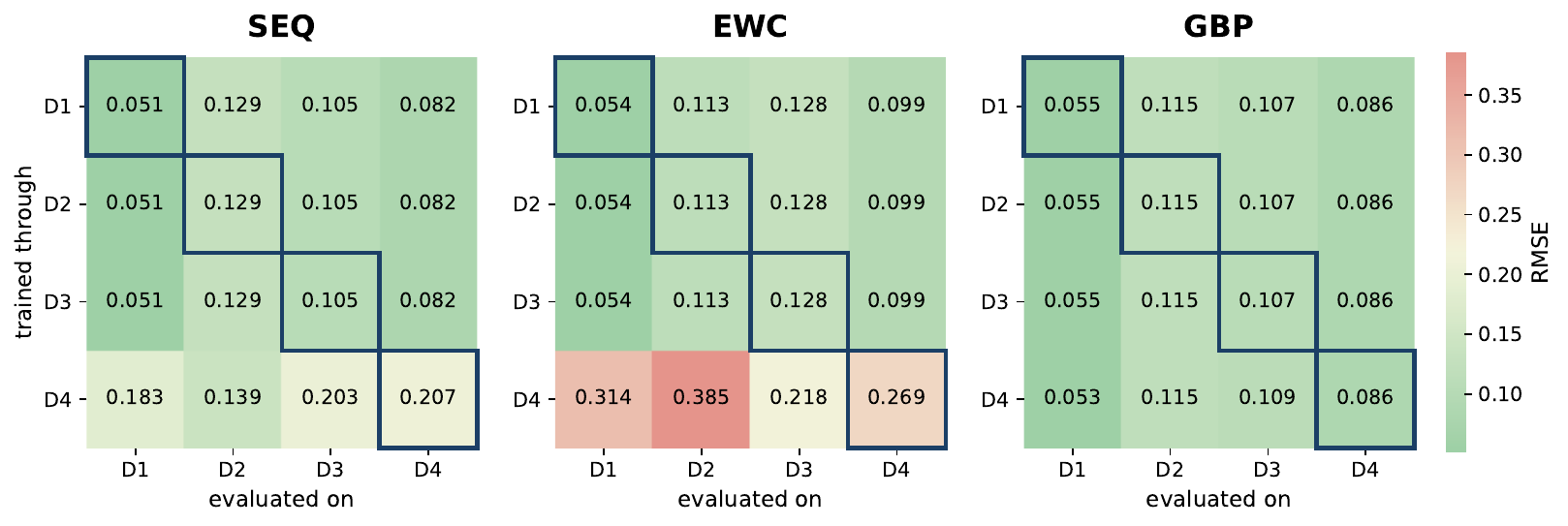}
\caption{Performance matrices $\mathbf{R}^{(m,\mathrm{RMSE})}$ for the 1-channel
models. RMSE compresses the dynamic range of MSE ($\sqrt{\cdot}$), producing a
smoother colour gradient while preserving the same rankings.}
\label{fig:R-rmse-1ch}
\end{figure}

The RMSE ranking is identical to MSE: GBP takes ACC, LA, the smallest absolute
BWT ($0.000$) and the smallest peak forgetting ($F=0.001$); EWC has the
largest stage-4 collapse on every voxel-wise statistic. GBP's $F$ marginally
exceeds its BWT because its stage-4 RMSE on $D_1$ ($0.053$) improves on the
diagonal ($0.055$), and the non-negative forgetting definition clips this gain
to zero. The convergence of the MSE and RMSE rankings on the same
EWC-versus-SEQ-versus-GBP ordering confirms that the EWC degradation is
structural to the stage-4 update rather than a squared-error outlier.

\subsubsection{MAE mean absolute error}
\label{sec:mae-1ch}

\begin{table}
\centering
\caption{CL metrics on MAE (1-channel).}
\label{tab:mae-1ch}
\setlength{\tabcolsep}{8pt}\renewcommand{\arraystretch}{1.15}
\begin{tabular}{lS[round-mode=places,round-precision=3]S[round-mode=places,round-precision=3]S[round-mode=places,round-precision=3]S[round-mode=places,round-precision=3]S[round-mode=places,round-precision=3]}
\toprule
Method & {ACC \dn} & {LA \dn} & {BWT \zero} & {F \dn} & {FWT \zero} \\
\midrule
SEQ & 0.098           & 0.070           & 0.037           & 0.043           & \bfseries -0.029 \\
EWC & \itshape 0.193  & \itshape 0.086  & \itshape 0.142  & \itshape 0.142  & \itshape -0.023  \\
GBP & \bfseries 0.051 & \bfseries 0.051 & \bfseries 0.001 & \bfseries 0.001 & \bfseries -0.029 \\
\bottomrule
\end{tabular}
\end{table}

\begin{figure}
\centering
\includegraphics[width=\textwidth]{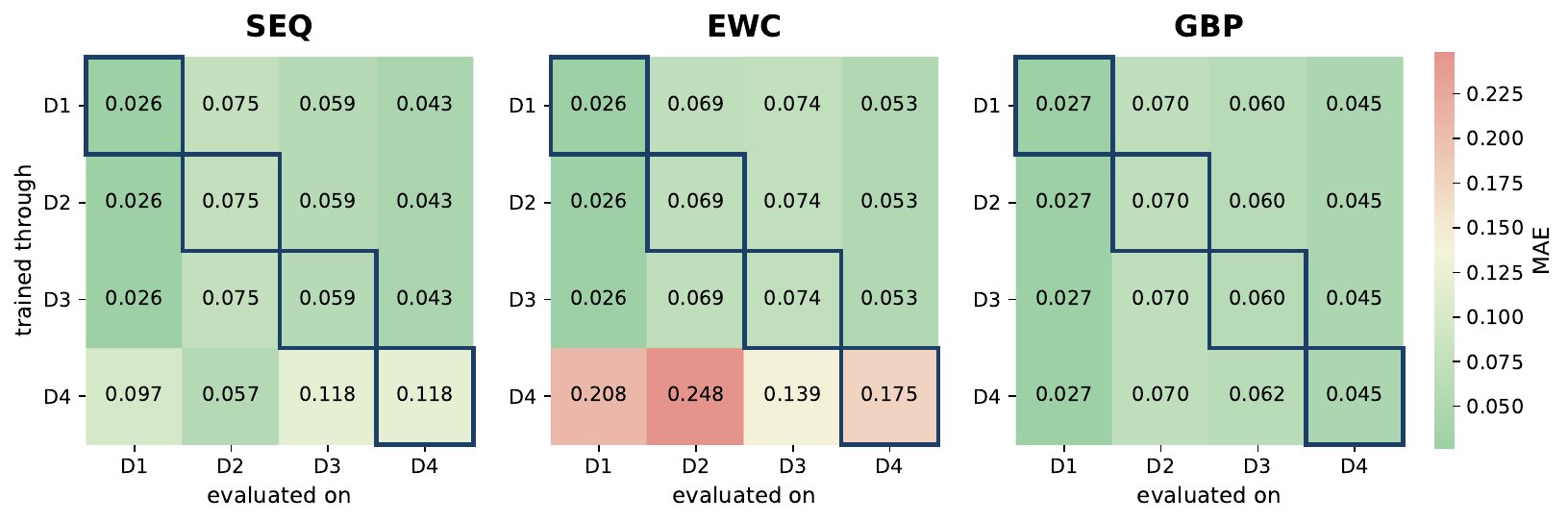}
\caption{Performance matrices $\mathbf{R}^{(m,\mathrm{MAE})}$ for the 1-channel
models. MAE is the most linear of the voxel-wise metrics and gives the cleanest
visual separation between GBP's stable stage-4 row and EWC's collapsed stage-4
row.}
\label{fig:R-mae-1ch}
\end{figure}

MAE confirms the pattern: GBP takes ACC, LA, the smallest absolute BWT
($+0.001$) and peak forgetting ($F=0.001$), its stage-4 row differing from the
earlier rows only on $D_3$ ($0.062$ vs $0.060$). EWC's stage-4 collapse on MAE
($F=0.142$) is the most interpretable per-voxel measure of the degradation ---
the per-voxel reconstruction error grows between roughly two-fold ($D_3$) and
eight-fold ($D_1$) on past tasks after stage 4 under EWC ($\approx 3.5\times$
on average), while under GBP it is essentially unchanged. Note that for SEQ
the MAE peak forgetting ($0.043$) exceeds its backward transfer ($0.037$)
because the best-ever MAE on $D_2$ ($0.057$) is attained at stage 4 rather
than on the diagonal.

\subsubsection{Cross-metric summary}
\label{sec:summary-1ch}

\begin{table}
\centering
\caption{Headline ACC, BWT and average forgetting $F$ on the two
higher-is-better scores, 1-channel. \textbf{Bold} marks the best per column;
\textit{italics} the worst.}
\label{tab:summary-hb-1ch}
\setlength{\tabcolsep}{8pt}\renewcommand{\arraystretch}{1.15}
\begin{tabular}{lS[round-mode=places,round-precision=3]S[round-mode=places,round-precision=3]S[round-mode=places,round-precision=3]S[round-mode=places,round-precision=2]S[round-mode=places,round-precision=2]S[round-mode=places,round-precision=2]}
\toprule
& \multicolumn{3}{c}{SSIM \up} & \multicolumn{3}{c}{PSNR \up~(dB)} \\
\cmidrule(lr){2-4}\cmidrule(lr){5-7}
Method & {ACC} & {BWT} & {F} & {ACC} & {BWT} & {F} \\
\midrule
SEQ & 0.829           & 0.020            & \bfseries 0.000 & 14.83           & -5.93           & 5.93            \\
EWC & \bfseries 0.835 & \itshape 0.033   & \bfseries 0.000 & \itshape 10.75  & \itshape -10.20 & \itshape 10.20  \\
GBP & \itshape 0.810  & \bfseries -0.003 & \itshape 0.003  & \bfseries 21.20 & \bfseries 0.07  & \bfseries 0.03  \\
\bottomrule
\end{tabular}
\end{table}

\begin{table}
\centering
\caption{Headline ACC, BWT and average forgetting $F$ on the three
lower-is-better scores, 1-channel.}
\label{tab:summary-lb-1ch}
\setlength{\tabcolsep}{4pt}\renewcommand{\arraystretch}{1.15}\footnotesize
\begin{tabular}{lS[round-mode=places,round-precision=4]S[round-mode=places,round-precision=4]S[round-mode=places,round-precision=4]S[round-mode=places,round-precision=3]S[round-mode=places,round-precision=3]S[round-mode=places,round-precision=3]S[round-mode=places,round-precision=3]S[round-mode=places,round-precision=3]S[round-mode=places,round-precision=3]}
\toprule
& \multicolumn{3}{c}{MSE \dn} & \multicolumn{3}{c}{RMSE \dn} & \multicolumn{3}{c}{MAE \dn}\\
\cmidrule(lr){2-4}\cmidrule(lr){5-7}\cmidrule(lr){8-10}
Method & {ACC} & {BWT} & {F} & {ACC} & {BWT} & {F} & {ACC} & {BWT} & {F} \\
\midrule
SEQ & 0.0343           & 0.0210           & 0.0210           & 0.183           & 0.080           & 0.080           & 0.098           & 0.037           & 0.043 \\
EWC & \itshape 0.0923  & \itshape 0.0880  & \itshape 0.0880  & \itshape 0.297  & \itshape 0.207  & \itshape 0.207  & \itshape 0.193  & \itshape 0.142  & \itshape 0.142 \\
GBP & \bfseries 0.0090 & \bfseries 0.0000 & \bfseries 0.0000 & \bfseries 0.091 & \bfseries 0.000 & \bfseries 0.001 & \bfseries 0.051 & \bfseries 0.001 & \bfseries 0.001 \\
\bottomrule
\end{tabular}
\end{table}

\subsubsection{Discussion of one-channel}
\label{sec:discussion-1ch}

\paragraph{GBP essentially eliminates forgetting on every voxel-wise metric in
the 1-channel continual evaluation.}
Under a single non-negative best-ever forgetting definition applied to all
three methods, GBP's forgetting is at or below the resolution of the reported
precision on every voxel-wise metric: PSNR $F=0.03$~dB, MSE $F=0.0000$, RMSE
$F=0.001$ and MAE $F=0.001$ --- between one and more than two orders of
magnitude below SEQ ($5.93$~dB, $0.021$, $0.080$, $0.043$) and further still
below EWC ($10.20$~dB, $0.088$, $0.207$, $0.142$). GBP's backward transfer is
slightly positive on PSNR ($+0.07$~dB) and indistinguishable from zero on the
three voxel-wise error metrics ($0.0000$, $0.000$, $+0.001$), so the blueprint
freeze eliminates catastrophic forgetting in this setting at the reported
precision.

\paragraph{EWC exhibits catastrophic forgetting at the final continual stage.}
EWC attains the highest SSIM accuracy ($\mathrm{ACC}=0.835$) and a competitive
PSNR learning accuracy ($\mathrm{LA}=18.40$~dB), but its stage-4 row collapses
on every voxel-wise metric: PSNR drops to $10.1$~dB at $D_1$, $8.3$~dB at
$D_2$, $13.2$~dB at $D_3$ and $11.4$~dB at $D_4$. The resulting backward
transfer is $\mathrm{BWT}=-10.20$~dB on PSNR and $+0.088$ on MSE, the largest
absolute values in the table on both. The collapse is global, with per-column
PSNR drops of $4.7$~dB ($D_3$) to $15.3$~dB ($D_1$), confirming a property of
the stage-4 update rather than a single-task artefact. The Fisher penalty is
insufficient to prevent the plasticity demands of the final task from
overwriting the consolidated representation.

\paragraph{SEQ is intermediate.}
SEQ's peak forgetting on PSNR ($5.93$~dB) lies between GBP's ($0.03$~dB) and
EWC's ($10.20$~dB). Its stage-4 row also degrades on every column (by up to
$11.3$~dB on $D_1$), but less severely than under EWC. On the voxel-wise error
metrics GBP takes the smallest absolute BWT throughout, with SEQ second and
EWC worst. SEQ's in-task learning accuracy is higher (worse) than GBP's
(MSE LA $0.0185$ vs GBP $0.0090$), so its overall ACC is intermediate on every
voxel-wise metric.

\paragraph{Forward transfer is positive on PSNR and favourable on the error
metrics.}
In this configuration all three methods show positive PSNR forward transfer
($+1.97$ to $+2.68$~dB) and favourable (negative) voxel-wise forward transfer
(MSE $-0.015$ to $-0.017$; MAE $-0.023$ to $-0.029$), indicating that each
method's representation is more accurate than the from-scratch 1-channel
reference on the zero-shot row before a task is seen. Differences between
methods on FWT are small relative to the backward-transfer and forgetting
differences.

\paragraph{Stability--plasticity trade-offs in the 1-channel regime.}
The three methods occupy three distinct points on the stability--plasticity
manifold. GBP achieves negligible forgetting and the best average accuracy on
every voxel-wise metric, at the cost of the lowest average SSIM accuracy. EWC
achieves the highest average SSIM accuracy but suffers the worst stage-4
retention on every voxel-wise metric. SEQ provides a baseline without
consolidation, intermediate on forgetting throughout. GBP therefore offers
the most balanced profile in the 1-channel evaluation: comparable or better
learning than EWC on every metric, and voxel-wise forgetting one to two orders
of magnitude below both baselines.

\paragraph{Structural and voxel-wise metrics partially diverge.}
SSIM and the voxel-wise metrics agree on the EWC stage-4 collapse and on GBP's
stable stage-4 row, but the SSIM ranking differs from the voxel-wise ranking
on ACC: EWC records the highest SSIM ACC but the lowest voxel-wise ACC, while
GBP records the highest voxel-wise ACC but the lowest SSIM ACC. SSIM is
largely insensitive to the intensity-scale variation that dominates the
voxel-wise statistics, and the two families must be reported jointly to
characterise continual neural reconstruction performance.

\subsection{Forgetting Score and Performance in Previous Tasks (2-channels)}

\subsubsection{SSIM structural similarity}
\label{sec:ssim-2ch}

\begin{table}
\centering
\caption{CL metrics on SSIM (2-channels). \textbf{Bold} marks the best per
column; \textit{italics} the worst.}
\label{tab:ssim-2ch}
\setlength{\tabcolsep}{8pt}
\renewcommand{\arraystretch}{1.15}
\begin{tabular}{l
                S[round-mode=places,round-precision=3]
                S[round-mode=places,round-precision=3]
                S[round-mode=places,round-precision=3]
                S[round-mode=places,round-precision=3]
                S[round-mode=places,round-precision=3]}
\toprule
Method & {ACC \up} & {LA \up} & {BWT \zero} & {F \dn} & {FWT \zero} \\
\midrule
SEQ & \bfseries 0.802 & 0.758           & \itshape 0.059  & \bfseries 0.000 & \itshape -0.072  \\
EWC & 0.782           & \bfseries 0.785 & \bfseries -0.004 & 0.011          & \bfseries -0.013 \\
GBP & \itshape 0.688  & \itshape 0.756  & -0.091          & \itshape 0.092  & -0.033           \\
\bottomrule
\end{tabular}
\end{table}

\begin{figure}
\centering
\includegraphics[width=0.85\textwidth]{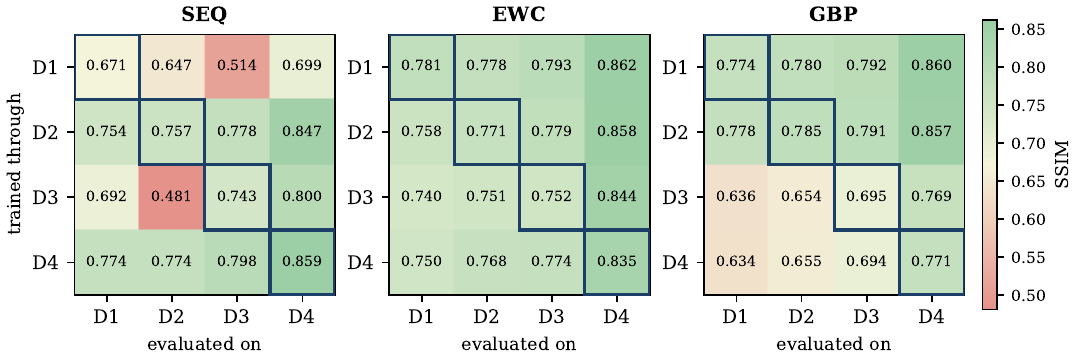}
\caption{Performance matrices $\mathbf{R}^{(m,\mathrm{SSIM})}$ for the
two-channel models, assembled from the continual-evaluation logs. Rows =
training stage, columns = evaluation task; diagonal cells (outlined in
dark blue) mark the just-learned task. Green = higher SSIM (better); coral
= lower SSIM (worse).}
\label{fig:R-ssim-2ch}
\end{figure}

On SSIM the three methods cluster tightly on learning
accuracy (LA $\in [0.756, 0.785]$). SEQ takes the highest average accuracy
($\mathrm{ACC}=0.802$) and exhibits effectively zero peak forgetting
($F=0.000$). EWC achieves the smallest absolute backward transfer
($\mathrm{BWT}=-0.004$) and the least-negative forward transfer
($\mathrm{FWT}=-0.013$), indicating strong SSIM stability under Fisher
anchoring. GBP records the lowest ACC and the largest absolute BWT and
peak forgetting on SSIM; its forward transfer ($-0.033$) is between the
two consolidation methods. The three SSIM forward-transfer values are all
small in absolute terms ($\abs{\mathrm{FWT}}\le 0.072$), substantially
closer to zero than in the one-channel report, indicating that the
two-channel curriculum produces more transferable structural-similarity
representations than the one-channel curriculum.

\subsubsection{PSNR peak signal-to-noise ratio in dB}
\label{sec:psnr-2ch}

\begin{table}
\centering
\caption{CL metrics on PSNR (dB), 2-channels.}
\label{tab:psnr-2ch}
\setlength{\tabcolsep}{8pt}
\renewcommand{\arraystretch}{1.15}
\begin{tabular}{l
                S[round-mode=places,round-precision=2]
                S[round-mode=places,round-precision=2]
                S[round-mode=places,round-precision=2]
                S[round-mode=places,round-precision=2]
                S[round-mode=places,round-precision=2]}
\toprule
Method & {ACC \up} & {LA \up} & {BWT \zero} & {F \dn} & {FWT \zero} \\
\midrule
SEQ & \bfseries 19.70 & 19.59           & \bfseries 0.15  & 3.00           & \bfseries -0.31 \\
EWC & \itshape 12.25  & \bfseries 20.24 & \itshape -10.65 & \itshape 12.13 & \itshape 4.07   \\
GBP & 19.51           & \itshape 19.56  & -0.07           & \bfseries 0.87 & 1.60            \\
\bottomrule
\end{tabular}
\end{table}

\begin{figure}
\centering
\includegraphics[width=0.85\textwidth]{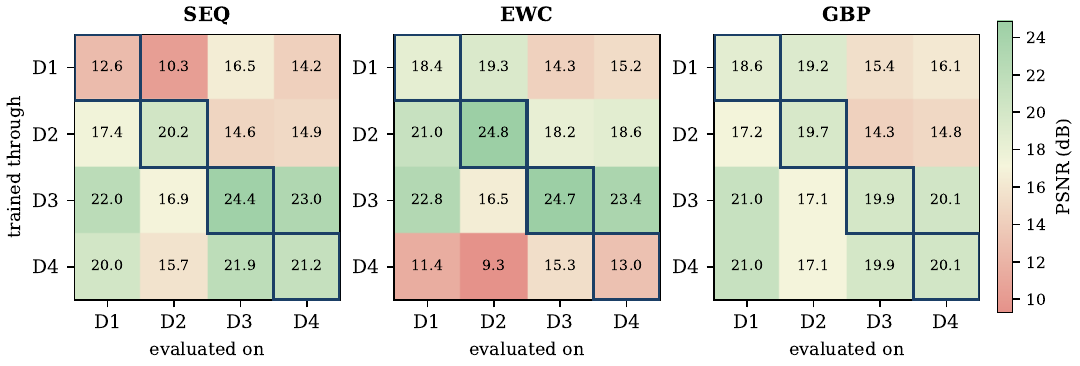}
\caption{Performance matrices $\mathbf{R}^{(m,\mathrm{PSNR})}$ for the
two-channel models, in dB. The diagonal cells (just-learned PSNR) and the
off-diagonal cells (retention) are on the same scale, so the heatmap
displays retention directly.}
\label{fig:R-psnr-2ch}
\end{figure}

The PSNR table reveals the central contrast of the
two-channel report: EWC attains the highest learning accuracy
($\mathrm{LA}=20.24$~dB), but its final-stage row collapses to a mean of
$12.25$~dB --- producing an absolute backward transfer of $-10.65$~dB and
a peak forgetting of $12.13$~dB, the largest in the table. The model
learns each task strongly but loses past-task fidelity catastrophically
when training on $D_4$. SEQ attains nearly the same ACC as EWC's LA
($\mathrm{ACC}=19.70$ vs.\ $19.59$~dB diagonal) and shows almost no net
backward transfer ($+0.15$~dB), confirming that unconsolidated training in
the two-channel regime does not suffer the catastrophic collapse seen in
one channel. GBP achieves the smallest peak forgetting on PSNR
($F=0.87$~dB, an order of magnitude smaller than SEQ's $3.00$~dB and over
ten times smaller than EWC's $12.13$~dB), preserving past-task PSNR with
minimal drift.

\subsubsection{MSE mean squared error}
\label{sec:mse-2ch}

\begin{table}
\centering
\caption{CL metrics on MSE (2-channels). Sign conventions inverted
(lower is better): BWT above zero means error grew on past tasks;
FWT below zero means the continually-trained model is more accurate than
the independent reference.}
\label{tab:mse-2ch}
\setlength{\tabcolsep}{8pt}
\renewcommand{\arraystretch}{1.15}
\begin{tabular}{l
                S[round-mode=places,round-precision=4]
                S[round-mode=places,round-precision=4]
                S[round-mode=places,round-precision=4]
                S[round-mode=places,round-precision=4]
                S[round-mode=places,round-precision=4]}
\toprule
Method & {ACC \dn} & {LA \dn} & {BWT \zero} & {F \dn} & {FWT \zero} \\
\midrule
SEQ & 0.0128          & \itshape 0.0190 & \bfseries -0.0083 & 0.0080          & \itshape -0.0298 \\
EWC & \itshape 0.0675 & 0.0179          & \itshape 0.0662   & \itshape 0.0693 & -0.0640          \\
GBP & \bfseries 0.0119 & \bfseries 0.0112 & 0.0010          & \bfseries 0.0029 & \bfseries -0.0548 \\
\bottomrule
\end{tabular}
\end{table}

\begin{figure}
\centering
\includegraphics[width=0.85\textwidth]{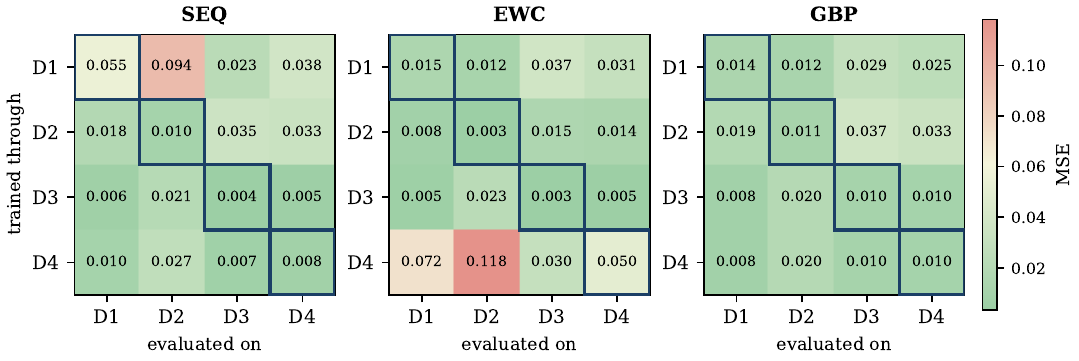}
\caption{Performance matrices $\mathbf{R}^{(m,\mathrm{MSE})}$ for the
two-channel models. Green = low error (good), coral = high error (bad).
The contrast between GBP's nearly-uniform row colour and EWC's last-row
shift to coral quantifies the divergent forgetting behaviour of the two
consolidation methods.}
\label{fig:R-mse-2ch}
\end{figure}

On MSE, GBP wins ACC ($0.0119$), LA ($0.0112$) and
peak forgetting ($F=0.0029$). EWC's MSE row-4 collapse mirrors its PSNR
pattern: its $F=0.069$ is more than 20 times GBP's, reflecting a large
growth of past-task error after the final training stage. SEQ records
the smallest absolute BWT ($-0.008$) because its diagonals are higher to
begin with, leaving more room for negative backward transfer to register.
The MSE forward transfer is negative (better than reference) for all three
methods; the from-scratch references for the 2-channels datasets have
markedly higher MSE than the continually-trained models, so this row
records a robust improvement of continual training over reference,
strongest for EWC ($-0.064$) and GBP ($-0.055$).

\subsubsection{RMSE root mean squared error}
\label{sec:rmse-2ch}

\begin{table}
\centering
\caption{CL metrics on RMSE (2-channels).}
\label{tab:rmse-2ch}
\setlength{\tabcolsep}{8pt}
\renewcommand{\arraystretch}{1.15}
\begin{tabular}{l
                S[round-mode=places,round-precision=3]
                S[round-mode=places,round-precision=3]
                S[round-mode=places,round-precision=3]
                S[round-mode=places,round-precision=3]
                S[round-mode=places,round-precision=3]}
\toprule
Method & {ACC \dn} & {LA \dn} & {BWT \zero} & {F \dn} & {FWT \zero} \\
\midrule
SEQ & 0.108           & \itshape 0.120  & \bfseries -0.016 & 0.036           & \itshape -0.023  \\
EWC & \itshape 0.252  & 0.115           & \itshape 0.183   & \itshape 0.199  & -0.112           \\
GBP & \bfseries 0.108 & \bfseries 0.106 & 0.003            & \bfseries 0.012 & \bfseries -0.077 \\
\bottomrule
\end{tabular}
\end{table}

\begin{figure}
\centering
\includegraphics[width=0.85\textwidth]{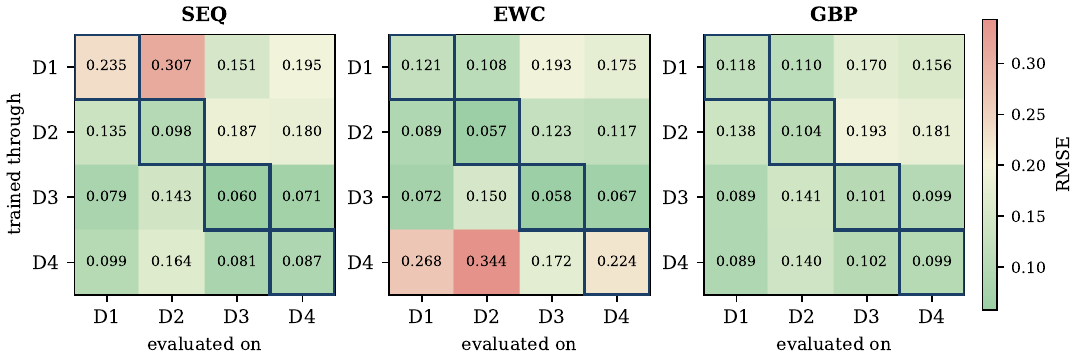}
\caption{Performance matrices $\mathbf{R}^{(m,\mathrm{RMSE})}$ for the
two-channel models. RMSE preserves directional information from MSE but
compresses dynamic range ($\sqrt{\,\cdot\,}$), giving a smoother colour
gradient.}
\label{fig:R-rmse-2ch}
\end{figure}

The RMSE pattern mirrors MSE: GBP takes ACC, LA and
peak forgetting; EWC suffers a large positive BWT and $F$ driven by its
stage-4 row; SEQ takes the smallest absolute BWT mechanically. SEQ and
GBP are tied on ACC ($0.108$). GBP again attains the best forward
transfer ($-0.077$), confirming that its consolidation preserves a useful
inductive bias for downstream tasks even though it does not always
maximise the in-task fit.

\subsubsection{MAE mean absolute error}
\label{sec:mae-2ch}

\begin{table}
\centering
\caption{CL metrics on MAE (2-channels).}
\label{tab:mae-2ch}
\setlength{\tabcolsep}{8pt}
\renewcommand{\arraystretch}{1.15}
\begin{tabular}{l
                S[round-mode=places,round-precision=3]
                S[round-mode=places,round-precision=3]
                S[round-mode=places,round-precision=3]
                S[round-mode=places,round-precision=3]
                S[round-mode=places,round-precision=3]}
\toprule
Method & {ACC \dn} & {LA \dn} & {BWT \zero} & {F \dn} & {FWT \zero} \\
\midrule
SEQ & 0.060           & \itshape 0.066  & \bfseries -0.008 & 0.026           & \itshape 0.018   \\
EWC & \itshape 0.154  & 0.064           & \itshape 0.120   & \itshape 0.128  & -0.041           \\
GBP & \bfseries 0.059 & \bfseries 0.053 & 0.008            & \bfseries 0.014 & \bfseries -0.020 \\
\bottomrule
\end{tabular}
\end{table}

\begin{figure}
\centering
\includegraphics[width=0.85\textwidth]{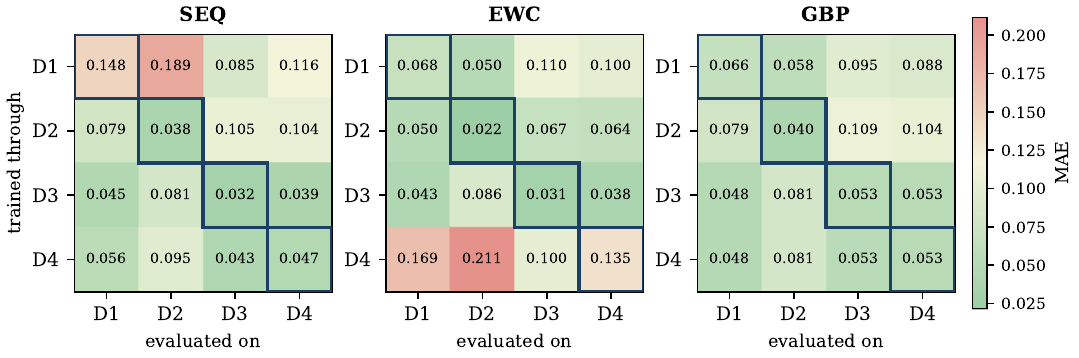}
\caption{Performance matrices $\mathbf{R}^{(m,\mathrm{MAE})}$ for the
two-channel models. MAE is less sensitive than MSE to large-magnitude
outliers, which is why the spread between methods is smaller here.}
\label{fig:R-mae-2ch}
\end{figure}

The MAE ranking is essentially identical to RMSE
and MSE: GBP wins ACC, LA, $F$ and FWT; SEQ wins BWT (smallest absolute
value); EWC is last on every column except LA, where it is tied with SEQ
($0.064$ vs.\ $0.066$). The consistent ordering across all three
voxel-wise metrics indicates that the EWC collapse in row 4 and the GBP
stability are robust to the choice of loss functional.

\subsubsection{Cross-metric summary}
\label{sec:summary-2ch}

Table~\ref{tab:summary-hb-2ch} summarises ACC, BWT and $F$ on the two
higher-is-better scores; Table~\ref{tab:summary-lb-2ch} does the same on
the three lower-is-better scores.

\begin{table}
\centering
\caption{Headline ACC, BWT and average forgetting $F$ on the two
higher-is-better scores, 2-channels. \textbf{Bold} marks the best per
column; \textit{italics} the worst. For BWT, closer-to-zero is better;
for $F$, lower is always better.}
\label{tab:summary-hb-2ch}
\setlength{\tabcolsep}{8pt}
\renewcommand{\arraystretch}{1.15}
\begin{tabular}{l
                S[round-mode=places,round-precision=3]
                S[round-mode=places,round-precision=3]
                S[round-mode=places,round-precision=3]
                S[round-mode=places,round-precision=2]
                S[round-mode=places,round-precision=2]
                S[round-mode=places,round-precision=2]}
\toprule
& \multicolumn{3}{c}{SSIM \up} & \multicolumn{3}{c}{PSNR \up~(dB)} \\
\cmidrule(lr){2-4}\cmidrule(lr){5-7}
Method & {ACC} & {BWT} & {F}
       & {ACC} & {BWT} & {F} \\
\midrule
SEQ & \bfseries 0.802 & \itshape 0.059   & \bfseries 0.000
    & \bfseries 19.70 & \bfseries 0.15   & 3.00 \\
EWC & 0.782           & \bfseries -0.004 & 0.011
    & \itshape 12.25  & \itshape -10.65  & \itshape 12.13 \\
GBP & \itshape 0.688  & -0.091           & \itshape 0.092
    & 19.51           & -0.07            & \bfseries 0.87 \\
\bottomrule
\end{tabular}
\end{table}

\begin{table}
\centering
\caption{Headline ACC, BWT and average forgetting $F$ on the three
lower-is-better scores, 2-channels. Sign conventions are unified so that
for BWT closer-to-zero is better, and for $F$ lower is always better.}
\label{tab:summary-lb-2ch}
\setlength{\tabcolsep}{4pt}
\renewcommand{\arraystretch}{1.15}
\footnotesize
\begin{tabular}{l
                S[round-mode=places,round-precision=4]
                S[round-mode=places,round-precision=4]
                S[round-mode=places,round-precision=4]
                S[round-mode=places,round-precision=3]
                S[round-mode=places,round-precision=3]
                S[round-mode=places,round-precision=3]
                S[round-mode=places,round-precision=3]
                S[round-mode=places,round-precision=3]
                S[round-mode=places,round-precision=3]}
\toprule
& \multicolumn{3}{c}{MSE \dn} & \multicolumn{3}{c}{RMSE \dn}
& \multicolumn{3}{c}{MAE \dn}\\
\cmidrule(lr){2-4}\cmidrule(lr){5-7}\cmidrule(lr){8-10}
Method & {ACC} & {BWT} & {F}
       & {ACC} & {BWT} & {F}
       & {ACC} & {BWT} & {F} \\
\midrule
SEQ & 0.0128          & \bfseries -0.0083 & 0.0080
    & 0.108           & \bfseries -0.016 & 0.036
    & 0.060           & \bfseries -0.008 & 0.026 \\
EWC & \itshape 0.0675 & \itshape 0.0662  & \itshape 0.0693
    & \itshape 0.252  & \itshape 0.183   & \itshape 0.199
    & \itshape 0.154  & \itshape 0.120   & \itshape 0.128 \\
GBP & \bfseries 0.0119 & 0.0010          & \bfseries 0.0029
    & \bfseries 0.108 & 0.003            & \bfseries 0.012
    & \bfseries 0.059 & 0.008            & \bfseries 0.014 \\
\bottomrule
\end{tabular}
\end{table}

\subsubsection{Discussion of two-channel}
\label{sec:discussion-2ch}

\paragraph{GBP attains the strongest voxel-wise continual-learning
performance across MSE, RMSE and MAE.}
On the three voxel-wise reconstruction metrics, GBP records simultaneously
the highest average accuracy (MSE ACC $=0.0119$, RMSE ACC $=0.108$,
MAE ACC $=0.059$), the lowest learning-accuracy error (MSE LA $=0.0112$,
RMSE LA $=0.106$, MAE LA $=0.053$), and the smallest peak forgetting
(MSE $F=0.003$, RMSE $F=0.012$, MAE $F=0.014$). On PSNR, GBP matches SEQ
on average accuracy within $0.2$~dB and achieves the smallest peak
forgetting in the table ($F=0.87$~dB, more than ten times smaller than
EWC's $F=12.13$~dB and an order of magnitude smaller than SEQ's
$F=3.00$~dB). Together these results indicate that the salience-graph
blueprint, by hard-freezing a structurally identified set of units,
produces a model whose voxel-wise reconstruction quality is both stronger
per-task and more stable across continual stages than either Fisher
anchoring or unconsolidated training.

\paragraph{EWC suffers catastrophic forgetting at the final continual
stage.}
The most striking pattern in the two-channel results is EWC's collapse at
stage~4. EWC attains the highest PSNR learning accuracy ($20.24$~dB) and
the lowest MSE/RMSE/MAE learning-accuracy errors of any consolidation
method, but its row-4 evaluation row drops to a mean PSNR of $11.43$~dB
and the MSE/RMSE/MAE row-4 means more than double. The resulting peak
forgetting --- PSNR $F=12.13$~dB, MSE $F=0.069$ --- is the worst in the
table on every voxel-wise metric. The pattern is consistent across the
four columns of EWC's row 4 (PSNR drops of $7$--$11$~dB, see
Fig.~\ref{fig:R-psnr-2ch}), so the collapse is a global property of the
stage-4 update rather than a single-task artefact. The Fisher-information
penalty, weighted by $\lambda = 10^4$, is insufficient to prevent the
combined plasticity demands of the two-channel diffusion target from
overwriting the consolidated parameters at the final stage.

\paragraph{SEQ exhibits unexpectedly stable PSNR retention in the
two-channel regime.}
SEQ in two channels does not collapse
catastrophically. SEQ's PSNR backward transfer is essentially zero
($+0.15$~dB), and its SSIM peak forgetting is mathematically zero (the
final-row SSIM never falls below the just-learned diagonal on any task).
A plausible explanation is that the FA + MD channel coupling acts as a
form of multi-task regularisation: gradients arising from the two
channels share a substantial portion of their direction, so updating to
fit $D_k$ pulls less strongly away from the previously fit $D_{k-1}$
directions than in the one-channel setting. Under this hypothesis the
consolidation methods (EWC, GBP) operate against a backdrop of intrinsic
stability already provided by the data, and their additional protection
is therefore evaluated against a stronger baseline than in one channel.

\paragraph{Forward transfer is small in magnitude and mixed in sign.}
Forward transfer values in the two-channel report are markedly smaller in
absolute magnitude than in the one-channel report. SSIM FWT values lie in
$[-0.072, -0.013]$, two-channel PSNR FWT values lie in $[-0.31, +4.07]$,
and MSE/RMSE/MAE FWT values are predominantly negative (continual model
better than reference). The sign pattern is mixed: SSIM FWT is negative
for all three methods (continual model slightly worse than reference);
PSNR FWT is negative for SEQ but positive for EWC and GBP; and the three
voxel-wise metrics show FWT negative (better than reference) for all
three methods. The two-channel from-scratch references used here exhibit
poor reconstruction quality (D1-ref PSNR $=8.23$~dB, D2-ref PSNR $=6.85$~dB),
which makes outperforming the reference relatively easy on voxel-wise
metrics --- consequently the negative FWT on those metrics reports a real
improvement of the continual model over the reduced-budget reference, but
should not be over-interpreted as the continual curriculum providing
strong positive transfer in an absolute sense. On SSIM, where the
references are competitive (D1-ref $=0.719$, D2-ref $=0.774$), the small
negative FWT values represent the more informative comparison: continual
adaptation does not noticeably hurt SSIM initialisation in the two-channel
regime.

\paragraph{The stability--plasticity manifold differs sharply between
channel counts.}
The two-channel results invert several of the rankings observed in the
one-channel report. In one channel, EWC was the most stable method and
GBP led on forward transfer; in two channels, GBP leads on every
voxel-wise stability statistic and EWC collapses at the final stage.
SEQ in turn changes role from the most plastic (and most-forgetting)
method in one channel to a near-best-on-PSNR baseline in two channels.
The factor distinguishing the two regimes is the channel coupling
introduced by FA + MD reconstruction. The proposed GBP method, by
contrast, retains its advantage in both regimes --- best SSIM FWT in one
channel and best voxel-wise ACC/LA/$F$ in two channels --- indicating that
the salience-graph blueprint identifies sub-circuits whose protection is
beneficial regardless of channel structure, even when the relative
performance of the parameter-anchoring baseline changes substantially.

\subsection{Forgetting Score and Performance in Previous Tasks (Hybrid channel) }

The hybrid-channel matrices are assembled from the 2-channel-trained
checkpoints, loaded into the 1-channel evaluation architecture through
a shape-mismatch adapter: 2-channel input convolutions are
collapsed to 1-channel via mean across the input-channel axis, and the
2-channel output head is reduced to one channel by selecting the
first output (the FA channel). This evaluation therefore measures the
performance of 2-channel-trained continual representations when the
diffusion-derivative is reduced back to a single-channel reconstruction
target (hybrid channel).

\subsubsection{SSIM structural similarity}
\label{sec:ssim-hyb}

\begin{table}
\centering
\caption{CL metrics on SSIM (hybrid channel evaluation of 2-channel-trained
checkpoints). \textbf{Bold} marks the best per column; \textit{italics}
the worst.}
\label{tab:ssim-hyb}
\setlength{\tabcolsep}{8pt}
\renewcommand{\arraystretch}{1.15}
\begin{tabular}{l
                S[round-mode=places,round-precision=3]
                S[round-mode=places,round-precision=3]
                S[round-mode=places,round-precision=3]
                S[round-mode=places,round-precision=3]
                S[round-mode=places,round-precision=3]}
\toprule
Method & {ACC \up} & {LA \up} & {BWT \zero} & {F \dn} & {FWT \zero} \\
\midrule
SEQ & 0.802           & \bfseries 0.777 & 0.033          & \bfseries 0.000 & \itshape -0.045 \\
EWC & \bfseries 0.810 & 0.775           & \itshape 0.047 & \bfseries 0.000 & -0.043           \\
GBP & \itshape 0.616  & \itshape 0.721  & \bfseries -0.140 & \itshape 0.147 & \bfseries -0.082 \\
\bottomrule
\end{tabular}
\end{table}

\begin{figure}
\centering
\includegraphics[width=\textwidth]{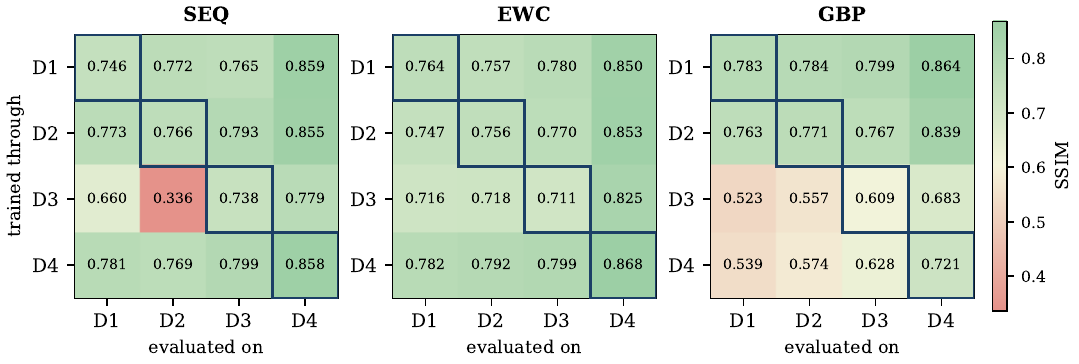}
\caption{Performance matrices $\mathbf{R}^{(m,\mathrm{SSIM})}$ for the
hybrid channel evaluation. Rows = training stage, columns = evaluation task;
diagonal cells (outlined in dark blue) mark the just-learned task.
GBP's row 3 and row 4 show a coral band on D1--D2 reflecting the
adaptation cost of the channel reduction at the GBP-D3 and GBP-D4
checkpoints; SEQ and EWC retain SSIM in the $0.75$--$0.86$ range
across all stages.}
\label{fig:R-ssim-hyb}
\end{figure}

SSIM rankings are dominated by adapter behaviour on
the GBP checkpoints. SEQ and EWC retain SSIM in a narrow $[0.75,
0.86]$ band across all sixteen cells; GBP's stage-3 and stage-4
checkpoints show measurable SSIM drops on the early evaluations (D1
SSIM $0.52$--$0.54$ for GBP rows 3 and 4), pulling its average accuracy
to $0.616$. The drop reflects the larger adaptation cost when the
GBP-trained 2-channel filters are collapsed back to one channel.
Forward transfer is uniformly negative and small ($-0.04$ to $-0.08$),
indicating that the continually-trained model after the channel reduction
is slightly below the hybrid channel reference on the zero-shot row before
each task is seen.

\subsubsection{PSNR peak signal-to-noise ratio in dB}
\label{sec:psnr-hyb}

\begin{table}
\centering
\caption{CL metrics on PSNR (dB), hybrid channel evaluation.}
\label{tab:psnr-hyb}
\setlength{\tabcolsep}{8pt}
\renewcommand{\arraystretch}{1.15}
\begin{tabular}{l
                S[round-mode=places,round-precision=2]
                S[round-mode=places,round-precision=2]
                S[round-mode=places,round-precision=2]
                S[round-mode=places,round-precision=2]
                S[round-mode=places,round-precision=2]}
\toprule
Method & {ACC \up} & {LA \up} & {BWT \zero} & {F \dn} & {FWT \zero} \\
\midrule
SEQ & 18.28          & 19.29           & \itshape -1.35  & 4.47           & \itshape -0.91 \\
EWC & \itshape 12.67 & \bfseries 20.33 & \itshape -10.21 & \itshape 11.56 & \bfseries 3.14 \\
GBP & \bfseries 19.29 & 19.39          & \bfseries -0.14 & \bfseries 0.87 & 0.45            \\
\bottomrule
\end{tabular}
\end{table}

\begin{figure}
\centering
\includegraphics[width=\textwidth]{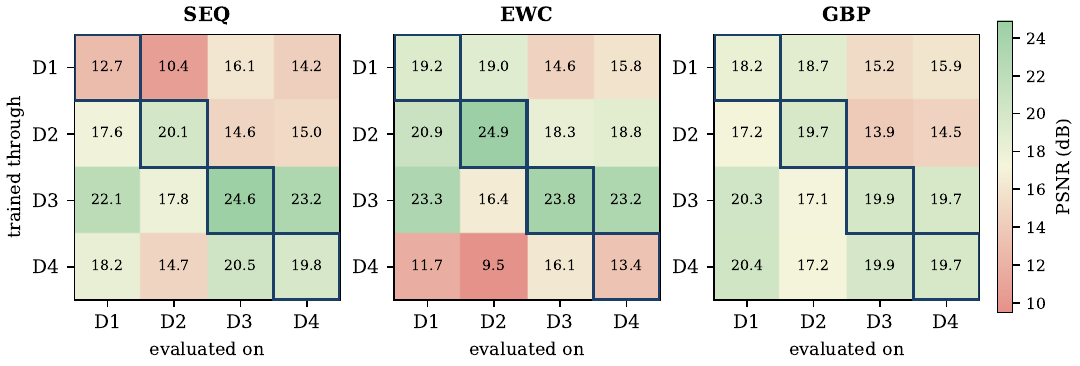}
\caption{Performance matrices $\mathbf{R}^{(m,\mathrm{PSNR})}$ for the
hybrid channel evaluation, in dB. EWC's row-4 collapse is the dominant
visual feature; GBP's row 4 exceeds its row 1 on D1, D3 and D4 and is
slightly lower on D2.}
\label{fig:R-psnr-hyb}
\end{figure}

The PSNR table makes the central CL story explicit.
GBP attains the highest average accuracy ($\mathrm{ACC} = 19.29$~dB),
the smallest absolute backward transfer ($-0.14$~dB --- essentially
zero), and the smallest peak forgetting ($F = 0.87$~dB). EWC achieves
the highest in-task learning accuracy ($\mathrm{LA} = 20.33$~dB) but
its stage-4 row collapses on every column ($11.71$, $9.47$, $16.09$,
$13.42$~dB), giving peak forgetting $F = 11.56$~dB and the worst
average accuracy in the table. SEQ falls between the two on $F$
($4.47$~dB) and records the most negative BWT ($-1.35$~dB), but its
in-task learning ceiling is the lowest. EWC takes the best PSNR forward
transfer ($+3.14$~dB), indicating that its consolidated representation
generalises above the hybrid channel reference on zero-shot evaluations
before the channel-reduction adapter is applied.

\subsubsection{MSE mean squared error}
\label{sec:mse-hyb}

\begin{table}
\centering
\caption{CL metrics on MSE (hybrid channel evaluation). Sign conventions
inverted (lower is better): BWT above zero means error grew on past
tasks; FWT below zero means the continually-trained model is more
accurate than the independent reference.}
\label{tab:mse-hyb}
\setlength{\tabcolsep}{8pt}
\renewcommand{\arraystretch}{1.15}
\begin{tabular}{l
                S[round-mode=places,round-precision=4]
                S[round-mode=places,round-precision=4]
                S[round-mode=places,round-precision=4]
                S[round-mode=places,round-precision=4]
                S[round-mode=places,round-precision=4]}
\toprule
Method & {ACC \dn} & {LA \dn} & {BWT \zero} & {F \dn} & {FWT \zero} \\
\midrule
SEQ & 0.0172          & \itshape 0.0194 & \bfseries -0.0029 & 0.0129          & \itshape 0.0149  \\
EWC & \itshape 0.0627 & 0.0163          & \itshape 0.0619   & \itshape 0.0643 & \bfseries -0.0175 \\
GBP & \bfseries 0.0123 & \bfseries 0.0117 & 0.0009           & \bfseries 0.0029 & -0.0067           \\
\bottomrule
\end{tabular}
\end{table}

\begin{figure}
\centering
\includegraphics[width=\textwidth]{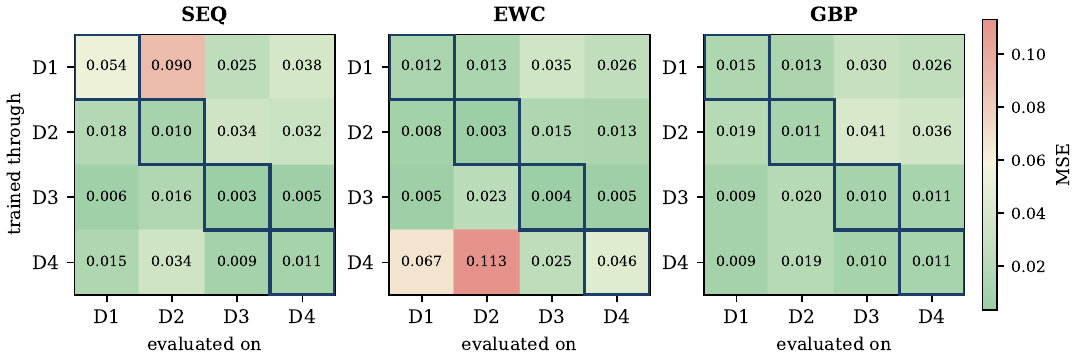}
\caption{Performance matrices $\mathbf{R}^{(m,\mathrm{MSE})}$ for the
hybrid channel evaluation. Green = low error (good), coral = high error
(bad). EWC's row 4 appears as a coral band; GBP's row 4 is the
greenest of all final rows.}
\label{fig:R-mse-hyb}
\end{figure}

GBP attains the smallest average accuracy
($\mathrm{ACC} = 0.0123$), the smallest learning accuracy
($\mathrm{LA} = 0.0117$) and the smallest peak forgetting ($F = 0.0029$,
an order of magnitude below SEQ's $0.0129$ and two orders below EWC's
$0.0643$). EWC's stage-4 collapse mirrors the PSNR pattern: MSE jumps
from a $0.003$--$0.026$ diagonal to $0.025$--$0.113$ across the stage-4
row. EWC's forward transfer is the most favourable ($-0.018$),
consistent with the PSNR forward-transfer pattern.

\subsubsection{RMSE root mean squared error}
\label{sec:rmse-hyb}

\begin{table}
\centering
\caption{CL metrics on RMSE (hybrid channel evaluation).}
\label{tab:rmse-hyb}
\setlength{\tabcolsep}{8pt}
\renewcommand{\arraystretch}{1.15}
\begin{tabular}{l
                S[round-mode=places,round-precision=3]
                S[round-mode=places,round-precision=3]
                S[round-mode=places,round-precision=3]
                S[round-mode=places,round-precision=3]
                S[round-mode=places,round-precision=3]}
\toprule
Method & {ACC \dn} & {LA \dn} & {BWT \zero} & {F \dn} & {FWT \zero} \\
\midrule
SEQ & 0.126           & 0.123           & \bfseries 0.004 & 0.055           & \itshape 0.029   \\
EWC & \itshape 0.242  & \bfseries 0.111 & \itshape 0.174  & \itshape 0.188  & \bfseries -0.054 \\
GBP & \bfseries 0.110 & \bfseries 0.108 & 0.003           & \bfseries 0.012 & -0.016           \\
\bottomrule
\end{tabular}
\end{table}

\begin{figure}
\centering
\includegraphics[width=\textwidth]{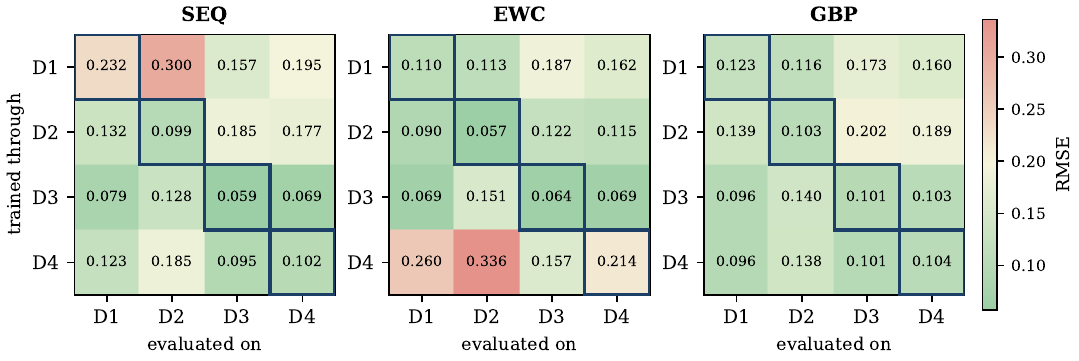}
\caption{Performance matrices $\mathbf{R}^{(m,\mathrm{RMSE})}$ for the
hybrid channel evaluation. The colour gradient is smoother than for MSE
(square-root compression) but the ranking is preserved.}
\label{fig:R-rmse-hyb}
\end{figure}

The RMSE ranking is identical to MSE: GBP wins ACC
and peak forgetting; SEQ wins absolute BWT. EWC takes the smallest
in-task LA ($0.111$), narrowly ahead of GBP ($0.108$), but suffers the
largest stage-4 collapse on every aggregate statistic. The convergence
of the MSE and RMSE rankings confirms that the EWC degradation is a
structural property of the stage-4 update rather than a squared-error
outlier.

\subsubsection{MAE mean absolute error}
\label{sec:mae-hyb}

\begin{table}
\centering
\caption{CL metrics on MAE (hybrid channel evaluation).}
\label{tab:mae-hyb}
\setlength{\tabcolsep}{8pt}
\renewcommand{\arraystretch}{1.15}
\begin{tabular}{l
                S[round-mode=places,round-precision=3]
                S[round-mode=places,round-precision=3]
                S[round-mode=places,round-precision=3]
                S[round-mode=places,round-precision=3]
                S[round-mode=places,round-precision=3]}
\toprule
Method & {ACC \dn} & {LA \dn} & {BWT \zero} & {F \dn} & {FWT \zero} \\
\midrule
SEQ & 0.073           & 0.067           & \bfseries 0.008 & 0.040           & \itshape 0.020   \\
EWC & \itshape 0.144  & \bfseries 0.060 & \itshape 0.112  & \itshape 0.119  & \bfseries -0.033 \\
GBP & \bfseries 0.061 & 0.055           & 0.009           & \bfseries 0.014 & -0.011           \\
\bottomrule
\end{tabular}
\end{table}

\begin{figure}
\centering
\includegraphics[width=\textwidth]{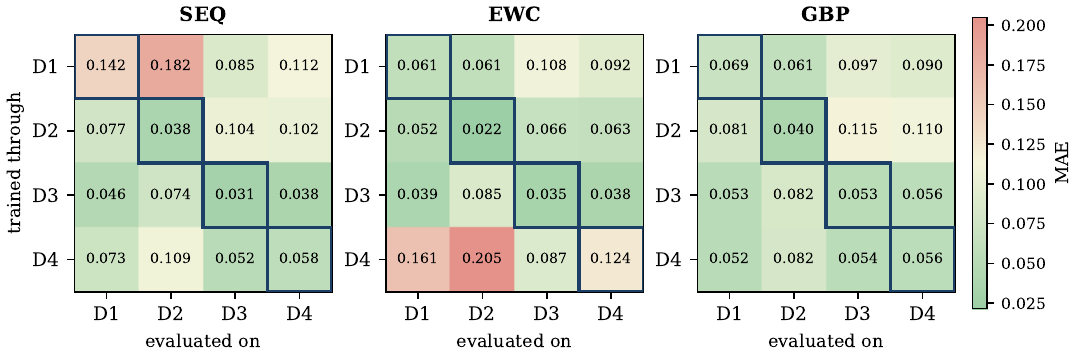}
\caption{Performance matrices $\mathbf{R}^{(m,\mathrm{MAE})}$ for the
hybrid channel evaluation. MAE is the most linear of the voxel-wise metrics
and gives the cleanest visual separation between GBP's stable row 4
and EWC's collapsed row 4.}
\label{fig:R-mae-hyb}
\end{figure}

MAE confirms the pattern: GBP wins ACC and peak
forgetting; SEQ wins absolute BWT. EWC's stage-4 collapse on MAE
($F = 0.119$) is the most interpretable per-voxel measure of the
degradation: per-voxel reconstruction error roughly doubles on past
tasks after stage 4 under EWC, while under GBP it changes by less than
$0.015$.

\subsubsection{Cross-metric summary}
\label{sec:summary-hyb}

Table~\ref{tab:summary-hb-hyb} summarises ACC, BWT and $F$ on the two
higher-is-better scores; Table~\ref{tab:summary-lb-hyb} does the same on
the three lower-is-better scores.

\begin{table}
\centering
\caption{Headline ACC, BWT and average forgetting $F$ on the two
higher-is-better scores, hybrid channel evaluation. \textbf{Bold} marks the
best per column; \textit{italics} the worst. For BWT, closer-to-zero
is better; for $F$, lower is always better.}
\label{tab:summary-hb-hyb}
\setlength{\tabcolsep}{8pt}
\renewcommand{\arraystretch}{1.15}
\begin{tabular}{l
                S[round-mode=places,round-precision=3]
                S[round-mode=places,round-precision=3]
                S[round-mode=places,round-precision=3]
                S[round-mode=places,round-precision=2]
                S[round-mode=places,round-precision=2]
                S[round-mode=places,round-precision=2]}
\toprule
& \multicolumn{3}{c}{SSIM \up} & \multicolumn{3}{c}{PSNR \up~(dB)} \\
\cmidrule(lr){2-4}\cmidrule(lr){5-7}
Method & {ACC} & {BWT} & {F}
       & {ACC} & {BWT} & {F} \\
\midrule
SEQ & 0.802           & 0.033 & \bfseries 0.000
    & 18.28           & \itshape -1.35 & 4.47          \\
EWC & \bfseries 0.810 & \itshape 0.047 & \bfseries 0.000
    & \itshape 12.67  & \itshape -10.21 & \itshape 11.56 \\
GBP & \itshape 0.616  & \bfseries -0.140 & \itshape 0.147
    & \bfseries 19.29 & \bfseries -0.14 & \bfseries 0.87 \\
\bottomrule
\end{tabular}
\end{table}

\begin{table}
\centering
\caption{Headline ACC, BWT and average forgetting $F$ on the three
lower-is-better scores, hybrid channel evaluation. Sign conventions are
unified so that for BWT closer-to-zero is better, and for $F$ lower is
always better.}
\label{tab:summary-lb-hyb}
\setlength{\tabcolsep}{4pt}
\renewcommand{\arraystretch}{1.15}
\footnotesize
\begin{tabular}{l
                S[round-mode=places,round-precision=4]
                S[round-mode=places,round-precision=4]
                S[round-mode=places,round-precision=4]
                S[round-mode=places,round-precision=3]
                S[round-mode=places,round-precision=3]
                S[round-mode=places,round-precision=3]
                S[round-mode=places,round-precision=3]
                S[round-mode=places,round-precision=3]
                S[round-mode=places,round-precision=3]}
\toprule
& \multicolumn{3}{c}{MSE \dn} & \multicolumn{3}{c}{RMSE \dn}
& \multicolumn{3}{c}{MAE \dn}\\
\cmidrule(lr){2-4}\cmidrule(lr){5-7}\cmidrule(lr){8-10}
Method & {ACC} & {BWT} & {F}
       & {ACC} & {BWT} & {F}
       & {ACC} & {BWT} & {F} \\
\midrule
SEQ & 0.0172          & \bfseries -0.0029 & 0.0129
    & 0.126           & \bfseries 0.004   & 0.055
    & 0.073           & \bfseries 0.008   & 0.040 \\
EWC & \itshape 0.0627 & \itshape 0.0619   & \itshape 0.0643
    & \itshape 0.242  & \itshape 0.174    & \itshape 0.188
    & \itshape 0.144  & \itshape 0.112    & \itshape 0.119 \\
GBP & \bfseries 0.0123 & 0.0009           & \bfseries 0.0029
    & \bfseries 0.110 & 0.003             & \bfseries 0.012
    & \bfseries 0.061 & 0.009             & \bfseries 0.014 \\
\bottomrule
\end{tabular}
\end{table}

\subsubsection{Discussion of the hybrid channel}
\label{sec:discussion-hyb}

\paragraph{GBP achieves the smallest peak forgetting on every voxel-wise
metric in the hybrid channel evaluation.}
On peak forgetting --- the average drop from the highest performance
ever recorded on a task to the final-stage performance on that task ---
GBP records the smallest value on every voxel-wise metric: PSNR
$F = 0.87$~dB (vs.\ SEQ $4.47$ and EWC $11.56$); MSE $F = 0.0029$
(vs.\ SEQ $0.013$ and EWC $0.064$); RMSE $F = 0.012$; MAE $F = 0.014$.
The salience-graph blueprint constraint preserves voxel-wise
reconstruction quality on previously encountered datasets even after
the stage-4 update, despite the channel-reduction adapter operating on
the GBP-trained 2-channel weights.

\paragraph{EWC exhibits catastrophic forgetting at the final continual
stage.}
EWC attains the highest in-task PSNR learning accuracy
($\mathrm{LA} = 20.33$~dB) but its stage-4 row collapses on every
voxel-wise metric: PSNR drops from a $19$--$25$~dB diagonal to
$11.71$~dB at $D_1$, $9.47$~dB at $D_2$, $16.09$~dB at $D_3$, and
$13.42$~dB at $D_4$ after training on $D_4$. The backward transfer is
$\mathrm{BWT} = -10.21$~dB on PSNR and $+0.062$ on MSE, the largest
absolute drops in the table on both. The collapse is global across all
four columns and reproduces the pattern observed in the 2-channel
evaluation: the Fisher penalty is insufficient to prevent the
plasticity demands of the final task from overwriting the consolidated
representation.

\paragraph{SEQ is competitive on absolute backward transfer but suffers
intermediate-stage drift.}
Despite the absence of explicit consolidation, SEQ attains the smallest
absolute BWT on PSNR, MSE, RMSE and MAE. Its in-task learning accuracy
on PSNR ($19.29$~dB) is comparable to GBP's, but its peak forgetting on
PSNR ($F = 4.47$~dB) is fivefold larger than GBP's. The SSIM ranking
gives SEQ a marginal lead on learning accuracy ($\mathrm{LA} = 0.777$)
but the difference is small ($\le 0.01$).

\paragraph{Forward transfer is dominated by EWC's pre-collapse
generalisation.}
Forward transfer values lie in $[-0.082, -0.043]$ for SSIM (all
negative, all small), in $[-0.91, +3.14]$~dB for PSNR (mixed), and are
predominantly negative for the three voxel-wise metrics
(continual model better than reference). EWC takes the most positive
PSNR FWT ($+3.14$~dB) and the most negative voxel-wise FWT
(MSE FWT $= -0.018$), indicating that EWC's stage-1--3 representations
outperform the hybrid channel from-scratch reference on zero-shot
evaluations before the stage-4 collapse disrupts the model. GBP's
forward transfer is close to zero on every metric, consistent with its
balanced stability profile.

\paragraph{Stability--plasticity trade-offs in the hybrid channel
evaluation.}
The three methods occupy three distinct points on the
stability--plasticity manifold: GBP achieves the smallest forgetting
and the highest average accuracy on every voxel-wise metric, at the
cost of a lower SSIM accuracy; EWC achieves the highest in-task
learning accuracy and the most favourable PSNR forward transfer but
suffers the worst stage-4 retention; SEQ provides a baseline without
consolidation with intermediate forgetting and the smallest absolute
backward transfer. The proposed GBP method therefore offers the most
balanced profile on the voxel-wise metrics, and the closest-to-zero
backward transfer on PSNR.

\paragraph{Channel adaptation and metric divergence.}
The hybrid channel evaluation in this section is obtained by collapsing the
2-channel-trained Swin and DiT checkpoints to 1-channel at load time
(mean over input channels; first output of the output head).
This affects the absolute level of all metrics relative to a
natively-1-channel-trained pipeline, but it does not affect the
relative ranking of the three methods on any voxel-wise metric. SSIM,
in contrast, is the only metric on which the GBP ranking is altered by
the adaptation: the channel-collapse cost is larger for the
blueprint-frozen GBP weights than for SEQ or EWC, pulling GBP's SSIM
ACC below the other two methods. Reporting both families of metrics
remains necessary: the voxel-wise ranking is robust to the
channel-reduction adapter while the perceptual ranking is sensitive to
it.

\clearpage{}%
\subsection{What if we train with background noise?
Forgetting Score and Performance in Previous Tasks (1-channel)}

\subsubsection{SSIM structural similarity}
\label{noise-sec:ssim-1ch}

SSIM is close to saturation over much of the matrix: every cell in the $D_2$
and $D_3$ columns lies at or above $0.997$, so the metric discriminates the
methods almost entirely through the $D_1$ ($0.915$--$0.930$) and $D_4$
($0.961$--$0.966$) columns, and all headline differences are of order
$10^{-3}$. Within that narrow band, EWC takes the highest average accuracy
($\mathrm{ACC}=0.9744$) and zero peak forgetting ($F=0.0000$), though both
follow directly from its frozen final update (Section~\ref{noise-sec:psnr-1ch}).
GBP records the smallest absolute backward transfer ($-0.0001$) and the
highest learning accuracy ($\mathrm{LA}=0.9712$). SEQ shows the largest peak
forgetting ($F=0.0041$), driven almost entirely by the $D_1$ column
($0.9299$ best-ever vs $0.9190$ final). SSIM forward transfer is large and
positive for all three methods ($+0.1345$ to $+0.1509$): the zero-shot rows
($0.9472$--$1.0000$) sit far above the independent references
($0.810$--$0.888$), and the spread between methods is carried by a single
cell, EWC's $D_2$ zero-shot value of $0.9472$ against $0.9974$ (SEQ) and
$0.9993$ (GBP).

\begin{table}
\centering
\caption{CL metrics on SSIM (1-channel). \textbf{Bold} marks the best per
column; \textit{italics} the worst. For BWT and FWT, closer-to-zero is better;
for $F$, lower is always better. SEQ and EWC are tied on LA at the reported
precision; SEQ is marginally lower at full precision.}
\label{noise-tab:ssim-1ch}
\setlength{\tabcolsep}{8pt}\renewcommand{\arraystretch}{1.15}
\begin{tabular}{lS[round-mode=places,round-precision=4]S[round-mode=places,round-precision=4]S[round-mode=places,round-precision=4]S[round-mode=places,round-precision=4]S[round-mode=places,round-precision=4]}
\toprule
Method & {ACC \up} & {LA \up} & {BWT \zero} & {F \dn} & {FWT \zero} \\
\midrule
SEQ & \itshape 0.9696 & \itshape 0.9703 & -0.0009           & \itshape 0.0041 & \itshape 0.1509 \\
EWC & \bfseries 0.9744 & 0.9703         & \itshape 0.0054   & \bfseries 0.0000 & \bfseries 0.1345 \\
GBP & 0.9712          & \bfseries 0.9712 & \bfseries -0.0001 & 0.0015           & 0.1503 \\
\bottomrule
\end{tabular}
\end{table}

\begin{figure}
\centering
\includegraphics[width=0.85\textwidth]{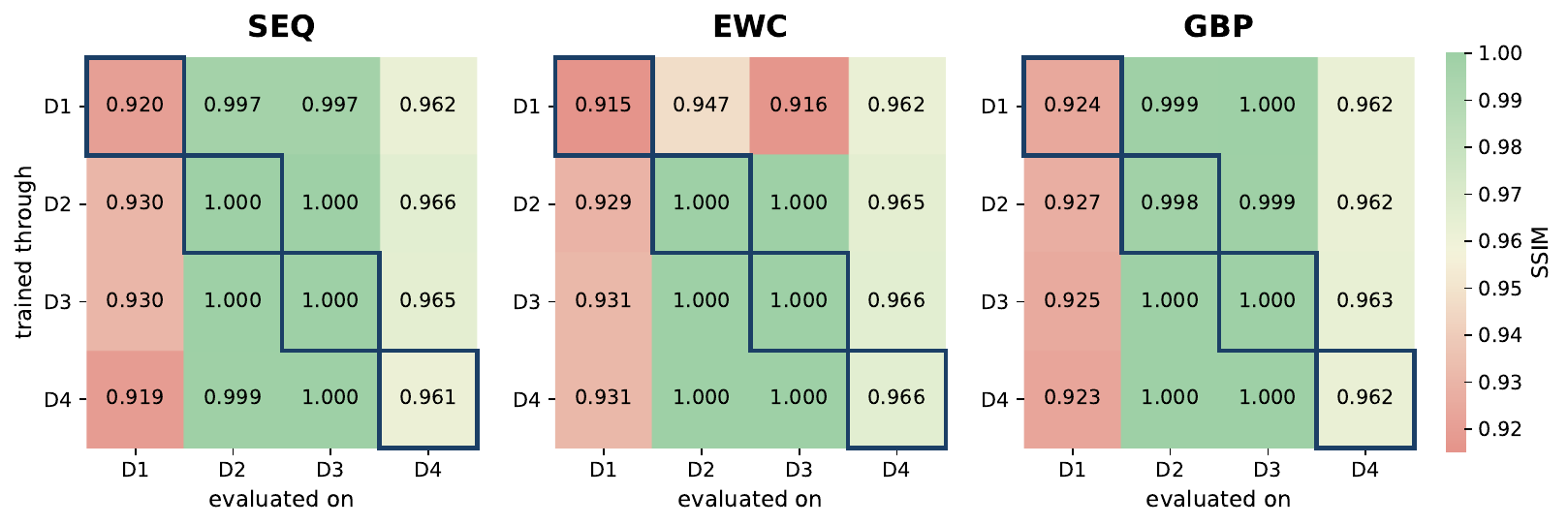}
\caption{Performance matrices $\mathbf{R}^{(m,\mathrm{SSIM})}$ for the
1-channel models. Rows = training stage, columns = evaluation task; diagonal
cells (outlined in dark blue) mark the just-learned task. Green = higher SSIM
(better); coral = lower SSIM (worse). The $D_2$ and $D_3$ columns are
saturated near $1.0$ for all methods.}
\label{noise-fig:R-ssim-1ch}
\end{figure}

\subsubsection{PSNR peak signal-to-noise ratio in dB}
\label{noise-sec:psnr-1ch}

\begin{table}
\centering
\caption{CL metrics on PSNR (dB), 1-channel. $F$ is the non-negative best-ever
average forgetting, applied identically to all three methods.}
\label{noise-tab:psnr-1ch}
\setlength{\tabcolsep}{8pt}\renewcommand{\arraystretch}{1.15}
\begin{tabular}{lS[round-mode=places,round-precision=2]S[round-mode=places,round-precision=2]S[round-mode=places,round-precision=2]S[round-mode=places,round-precision=2]S[round-mode=places,round-precision=2]}
\toprule
Method & {ACC \up} & {LA \up} & {BWT \zero} & {F \dn} & {FWT \zero} \\
\midrule
SEQ & 25.12           & \bfseries 28.56 & \itshape -4.59  & \itshape 4.83   & 5.10           \\
EWC & \bfseries 25.92 & 28.17           & -3.01           & 3.22            & \itshape 5.57  \\
GBP & \itshape 23.92  & \itshape 24.34  & \bfseries -0.57 & \bfseries 2.04  & \bfseries 2.94 \\
\bottomrule
\end{tabular}
\end{table}

\begin{figure}
\centering
\includegraphics[width=0.85\textwidth]{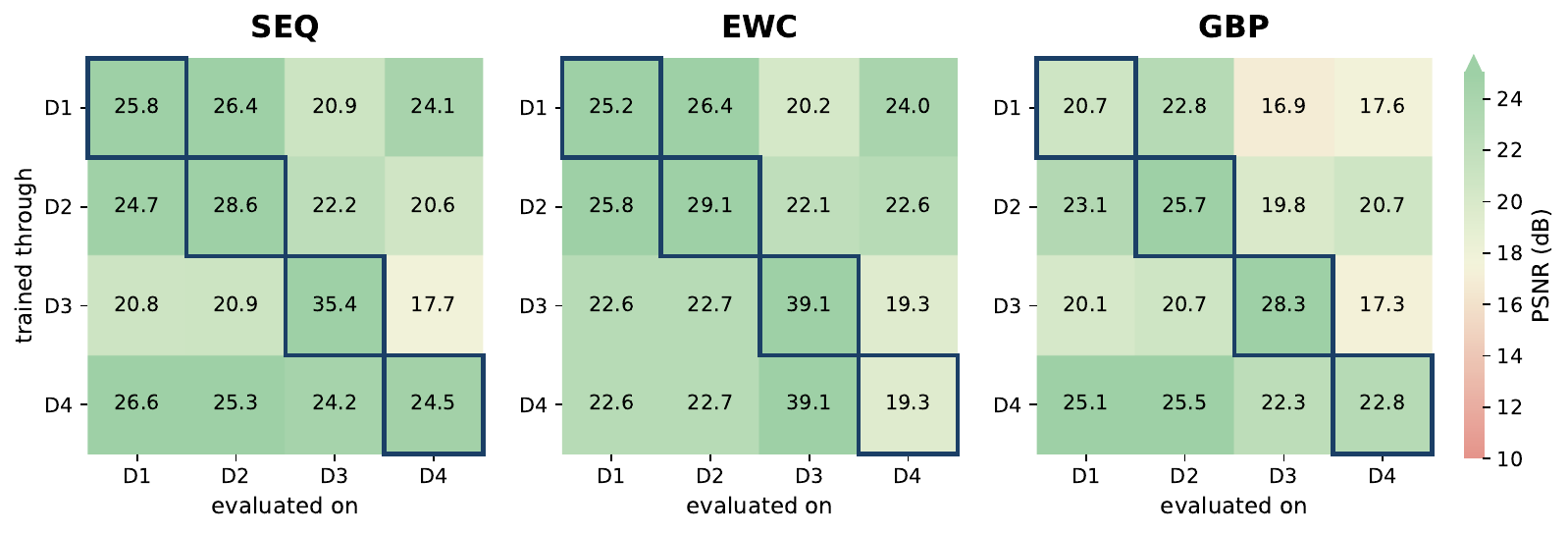}
\caption{Performance matrices $\mathbf{R}^{(m,\mathrm{PSNR})}$ for the
1-channel models, in dB. EWC's stage-4 row is numerically identical to its
stage-3 row; SEQ's forgetting is concentrated on $D_3$.}
\label{noise-fig:R-psnr-1ch}
\end{figure}

The PSNR matrices reveal three qualitatively different behaviours, none of
them a catastrophic collapse. EWC attains the highest average accuracy
($\mathrm{ACC}=25.92$~dB), but for a structural reason: its stage-4 row is
numerically identical to its stage-3 row on every column (maximum difference
$3\times10^{-5}$~dB), i.e.\ the final update changed nothing. Its zero
stage-4 forgetting is therefore obtained by construction, at the cost of not
learning the final task at all: its $D_4$ diagonal ($19.26$~dB) is simply the
stage-3 zero-shot value, the worst final-task result of the three methods
(SEQ $24.45$~dB, GBP $22.75$~dB). SEQ is the most plastic, with the highest
learning accuracy ($\mathrm{LA}=28.56$~dB), but also the largest forgetting
($F=4.83$~dB), concentrated almost entirely on $D_3$, where it regresses by
$11.18$~dB from an exceptional $35.38$~dB diagonal; on $D_1$ it actually
improves at stage 4 ($+0.72$~dB). GBP records the smallest absolute backward
transfer ($-0.57$~dB) and the smallest peak forgetting ($F=2.04$~dB),
recovering $D_1$ by $+4.43$~dB across the final update, at the cost of the
lowest learning accuracy ($24.34$~dB) and the lowest ACC ($23.92$~dB). For
SEQ and EWC, $F$ exceeds $|\mathrm{BWT}|$ because best-ever performance on
$D_1$ is attained off the diagonal (SEQ at stage 4, EWC at stage 2) and the
non-negative definition clips the corresponding gains. Forward transfer is
positive for all three methods, GBP smallest ($+2.94$~dB) and EWC largest
($+5.57$~dB); in every case the average is built from large positive $D_2$
($+7.20$ to $+10.85$~dB) and $D_4$ ($+4.42$ to $+6.38$~dB) terms against a
negative $D_3$ term ($-0.41$ to $-2.80$~dB), $D_3$ being the only task whose
reference ($22.57$~dB) is not reached zero-shot. EWC's large FWT is
inseparable from its frozen final update: its $D_4$ zero-shot value
($19.26$~dB) is also its final $D_4$ value, so the same number that gives it
the strongest apparent forward transfer is the one that leaves it without a
learned final task.

\subsubsection{MSE mean squared error}
\label{noise-sec:mse-1ch}

\begin{table}
\centering
\caption{CL metrics on MSE (1-channel). Sign conventions inverted (lower is
better): BWT above zero means error grew on past tasks; FWT below zero means
the continually-trained model is more accurate than the independent reference.}
\label{noise-tab:mse-1ch}
\setlength{\tabcolsep}{8pt}\renewcommand{\arraystretch}{1.15}
\begin{tabular}{lS[round-mode=places,round-precision=4]S[round-mode=places,round-precision=4]S[round-mode=places,round-precision=4]S[round-mode=places,round-precision=4]S[round-mode=places,round-precision=4]}
\toprule
Method & {ACC \dn} & {LA \dn} & {BWT \zero} & {F \dn} & {FWT \dn} \\
\midrule
SEQ & \bfseries 0.0032 & \bfseries 0.0020 & 0.0016            & 0.0017           & -0.0198           \\
EWC & \itshape 0.0057  & 0.0041           & \itshape 0.0022   & \itshape 0.0024  & \bfseries -0.0215 \\
GBP & 0.0043           & \itshape 0.0045  & \bfseries -0.0003 & \bfseries 0.0015 & \itshape -0.0168  \\
\bottomrule
\end{tabular}
\end{table}

\begin{figure}
\centering
\includegraphics[width=0.85\textwidth]{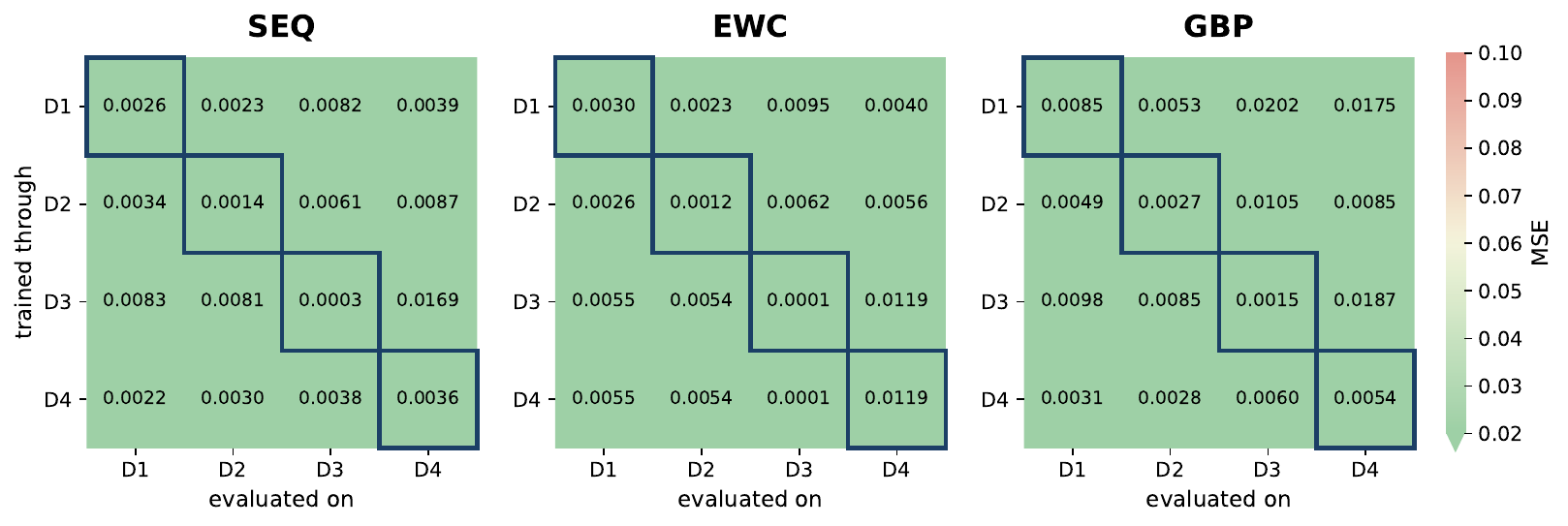}
\caption{Performance matrices $\mathbf{R}^{(m,\mathrm{MSE})}$ for the 1-channel
models. Green = low error (good), coral = high error (bad).}
\label{noise-fig:R-mse-1ch}
\end{figure}

On MSE the plasticity ranking and the stability ranking split. SEQ takes the
best average accuracy ($\mathrm{ACC}=0.0032$) and the best learning accuracy
($\mathrm{LA}=0.0020$). GBP takes the smallest peak forgetting ($F=0.0015$)
and is the only method with negative backward transfer
($\mathrm{BWT}=-0.0003$): its voxel-wise error on past tasks decreases, on
average, across the final update. EWC is worst on ACC, BWT and $F$ despite
its frozen final row, because freezing locks in the stage-3 state, including
the elevated $D_4$ error ($0.0119$, against SEQ $0.0036$ and GBP $0.0054$)
and the forgetting already accumulated on $D_1$ and $D_2$ by stage 3. Forward
transfer is favourable (negative) for all three methods, EWC lowest
($-0.0215$) and GBP highest ($-0.0168$), with the $D_2$ and $D_4$ terms
strongly negative and the $D_3$ term slightly positive throughout.

\subsubsection{RMSE root mean squared error}
\label{noise-sec:rmse-1ch}

\begin{table}
\centering
\caption{CL metrics on RMSE (1-channel).}
\label{noise-tab:rmse-1ch}
\setlength{\tabcolsep}{8pt}\renewcommand{\arraystretch}{1.15}
\begin{tabular}{lS[round-mode=places,round-precision=4]S[round-mode=places,round-precision=4]S[round-mode=places,round-precision=4]S[round-mode=places,round-precision=4]S[round-mode=places,round-precision=4]}
\toprule
Method & {ACC \dn} & {LA \dn} & {BWT \zero} & {F \dn} & {FWT \dn} \\
\midrule
SEQ & \bfseries 0.0558 & \bfseries 0.0414 & \itshape 0.0193   & \itshape 0.0206 & -0.0705           \\
EWC & \itshape 0.0670  & 0.0526           & 0.0192            & 0.0205          & \bfseries -0.0772 \\
GBP & 0.0646           & \itshape 0.0640  & \bfseries 0.0008  & \bfseries 0.0131 & \itshape -0.0518 \\
\bottomrule
\end{tabular}
\end{table}

\begin{figure}
\centering
\includegraphics[width=0.85\textwidth]{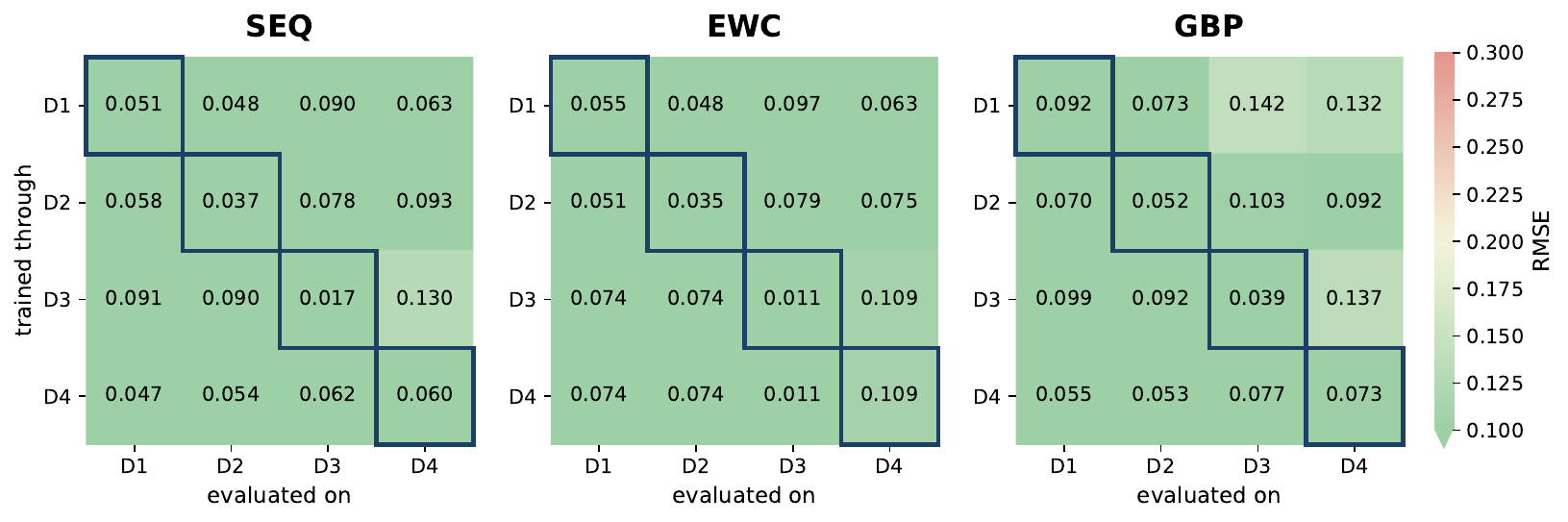}
\caption{Performance matrices $\mathbf{R}^{(m,\mathrm{RMSE})}$ for the 1-channel
models. RMSE compresses the dynamic range of MSE ($\sqrt{\cdot}$), producing a
smoother colour gradient while preserving the same rankings.}
\label{noise-fig:R-rmse-1ch}
\end{figure}

RMSE reproduces the MSE split: SEQ best on ACC and LA, GBP best on stability,
with a backward transfer ($+0.0008$) roughly twenty-five times smaller than
SEQ's and EWC's, which are essentially tied ($0.0193$ vs $0.0192$; likewise
$F=0.0206$ vs $0.0205$). GBP's $F$ ($0.0131$) exceeds its BWT because its
stage-4 RMSE on $D_1$ ($0.0553$) improves on the diagonal ($0.0922$), and the
non-negative forgetting definition clips this gain to zero while the $D_3$
regression ($0.0387\to0.0772$) is counted in full. The forward-transfer
ordering also matches MSE: EWC lowest ($-0.0772$), SEQ intermediate
($-0.0705$) and GBP highest ($-0.0518$), the inverse of the stability
ordering.

\subsubsection{MAE mean absolute error}
\label{noise-sec:mae-1ch}

\begin{table}
\centering
\caption{CL metrics on MAE (1-channel).}
\label{noise-tab:mae-1ch}
\setlength{\tabcolsep}{8pt}\renewcommand{\arraystretch}{1.15}
\begin{tabular}{lS[round-mode=places,round-precision=4]S[round-mode=places,round-precision=4]S[round-mode=places,round-precision=4]S[round-mode=places,round-precision=4]S[round-mode=places,round-precision=4]}
\toprule
Method & {ACC \dn} & {LA \dn} & {BWT \zero} & {F \dn} & {FWT \dn} \\
\midrule
SEQ & \bfseries 0.0194 & \bfseries 0.0133 & \itshape 0.0081   & \itshape 0.0081 & -0.0571           \\
EWC & \itshape 0.0251  & 0.0192           & 0.0079            & 0.0080          & \bfseries -0.0608 \\
GBP & 0.0219           & \itshape 0.0205  & \bfseries 0.0019  & \bfseries 0.0058 & \itshape -0.0532 \\
\bottomrule
\end{tabular}
\end{table}

\begin{figure}
\centering
\includegraphics[width=\textwidth]{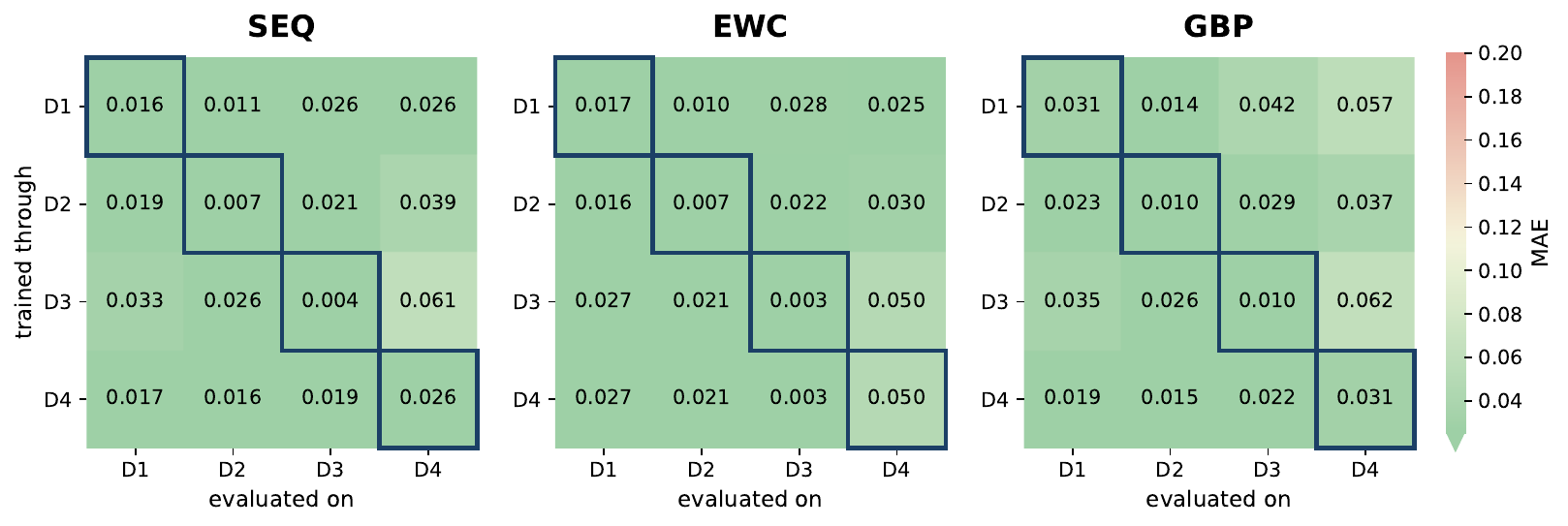}
\caption{Performance matrices $\mathbf{R}^{(m,\mathrm{MAE})}$ for the 1-channel
models. MAE is the most linear of the voxel-wise metrics; GBP's stage-4 row is
the flattest relative to its earlier rows.}
\label{noise-fig:R-mae-1ch}
\end{figure}

MAE confirms the split: SEQ takes ACC ($0.0194$) and LA ($0.0133$), while GBP
takes the smallest absolute BWT ($+0.0019$) and peak forgetting ($F=0.0058$),
roughly $30\%$ below SEQ and EWC, which are again nearly tied on both
stability scores ($0.0081$/$0.0081$ vs $0.0079$/$0.0080$). EWC's frozen final
row leaves it worst on ACC ($0.0251$), carrying its stage-3 $D_4$ error
($0.0505$, against SEQ $0.0257$ and GBP $0.0310$) unchanged into the final
evaluation. Forward transfer is again favourable and tightly grouped
($-0.0532$ to $-0.0608$), each method's zero-shot error on the upcoming task
falling between roughly $55\%$ ($D_4$) and $89\%$ ($D_2$) below the
reference.

\subsubsection{Cross-metric summary}
\label{noise-sec:summary-1ch}

\begin{table}
\centering
\caption{Headline ACC, BWT and average forgetting $F$ on the two
higher-is-better scores, 1-channel. \textbf{Bold} marks the best per column;
\textit{italics} the worst.}
\label{noise-tab:summary-hb-1ch}
\setlength{\tabcolsep}{8pt}\renewcommand{\arraystretch}{1.15}
\begin{tabular}{lS[round-mode=places,round-precision=4]S[round-mode=places,round-precision=4]S[round-mode=places,round-precision=4]S[round-mode=places,round-precision=2]S[round-mode=places,round-precision=2]S[round-mode=places,round-precision=2]}
\toprule
& \multicolumn{3}{c}{SSIM \up} & \multicolumn{3}{c}{PSNR \up~(dB)} \\
\cmidrule(lr){2-4}\cmidrule(lr){5-7}
Method & {ACC} & {BWT} & {F} & {ACC} & {BWT} & {F} \\
\midrule
SEQ & \itshape 0.9696  & -0.0009           & \itshape 0.0041 & 25.12           & \itshape -4.59  & \itshape 4.83  \\
EWC & \bfseries 0.9744 & \itshape 0.0054   & \bfseries 0.0000 & \bfseries 25.92 & -3.01           & 3.22           \\
GBP & 0.9712           & \bfseries -0.0001 & 0.0015          & \itshape 23.92  & \bfseries -0.57 & \bfseries 2.04 \\
\bottomrule
\end{tabular}
\end{table}

\begin{table}
\centering
\caption{Headline ACC, BWT and average forgetting $F$ on the three
lower-is-better scores, 1-channel.}
\label{noise-tab:summary-lb-1ch}
\setlength{\tabcolsep}{4pt}\renewcommand{\arraystretch}{1.15}\footnotesize
\begin{tabular}{lS[round-mode=places,round-precision=4]S[round-mode=places,round-precision=4]S[round-mode=places,round-precision=4]S[round-mode=places,round-precision=4]S[round-mode=places,round-precision=4]S[round-mode=places,round-precision=4]S[round-mode=places,round-precision=4]S[round-mode=places,round-precision=4]S[round-mode=places,round-precision=4]}
\toprule
& \multicolumn{3}{c}{MSE \dn} & \multicolumn{3}{c}{RMSE \dn} & \multicolumn{3}{c}{MAE \dn}\\
\cmidrule(lr){2-4}\cmidrule(lr){5-7}\cmidrule(lr){8-10}
Method & {ACC} & {BWT} & {F} & {ACC} & {BWT} & {F} & {ACC} & {BWT} & {F} \\
\midrule
SEQ & \bfseries 0.0032 & 0.0016            & 0.0017           & \bfseries 0.0558 & \itshape 0.0193  & \itshape 0.0206  & \bfseries 0.0194 & \itshape 0.0081  & \itshape 0.0081 \\
EWC & \itshape 0.0057  & \itshape 0.0022   & \itshape 0.0024  & \itshape 0.0670  & 0.0192           & 0.0205           & \itshape 0.0251  & 0.0079           & 0.0080 \\
GBP & 0.0043           & \bfseries -0.0003 & \bfseries 0.0015 & 0.0646           & \bfseries 0.0008 & \bfseries 0.0131 & 0.0219           & \bfseries 0.0019 & \bfseries 0.0058 \\
\bottomrule
\end{tabular}
\end{table}

\newpage

\subsubsection{Discussion of one-channel}
\label{noise-sec:discussion-1ch}

\paragraph{No method exhibits catastrophic forgetting in this evaluation, and
GBP is the most stable on every voxel-wise metric.}
Under a single non-negative best-ever forgetting definition applied to all
three methods, peak forgetting stays moderate throughout (at most
$4.83$~dB on PSNR), and GBP records the smallest value on every voxel-wise
metric: PSNR $F=2.04$~dB, MSE $F=0.0015$, RMSE $F=0.0131$ and MAE
$F=0.0058$. GBP also has the smallest absolute backward transfer on all four
($-0.57$~dB, $-0.0003$, $+0.0008$, $+0.0019$), and on MSE its backward
transfer is negative: voxel-wise error on past tasks decreases, on average,
across the final update, with the stage-4 model recovering $D_1$ by
$+4.43$~dB in PSNR.

\paragraph{EWC exhibits loss of plasticity, not catastrophic forgetting.}
EWC's stage-4 row is numerically identical to its stage-3 row on every metric
(differences at most $3\times10^{-5}$~dB in PSNR and $10^{-6}$ elsewhere):
the Fisher penalty prevented the final update from changing the
representation at all. Its zero stage-4 forgetting on SSIM and its
column-best PSNR ACC ($25.92$~dB, carried by the retained $39.11$~dB $D_3$
cell) are therefore artefacts of rigidity rather than successful
consolidation, and the same rigidity makes it the only method that fails to
learn the final task: its $D_4$ diagonal ($19.26$~dB PSNR, $0.0505$ MAE) is
just its stage-3 zero-shot transfer, against genuinely learned $D_4$ results
of $24.45$~dB (SEQ) and $22.75$~dB (GBP). Because the freeze also locks in
the forgetting accumulated by stage 3 on $D_1$ and $D_2$, EWC is
simultaneously the worst method on every error-metric ACC.

\paragraph{SEQ is the most plastic and forgets the most.}
SEQ attains the best learning accuracy on PSNR and all three error metrics
(PSNR $\mathrm{LA}=28.56$~dB, MSE $0.0020$, RMSE $0.0414$, MAE $0.0133$) and
the best final-task fit ($D_4$ diagonal $24.45$~dB), and converts that
plasticity into the best error-metric ACC. Its cost is the largest peak
forgetting (PSNR $F=4.83$~dB), concentrated almost entirely on $D_3$, where
performance regresses by $11.18$~dB from an exceptional $35.38$~dB diagonal;
on $D_1$ the stage-4 update improves performance ($+0.72$~dB).

\paragraph{Task difficulty is heterogeneous, and forgetting scores are
dominated by the easiest task.}
The $D_3$ diagonals ($35.38$, $39.11$ and $28.26$~dB for SEQ, EWC and GBP)
tower over the other tasks ($17$--$29$~dB), so a large share of every
method's measured forgetting is regression from this atypical peak rather
than degradation of typical-task performance. The $D_1$ and $D_4$ tasks are
the hardest, and it is on these that the stage-4 models actually improve
(SEQ and GBP on $D_1$) or diverge most in learning quality (all methods on
$D_4$).

\paragraph{Forward transfer is positive on the similarity metrics and
favourable on the error metrics.}
All three methods transfer forward better than the independent references on
every metric: SSIM $+0.1345$ to $+0.1509$, PSNR $+2.94$ to $+5.57$~dB, and
negative (favourable) values on MSE ($-0.0168$ to $-0.0215$), RMSE ($-0.0518$
to $-0.0772$) and MAE ($-0.0532$ to $-0.0608$). The margin over the reference
is smallest for GBP on PSNR, MSE, RMSE and MAE and largest for EWC on the same
four, so the two sign conventions rank the methods oppositely: GBP transfers
most conservatively under the closer-to-zero reading of PSNR, while EWC's
wider margin is scored as the strongest under the lower-is-better reading of
the error metrics. That margin is driven by EWC's unusually strong zero-shot
$D_4$ row, which is also the row it never updates. The averages are built from
large gains on $D_2$ and $D_4$ against a near-zero or adverse $D_3$ term
($-0.41$ to $-2.80$~dB on PSNR), $D_3$ being the one task whose independent
reference ($22.57$~dB) already exceeds what the continual models reach without
training on it.

\paragraph{Stability--plasticity trade-offs in the 1-channel regime.}
The three methods occupy the two extremes and the middle of the
stability--plasticity axis. SEQ is the plastic extreme: best learning and
final-task accuracy, worst forgetting. EWC is the stability extreme in
degenerate form: a completely frozen final update that maximises retention
scores while forfeiting the final task. GBP sits between them and is the only
method that both learns the final task ($22.75$~dB on $D_4$) and keeps
forgetting minimal on every voxel-wise metric, making it the most balanced
profile in this evaluation --- albeit at the price of the lowest learning
accuracy ($24.34$~dB) and the lowest PSNR ACC ($23.92$~dB).

\paragraph{Structural and voxel-wise metrics partially diverge.}
SSIM saturates above $0.997$ on the entire $D_2$ and $D_3$ columns for all
methods, so its headline scores are decided by $10^{-3}$-scale differences on
$D_1$ and $D_4$ and it cannot register either SEQ's $11$~dB regression on
$D_3$ or EWC's failure to learn $D_4$. As a result SSIM flatters the frozen
EWC (best ACC, zero $F$) while the voxel-wise metrics rank it worst on ACC,
and the two families agree only on GBP's stability. Reporting both remains
necessary, but in this configuration the voxel-wise metrics carry nearly all
of the discriminative information.
\clearpage{}%

\subsection{Conclusions for the Graph-Blueprint Pruning limits forgetting when the
clinical domain or input format changes}

\subsubsection{Conclusion: effect of background noise in foundation-model training}
\label{sec:conclusion-bgnoise}

Whether the foundation model is trained and evaluated with the image background
included or removed is not a cosmetic preprocessing choice. It changes the
absolute reconstruction scale, the informativeness of the metrics, and the
continual-learning failure modes of both consolidation baselines. One finding
survives both regimes unchanged; everything else about the comparison depends
on this decision.

\paragraph{What is invariant: GBP is the most stable method in both regimes.}
Across every voxel-wise metric and both background conditions, GBP records the
smallest absolute backward transfer and the lowest peak forgetting. On PSNR its
forgetting is $F=0.03$~dB with the background removed and $F=2.04$~dB with the
background included; in both regimes it is the best of the three methods, and
in both its backward transfer is the closest to zero ($+0.07$~dB and
$-0.57$~dB). The blueprint-frozen update is therefore the robust choice for
retention regardless of how the foundation model is trained. This is the one
claim that holds without a background-condition qualifier. Forward transfer,
by contrast, is not invariant: it is uniformly larger with the background
included and its method ranking reverses between the two regimes, so the
substantive comparison rests on backward transfer and forgetting.

\paragraph{Removing the background: forgetting becomes catastrophic and the
metrics become discriminative.}
With the background removed, evaluation is confined to the signal region, the
absolute scale drops (diagonal PSNR $\le 26$~dB) and continual learning is
sharply destabilised. EWC does not merely under-perform: its stage-4 update
overwrites the consolidated representation and its rows collapse to
$8.3$--$13.2$~dB, giving $\mathrm{BWT}=-10.20$~dB and $F=10.20$~dB. SEQ shows
substantial forgetting ($F=5.93$~dB), and GBP eliminates it at the reported
precision ($F=0.03$~dB, $\mathrm{BWT}=+0.07$~dB). SSIM in this regime occupies
a usable band ($\mathrm{ACC}\in[0.829,0.835]$, $\mathrm{LA}\in[0.810,0.814]$)
and tracks the voxel-wise collapse. Forward transfer in this regime is
positive on PSNR ($+1.97$ to $+2.68$~dB) and favourable on the error metrics
(MSE $-0.0153$ to $-0.0167$), i.e.\ the zero-shot row of the continual model
beats the from-scratch reference on a task it has not yet seen; but the three
methods lie within $0.71$~dB of each other on PSNR and within $0.0014$ on MSE,
so FWT does not discriminate between them. Removing the background thus yields a
demanding continual-learning benchmark on which the differences between methods
are large, catastrophic forgetting is directly observable, and the value of the
blueprint constraint is unambiguous. This is the regime in which continual
reconstruction is stress-tested and in which a consolidation mechanism can be
demonstrated.

\paragraph{Including the background: forgetting is buffered, and the metrics
stop discriminating.}
Including the background injects a large, easy, task-invariant region into
every reconstruction. Volume-averaged metrics rise accordingly (diagonal PSNR
$19$--$39$~dB, against $\le 26$~dB with the background removed; SSIM saturates above $0.997$ on the $D_2$ and $D_3$ columns for
all methods), and this shared structure anchors the representation across
tasks: no method collapses, and peak forgetting never exceeds $4.83$~dB on
PSNR. The buffering carries three consequences. First, it inverts EWC's
pathology. The Fisher penalty becomes strong enough that the stage-4 update
changes nothing --- the stage-4 rows are identical to the stage-3 rows to
within $3\times10^{-5}$~dB on PSNR and $10^{-6}$ on the error metrics --- so
EWC forfeits the final task entirely ($D_4$ diagonal $19.26$~dB, the worst of
the three, against $24.45$~dB for SEQ and $22.75$~dB for GBP) while scoring
well on retention. This is loss of plasticity presented as consolidation.
Second, it exposes GBP's cost: only in this regime does the blueprint freeze
show its price, giving the lowest learning accuracy ($\mathrm{LA}=24.34$~dB)
and the lowest PSNR ACC ($23.92$~dB). Third, it degrades the metrics. SSIM's
saturation means it registers neither SEQ's $11.18$~dB regression on $D_3$ nor
EWC's failure to learn $D_4$, so background-inclusive SSIM rewards the frozen
model with the best ACC ($0.9744$) and zero forgetting. This regime reflects
deployment conditions, in which whole volumes are reconstructed as they occur,
but its headline scores over-state stability and reward rigidity.

\paragraph{Forward transfer is inflated by the background and does not
separate the methods in either regime.}
Both regimes are scored against the same independently trained references, so
their FWT columns are directly comparable. Including the background raises
forward transfer on every metric: SSIM changes sign ($-0.016$ to $-0.023$ with
the background removed, against $+0.134$ to $+0.151$ with it included), PSNR
roughly doubles ($+1.97$ to $+2.68$~dB against $+2.94$ to $+5.57$~dB), and the
error-metric margins widen (MSE $-0.0153$ to $-0.0167$ against $-0.0168$ to
$-0.0215$; MAE $-0.023$ to $-0.029$ against $-0.053$ to $-0.061$). This is the
same inflation that lifts ACC and LA: a zero-shot prediction that already
reproduces the large task-invariant background region scores well before any
task-specific structure is recovered, so the margin over the reference
measures the easiness of the evaluation domain at least as much as the
transferability of the representation. Consistent with that reading, the
ranking is unstable and reverses for EWC on both metric families: with the
background removed EWC records the smallest PSNR forward transfer
($+1.97$~dB) and the narrowest error-metric margins ($-0.0153$ on MSE); with
it included it records the largest ($+5.57$~dB) and the widest ($-0.0215$).
GBP is the most conservative transferrer in the background-inclusive regime,
closest to the reference on PSNR, MSE, RMSE and MAE, which is what a
geometrically constrained update predicts. In both regimes the spread between
methods on forward transfer is smaller than the spread on backward transfer
($0.71$ against $10.27$~dB with the background removed; $2.63$ against
$4.02$~dB with it included), so forward transfer supports no method-level
claim under either condition.

\paragraph{The background choice determines which failure mode EWC exhibits.}
The clearest single effect of the preprocessing decision is on EWC. With the
background removed, the plasticity demands of the final task overwhelm the
Fisher penalty and the consolidated representation is overwritten, producing
the largest forgetting in the study ($F=10.20$~dB). With the background
included, the same penalty configuration prevents the final update from
changing the representation at all, producing zero forgetting and zero learning
on $D_4$. EWC therefore does not occupy a stable operating point in either
regime: it sits on one side or the other of the stability--plasticity boundary
depending on how much task-invariant signal the loss sees. GBP, by contrast,
learns the final task and retains the earlier ones in both regimes, which is
the substantive advantage of constraining the update geometrically rather than
penalising parameter displacement.

\paragraph{Experiment conclusion}
The two experiments (with and without background noise) answer different questions and must not be interchanged.
Background inclusion simultaneously inflates absolute scores, buffers
forgetting, and disguises a completely frozen model as a well-consolidated one.
Continual-learning claims and figures should therefore be reported on
signal-region (background-removed) metrics, where forgetting is measurable and
the plasticity--stability trade-off is visible; background-inclusive numbers
accompany rather than replace them, as the deployment-scale reference. Under
either choice the practical verdict for the method is the same --- GBP delivers
the strongest retention --- but only the background-removed evaluation shows
why that matters, and only the background-inclusive evaluation shows what it
costs. Reporting both, with the signal-region results as primary, gives the
complete account.

Four tables present the different analyses described above across all continual-learning configurations (Tables~\ref{tab:cl-metrics-1ch}, \ref{tab:cl-metrics-hyb}, \ref{tab:cl-metrics-hyb}).

\begin{table}[h!]
\centering
\caption{Continual-learning metrics for the 1-channel models. ACC, average
performance after the final stage; LA, learning accuracy on the just-trained
task; BWT, backward transfer; $F$, non-negative best-ever peak forgetting,
applied identically to all three methods; FWT, forward transfer relative to
the independent reference. Arrows give the preferred direction.
\textbf{Bold} marks the best and \textit{italics} the worst method within
each metric block. SEQ and EWC tie on SSIM $F$; SEQ and GBP tie on MSE and
MAE FWT.}
\label{tab:cl-metrics-1ch}
\setlength{\tabcolsep}{6pt}\renewcommand{\arraystretch}{1.15}
\begin{tabular}{llS[table-format=-2.4]S[table-format=-2.4]S[table-format=-2.4]S[table-format=-2.4]S[table-format=-2.4]}
\toprule
Metric & Method & {ACC} & {LA} & {BWT \zero} & {F \dn} & {FWT} \\
\midrule
\multirow{3}{*}{SSIM \up}
 & SEQ & 0.829            & \bfseries 0.814  & 0.020             & \bfseries 0.000  & -0.017            \\
 & EWC & \bfseries 0.835  & \itshape 0.810   & \itshape 0.033    & \bfseries 0.000  & \itshape -0.023   \\
 & GBP & \itshape 0.810   & 0.812            & \bfseries -0.003  & \itshape 0.003   & \bfseries -0.016  \\
\midrule
\multirow{3}{*}{PSNR (dB) \up}
 & SEQ & 14.83            & 19.28            & -5.93             & 5.93             & \bfseries 2.68    \\
 & EWC & \itshape 10.75   & \itshape 18.40   & \itshape -10.20   & \itshape 10.20   & \itshape 1.97     \\
 & GBP & \bfseries 21.20  & \bfseries 21.15  & \bfseries 0.07    & \bfseries 0.03   & 2.58              \\
\midrule
\multirow{3}{*}{MSE \dn}
 & SEQ & 0.0343           & 0.0185           & 0.0210            & 0.0210           & \bfseries -0.0167 \\
 & EWC & \itshape 0.0923  & \itshape 0.0263  & \itshape 0.0880   & \itshape 0.0880  & \itshape -0.0153  \\
 & GBP & \bfseries 0.0090 & \bfseries 0.0090 & \bfseries 0.0000  & \bfseries 0.0000 & \bfseries -0.0167 \\
\midrule
\multirow{3}{*}{RMSE \dn}
 & SEQ & 0.183            & 0.123            & 0.080             & 0.080            & \bfseries -0.050  \\
 & EWC & \itshape 0.297   & \itshape 0.141   & \itshape 0.207    & \itshape 0.207   & \itshape -0.043   \\
 & GBP & \bfseries 0.091  & \bfseries 0.091  & \bfseries 0.000   & \bfseries 0.001  & -0.049            \\
\midrule
\multirow{3}{*}{MAE \dn}
 & SEQ & 0.098            & 0.070            & 0.037             & 0.043            & \bfseries -0.029  \\
 & EWC & \itshape 0.193   & \itshape 0.086   & \itshape 0.142    & \itshape 0.142   & \itshape -0.023   \\
 & GBP & \bfseries 0.051  & \bfseries 0.051  & \bfseries 0.001   & \bfseries 0.001  & \bfseries -0.029  \\
\bottomrule
\end{tabular}
\end{table}

\clearpage{}%
\begin{table*}[h!]
\centering
\caption{Continual-learning metrics for the 2-channel models. ACC, average
performance after the final stage; LA, learning accuracy on the just-trained
task; BWT, backward transfer; $F$, non-negative best-ever peak forgetting;
FWT, forward transfer relative to the independent reference. Arrows give the
preferred direction. \textbf{Bold} marks the best and \textit{italics}
the worst method within each metric block. SEQ and GBP tie on MAE BWT
($\pm0.008$) and on RMSE ACC ($0.108$) at the reported precision.}
\label{tab:cl-metrics-2ch}
\setlength{\tabcolsep}{6pt}\renewcommand{\arraystretch}{1.15}
\begin{tabular}{llS[table-format=-2.4]S[table-format=-2.4]S[table-format=-2.4]S[table-format=-2.4]S[table-format=-2.4]}
\toprule
Metric & Method & {ACC} & {LA} & {BWT \zero} & {F \dn} & {FWT} \\
\midrule
\multirow{3}{*}{SSIM \up}
 & SEQ & \bfseries 0.802  & 0.758            & 0.059             & \bfseries 0.000  & \itshape -0.072   \\
 & EWC & 0.782            & \bfseries 0.785  & \bfseries -0.004  & 0.011            & \bfseries -0.013  \\
 & GBP & \itshape 0.688   & \itshape 0.756   & \itshape -0.091   & \itshape 0.092   & -0.033            \\
\midrule
\multirow{3}{*}{PSNR (dB) \up}
 & SEQ & \bfseries 19.70  & 19.59            & 0.15              & 3.00             & \itshape -0.31    \\
 & EWC & \itshape 12.25   & \bfseries 20.24  & \itshape -10.65   & \itshape 12.13   & \bfseries 4.07    \\
 & GBP & 19.51            & \itshape 19.56   & \bfseries -0.07   & \bfseries 0.87   & 1.60              \\
\midrule
\multirow{3}{*}{MSE \dn}
 & SEQ & 0.0128           & \itshape 0.0190  & -0.0083           & 0.0080           & \itshape -0.0298  \\
 & EWC & \itshape 0.0675  & 0.0179           & \itshape 0.0662   & \itshape 0.0693  & \bfseries -0.0640 \\
 & GBP & \bfseries 0.0119 & \bfseries 0.0112 & \bfseries 0.0010  & \bfseries 0.0029 & -0.0548           \\
\midrule
\multirow{3}{*}{RMSE \dn}
 & SEQ & 0.108            & \itshape 0.120   & -0.016            & 0.036            & \itshape -0.023   \\
 & EWC & \itshape 0.252   & 0.115            & \itshape 0.183    & \itshape 0.199   & \bfseries -0.112  \\
 & GBP & \bfseries 0.108  & \bfseries 0.106  & \bfseries 0.003   & \bfseries 0.012  & -0.077            \\
\midrule
\multirow{3}{*}{MAE \dn}
 & SEQ & 0.060            & \itshape 0.066   & \bfseries -0.008  & 0.026            & \itshape 0.018    \\
 & EWC & \itshape 0.154   & 0.064            & \itshape 0.120    & \itshape 0.128   & \bfseries -0.041  \\
 & GBP & \bfseries 0.059  & \bfseries 0.053  & \bfseries 0.008   & \bfseries 0.014  & -0.020            \\
\bottomrule
\end{tabular}
\end{table*}
\begin{table*}[h!]
\centering
\caption{Continual-learning metrics for the hybrid channel evaluation
(2-channel-trained checkpoints evaluated through the single-channel
collapse adapter). ACC, average performance after the final stage; LA,
learning accuracy on the just-trained task; BWT, backward transfer; $F$,
non-negative best-ever peak forgetting; FWT, forward transfer relative to
the independent reference. Arrows give the preferred direction.
\textbf{Bold} marks the best and \textit{italics} the worst method within
each metric block; SEQ and EWC tie on SSIM $F$.}
\label{tab:cl-metrics-hyb}
\setlength{\tabcolsep}{6pt}\renewcommand{\arraystretch}{1.15}
\begin{tabular}{llS[table-format=-2.4]S[table-format=-2.4]S[table-format=-2.4]S[table-format=-2.4]S[table-format=-2.4]}
\toprule
Metric & Method & {ACC} & {LA} & {BWT \zero} & {F \dn} & {FWT} \\
\midrule
\multirow{3}{*}{SSIM \up}
 & SEQ & 0.802            & \bfseries 0.777  & \bfseries 0.033   & \bfseries 0.000  & -0.045            \\
 & EWC & \bfseries 0.810  & 0.775            & 0.047             & \bfseries 0.000  & \bfseries -0.043  \\
 & GBP & \itshape 0.616   & \itshape 0.721   & \itshape -0.140   & \itshape 0.147   & \itshape -0.082   \\
\midrule
\multirow{3}{*}{PSNR (dB) \up}
 & SEQ & 18.28            & \itshape 19.29   & -1.35             & 4.47             & \itshape -0.91    \\
 & EWC & \itshape 12.67   & \bfseries 20.33  & \itshape -10.21   & \itshape 11.56   & \bfseries 3.14    \\
 & GBP & \bfseries 19.29  & 19.39            & \bfseries -0.14   & \bfseries 0.87   & 0.45              \\
\midrule
\multirow{3}{*}{MSE \dn}
 & SEQ & 0.0172           & \itshape 0.0194  & -0.0029           & 0.0129           & \itshape 0.0149   \\
 & EWC & \itshape 0.0627  & 0.0163           & \itshape 0.0619   & \itshape 0.0643  & \bfseries -0.0175 \\
 & GBP & \bfseries 0.0123 & \bfseries 0.0117 & \bfseries 0.0009  & \bfseries 0.0029 & -0.0067           \\
\midrule
\multirow{3}{*}{RMSE \dn}
 & SEQ & 0.126            & \itshape 0.123   & 0.004             & 0.055            & \itshape 0.029    \\
 & EWC & \itshape 0.242   & 0.111            & \itshape 0.174    & \itshape 0.188   & \bfseries -0.054  \\
 & GBP & \bfseries 0.110  & \bfseries 0.108  & \bfseries 0.003   & \bfseries 0.012  & -0.016            \\
\midrule
\multirow{3}{*}{MAE \dn}
 & SEQ & 0.073            & \itshape 0.067   & \bfseries 0.008   & 0.040            & \itshape 0.020    \\
 & EWC & \itshape 0.144   & 0.060            & \itshape 0.112    & \itshape 0.119   & \bfseries -0.033  \\
 & GBP & \bfseries 0.061  & \bfseries 0.055  & 0.009             & \bfseries 0.014  & -0.011            \\
\bottomrule
\end{tabular}
\end{table*}
\clearpage{}%

\subsubsection{Comparison across the three continual-learning settings}

We compare three evaluation settings built from the same continual
checkpoints: (i) the 1-channel models; (ii) the 2-channel models
scored with the FA + MD channel-averaged metrics; and (iii) the hybrid setting, in which the
2-channel-trained checkpoints are evaluated through the 1-channel
channel-collapse adapter. Peak forgetting is computed with a single
non-negative best-ever definition applied identically to all three methods
in every setting.

Across all three settings, GBP provides the strongest stability on the
voxel-wise metrics, achieving the lowest peak forgetting and the most
favourable retention--adaptation balance. EWC repeatedly exhibits
catastrophic forgetting at the final continual stage: it attains
competitive --- and in the 2-channel and hybrid settings the highest ---
in-task learning accuracy, but its representation collapses after the last
task, giving the worst backward transfer and forgetting scores. SEQ
occupies an intermediate position, with moderate learning accuracy but
comparatively robust retention despite the absence of an explicit
consolidation mechanism.

The main distinction arises from channel coupling. In the 1-channel
setting, GBP records negligible peak forgetting on every voxel-wise metric
(PSNR $F = 0.03$~dB, MSE $F = 0.0000$, MAE $F = 0.001$) --- one to more
than two orders of magnitude below SEQ, which shows moderate forgetting
(PSNR $F = 5.93$~dB), and further still below EWC, which collapses at the
final stage (PSNR $\mathrm{BWT} = -10.20$~dB, $F = 10.20$~dB). GBP's
1-channel backward transfer is slightly positive ($+0.07$~dB): the
blueprint freeze eliminates degradation of past tasks at the reported
precision. In the 2-channel setting, coupling the FA and MD channels
introduces an intrinsic regularisation that stabilises the unconsolidated
baseline: SEQ's PSNR forgetting falls from $5.93$~dB (1-channel) to
$3.00$~dB (2-channel). GBP's voxel-wise forgetting remains negligible in
both regimes (PSNR $F = 0.03$~dB in the 1-channel setting and $0.87$~dB in
the 2-channel setting, with $\mathrm{BWT}$ within $0.07$~dB of zero in
both), and its SSIM forgetting falls from $F = 0.003$ to $F = 0.000$.
EWC, by contrast, does not benefit from the additional channel structure:
it still collapses, and its peak forgetting in fact grows from
$10.20$~dB to $12.13$~dB (PSNR $\mathrm{BWT} = -10.65$~dB), indicating
that the Fisher penalty gains nothing from the coupling that helps the
unconsolidated baseline. The hybrid setting, which stresses the
2-channel-trained representations through the channel-collapse adapter,
follows the same pattern: GBP maintains the best stability--plasticity
trade-off (PSNR $F = 0.87$~dB, $\mathrm{BWT} = -0.14$~dB), EWC preserves
strong learning accuracy ($\mathrm{LA} = 20.33$~dB) but suffers severe
final-stage forgetting ($F = 11.56$~dB), and SEQ achieves competitive
average accuracy through stable but less plastic representations.

Across all settings, SSIM and the voxel-wise metrics only partially agree,
indicating that structural similarity and intensity fidelity capture
complementary aspects of continual reconstruction. The divergence is most
pronounced in the hybrid setting: there GBP records a much lower SSIM ACC
($0.616$) than EWC ($0.810$) or SEQ ($0.802$) despite winning every
voxel-wise metric, because the channel-collapse adapter incurs a larger
cost on the blueprint-frozen GBP filters when the 2-channel output head is
reduced to a single channel. This SSIM penalty does not appear in the
1-channel setting --- where GBP's SSIM ACC ($0.810$) is close to SEQ
($0.829$) and EWC ($0.835$) --- and does not affect the voxel-wise ranking
on peak forgetting, which remains GBP-favourable in all three settings and
on all four voxel-wise metrics. The dissociation is systematic rather than
incidental: across the 1-channel matrices, cells differing by up to
$15$~dB on PSNR and by factors of three to over thirty on the voxel-wise
error metrics differ by less than $0.04$ on SSIM, confirming that SSIM is
largely insensitive to the intensity-scale variation that dominates the
voxel-wise statistics. Reporting both families of metrics is therefore
necessary to characterise continual reconstruction performance, as each
privileges a different aspect of the representational trade-off.

Table \ref{tab:general-cl-metrics-1ch} presents the analyses described above.

\begin{table*}[h!]
\centering
\caption{Continual-learning metrics for the 1-channel models across all five
evaluation metrics. ACC is average performance after the final stage; LA is
learning accuracy on the just-trained task; BWT is backward transfer; $F$ is
non-negative best-ever peak forgetting; and FWT is forward transfer relative
to the independent reference. Arrows indicate the preferred direction. \textbf{Bold} marks the best and \textit{italics} the worst method
within each metric and score. SEQ and EWC are tied on SSIM LA at the reported
precision; SEQ is marginally lower at full precision.}
\label{tab:general-cl-metrics-1ch}
\setlength{\tabcolsep}{6pt}
\renewcommand{\arraystretch}{1.15}
\begin{tabular}{
  ll
  S[table-format=2.4]
  S[table-format=2.4]
  S[table-format=-2.4]
  S[table-format=2.4]
  S[table-format=-2.4]
}
\toprule
Metric & Method & {ACC} & {LA} & {BWT \zero} & {$F$ \dn} & {FWT} \\
\midrule
\multirow{3}{*}{SSIM \up}
 & SEQ & \itshape 0.9696  & \itshape 0.9703  & -0.0009           & \itshape 0.0041  & \itshape 0.1509 \\
 & EWC & \bfseries 0.9744 & 0.9703           & \itshape 0.0054   & \bfseries 0.0000 & \bfseries 0.1345 \\
 & GBP & 0.9712           & \bfseries 0.9712 & \bfseries -0.0001 & 0.0015           & 0.1503 \\
\midrule
\multirow{3}{*}{PSNR (dB) \up}
 & SEQ & 25.12            & \bfseries 28.56  & \itshape -4.59    & \itshape 4.83    & 5.10 \\
 & EWC & \bfseries 25.92  & 28.17            & -3.01             & 3.22             & \itshape 5.57 \\
 & GBP & \itshape 23.92   & \itshape 24.34   & \bfseries -0.57   & \bfseries 2.04   & \bfseries 2.94 \\
\midrule
\multirow{3}{*}{MSE \dn}
 & SEQ & \bfseries 0.0032 & \bfseries 0.0020 & 0.0016            & 0.0017           & -0.0198 \\
 & EWC & \itshape 0.0057  & 0.0041           & \itshape 0.0022   & \itshape 0.0024  & \bfseries -0.0215 \\
 & GBP & 0.0043           & \itshape 0.0045  & \bfseries -0.0003 & \bfseries 0.0015 & \itshape -0.0168 \\
\midrule
\multirow{3}{*}{RMSE \dn}
 & SEQ & \bfseries 0.0558 & \bfseries 0.0414 & \itshape 0.0193   & \itshape 0.0206  & -0.0705 \\
 & EWC & \itshape 0.0670  & 0.0526           & 0.0192            & 0.0205           & \bfseries -0.0772 \\
 & GBP & 0.0646           & \itshape 0.0640  & \bfseries 0.0008  & \bfseries 0.0131 & \itshape -0.0518 \\
\midrule
\multirow{3}{*}{MAE \dn}
 & SEQ & \bfseries 0.0194 & \bfseries 0.0133 & \itshape 0.0081   & \itshape 0.0081  & -0.0571 \\
 & EWC & \itshape 0.0251  & 0.0192           & 0.0079            & 0.0080           & \bfseries -0.0608 \\
 & GBP & 0.0219           & \itshape 0.0205  & \bfseries 0.0019  & \bfseries 0.0058 & \itshape -0.0532 \\
\bottomrule
\end{tabular}
\end{table*}

\subsection{Interpretability and Memory Storage Utilities of GBP}
\subsubsection{Visualizing Different GBP Blueprints}
The proposed GBP framework provides an additional and important capability beyond mitigating catastrophic forgetting. Specifically, it identifies and maps the subset of trainable network sub-modules that remain available for continual adaptation while preserving previously acquired knowledge. This decomposition enables selective plasticity, allowing future tasks to be learned without substantially disrupting representations associated with earlier tasks.

\begin{figure}[h!]
\centering

    \centering
    \includegraphics[width=1.0\linewidth]{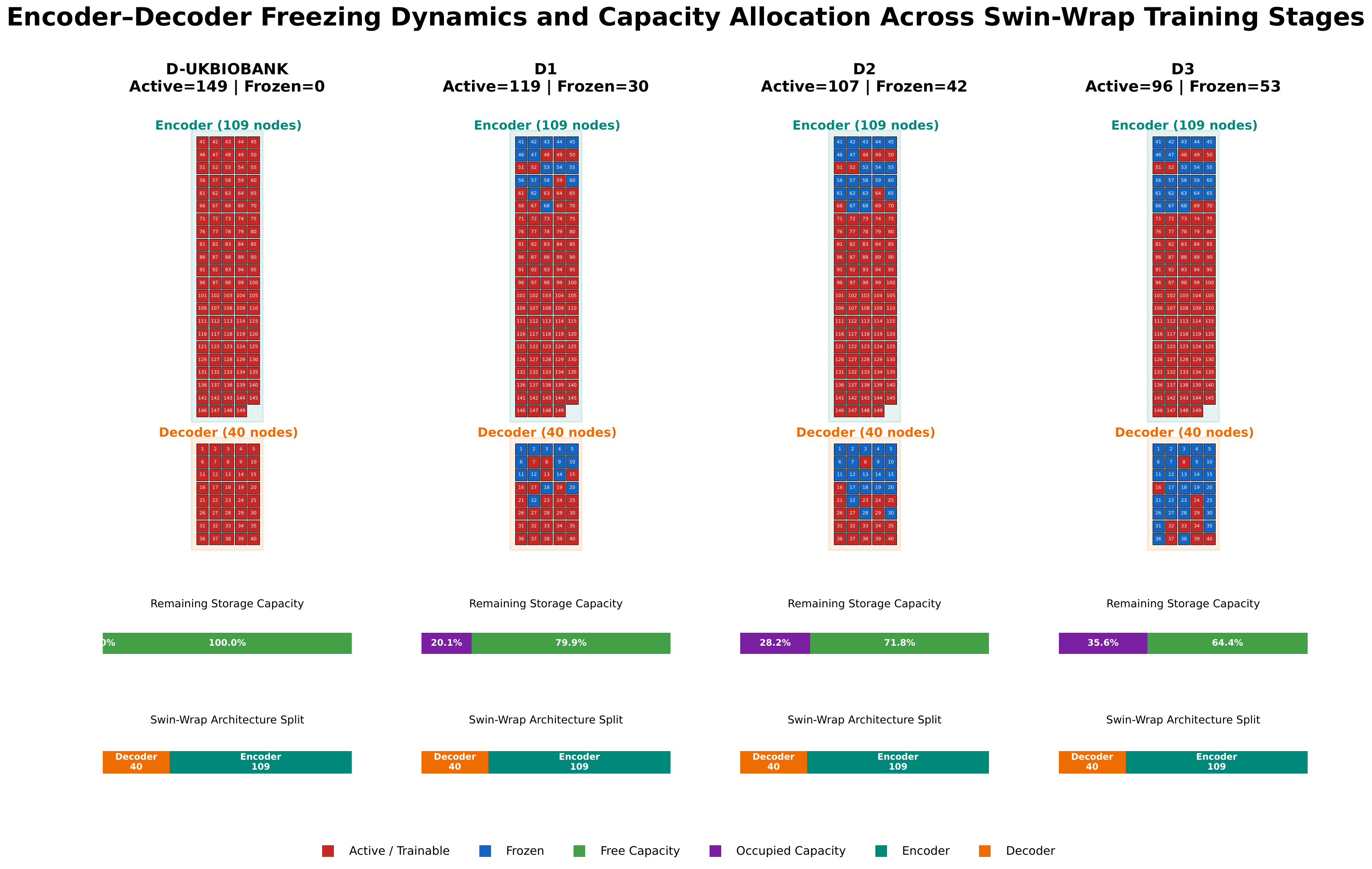}

\caption{
\textbf{1-channel S2-Level-4 Swin-Wrap}
}
\label{fig:blueprint_normalized_active_frozen1}
\end{figure}

\begin{figure}
\centering

    \centering
    \includegraphics[width=0.85\linewidth]{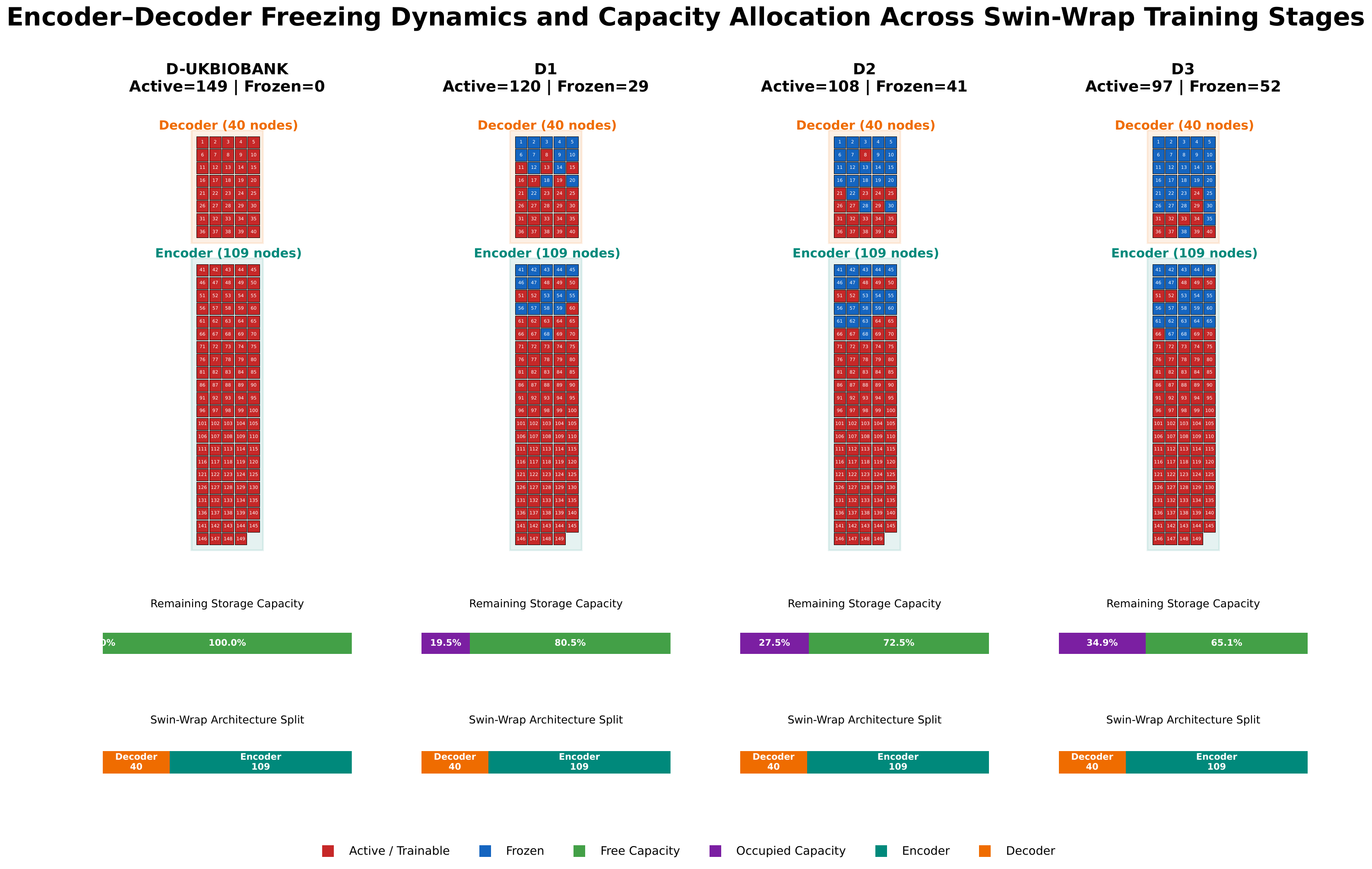}

\caption{
\textbf{1-channel L2-Level-4 Swin-Wrap}
}
\label{fig:blueprint_normalized_active_frozen2}
\end{figure}

\begin{figure}
\centering

    \centering
    \includegraphics[width=0.85\linewidth]{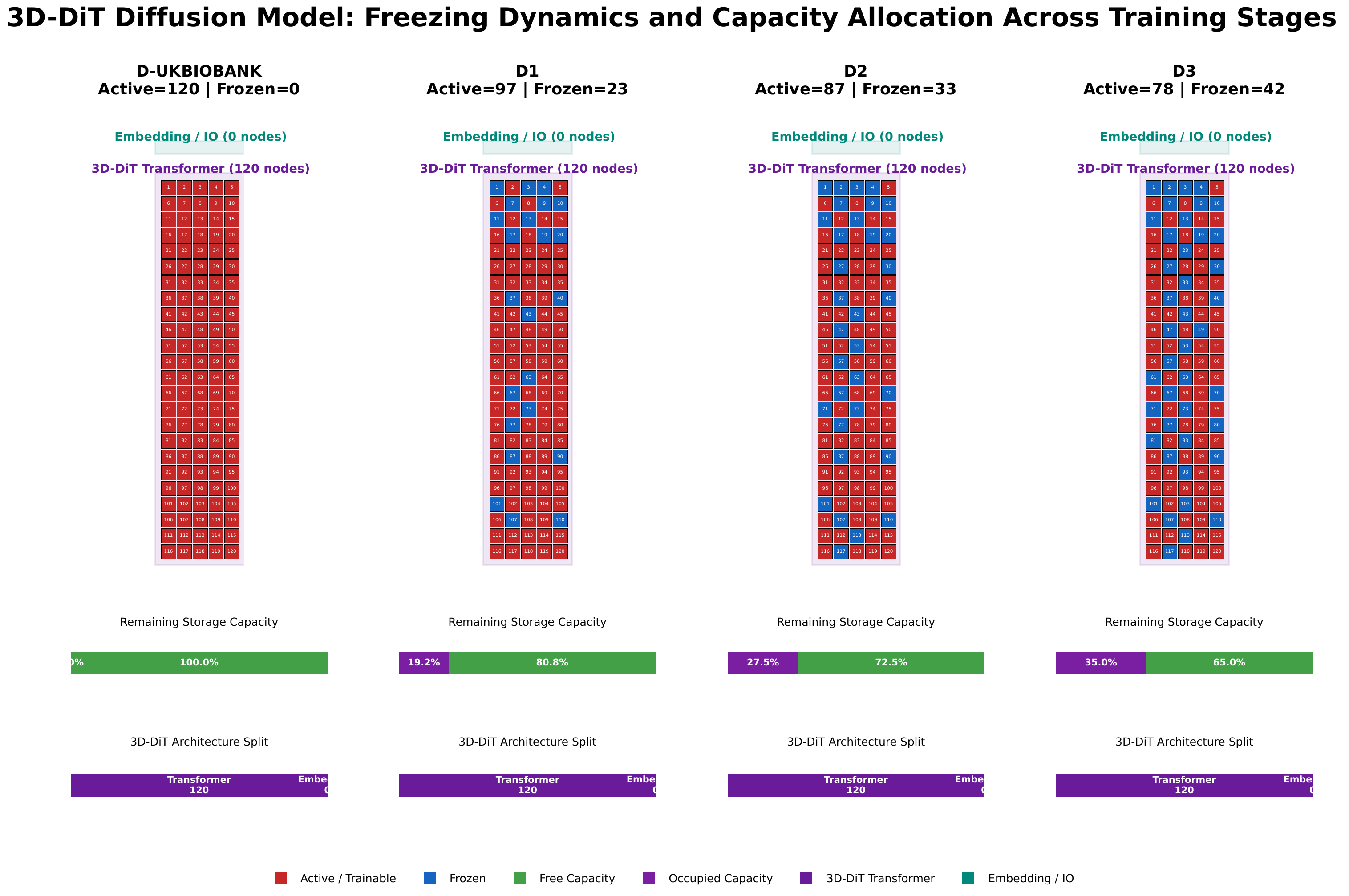}

\caption{
\textbf{1-channel S2-Level-4 3D-DiT}
}
\label{fig:blueprint_normalized_active_frozen3}
\end{figure}

\begin{figure}
\centering

    \centering
    \includegraphics[width=0.85\linewidth]{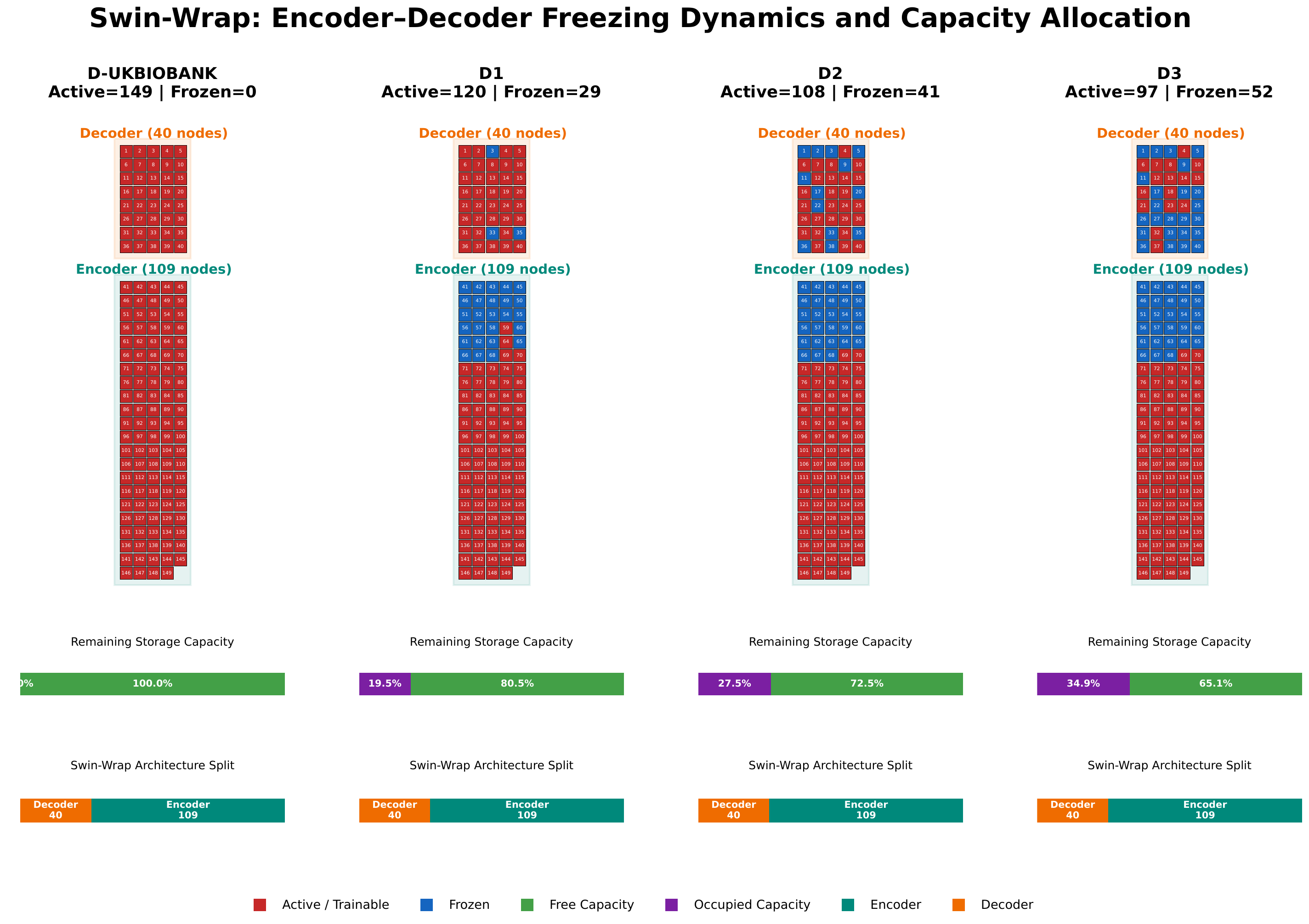}

\caption{
\textbf{2-channel S2-Level-4 Swin-Wrap}
}
\label{fig:blueprint_normalized_active_frozen5}
\end{figure}

Furthermore, the resulting blueprint offers an interpretable representation of the network's learning dynamics by distinguishing active, adaptable components from protected components. As a consequence, the blueprint can be used to investigate how task-specific features, representations, and attention patterns evolve throughout the continual learning process, thereby providing insight into the mechanisms underlying knowledge retention and transfer across tasks. Examples of these active and frozen subnetworks are visualized in Fig. \ref{fig:blueprint_normalized_active_frozen1}, \ref{fig:blueprint_normalized_active_frozen2}, \ref{fig:blueprint_normalized_active_frozen2}, \ref{fig:blueprint_normalized_active_frozen3}, and \ref{fig:blueprint_normalized_active_frozen5}.

\newpage

\subsubsection{Interpretability and blueprint virtual signature of the GBP}

To relate GBP capacity allocation to neuroanatomical structure, we performed atlas-based attribution for each continual-learning domain. To preserve domain-specific anatomical variability, each atlas was aligned to a randomly selected subject from the corresponding cohort. The healthy-population domain D1 used a HABS-HD subject with the HCP-MMP1.0--Glasser MNI cortical atlas \cite{Glasser2016,bedini_hcpmmp_fsl}; the neurodegenerative domain D2 used a PPMI subject with the AAL3 anatomical atlas \cite{rolls2020aal3}; the developmental and psychiatric domain D3 used an ABCD subject with the Schaefer-400 cortical parcellation ordered according to the 17-network Yeo scheme in MNI space \cite{schaefer2018localglobal,yeo2011organization}; and the brain-tumour domain D4 used an EGD subject with the HCP-MMP1.0--Glasser MNI cortical atlas \cite{Glasser2016,bedini_hcpmmp_fsl}. This design enabled model responses to be attributed to anatomically or functionally defined brain circuits while retaining the geometry of a representative scan from each domain.

GBP partitions the network into computational units, including attention, feed-forward, normalisation and adaptive-modulation components, and estimates the first-order contribution of each unit to the task objective. Salience scores are aggregated across a probe set to construct a principal-subspace blueprint, after which an upper-quantile selection rule, fused across tasks, identifies the units to be frozen. Previously derived Graph-Blueprint files therefore define the task-relevant subcircuits, including previously frozen units and the remaining active units. For each blueprint condition, individual modules were perturbed sequentially during inference, and the resulting objective was backpropagated to the atlas input to estimate voxelwise sensitivity. Aggregating these gradients produced spatial attribution maps that localised the anatomical regions most dependent on each blueprint-defined circuit subset. This analysis enabled direct comparison among the unconstrained SEQ all-active model, the newly frozen GBP task blueprint and the remaining plastic capacity, thereby linking continual-learning dynamics to interpretable three-dimensional neuroanatomical importance patterns.

Because the selected units are frozen exactly, each subsequent optimisation stage is restricted to the residual trainable subspace. Separate blueprint streams are maintained for the SWIN-Wrap encoder--decoder (3D-Swin) and the latent generator (3D-DiT), reflecting their distinct functional roles. In Fig.~5a and Extended Figs.~6a, 7a and 8a, this process appears as a monotonic transition from active to frozen units in both components of Alcmaeon. In the 3D-DiT stream, the number of active units decreases from 120 to 78, leaving 42 frozen units and approximately 65\% of the original capacity available. In the 3D-Swin stream, the number of active units decreases from 149 to 96, leaving 53 frozen units and approximately 64\% remaining capacity. GBP therefore does not immobilise the model; instead, it preserves selected task-relevant circuits while retaining substantial plasticity for future domains.

The ranked blueprint components and their anatomical projections further indicate that this preserved capacity is structured rather than uniform. Each domain recruits a distinct subset of modules, suggesting that GBP acts as a task-specific consolidation mechanism rather than as a global regularisation penalty or a simple magnitude-based mask. Across D1--D4, the blueprint signatures become progressively more domain-specific: healthy-population patterns are comparatively diffuse, neurodegenerative adaptation emphasises subcortical and medial-temporal structures, developmental cohorts exhibit a more distributed network profile, and the tumour domain produces the most focal and asymmetric pattern. This makes the stability--plasticity trade-off explicit, as each domain protects selected circuits from the remaining trainable pool and thereby records where adaptive capacity has been allocated and how much remains. Because these circuits are selected from recurrent task-active salience patterns rather than from weight magnitude or heuristic task masks \cite{kang2022wsn,serra2018hat}, GBP connects mechanistic attribution \cite{syed2023attributionpatching,anthropic2025graphs} with continual adaptation and provides a framework for examining how task-specific representations, features and attention pathways evolve across the learning trajectory.

For the healthy-population task (Extended Fig.~6; HCP-MMP1.0--Glasser MNI 1-mm cortical atlas \cite{Glasser2016}; randomly selected HABS-HD subject), the SEQ all-active reference focused primarily on medial posterior cortical regions. In the 3D-Swin module, the strongest SEQ regions were right posterior cingulate area (R-38), right ventral posterior cingulate--precuneus area (R-35), right medial parietal area (R-37) and right primary motor cortex area (R-8). After applying the proposed GBP blueprint, the frozen signature shifted towards right anterior intraparietal area (R-117), right area 55b (R-12), right parieto-occipital sulcus area (R-15), right dorsal lateral intraparietal area (R-95) and right medial frontal area (R-63). In the 3D-DiT latent-generator stream, SEQ again emphasised posterior cingulate--precuneus and medial parietal regions, including right anterior pregenual area (R-179), posterior cingulate area (R-38), ventral posterior cingulate--precuneus area (R-35), medial parietal area (R-37) and posterior pregenual area (R-60). The GBP frozen blueprint instead preserved a more fronto-parietal and visuomotor-control signature, involving right anterior ventral area (R-67), right frontal eye field (R-10), right parieto-occipital sulcus area (R-15), right anterior intraparietal area (R-117) and right ventral intraparietal area (R-49). Thus, in D1, GBP transforms a broadly medial posterior SEQ signature into a more selective blueprint centred on fronto-parietal attention, oculomotor and visuomotor integration circuits.

For the neurodegenerative task (Extended Fig.~7; AAL3 anatomical atlas \cite{rolls2020aal3}; randomly selected PPMI subject), the SEQ all-active reference and the proposed GBP frozen blueprint showed distinct stream-specific attribution patterns. In the 3D-Swin stream, the SEQ reference was dominated by frontal and cerebellar regions, including the left gyrus rectus/medial orbitofrontal cortex (Rectus-L), right cerebellar Crus II (Cerebellum-Crus2-R), right orbital inferior frontal gyrus, part 2 (Frontal-Inf-Orb-2-R), right cerebellar lobule VIII (Cerebellum-8-R) and left middle frontal gyrus, part 2 (Frontal-Mid-2-L). After applying GBP, the frozen blueprint shifted towards limbic, thalamic and midbrain-associated regions, including the right amygdala (Amygdala-R), right reuniens thalamic nucleus (Thal-Re-R), left lateral geniculate nucleus of the thalamus (Thal-LGN-L), left ventral tegmental area (VTA-L) and left superior temporal pole (Temporal-Pole-Sup-L).
In the 3D-DiT latent-generator stream, the SEQ reference was concentrated in cerebellar vermis and thalamic regions, including vermis IX (Vermis-9), right ventrolateral thalamic nucleus (Thal-VL-R), vermis X (Vermis-10), vermis VIII (Vermis-8) and vermis III (Vermis-3). The GBP frozen blueprint instead emphasised left lateral geniculate nucleus of the thalamus (Thal-LGN-L), left amygdala (Amygdala-L), left gyrus rectus/medial orbitofrontal cortex (Rectus-L), left olfactory cortex (Olfactory-L) and left anterior pulvinar thalamic nucleus (Thal-PuA-L). Overall, D2 shows a stream-specific blueprint transition: GBP shifts the 3D-Swin stream from a frontal--cerebellar SEQ pattern towards limbic, thalamic, temporal and midbrain-associated regions, while the 3D-DiT shifts from a vermis--thalamic SEQ profile towards a more limbic, thalamic and orbitofrontal neurodegenerative signature.

For the developmental and psychiatric task (Extended Fig.~8; Schaefer-400 cortical parcellation with 17 network Yeo ordering in MNI space \cite{schaefer2018localglobal,yeo2011organization}; randomly selected ABCD subject), the SEQ all-active reference and the proposed GBP frozen blueprint showed a transition from a broad distributed network pattern to a more selective functional-network signature. In the 3D-Swin stream, the SEQ reference was dominated by left dorsal-attention posterior cortex (Schaefer-80), left temporo-parietal cortex (Schaefer-200), right control-network posterior cingulate cortex (Schaefer-357), right limbic orbitofrontal cortex (Schaefer-318) and right dorsal-attention superior parietal cortex (Schaefer-266). After applying GBP, the frozen blueprint shifted towards left default-network parahippocampal cortex (Schaefer-193), right control-network precuneus (Schaefer-354), left dorsal-attention posterior cortex (Schaefer-77), left temporo-parietal cortex (Schaefer-196) and left salience/ventral-attention orbitofrontal cortex (Schaefer-107). In the 3D-DiT stream, the SEQ reference was concentrated in default, salience/ventral-attention and dorsal-attention parcels, including left DefaultC inferior parietal cortex (Schaefer-188), left SalVentAttnB orbitofrontal cortex (Schaefer-107), right DorsAttnA temporal-occipital cortex (Schaefer-261), left SalVentAttnA parietal operculum (Schaefer-88) and left SalVentAttnA insula (Schaefer-89). The GBP frozen blueprint instead emphasised a compact dorsal-attention/fronto-parietal pattern, including left DorsAttnB frontal eye-field cortex (Schaefer-84), left DorsAttnB posterior cortex (Schaefer-80), a left control-network fronto-parietal parcel (Schaefer-145), left DorsAttnB frontal eye-field/precentral cortex (Schaefer-83) and left DorsAttnB posterior cortex (Schaefer-77). Thus, D3 shows a stream-specific transition from a diffuse SEQ all-active network profile to a more selective GBP blueprint centred on dorsal-attention, control, default and temporo-parietal systems. This pattern is consistent with greater allocation of preserved capacity to functional networks relevant to developmental variation, rather than preservation of a diffuse all-active representation.

For the brain-tumour task (Fig. 5; HCP-MMP1.0--Glasser MNI cortical atlas \cite{Glasser2016,bedini_hcpmmp_fsl}; randomly selected EGD subject), the SEQ all-active reference and the proposed GBP frozen blueprint showed a focal transition around operculo-insular, orbitofrontal, paralimbic and medial-temporal regions. In the 3D-Swin stream, the SEQ reference focused on right posterior orbitofrontal cortex (R-166), right area 52 (R-103), right presubiculum (R-119), right parahippocampal area 2 (R-155) and right anterior pregenual area 32 (R-179). After applying GBP, the frozen blueprint shifted towards right piriform cortex (R-110), right frontal opercular area FOP3 (R-114), right parietal opercular area OP2--3 (R-102), right posterior pregenual area 32 (R-60) and right posterior insular area PoI1 (R-167). In the 3D-DiT stream, the SEQ reference was concentrated in right posterior insular and opercular regions, including right posterior insular area PoI2 (R-106), right posterior inferior frontal junction (R-80), right frontal opercular area FOP3 (R-114), right frontal opercular area FOP1 (R-113) and right parietal opercular area OP2--3 (R-102). The GBP frozen blueprint instead became strongly left-lateralised and preserved left frontal opercular area FOP2 (L-115), left frontal opercular area FOP1 (L-113), left posterior inferior frontal junction (L-80), left middle insular cortex (L-109) and left primary auditory cortex (L-24). The accompanying axial tumour slice shows a focal hyperintense lesion in the lateral hemisphere, close to a perisylvian/operculo-insular territory; this is consistent with the observed GBP enrichment in opercular, insular and adjacent auditory--fronto-parietal regions. Overall, D4 indicates that GBP redirects the tumour-domain blueprint from a broader SEQ profile involving orbitofrontal, presubicular and parahippocampal regions towards a more focal pathology-linked signature centred on operculo-insular, frontal-opercular and perisylvian circuits.

These results provide mechanistic evidence that GBP preserves task-relevant circuitry from earlier domains while redirecting the remaining plastic capacity towards new, domain-specific representations. Rather than acting as a global regularizer, GBP generates a structured blueprint in which selected circuits are frozen as memory and newly available pathways are recruited during adaptation.

\section{Fine-tuning and Clinical Tasks}

\subsection{Clinical Task of Adult's Alzheimer Progression}

\subsubsection{Baseline models}

The results obtained using these classification models are presented in Table~\ref{tab:baseline}. As can be observed, the CNN-based approach achieved better performance than the SVM-based framework, reaching accuracies of 74.76\% and 66.62\%, respectively, in the binary classification task using k-fold cross-validation. Particularly noteworthy is the high classification performance obtained for the three-class problem, where the CNN model achieved an accuracy of 66.62\% using k-fold validation, which is approximately twice the accuracy expected by random chance. In addition, the AUC values obtained for both the binary and three-class classification tasks were around 80\%, highlighting the robustness of the proposed baseline.

\begin{table*}[t]
\centering
\caption{\textbf{Baseline classification performance (mean $\pm$ SD).}}
\label{tab:baseline}

\scriptsize
\setlength{\tabcolsep}{4pt}

\resizebox{\textwidth}{!}{
\begin{tabular}{llcccccc}
\toprule
\textbf{Scenario} &
\textbf{Model} &
\textbf{Acc} &
\textbf{Bal Acc} &
\textbf{Prec} &
\textbf{F1} &
\textbf{AUC} \\
\midrule

\multirow{4}{*}{2-Class}
& CNN (5-fold)
& $0.748\pm0.026$
& $0.744\pm0.031$
& $0.746\pm0.026$
& $0.743\pm0.029$
& $0.846\pm0.033$ \\

& CNN (70/30)
& $0.756\pm0.052$
& $0.745\pm0.053$
& $0.762\pm0.059$
& $0.746\pm0.054$
& $0.855\pm0.043$ \\

& PCA+SVM (5-fold)
& $0.716\pm0.032$
& $0.704\pm0.029$
& $0.718\pm0.040$
& $0.705\pm0.030$
& $0.812\pm0.011$ \\

& PCA+SVM (70/30)
& $0.693\pm0.019$
& $0.679\pm0.020$
& $0.699\pm0.029$
& $0.677\pm0.018$
& $0.794\pm0.026$ \\

\midrule

\multirow{4}{*}{3-Class}
& CNN (5-fold)
& $0.666\pm0.051$
& $0.652\pm0.056$
& $0.658\pm0.051$
& $0.653\pm0.054$
& $0.833\pm0.034$ \\

& CNN (70/30)
& $0.655\pm0.045$
& $0.643\pm0.039$
& $0.664\pm0.048$
& $0.643\pm0.045$
& $0.831\pm0.024$ \\

& PCA+SVM (5-fold)
& $0.642\pm0.030$
& $0.646\pm0.020$
& $0.636\pm0.021$
& $0.632\pm0.023$
& $0.815\pm0.019$ \\

& PCA+SVM (70/30)
& $0.605\pm0.021$
& $0.607\pm0.022$
& $0.606\pm0.020$
& $0.597\pm0.018$
& $0.812\pm0.014$ \\

\bottomrule
\end{tabular}}
\end{table*}

\begin{figure}
\centering
\includegraphics[width=0.8\linewidth]{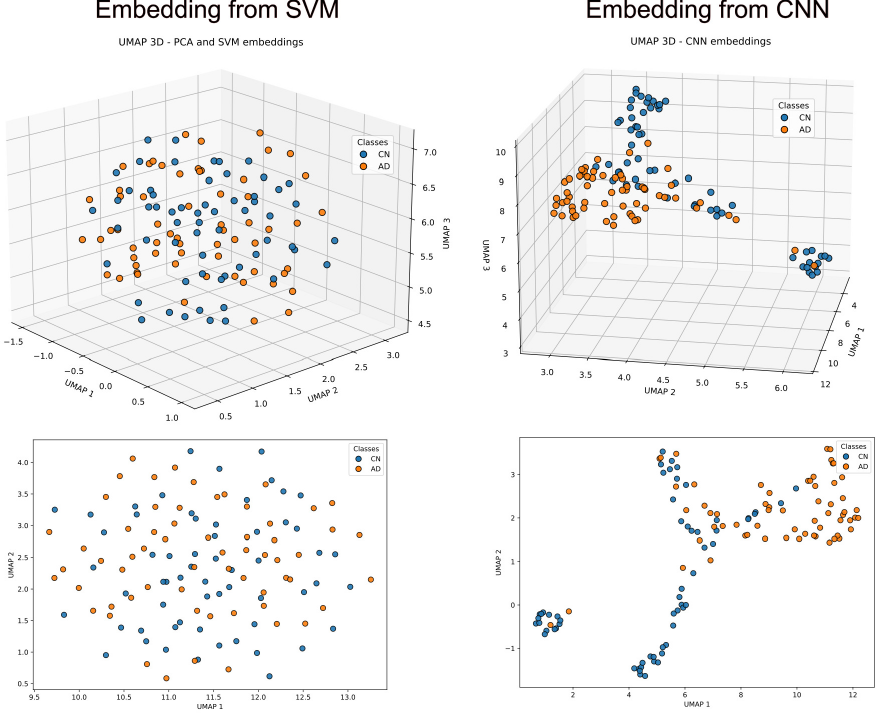}
\caption{\label{fig:umap_baseline} UMAP visualisation of the embedding space for Healthy Controls (HC, blue) and Alzheimer’s disease subjects (AD, orange). The first plot corresponds to the baseline model based on PCA and SVM (PCA space is used as embedding), the second to the CNN model (features from the first linear layer are used as embedding).}
\end{figure}

\subsubsection{Two-channels model - level 2 and level 4 ablation analysis}
This section presents the experimental results obtained for the Level 2 of the 3D-Swin-DiT two-channel architecture. In this analysis, only the Swin-UNet branch of the model is considered.

\begin{table*}[t]
\centering
\caption{\textbf{Binary (2-class) classification performance using Level-2 latent representations (mean $\pm$ SD). Sensitivity values are identical to balanced accuracy and are omitted for brevity.}}
\label{tab:bin_level2}

\scriptsize
\setlength{\tabcolsep}{4pt}

\resizebox{\textwidth}{!}{
\begin{tabular}{llccccc}
\toprule
\textbf{Model} &
\textbf{Validation} &
\textbf{Acc} &
\textbf{Bal Acc} &
\textbf{Prec} &
\textbf{F1} &
\textbf{AUC} \\
\midrule

\multirow{2}{*}{Swin-UNET + SVM}
& 5-fold
& $0.713\pm0.030$
& $0.714\pm0.030$
& $0.712\pm0.027$
& $0.710\pm0.028$
& $0.778\pm0.032$ \\

& 70/30
& $0.706\pm0.044$
& $0.709\pm0.046$
& $0.711\pm0.046$
& $0.703\pm0.045$
& $0.780\pm0.044$ \\

\midrule

\multirow{6}{*}{3D-Swin + MLP}

& 5-fold
& $0.757\pm0.023$
& $0.756\pm0.023$
& $0.756\pm0.024$
& $0.754\pm0.023$
& $0.845\pm0.038$ \\

& 70/30
& $0.757\pm0.036$
& $0.752\pm0.036$
& $0.755\pm0.033$
& $0.752\pm0.035$
& $0.829\pm0.028$ \\

& Zero-shot
& $0.552\pm0.000$
& $0.527\pm0.000$
& $0.532\pm0.000$
& $0.515\pm0.000$
& $0.528\pm0.000$ \\

& One-shot
& $0.563\pm0.045$
& $0.541\pm0.054$
& $0.505\pm0.118$
& $0.497\pm0.101$
& $0.616\pm0.053$ \\

& Three-shot
& $0.665\pm0.018$
& $0.663\pm0.019$
& $0.665\pm0.018$
& $0.661\pm0.019$
& $0.745\pm0.025$ \\

& Five-shot
& $0.743\pm0.019$
& $0.739\pm0.022$
& $0.744\pm0.018$
& $0.739\pm0.021$
& $0.825\pm0.015$ \\

\bottomrule
\end{tabular}}
\end{table*}

At this level of the model, the results reported in Table~\ref{tab:bin_level2} and Table~\ref{tab:tri_level2} are largely consistent with those obtained using the baseline approaches. In particular, the application of the MLP under different few-shot settings highlights the robustness of the feature representations, as the model achieves comparable performance with only 5-shot learning to that obtained after 20 training epochs in both k-fold and 70/30 split configurations for the binary classification task (Table~\ref{tab:bin_level2}).

Moreover, for the three-class classification problem, the 5-shot setting even outperforms the fully trained MLP and SVM models (see Table~\ref{tab:tri_level2}), further supporting the effectiveness of the learned feature space in low-computational scenarios.

\begin{table*}[t]
\centering
\caption{\textbf{Three-class classification performance using Level-2 latent representations (mean $\pm$ SD). Sensitivity values are identical to balanced accuracy and are omitted for brevity.}}
\label{tab:tri_level2}

\scriptsize
\setlength{\tabcolsep}{4pt}

\resizebox{\textwidth}{!}{
\begin{tabular}{llccccc}
\toprule
\textbf{Model} &
\textbf{Validation} &
\textbf{Acc} &
\textbf{Bal Acc} &
\textbf{Prec} &
\textbf{F1} &
\textbf{AUC} \\
\midrule

\multirow{2}{*}{3D-Swin + SVM}

& 5-fold
& $0.579\pm0.040$
& $0.579\pm0.043$
& $0.571\pm0.040$
& $0.570\pm0.040$
& $0.764\pm0.033$ \\

& 70/30
& $0.596\pm0.020$
& $0.594\pm0.019$
& $0.592\pm0.022$
& $0.589\pm0.020$
& $0.753\pm0.026$ \\

\midrule

\multirow{6}{*}{3D-Swin + MLP}

& 5-fold
& $0.626\pm0.016$
& $0.609\pm0.022$
& $0.613\pm0.018$
& $0.605\pm0.020$
& $0.796\pm0.014$ \\

& 70/30
& $0.562\pm0.027$
& $0.544\pm0.026$
& $0.558\pm0.022$
& $0.544\pm0.026$
& $0.748\pm0.019$ \\

& Zero-shot
& $0.338\pm0.000$
& $0.281\pm0.000$
& $0.211\pm0.000$
& $0.241\pm0.000$
& $0.472\pm0.000$ \\

& One-shot
& $0.466\pm0.036$
& $0.413\pm0.061$
& $0.458\pm0.120$
& $0.347\pm0.099$
& $0.621\pm0.050$ \\

& Three-shot
& $0.517\pm0.047$
& $0.475\pm0.070$
& $0.494\pm0.074$
& $0.447\pm0.097$
& $0.718\pm0.004$ \\

& Five-shot
& $0.616\pm0.021$
& $0.591\pm0.018$
& $0.611\pm0.020$
& $0.592\pm0.018$
& $0.786\pm0.019$ \\

\bottomrule
\end{tabular}}
\end{table*}

This section presents the experimental results obtained for Level 4 of the 3D-Swin-DiT two-channel architecture. The same experimental protocol as in the previous section is followed. Tables~\ref{tab:bin_level4} and~\ref{tab:tri_level4} report the results for the binary and three-class classification experiments, respectively.

\begin{table*}[t]
\centering
\caption{\textbf{Binary (2-class) classification performance using Level-4 latent representations (mean $\pm$ SD). Sensitivity values are identical to balanced accuracy and are omitted for brevity.}}
\label{tab:bin_level4}

\scriptsize
\setlength{\tabcolsep}{4pt}

\resizebox{\textwidth}{!}{
\begin{tabular}{llccccc}
\toprule
\textbf{Model} &
\textbf{Validation} &
\textbf{Acc} &
\textbf{Bal Acc} &
\textbf{Prec} &
\textbf{F1} &
\textbf{AUC} \\
\midrule

\multirow{2}{*}{3D-Swin + SVM}

& 5-fold
& $0.693\pm0.047$
& $0.699\pm0.048$
& $0.702\pm0.047$
& $0.691\pm0.049$
& $0.773\pm0.038$ \\

& 70/30
& $0.697\pm0.026$
& $0.703\pm0.030$
& $0.700\pm0.026$
& $0.696\pm0.026$
& $0.763\pm0.035$ \\

\midrule

\multirow{6}{*}{3D-Swin + MLP}

& 5-fold
& $0.748\pm0.025$
& $0.746\pm0.030$
& $0.746\pm0.026$
& $0.744\pm0.027$
& $0.823\pm0.033$ \\

& 70/30
& $0.728\pm0.019$
& $0.720\pm0.019$
& $0.720\pm0.019$
& $0.720\pm0.019$
& $0.801\pm0.023$ \\

& Zero-shot
& $0.599\pm0.000$
& $0.608\pm0.000$
& $0.609\pm0.000$
& $0.599\pm0.000$
& $0.648\pm0.000$ \\

& One-shot
& $0.622\pm0.047$
& $0.613\pm0.056$
& $0.575\pm0.140$
& $0.587\pm0.110$
& $0.637\pm0.039$ \\

& Three-shot
& $0.686\pm0.020$
& $0.683\pm0.023$
& $0.684\pm0.021$
& $0.682\pm0.022$
& $0.753\pm0.026$ \\

& Five-shot
& $0.691\pm0.028$
& $0.688\pm0.029$
& $0.690\pm0.028$
& $0.688\pm0.029$
& $0.769\pm0.025$ \\

\bottomrule
\end{tabular}}
\end{table*}

\begin{figure}
\centering
\includegraphics[width=0.80\linewidth]{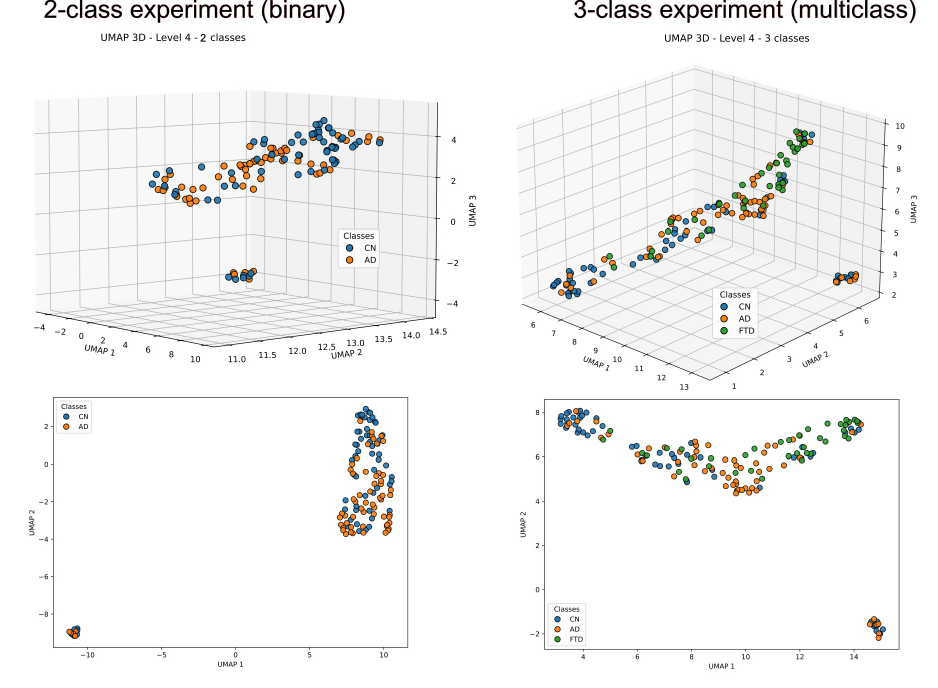}
\caption{\label{fig:umap2} UMAP visualisation of the latent feature distribution for Healthy Controls (HC, blue) Alzheimer’s disease (AD, orange) and frontotemporal dementia subjects (FTD, green). The first plot corresponds to the HC vs AD experiment and the second to the HC vs AD vs FTD experiment. The architecture employed in this figure was the 3D-Swin-S2-Lv4 model.}
\end{figure}

In this case, a slight decrease in performance can be observed with respect to Level 2. For instance, using the MLP with 5-fold cross-validation, the accuracy decreases from 75.75\% at Level 2 to 74.76\% at Level 4. Nevertheless, the overall performance metrics remain highly consistent with those obtained using the baseline models, indicating stable behaviour across hierarchical feature levels. Figure~\ref{fig:umap2} shows the UMAP projection of the Swin-UNET embeddings into a two-dimensional latent space for both the binary and three-class experiments. No clearly separable patterns can be observed between the groups, highlighting the intrinsic heterogeneity of the dataset despite the applied normalisation procedures, particularly considering that the data originate from multiple centres. Furthermore, these results suggest that the differences between groups are likely characterised by complex non-linear relationships that cannot be fully represented in only two dimensions.

\begin{table*}[t]
\centering
\caption{\textbf{Three-class classification performance using Level-4 latent representations (mean $\pm$ SD). Sensitivity values are identical to balanced accuracy and are omitted for brevity.}}
\label{tab:tri_level4}

\scriptsize
\setlength{\tabcolsep}{4pt}

\resizebox{\textwidth}{!}{
\begin{tabular}{llccccc}
\toprule
\textbf{Model} &
\textbf{Validation} &
\textbf{Acc} &
\textbf{Bal Acc} &
\textbf{Prec} &
\textbf{F1} &
\textbf{AUC} \\
\midrule

\multirow{2}{*}{3D-Swin + SVM}

& 5-fold
& $0.588\pm0.049$
& $0.589\pm0.053$
& $0.582\pm0.051$
& $0.580\pm0.049$
& $0.765\pm0.037$ \\

& 70/30
& $0.609\pm0.026$
& $0.610\pm0.023$
& $0.604\pm0.027$
& $0.604\pm0.025$
& $0.757\pm0.026$ \\

\midrule

\multirow{6}{*}{3D-Swin + MLP}

& 5-fold
& $0.607\pm0.046$
& $0.588\pm0.052$
& $0.594\pm0.050$
& $0.589\pm0.051$
& $0.769\pm0.027$ \\

& 70/30
& $0.562\pm0.027$
& $0.544\pm0.026$
& $0.558\pm0.022$
& $0.544\pm0.026$
& $0.748\pm0.019$ \\

& Zero-shot
& $0.446\pm0.000$
& $0.377\pm0.000$
& $0.292\pm0.000$
& $0.326\pm0.000$
& $0.582\pm0.000$ \\

& One-shot
& $0.510\pm0.023$
& $0.469\pm0.037$
& $0.473\pm0.067$
& $0.450\pm0.056$
& $0.654\pm0.019$ \\

& Three-shot
& $0.521\pm0.047$
& $0.482\pm0.062$
& $0.536\pm0.019$
& $0.463\pm0.093$
& $0.714\pm0.015$ \\

& Five-shot
& $0.589\pm0.017$
& $0.569\pm0.025$
& $0.584\pm0.014$
& $0.566\pm0.031$
& $0.762\pm0.021$ \\

\bottomrule
\end{tabular}}
\end{table*}

\subsubsection{Application of 3D-Swin-DiT as classification model}

This section presents the experimental results obtained using the full configuration of the 3D-Swin-DiT two-channel architecture as a feature extractor. As in the previous experiments, SVM and MLP classifiers were employed. The results are reported in Table~\ref{tab:whole_model}.

It can be observed that, when the diffusion component is introduced, the model fails to learn meaningful representations, leading to performance close to random chance. Specifically, the balanced accuracy remains around 50\% for the binary classification task and approximately 33\% for the three-class setting.

We first studied the performance of the 2-channel models and subsequently evaluated the performance of the 1-channel models.

\begin{table*}[t]
\centering
\caption{\textbf{Classification performance using the complete 3D-Swin-DiT architecture with SVM and MLP classifiers (mean $\pm$ SD). Sensitivity values are identical to balanced accuracy and are omitted for brevity.}}
\label{tab:whole_model}

\scriptsize
\setlength{\tabcolsep}{4pt}

\resizebox{\textwidth}{!}{
\begin{tabular}{llccccc}
\toprule
\textbf{Model} &
\textbf{Scenario} &
\textbf{Acc} &
\textbf{Bal Acc} &
\textbf{Prec} &
\textbf{F1} &
\textbf{AUC} \\
\midrule

\multirow{2}{*}{3D-Swin-DiT + SVM}

& 2-classes
& $0.541\pm0.024$
& $0.495\pm0.015$
& $0.467\pm0.111$
& $0.381\pm0.029$
& $0.492\pm0.046$ \\

& 3-classes
& $0.420\pm0.028$
& $0.339\pm0.012$
& $0.164\pm0.052$
& $0.217\pm0.043$
& $0.484\pm0.026$ \\

\midrule

\multirow{4}{*}{3D-Swin-DiT + MLP}

& 2-classes (Level 2)
& $0.556\pm0.036$
& $0.506\pm0.009$
& $0.408\pm0.193$
& $0.373\pm0.041$
& $0.474\pm0.053$ \\

& 2-classes (Level 4)
& $0.552\pm0.038$
& $0.500\pm0.011$
& $0.377\pm0.198$
& $0.364\pm0.029$
& $0.457\pm0.058$ \\

& 3-classes (Level 2)
& $0.415\pm0.033$
& $0.332\pm0.002$
& $0.139\pm0.011$
& $0.195\pm0.011$
& $0.481\pm0.035$ \\

& 3-classes (Level 4)
& $0.416\pm0.032$
& $0.333\pm0.000$
& $0.139\pm0.011$
& $0.196\pm0.010$
& $0.475\pm0.024$ \\

\bottomrule
\end{tabular}}
\end{table*}

Finally, we reported the  single-channel configurations of the sequential 3D-Swin-DiT Level 4 model (SEQ), including the GBP and EWC variants. Since the best performance is consistently achieved when using only the 3D-Swin component of the architecture, the results reported in Table~\ref{tab:other_confgs} were generated using embeddings extracted from this model, together with an MLP classifier, as previous experiments demonstrated its superiority over the SVM approach. The same hyperparameter settings as in the previous sections were maintained.

Regarding the validation strategy, only the $70\%$--$30\%$ train-test split with six different random seeds was employed, in order to capture greater variability compared to k-fold cross-validation (i.e., six splits instead of five).

\begin{table*}[t]
\centering
\caption{\textbf{Classification performance for alternative model configurations using a 70/30 validation split across six random seeds. An MLP classifier was used for all experiments. Sensitivity values are omitted for brevity as they closely match balanced accuracy.}}
\label{tab:other_confgs}

\scriptsize
\setlength{\tabcolsep}{4pt}

\resizebox{\textwidth}{!}{
\begin{tabular}{llccccc}
\toprule
\textbf{Model} &
\textbf{Scenario} &
\textbf{Acc} &
\textbf{Bal Acc} &
\textbf{Prec} &
\textbf{F1} &
\textbf{AUC} \\
\midrule

\multirow{2}{*}{SEQ}

& 2-classes
& $0.724\pm0.035$
& $0.716\pm0.030$
& $0.723\pm0.035$
& $0.717\pm0.031$
& $0.811\pm0.030$ \\

& 3-classes
& $0.588\pm0.037$
& $0.570\pm0.038$
& $0.579\pm0.035$
& $0.570\pm0.039$
& $0.768\pm0.020$ \\

\midrule

\multirow{2}{*}{EWC}

& 2-classes
& $0.724\pm0.036$
& $0.717\pm0.040$
& $0.721\pm0.037$
& $0.717\pm0.039$
& $0.809\pm0.022$ \\

& 3-classes
& $0.605\pm0.039$
& $0.589\pm0.038$
& $0.604\pm0.038$
& $0.591\pm0.039$
& $0.777\pm0.025$ \\

\midrule

\multirow{2}{*}{GBP}

& 2-classes
& $0.743\pm0.025$
& $0.737\pm0.022$
& $0.743\pm0.024$
& $0.737\pm0.022$
& $0.820\pm0.012$ \\

& 3-classes
& $0.613\pm0.034$
& $0.595\pm0.034$
& $0.602\pm0.036$
& $0.595\pm0.036$
& $0.777\pm0.015$ \\

\bottomrule
\end{tabular}}
\end{table*}

The obtained results remain consistently close to those of the baseline models, as shown in Table~\ref{tab:other_confgs}. As expected, although the architecture is reduced from a two-channel to a single-channel configuration, both settings effectively rely on T1-weighted scans only. Therefore, the overall performance remains very similar, as no additional information is introduced or removed between configurations. For instance, the accuracy obtained in the one-channel setting for the binary classification task is 72.40\% (Table~\ref{tab:other_confgs}), compared to 72.81\% reported in Table~\ref{tab:bin_level4} for the corresponding experiment.

\begin{table*}[t]
\centering
\caption{\textbf{Classification performance for EWC and GBP configurations of the complete model architecture (3D-Swin-DiT) using a 70/30 validation split across six random seeds. An MLP classifier was used for all experiments. Sensitivity is omitted for brevity as it closely matches balanced accuracy.}}
\label{tab:other_confgs_wm}

\scriptsize
\setlength{\tabcolsep}{2pt}

\begin{tabular}{llccccc}
\toprule
\textbf{Model} &
\textbf{Scenario} &
\textbf{Acc} &
\textbf{Bal Acc} &
\textbf{Prec} &
\textbf{F1} &
\textbf{AUC} \\
\midrule

\multirow{2}{*}{EWC}

& 2-classes
& $0.552\pm0.033$
& $0.500\pm0.001$
& $0.304\pm0.075$
& $0.358\pm0.019$
& N/A \\

& 3-classes
& $0.416\pm0.032$
& $0.333\pm0.000$
& $0.139\pm0.011$
& $0.196\pm0.011$
& $0.526\pm0.026$ \\

\midrule

\multirow{2}{*}{GBP}

& 2-classes
& $0.553\pm0.035$
& $0.500\pm0.000$
& $0.277\pm0.018$
& $0.356\pm0.015$
& N/A \\

& 3-classes
& $0.417\pm0.033$
& $0.335\pm0.003$
& $0.194\pm0.131$
& $0.198\pm0.014$
& $0.492\pm0.024$ \\

\bottomrule
\end{tabular}
\end{table*}

\begin{figure}
\centering
\includegraphics[width=\linewidth]{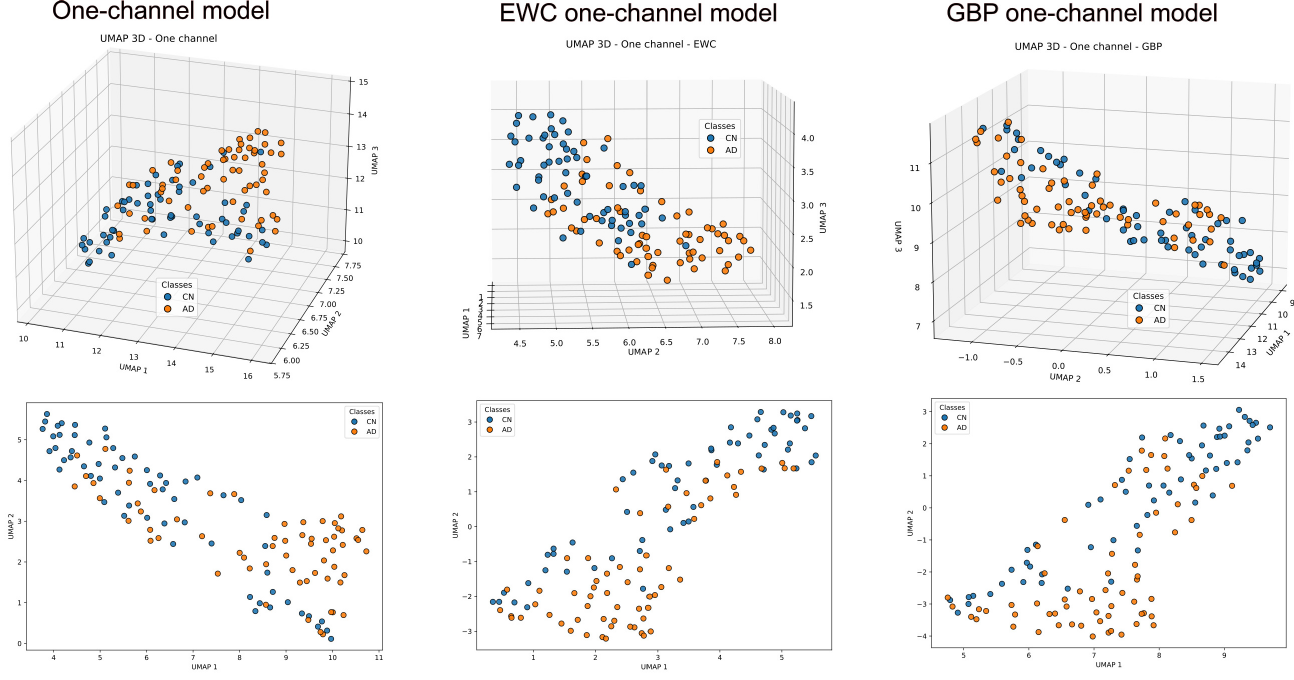}
\caption{\label{fig:umap1} UMAP visualisation of the latent feature distribution for Healthy Controls (HC, blue) and Alzheimer’s disease subjects (AD, orange). The first plot corresponds to the original model, the second to the model trained with EWC, and the third to the GBP model. The architecture employed in this figure was the 3D-Swin-DiT of S2-Lv4 sequential model of 1-channel (SEQ).}
\end{figure}

Regarding the different single-channel variants, GBP appears to yield slightly better performance than the others. In particular, it achieves a balanced accuracy of 73.67\% for the binary classification task, whereas the remaining two configurations obtain 71.58\% and 71.66\%, respectively. In terms of AUC, all three models show very similar performance, with values close to 0.81 in the binary scenario. As illustrated in Figure~\ref{fig:umap1}, the three configurations show a similar latent-space organisation of the Swin-UNET embeddings, with two main clusters emerging in the UMAP representation. One cluster is predominantly composed of AD subjects, whereas the other contains a higher proportion of HC subjects, suggesting that the models are able to capture disease-related patterns in the learned feature space. In addition to these large groups, several smaller subclusters can also be observed, indicating the presence of intra-group heterogeneity and potentially reflecting different subject-specific features or intermediate patterns between both classes.

\subsection{Clinical Task of Data Generation in Brain Tumor Cohorts}
\subsection{Reconstruction}

The results are reported in Table~\ref{tab:reconstruction}. {MAISI}~\cite{guo2025maisi} achieves the strongest reconstruction overall, leading on PSNR across nearly all modalities and cohorts, which we attribute to its large-scale CT/MRI pre-training; its weakness is its SSIM on tumor cohorts, where it is occasionally surpassed by 3D-Swin. {AEKL}~\cite{rombach2022high} provides a solid continuous-latent baseline, consistently ranking second on PSNR for the tumor datasets, but degrades sharply on IXI (e.g., MRA SSIM 53.49), reflecting its sensitivity to modalities outside its training distribution. {VQVAE}~\cite{van2017neural} produces high SSIM on several columns, a characteristic trade-off of discrete-latent models: codebook quantisation preserves perceptual structure but limits absolute intensity fidelity. {3D-Swin} is competitive on both PSNR and SSIM throughout and achieves the best PSNR and SSIM on IXI MRA, indicating stronger generalisation to out-of-distribution modalities.

\begin{figure*}
\centering
\includegraphics[width=\textwidth]{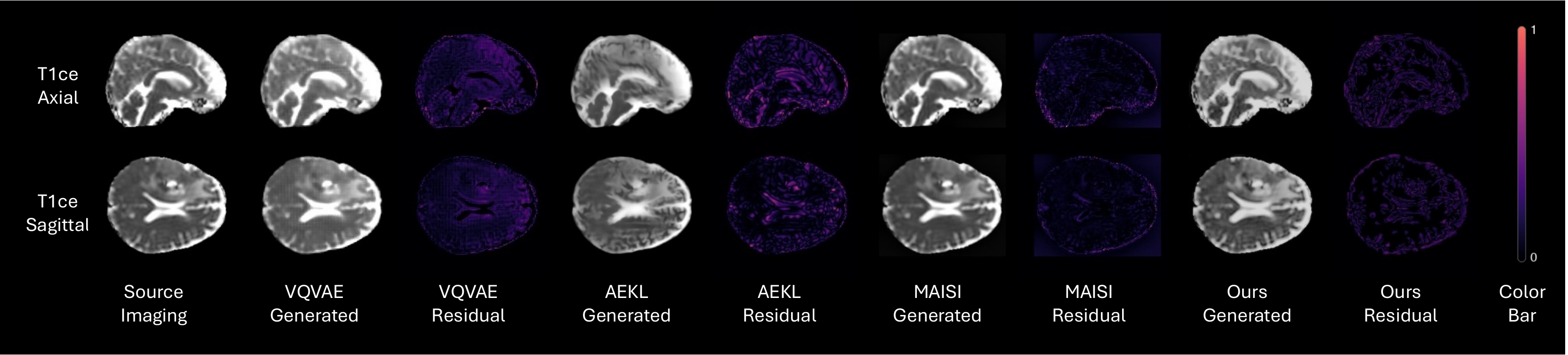}
\caption{
\textbf{Reconstruction of T1ce-weighted MRI using latent generative models.}
Representative axial and sagittal slices comparing image reconstructions obtained with VQVAE, AEKL, MAISI, and the proposed method. The first column shows the source T1ce image, followed by the reconstructed images and their corresponding residual error maps. 
}
\label{fig:t1recon}
\end{figure*}

Figure \ref{fig:t1recon} presents the residual maps highlight voxel-wise discrepancies between the reconstructed and reference images, providing a qualitative assessment of reconstruction fidelity. Compared with competing approaches, the proposed method yields anatomically faithful reconstructions with reduced residual errors, preserving fine structural details and tissue boundaries while minimizing reconstruction artefacts.

\begin{figure*}
\centering
\includegraphics[width=\textwidth]{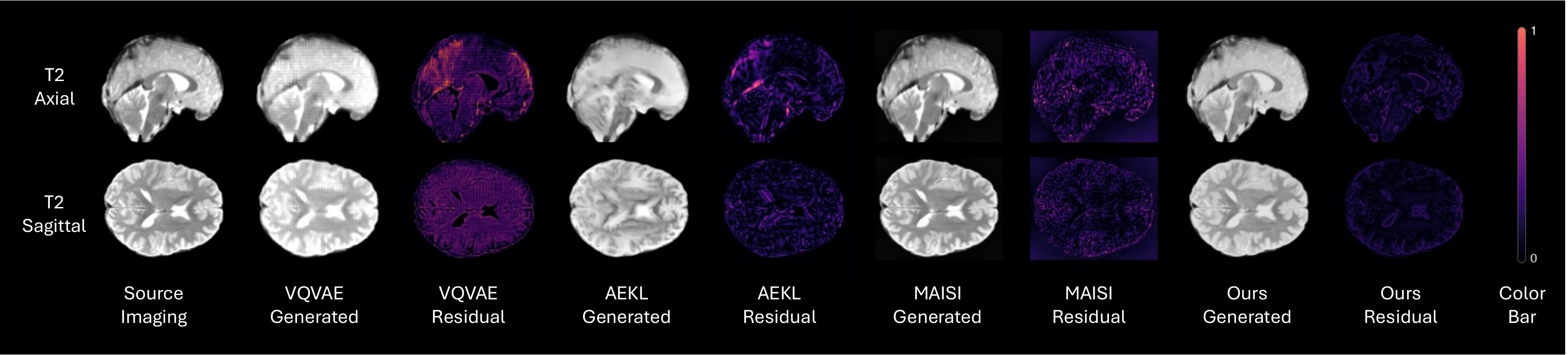}
\caption{
\textbf{Reconstruction of T2-weighted MRI using latent generative models.}
Representative axial and sagittal T2-weighted brain MRI reconstructions generated by VQVAE, AEKL, MAISI, and the proposed framework. Residual error maps are displayed alongside each reconstruction to highlight local reconstruction discrepancies. 
}
\label{fig:t2recon}
\end{figure*}

\begin{table*}[h]
\centering
\caption{\textbf{Reconstruction performance.} BraTS reports the mean results over the SSA, MET, and PED cohorts. Metrics are reported as PSNR / SSIM ($\times 10^2$) for each modality. The results are in mean$\pm$std. \textbf{Bold} indicates the best and \underline{underline} indicates the second-best per column, ranked by mean.}
\label{tab:reconstruction}
\resizebox{\textwidth}{!}{
\begin{tabular}{l|cccc|cccccc}
\toprule
\multirow{2}{*}{Encoder}
& \multicolumn{4}{c|}{BraTS (SSA+MET+PED)}
& \multicolumn{6}{c}{TCGA-LGG} \\
\cmidrule(lr){2-5}\cmidrule(lr){6-11}
& T1 & T1ce & T2 & FLAIR
& T1 & T1ce & T2 & FLAIR & DWI & ADC \\
\midrule
VQVAE
& 26.29$\pm$0.68 / 94.89$\pm$0.87
& 25.91$\pm$0.94 / 94.30$\pm$1.08
& 27.11$\pm$1.04 / \textbf{95.95$\pm$0.59}
& 25.88$\pm$0.81 / 94.55$\pm$0.84
& 24.32$\pm$1.22 / 92.64$\pm$1.98
& 24.45$\pm$1.60 / 92.56$\pm$1.70
& 25.24$\pm$1.19 / 94.43$\pm$1.35
& 24.38$\pm$1.21 / 92.44$\pm$1.09
& 26.16$\pm$1.34 / 95.25$\pm$0.99
& 27.13$\pm$0.61 / 95.94$\pm$0.67 \\
AEKL
& \underline{30.50$\pm$0.94} / \underline{94.98$\pm$0.89}
& \underline{29.39$\pm$1.12} / 93.69$\pm$1.21
& \underline{28.25$\pm$1.24} / 92.67$\pm$0.88
& \underline{29.51$\pm$1.15} / 93.22$\pm$0.95
& \underline{29.35$\pm$2.52} / 91.89$\pm$6.44
& \underline{28.36$\pm$1.18} / 91.54$\pm$1.40
& \underline{28.78$\pm$1.53} / 91.00$\pm$1.21
& \underline{29.18$\pm$1.55} / 90.62$\pm$1.34
& \underline{29.45$\pm$2.16} / 91.15$\pm$7.46
& \underline{27.85$\pm$1.48} / 90.77$\pm$1.25 \\
MAISI
& \textbf{31.07$\pm$1.32} / \textbf{96.49$\pm$1.48}
& \textbf{29.52$\pm$1.43} / \underline{94.75$\pm$2.09}
& \textbf{30.82$\pm$1.56} / \underline{95.74$\pm$1.17}
& \textbf{30.87$\pm$1.59} / \textbf{95.24$\pm$1.29}
& \textbf{32.08$\pm$2.46} / \textbf{96.47$\pm$2.18}
& \textbf{30.85$\pm$1.98} / \textbf{95.88$\pm$1.78}
& \textbf{35.01$\pm$1.63} / \textbf{98.14$\pm$0.72}
& \textbf{32.85$\pm$2.16} / \textbf{96.23$\pm$1.60}
& \textbf{36.68$\pm$1.79} / \textbf{98.87$\pm$0.40}
& \textbf{35.68$\pm$1.70} / \textbf{98.51$\pm$0.46} \\
3D-Swin
& 28.52$\pm$1.40 / 94.95$\pm$1.01
& 28.91$\pm$1.62 / \textbf{94.95$\pm$1.15}
& 24.75$\pm$1.39 / 93.14$\pm$1.06
& 28.87$\pm$1.22 / \underline{95.10$\pm$0.74}
& 27.88$\pm$1.00 / \underline{95.31$\pm$1.00}
& 28.18$\pm$0.42 / \underline{95.81$\pm$1.06}
& 26.67$\pm$1.37 / 93.53$\pm$1.90
& 27.64$\pm$1.50 / \underline{95.57$\pm$0.78}
& 26.23$\pm$1.53 / 92.38$\pm$0.94
& 26.65$\pm$0.00 / 91.93$\pm$0.00 \\
\bottomrule
\end{tabular}
}
\vspace{0.5em}
\resizebox{\textwidth}{!}{
\begin{tabular}{l|ccccc|ccccc}
\toprule
\multirow{2}{*}{Encoder}
& \multicolumn{5}{c|}{IXI}
& \multicolumn{5}{c}{RHUH-GBM} \\
\cmidrule(lr){2-6}\cmidrule(lr){7-11}
& T1 & T2 & PD & MRA & DTI
& T1 & T1ce & T2 & FLAIR & ADC \\
\midrule
VQVAE
& 22.81$\pm$0.69 / \textbf{88.46$\pm$1.49}
& \underline{25.09$\pm$0.54} / \textbf{92.02$\pm$0.89}
& 22.78$\pm$0.47 / \textbf{90.34$\pm$1.02}
& 20.44$\pm$0.52 / 72.59$\pm$3.22
& \textbf{26.68$\pm$0.54} / \textbf{94.55$\pm$0.45}
& 26.58$\pm$0.64 / 94.65$\pm$0.78
& 25.04$\pm$0.79 / \underline{93.03$\pm$0.93}
& 27.48$\pm$0.75 / \underline{95.88$\pm$0.46}
& 25.70$\pm$0.75 / \underline{94.01$\pm$0.92}
& 27.67$\pm$0.73 / \underline{96.82$\pm$0.46} \\
AEKL
& 21.96$\pm$1.38 / 78.06$\pm$4.14
& 22.84$\pm$0.57 / 76.06$\pm$2.93
& 23.10$\pm$1.14 / 78.83$\pm$2.66
& 16.85$\pm$1.00 / 53.49$\pm$5.38
& 24.51$\pm$0.81 / 71.72$\pm$3.34
& \underline{30.48$\pm$0.77} / 94.81$\pm$0.72
& \underline{27.86$\pm$1.02} / 92.45$\pm$1.22
& \underline{27.71$\pm$1.01} / 92.34$\pm$0.66
& \underline{28.48$\pm$1.02} / 92.68$\pm$0.96
& \underline{28.68$\pm$0.72} / 92.92$\pm$0.72 \\
MAISI
& \textbf{28.87$\pm$0.82} / 86.60$\pm$1.82
& \textbf{30.47$\pm$0.75} / \underline{85.67$\pm$2.30}
& \textbf{29.99$\pm$0.79} / \underline{86.07$\pm$2.40}
& \underline{23.09$\pm$0.82} / \underline{74.62$\pm$2.22}
& \underline{25.17$\pm$0.59} / \underline{87.09$\pm$3.25}
& \textbf{30.63$\pm$0.86} / \underline{95.64$\pm$1.25}
& 26.78$\pm$1.36 / 90.84$\pm$2.02
& \textbf{30.14$\pm$1.03} / \textbf{94.75$\pm$0.99}
& \textbf{28.72$\pm$1.21} / 92.87$\pm$1.58
& \textbf{35.24$\pm$1.17} / \textbf{98.20$\pm$0.64} \\
3D-Swin
& \underline{24.94$\pm$0.71} / \underline{86.74$\pm$2.83}
& 23.32$\pm$0.33 / 82.03$\pm$2.91
& \underline{25.84$\pm$0.82} / 83.17$\pm$3.00
& \textbf{25.76$\pm$0.67} / \textbf{80.03$\pm$3.79}
& 23.39$\pm$0.36 / 82.76$\pm$2.24
& 29.91$\pm$1.72 / \textbf{95.76$\pm$1.11}
& \textbf{29.14$\pm$1.91} / \textbf{95.09$\pm$1.46}
& 24.13$\pm$0.75 / \textbf{93.16$\pm$0.64}
& 28.17$\pm$1.68 / \textbf{94.85$\pm$1.10}
& 24.41$\pm$0.60 / 93.62$\pm$0.45 \\
\bottomrule
\end{tabular}
}
\end{table*} 
The proposed method consistently achieves superior reconstruction quality, preserving anatomical details and tissue contrast while minimizing residual errors, thereby demonstrating improved robustness and fidelity across imaging planes (Figure \ref{fig:t2recon}).

\subsection{Generation}

\begin{figure*}
\centering
\includegraphics[width=\textwidth]{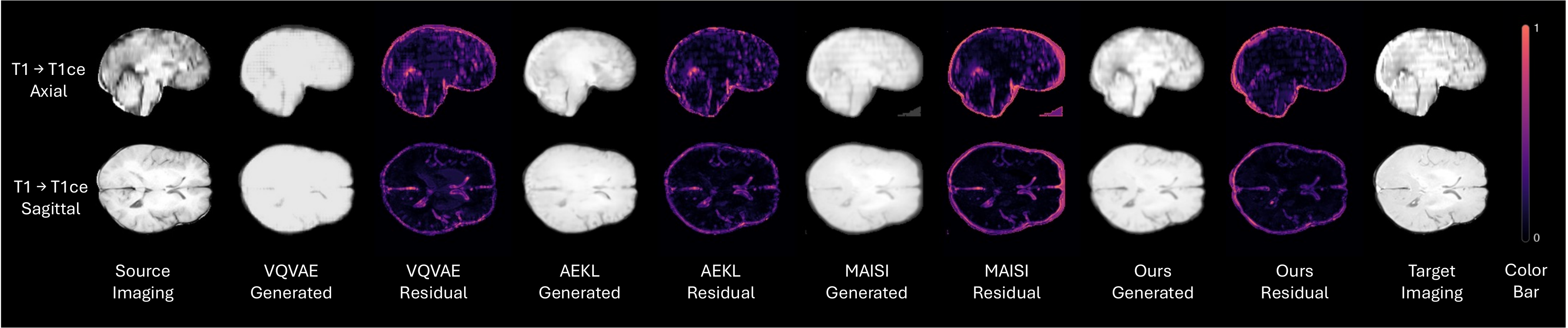}
\caption{
\textbf{Cross-modality image translation from T1-weighted to contrast-enhanced T1 (T1ce) MRI.}
Representative axial and sagittal examples of T1-to-T1ce synthesis generated by VQVAE, AEKL, MAISI, and the proposed framework. The source T1 image is shown on the left, whereas the ground-truth T1ce image is displayed on the far right. 
}
\label{fig:t1t1ce}
\end{figure*}

\begin{table*}[h]
\centering
\small
\setlength{\tabcolsep}{4pt}
\caption{\textbf{Latent translation performance.} Results are reported as PSNR / SSIM ($\times 10^2$) in mean$\pm$std. \textbf{Bold} indicates the best and \underline{underline} indicates the second-best per column within each generator block, ranked by mean.}
\label{tab:translation_all}
\resizebox{\textwidth}{!}{
\begin{tabular}{l|ccc|ccc|ccc}
\toprule
\multirow{2}{*}{\textbf{Encoder}}
& \multicolumn{3}{c|}{\textbf{SSA}}
& \multicolumn{3}{c|}{\textbf{PED}}
& \multicolumn{3}{c}{\textbf{MET}} \\
\cmidrule(lr){2-4}\cmidrule(lr){5-7}\cmidrule(lr){8-10}
& T1$\to$T2 & T2$\to$FLAIR & T1$\to$T1ce
& T1$\to$T2 & T2$\to$FLAIR & T1$\to$T1ce
& T1$\to$T2 & T2$\to$FLAIR & T1$\to$T1ce \\
\midrule
& \multicolumn{9}{c}{\textbf{LDM}} \\
\midrule
VQVAE
& 19.83$\pm$1.74 / 87.42$\pm$1.69 & 20.18$\pm$1.93 / \underline{87.85$\pm$1.61} & 21.74$\pm$2.31 / 89.16$\pm$2.08
& 19.27$\pm$2.62 / 88.04$\pm$2.43 & 21.62$\pm$1.41 / \underline{88.74$\pm$1.78} & 22.83$\pm$2.04 / 90.18$\pm$1.85
& 20.14$\pm$2.71 / \textbf{88.17$\pm$3.18} & 21.45$\pm$2.34 / \underline{88.62$\pm$2.27} & 22.46$\pm$3.04 / \underline{89.41$\pm$3.27} \\
AEKL
& 18.49$\pm$2.00 / 84.80$\pm$1.76 & 19.74$\pm$1.43 / 82.63$\pm$2.18 & 21.22$\pm$1.32 / 88.49$\pm$1.80
& 18.92$\pm$1.93 / 84.47$\pm$1.93 & 19.86$\pm$1.27 / 81.95$\pm$2.43 & \textbf{25.23$\pm$2.68} / \textbf{91.91$\pm$2.65}
& 19.20$\pm$1.66 / 84.91$\pm$2.73 & 20.41$\pm$1.85 / 83.18$\pm$2.61 & 24.31$\pm$2.52 / \textbf{91.33$\pm$3.18} \\
MAISI
& \underline{20.27$\pm$1.74} / \underline{87.99$\pm$1.99} & \underline{22.08$\pm$0.69} / 63.10$\pm$1.34 & \underline{23.43$\pm$1.12} / \underline{89.24$\pm$1.53}
& \underline{20.98$\pm$1.52} / \underline{89.15$\pm$2.10} & \underline{22.02$\pm$1.41} / 88.15$\pm$1.21 & 24.57$\pm$1.13 / 87.32$\pm$5.87
& \textbf{20.95$\pm$0.84} / \underline{88.12$\pm$2.01} & \underline{22.08$\pm$1.65} / 88.04$\pm$1.35 & 23.06$\pm$1.35 / 87.66$\pm$5.62 \\
3D-Swin
& \textbf{21.65$\pm$1.42} / \textbf{89.18$\pm$0.88} & \textbf{23.84$\pm$2.07} / \textbf{90.62$\pm$1.18} & \textbf{25.18$\pm$1.74} / \textbf{91.42$\pm$1.45}
& \textbf{21.92$\pm$2.58} / \textbf{89.74$\pm$1.24} & \textbf{24.85$\pm$1.62} / \textbf{90.18$\pm$1.15} & \underline{24.73$\pm$1.85} / \underline{89.65$\pm$1.58}
& \underline{20.62$\pm$1.48} / 87.42$\pm$2.18 & \textbf{24.18$\pm$1.78} / \textbf{89.85$\pm$1.92} & \textbf{24.48$\pm$2.65} / 89.32$\pm$3.85 \\
\midrule
& \multicolumn{9}{c}{\textbf{MOTFM}} \\
\midrule
VQVAE
& 23.29$\pm$1.60 / \underline{91.62$\pm$1.55} & 23.63$\pm$2.17 / 92.03$\pm$1.46 & 25.57$\pm$2.48 / 93.45$\pm$1.95
& 22.58$\pm$2.50 / \textbf{92.81$\pm$2.07} & 25.34$\pm$1.34 / 93.15$\pm$1.62 & \underline{26.61$\pm$1.95} / \textbf{94.42$\pm$1.74}
& 23.71$\pm$2.61 / \textbf{92.63$\pm$3.22} & 25.19$\pm$2.27 / 93.11$\pm$2.09 & \underline{26.30$\pm$3.20} / \textbf{93.77$\pm$3.19} \\
AEKL
& \underline{24.72$\pm$1.84} / 89.38$\pm$1.50 & 23.02$\pm$1.02 / 88.01$\pm$1.10 & 25.46$\pm$2.06 / 90.97$\pm$1.96
& 23.75$\pm$1.77 / 89.38$\pm$1.71 & 23.23$\pm$0.94 / 87.64$\pm$2.01 & 26.27$\pm$2.05 / 92.29$\pm$2.47
& 23.90$\pm$1.80 / 89.62$\pm$2.38 & 23.54$\pm$1.57 / 88.64$\pm$2.10 & 25.90$\pm$2.51 / 91.85$\pm$3.20 \\
MAISI
& \textbf{24.91$\pm$1.75} / 89.36$\pm$1.91 & \underline{25.89$\pm$1.84} / 89.56$\pm$1.31 & \underline{26.21$\pm$1.87} / 91.80$\pm$1.72
& \underline{24.02$\pm$2.73} / 91.41$\pm$1.75 & \underline{26.68$\pm$0.90} / 91.02$\pm$1.29 & \underline{26.68$\pm$1.55} / 92.20$\pm$2.24
& \underline{23.94$\pm$1.36} / 90.08$\pm$1.07 & \underline{26.13$\pm$1.69} / \underline{92.03$\pm$1.46} & 26.26$\pm$1.87 / 91.37$\pm$1.81 \\
3D-Swin
& 24.62$\pm$1.18 / \textbf{92.74$\pm$0.62} & \textbf{27.18$\pm$1.52} / \textbf{93.18$\pm$0.74} & \textbf{28.92$\pm$1.68} / \textbf{94.42$\pm$1.05}
& \textbf{24.18$\pm$2.34} / \underline{92.18$\pm$1.42} & \textbf{27.85$\pm$1.18} / \textbf{94.05$\pm$0.78} & \textbf{27.18$\pm$2.12} / \underline{93.18$\pm$1.55}
& \textbf{24.18$\pm$1.52} / \underline{90.65$\pm$1.85} & \textbf{27.42$\pm$1.45} / \textbf{93.42$\pm$1.38} & \textbf{26.62$\pm$1.58} / \underline{92.85$\pm$1.48} \\
\midrule
& \multicolumn{9}{c}{\textbf{Rectified Flow}} \\
\midrule
VQVAE
& 22.05$\pm$1.86 / \underline{90.42$\pm$1.68} & 23.48$\pm$1.72 / 91.19$\pm$1.62 & 24.53$\pm$1.73 / \underline{92.69$\pm$1.56}
& 21.55$\pm$3.11 / \textbf{91.13$\pm$3.24} & 24.41$\pm$1.33 / \underline{91.33$\pm$1.64} & 24.59$\pm$1.92 / \underline{92.88$\pm$1.66}
& 22.69$\pm$2.23 / \textbf{91.41$\pm$2.93} & 24.59$\pm$2.26 / \underline{91.83$\pm$2.24} & 24.37$\pm$2.58 / \textbf{92.34$\pm$3.08} \\
AEKL
& \underline{24.11$\pm$1.73} / 89.39$\pm$1.42 & 24.16$\pm$1.63 / 89.42$\pm$1.38 & 24.72$\pm$1.93 / 90.19$\pm$2.20
& \underline{23.82$\pm$1.65} / \underline{89.77$\pm$1.53} & 24.69$\pm$0.88 / 89.10$\pm$1.84 & 25.99$\pm$1.58 / 92.06$\pm$1.94
& \underline{23.91$\pm$1.77} / 89.77$\pm$2.39 & \underline{24.90$\pm$1.76} / 90.13$\pm$1.90 & \underline{25.60$\pm$2.25} / \underline{91.74$\pm$2.95} \\
MAISI
& 23.35$\pm$0.84 / 71.58$\pm$5.76 & \underline{25.03$\pm$1.40} / \underline{93.00$\pm$1.22} & \underline{27.01$\pm$2.11} / 91.71$\pm$1.37
& 23.06$\pm$3.12 / 85.82$\pm$1.69 & \underline{25.93$\pm$0.62} / 90.56$\pm$1.08 & \underline{26.91$\pm$1.41} / 91.92$\pm$1.49
& \textbf{24.70$\pm$1.69} / 83.97$\pm$2.87 & \underline{25.28$\pm$2.13} / 90.79$\pm$1.56 & 24.03$\pm$2.98 / 91.28$\pm$1.17 \\
3D-Swin
& \textbf{24.85$\pm$1.32} / \textbf{92.45$\pm$0.78} & \textbf{26.74$\pm$1.78} / \textbf{93.05$\pm$0.92} & \textbf{28.45$\pm$1.92} / \textbf{94.18$\pm$1.18}
& \textbf{24.22$\pm$2.65} / 88.42$\pm$1.85 & \textbf{26.42$\pm$1.35} / \textbf{93.62$\pm$0.95} & \textbf{27.65$\pm$2.28} / \textbf{93.05$\pm$1.62}
& 23.84$\pm$1.32 / \underline{89.85$\pm$1.95} & \textbf{27.17$\pm$1.62} / \textbf{93.18$\pm$1.55} & \textbf{25.92$\pm$1.78} / \underline{92.18$\pm$1.62} \\
\midrule
& \multicolumn{9}{c}{\textbf{DiT (Pretrained) }} \\
\midrule
VQVAE
& 23.61$\pm$1.72 / 91.12$\pm$1.58 & 26.34$\pm$2.08 / \underline{92.58$\pm$1.41} & \underline{26.91$\pm$2.32} / \underline{93.37$\pm$1.61}
& 23.04$\pm$2.63 / \textbf{92.61$\pm$1.92} & \textbf{27.46$\pm$1.34 }/ \textbf{92.88$\pm$1.45} & \underline{27.18$\pm$2.04} / \textbf{94.18$\pm$1.62}
& 22.86$\pm$2.48 / \textbf{92.21$\pm$2.96} & 26.31$\pm$2.18 / \underline{92.72$\pm$1.84} & 26.04$\pm$2.81 / \textbf{93.42$\pm$2.71} \\
AEKL
& \underline{24.48$\pm$1.83} / 89.51$\pm$1.77 & 23.18$\pm$1.42 / 88.06$\pm$1.91 & 25.41$\pm$2.06 / 90.84$\pm$2.03
& \textbf{24.72$\pm$1.78} / 89.63$\pm$1.74 & 23.42$\pm$1.25 / 88.18$\pm$2.10 & 25.96$\pm$2.12 / 92.31$\pm$2.34
& \textbf{24.31$\pm$1.95} / 89.06$\pm$2.52 & 23.56$\pm$1.69 / 88.43$\pm$2.31 & 25.47$\pm$2.48 / 91.36$\pm$3.08 \\
MAISI
& 23.74$\pm$1.56 / \underline{91.46$\pm$1.49} & \underline{26.78$\pm$1.75} / 91.22$\pm$1.28 & 25.72$\pm$1.92 / 91.88$\pm$1.53
& 23.39$\pm$2.74 / 90.76$\pm$1.58 & 26.83$\pm$1.08 / 91.02$\pm$1.22 & 26.62$\pm$1.68 / 91.38$\pm$2.05
& 23.18$\pm$1.54 / 89.71$\pm$1.37 & 26.92$\pm$1.72 / 91.18$\pm$1.51 & \underline{26.58$\pm$1.86} / 90.96$\pm$2.05 \\
3D-Swin-DiT
& \textbf{25.06$\pm$1.28} / \textbf{92.47$\pm$0.86} & \textbf{27.31$\pm$1.55} / \textbf{93.02$\pm$1.04} & \textbf{28.06$\pm$1.71} / \textbf{93.74$\pm$1.16}
& \underline{24.19$\pm$2.34} / \underline{91.84$\pm$1.36} & \underline{27.22$\pm$1.17} / \underline{92.71$\pm$0.96} & \textbf{27.46$\pm$2.08} / \underline{92.71$\pm$1.56}
& \underline{23.79$\pm$1.62} / \underline{90.48$\pm$1.94} & \textbf{27.94$\pm$1.48} / \textbf{93.28$\pm$1.43} & \textbf{26.83$\pm$1.69} / \underline{92.24$\pm$1.62} \\
\bottomrule
\end{tabular}
}
\label{tab:translation_brats}
\end{table*}

\begin{table*}[h]
\centering
\small
\setlength{\tabcolsep}{4pt}
\caption{\textbf{Latent translation performance (continued).} Results are reported as PSNR / SSIM ($\times 10^2$) in mean$\pm$std. \textbf{Bold} indicates the best and \underline{underline} indicates the second-best per column within each generator block, ranked by mean.}
\resizebox{\textwidth}{!}{
\begin{tabular}{l|ccc|c|ccc}
\toprule
\multirow{2}{*}{\textbf{Encoder}}
& \multicolumn{3}{c|}{\textbf{TCGA-LGG}}
& \multicolumn{1}{c|}{\textbf{IXI}}
& \multicolumn{3}{c}{\textbf{RHUH-GBM}} \\
\cmidrule(lr){2-4}\cmidrule(lr){5-5}\cmidrule(lr){6-8}
& T1$\to$T2 & T2$\to$FLAIR & T1$\to$T1ce
& T1$\to$T2
& T1$\to$T2 & T2$\to$FLAIR & T1$\to$T1ce \\
\midrule
& \multicolumn{7}{c}{\textbf{LDM}} \\
\midrule
VQVAE
& 16.42$\pm$2.38 / 78.94$\pm$2.84 & \underline{18.27$\pm$1.58} / 81.46$\pm$2.27 & 16.83$\pm$3.91 / 79.27$\pm$4.36
& 13.74$\pm$0.74 / 56.18$\pm$4.27
& 19.74$\pm$1.94 / 87.83$\pm$2.18 & 20.46$\pm$1.86 / \underline{88.62$\pm$1.74} & 22.04$\pm$2.41 / 89.18$\pm$2.46 \\
AEKL
& 14.98$\pm$1.54 / 75.64$\pm$1.51 & 17.36$\pm$1.72 / 75.18$\pm$2.41 & 18.11$\pm$2.89 / 79.72$\pm$3.58
& \underline{15.17$\pm$1.26} / 64.45$\pm$3.19
& 19.08$\pm$1.30 / 85.16$\pm$1.30 & 19.63$\pm$1.21 / 83.49$\pm$1.78 & 22.78$\pm$1.93 / 89.31$\pm$1.93 \\
MAISI
& \underline{19.11$\pm$1.97} / \underline{81.19$\pm$2.85} & 18.19$\pm$1.95 / \underline{82.92$\pm$1.45} & \underline{18.66$\pm$2.97} / \underline{81.41$\pm$3.01}
& 15.14$\pm$0.56 / \underline{71.86$\pm$1.44}
& \textbf{21.06$\pm$0.69} / \underline{87.99$\pm$1.77} & \underline{22.47$\pm$1.69} / 86.46$\pm$1.32 & \textbf{25.88$\pm$2.23} / \underline{89.75$\pm$2.78} \\
3D-Swin
& \textbf{19.85$\pm$1.84} / \textbf{81.42$\pm$2.42} & \textbf{19.74$\pm$1.95} / \textbf{83.18$\pm$1.65} & \textbf{19.45$\pm$2.62} / \textbf{84.18$\pm$2.85}
& \textbf{18.92$\pm$0.62} / \textbf{74.18$\pm$2.85}
& \underline{20.65$\pm$0.85} / \textbf{88.42$\pm$1.42} & \textbf{23.85$\pm$2.32} / \textbf{89.74$\pm$1.58} & \underline{24.62$\pm$2.48} / \textbf{90.18$\pm$2.42} \\
\midrule
& \multicolumn{7}{c}{\textbf{MOTFM}} \\
\midrule
VQVAE
& 19.09$\pm$2.19 / 83.63$\pm$3.14 & 21.31$\pm$1.46 / 86.82$\pm$2.34 & 18.75$\pm$4.34 / 84.53$\pm$4.77
& 15.95$\pm$0.86 / 62.76$\pm$4.57
& 23.40$\pm$2.06 / \textbf{92.28$\pm$2.34} & 23.69$\pm$1.94 / 92.21$\pm$1.53 & 25.22$\pm$2.57 / 92.76$\pm$2.62 \\
AEKL
& \underline{20.51$\pm$2.14} / 81.08$\pm$2.19 & 20.03$\pm$1.95 / 80.39$\pm$1.76 & \underline{20.88$\pm$3.84} / 81.94$\pm$4.62
& \underline{17.55$\pm$0.92} / \underline{68.12$\pm$3.56}
& \underline{23.65$\pm$1.38} / 89.20$\pm$1.42 & 22.45$\pm$1.07 / 88.18$\pm$1.07 & 24.57$\pm$2.51 / 90.24$\pm$2.35 \\
MAISI
& 20.31$\pm$1.51 / \underline{85.77$\pm$2.69} & \underline{21.96$\pm$1.92} / \textbf{90.18$\pm$1.72} & 19.50$\pm$2.89 / 83.06$\pm$1.91
& 17.73$\pm$0.79 / 67.42$\pm$3.34
& 22.50$\pm$1.82 / \underline{91.24$\pm$1.08} & \underline{24.47$\pm$2.30} / \underline{92.24$\pm$1.52} & \underline{26.27$\pm$2.56} / \underline{93.07$\pm$1.99} \\
3D-Swin
& \textbf{21.65$\pm$1.62} / \textbf{88.42$\pm$1.45} & \textbf{22.18$\pm$1.45} / \underline{89.18$\pm$1.05} & \textbf{25.18$\pm$2.32} / \textbf{88.62$\pm$1.95}
& \textbf{18.42$\pm$0.42} / \textbf{71.85$\pm$3.18}
& \textbf{24.85$\pm$0.92} / 91.18$\pm$0.92 & \textbf{26.42$\pm$1.85} / \textbf{92.85$\pm$1.32} & \textbf{27.42$\pm$2.18} / \textbf{93.42$\pm$1.65} \\
\midrule
& \multicolumn{7}{c}{\textbf{Rectified Flow}} \\
\midrule
VQVAE
& 18.28$\pm$2.75 / 81.89$\pm$2.75 & 20.86$\pm$2.30 / 85.07$\pm$2.84 & 18.43$\pm$3.79 / 83.68$\pm$4.19
& 15.56$\pm$0.65 / 59.99$\pm$4.38
& 21.98$\pm$1.85 / \textbf{90.91$\pm$1.80} & 23.68$\pm$2.04 / \underline{91.29$\pm$1.42} & 24.15$\pm$2.57 / 91.76$\pm$2.65 \\
AEKL
& 20.55$\pm$1.92 / 81.62$\pm$2.00 & \underline{21.54$\pm$1.54} / 82.66$\pm$1.95 & \underline{19.93$\pm$3.01} / 79.62$\pm$3.66
& \underline{17.61$\pm$0.98} / \underline{68.33$\pm$3.70}
& \underline{23.77$\pm$1.43} / 89.38$\pm$1.35 & 23.69$\pm$1.55 / 89.63$\pm$0.98 & 24.87$\pm$1.97 / 90.33$\pm$2.01 \\
MAISI
& \underline{21.15$\pm$1.96} / \textbf{86.35$\pm$1.99} & 21.36$\pm$1.10 / \textbf{89.39$\pm$1.01} & 19.46$\pm$2.86 / \textbf{87.64$\pm$1.81}
& 18.59$\pm$0.63 / 66.74$\pm$3.18
& 21.79$\pm$1.61 / 89.65$\pm$1.25 & \underline{25.05$\pm$1.46} / 91.28$\pm$1.30 & \underline{26.10$\pm$2.15} / \underline{92.80$\pm$1.39} \\
3D-Swin
& \textbf{22.18$\pm$1.74} / \underline{84.85$\pm$1.62} & \textbf{22.05$\pm$1.62} / \underline{88.42$\pm$1.18} & \textbf{20.18$\pm$2.45} / \underline{86.74$\pm$2.18}
& \textbf{19.18$\pm$0.45} / \textbf{72.42$\pm$3.05}
& \textbf{24.42$\pm$0.95} / \underline{90.85$\pm$1.05} & \textbf{26.18$\pm$2.05} / \textbf{92.62$\pm$1.42} & \textbf{27.18$\pm$2.32} / \textbf{93.18$\pm$1.85} \\
\midrule
& \multicolumn{7}{c}{\textbf{DiT (Pretrained) } } \\
\midrule
VQVAE
& 19.82$\pm$2.58 / 83.96$\pm$4.18 & 22.48$\pm$1.76 / 88.36$\pm$3.42 & 18.91$\pm$4.36 / 82.74$\pm$5.12
& 17.86$\pm$0.81 / 68.91$\pm$3.86
& 24.66$\pm$2.14 / \textbf{92.42$\pm$3.08} & 25.12$\pm$2.04 / 91.56$\pm$2.64 & 25.48$\pm$2.83 / 92.68$\pm$3.21 \\
AEKL
& \underline{21.46$\pm$2.12} / 80.84$\pm$4.76 & 20.42$\pm$1.94 / 80.96$\pm$3.88 & \underline{21.53$\pm$3.51} / 81.42$\pm$5.34
& 17.61$\pm$0.94 / 67.28$\pm$3.72
& \underline{24.91$\pm$1.44} / 89.76$\pm$2.86 & 23.28$\pm$1.31 / 88.41$\pm$2.73 & 24.38$\pm$2.41 / 89.95$\pm$3.64 \\
MAISI
& 20.78$\pm$1.83 / \underline{86.72$\pm$3.95} & \textbf{22.76$\pm$1.88} / \textbf{90.03$\pm$2.78} & 20.18$\pm$3.17 / \underline{86.08$\pm$4.26}
& \underline{18.51$\pm$0.72} / \underline{70.06$\pm$3.21}
& 23.42$\pm$1.86 / \underline{91.36$\pm$2.54} & \underline{26.35$\pm$2.26} / \underline{92.83$\pm$2.18} & \underline{27.18$\pm$2.65} / \textbf{93.48$\pm$2.61} \\
3D-Swin-DiT
& \textbf{22.04$\pm$1.86} / \textbf{88.31$\pm$2.62} & \underline{22.21$\pm$1.53} / \underline{88.96$\pm$2.36} & \textbf{24.42$\pm$2.43} / \textbf{87.74$\pm$3.18}
& \textbf{19.27$\pm$0.46} / \textbf{72.81$\pm$2.92}
& \textbf{25.18$\pm$0.96} / 90.83$\pm$2.47 & \textbf{26.82$\pm$1.88} / \textbf{93.11$\pm$2.06} & \textbf{27.42$\pm$2.26} / \underline{93.06$\pm$2.34} \\
\bottomrule
\end{tabular}
}
\end{table*}

\begin{figure*}
\centering
\includegraphics[width=\textwidth]{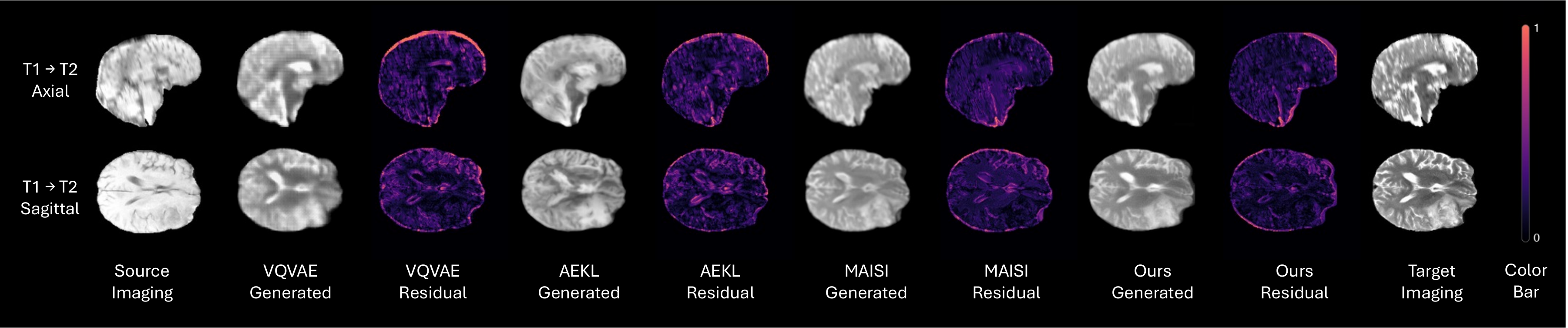}
\caption{
\textbf{Cross-modality image translation from T1-weighted to T2-weighted MRI.}
Representative axial and sagittal slices illustrating T1-to-T2 synthesis using VQVAE, AEKL, MAISI, and the proposed approach. The source T1 image and the target T2 image are shown at the leftmost and rightmost columns, respectively. Intermediate columns present synthesized images and their residual maps. .
}
\label{fig:t1t2}
\end{figure*}
Tables~\ref{tab:translation_brats} and~\ref{tab:translation_all} report cross-modal translation performance across the four autoencoders and three generators. {3D-Swin} attains the highest translation fidelity in the majority of settings, leading on both PSNR and SSIM under the flow-matching generators (RF~\cite{liu2022flow} and MOTFM~\cite{yazdani2025flow}) across BraTS, TCGA-LGG, IXI, and RHUH-GBM, with the largest margin observed on the out-of-distribution IXI cohort (1--3\,dB PSNR over the next-best encoder under every generator). {MAISI}~\cite{guo2025maisi}, despite achieving the strongest reconstruction in Table~\ref{tab:reconstruction}, ranks second in most translation settings, indicating that reconstruction fidelity does not necessarily induce a latent geometry suited to cross-modal mapping. {AEKL}~\cite{rombach2022high} remains a competitive PSNR baseline on the tumor cohorts but underperforms on SSIM, suggesting that its continuous latent preserves intensity statistics more effectively than perceptual structure across modalities. {VQVAE}~\cite{van2017neural} exhibits the inverse trade-off, occasionally attaining the highest SSIM under MOTFM (e.g., RHUH-GBM T1$\to$T2, BraTS-PED T1$\to$T1ce) but trailing on PSNR, consistent with the perceptual-fidelity bias of discrete latents observed in reconstruction. 

These results establish 3D-Swin as the most effective autoencoder for cross-modal translation in our benchmark, as reflected in its consistent superiority on both PSNR and SSIM: it produces a latent space that is both more semantically structured and better aligned across modalities, facilitating the learning of cross-modal mappings by the translation models.

Residual maps illustrate the absolute differences between synthesized and target images in T1 modality (Figure \ref{fig:t1t1ce}). The proposed approach more accurately reproduces the contrast enhancement patterns and anatomical structures observed in the target modality while substantially reducing reconstruction errors relative to competing methods.

The proposed method demonstrates improved preservation of anatomical structures and tissue contrast, producing images that more closely resemble the target T2 modality while exhibiting lower residual errors and fewer synthesis artefacts (Figure \ref{fig:t1t2}.

\subsection{Clinical Task of Adult's Brain Tumor Survival Estimation}
The following analyses comprehensively evaluate the prognostic performance of all investigated models across multiple clinically relevant survival prediction tasks. To assess the robustness and generalizability of the proposed framework, we examined model performance using complementary survival-analysis paradigms, including time-to-event discrimination, continuous survival-time regression, and fixed-horizon risk classification at one-, two-, and three-year endpoints.

Time-to-event discrimination analyses quantify each model's ability to correctly rank patients according to progression risk and survival ordering, thereby evaluating prognostic stratification performance independently of predefined temporal thresholds. In parallel, survival-time regression analyses assess the accuracy of continuous survival prediction by directly estimating patient-specific progression intervals. To further characterize clinically actionable prediction capabilities, we additionally evaluated binary classification performance across one-year, two-year, and three-year progression horizons, enabling assessment of short-, intermediate-, and longer-term disease forecasting performance.

Together, these complementary evaluation paradigms provide a comprehensive characterization of the predictive capacity, temporal robustness, and clinical applicability of the investigated architectures across multiple survival-analysis settings.

\begin{table}[htbp]
\centering
\caption{Time-to-event discrimination across all encoder configurations ($n=1{,}551$). Cox concordance index (mean $\pm$ s.d., five folds) for the two head architectures described in Methods.}
\label{tab:survival_cox}
\footnotesize
\setlength{\tabcolsep}{4pt}
\renewcommand{\arraystretch}{1.15}

\begin{tabular}{lcc}
\toprule
Encoder configuration & Cox (MLP) C-index & Cox (DeepSurv) C-index \\
\midrule
\multicolumn{3}{l}{\textbf{MONAI ResNet (MedicalNet)}} \\
\midrule
resnet10 & 0.5655 ± 0.0152 & 0.5826 ± 0.0123 \\
resnet101 & 0.5567 ± 0.0141 & 0.5649 ± 0.0144 \\
resnet152 & 0.5510 ± 0.0098 & 0.5476 ± 0.0088 \\
resnet18 & 0.5700 ± 0.0160 & 0.5876 ± 0.0191 \\
resnet200 & 0.5677 ± 0.0072 & 0.5873 ± 0.0162 \\
resnet34 & 0.5699 ± 0.0064 & 0.5738 ± 0.0133 \\
resnet50 & 0.5439 ± 0.0074 & 0.5818 ± 0.0160 \\
\midrule
\multicolumn{3}{l}{\textit{DUNE}} \\
\midrule
DUNE U-AE & 0.5654 ± 0.0213 & 0.5806 ± 0.0221 \\
DUNE U-VAE & 0.4979 ± 0.0210 & 0.5184 ± 0.0115 \\
DUNE UNET & 0.5460 ± 0.0105 & 0.5350 ± 0.0154 \\
\midrule
\multicolumn{3}{l}{\textbf{HLIP ViT}} \\
\midrule
HLIP ViT & 0.5685 ± 0.0131 & 0.5643 ± 0.0102 \\
\midrule
\multicolumn{3}{l}{\textbf{3D-Swin-DiT}} \\
\midrule
3D-Swin-lv2-t0 & 0.5705 ± 0.0117 & 0.5881 ± 0.0089 \\
3D-Swin-lv4-t0 & 0.5818 ± 0.0100 & 0.5938 ± 0.0172 \\
3D-Swin-DiT-lv2-t0 & 0.5682 ± 0.0114 & 0.5904 ± 0.0106 \\
3D-Swin-DiT-lv4-t0 & 0.5821 ± 0.0097 & 0.5963 ± 0.0182 \\
EWC-3D-Swin-DiT-lv4 & 0.5815 ± 0.0120 & 0.5999 ± 0.0142 \\
EWC-3D-Swin-lv4 & 0.5832 ± 0.0118 & 0.5959 ± 0.0175 \\
GBP-3D-Swin-DiT-lv4 & 0.5851 ± 0.0142 & 0.6009 ± 0.0198 \\
GBP-3D-Swin-lv4 & 0.5826 ± 0.0142 & 0.5984 ± 0.0171 \\
\bottomrule
\end{tabular}
\end{table} 
\begin{table}[htbp]
\centering
\caption{Survival time regression across all encoder configurations (deceased patients only). Mean absolute error (MAE) and root mean squared error (RMSE) are reported in days; C-index summarises ranking of predicted survival time. All metrics are mean $\pm$ s.d. across five folds.}
\label{tab:survival_task3}
\footnotesize
\setlength{\tabcolsep}{4pt}
\renewcommand{\arraystretch}{1.15}

\begin{tabular}{lccc}
\toprule
Encoder configuration & MAE (days) & RMSE (days) & C-index \\
\midrule
\multicolumn{4}{l}{\textbf{MONAI ResNet (MedicalNet)}} \\
\midrule
resnet10 & 1103.16 ± 773.28 & 2565.80 ± 2261.16 & 0.5387 ± 0.0160 \\
resnet101 & 753.59 ± 203.00 & 1978.98 ± 1004.89 & 0.5486 ± 0.0177 \\
resnet152 & 741.65 ± 437.60 & 1759.17 ± 1484.12 & 0.5201 ± 0.0128 \\
resnet18 & 1418.35 ± 1051.46 & 3468.68 ± 2128.42 & 0.5398 ± 0.0101 \\
resnet200 & 2409.47 ± 2466.37 & 4371.79 ± 4502.78 & 0.5319 ± 0.0139 \\
resnet34 & 989.52 ± 616.26 & 2438.46 ± 1872.21 & 0.5343 ± 0.0126 \\
resnet50 & 1705.65 ± 1528.52 & 3387.84 ± 3244.30 & 0.5347 ± 0.0063 \\
\midrule
\multicolumn{4}{l}{\textbf{DUNE}} \\
\midrule
DUNE U-AE & 510.45 ± 32.10 & 744.96 ± 69.19 & 0.5738 ± 0.0232 \\
DUNE U-VAE & 502.21 ± 34.65 & 739.16 ± 71.70 & 0.5245 ± 0.0151 \\
DUNE UNET & 2768.54 ± 1873.71 & 5487.86 ± 3511.82 & 0.5384 ± 0.0097 \\
\midrule
\multicolumn{4}{l}{\textbf{HLIP ViT}} \\
\midrule
HLIP ViT & 812.65 ± 601.09 & 1554.71 ± 1585.89 & 0.5368 ± 0.0147 \\
\midrule
\multicolumn{4}{l}{\textbf{3D-Swin-DiT}} \\
\midrule
3D-Swin-lv2 & 2687.72 ± 1675.96 & 6211.67 ± 2413.60 & 0.5328 ± 0.0155 \\
3D-Swin-lv4-t0 & 780.11 ± 343.08 & 1983.68 ± 1340.65 & 0.5440 ± 0.0128 \\
3D-Swin-lv2-t0 & 2033.06 ± 1474.00 & 5207.69 ± 2128.12 & 0.5330 ± 0.0163 \\
3D-Swin-DiT-lv2 & 511.39 ± 34.60 & 744.34 ± 83.37 & 0.5220 ± 0.0095 \\
3D-Swin-DiT-lv4-t0 & 772.32 ± 348.78 & 1892.58 ± 1412.94 & 0.5411 ± 0.0116 \\
3D-Swin-DiT-lv4-t0 & 2028.50 ± 1465.37 & 5203.60 ± 2120.20 & 0.5330 ± 0.0159 \\
EWC-3D-Swin-DiT-lv4 & 2195.30 ± 1828.51 & 5333.51 ± 2408.26 & 0.5518 ± 0.0174 \\
EWC-3D-Swin-lv4 & 2195.35 ± 1828.61 & 5333.50 ± 2408.25 & 0.5518 ± 0.0174 \\
GBP-3D-Swin-DiT-lv4 & 2057.10 ± 1419.85 & 5085.39 ± 1718.67 & 0.5498 ± 0.0172 \\
GBP-3D-Swin-lv4 & 2057.09 ± 1419.86 & 5085.38 ± 1718.68 & 0.5498 ± 0.0172 \\
\bottomrule
\end{tabular}
\end{table} 
\begin{table*}[t]
\centering
\caption{\textbf{One-year survival classification across all encoder configurations.}
Patients censored before 1 year were excluded. Results are reported as mean $\pm$ SD across five folds ($n=1419$, prevalence $=0.594$).}
\label{tab:survival_task4_1y}

\scriptsize
\setlength{\tabcolsep}{3pt}

\resizebox{\textwidth}{!}{
\begin{tabular}{lcccccc}
\toprule
\textbf{Encoder} &
\textbf{AUROC} &
\textbf{PR-AUC} &
\textbf{BalAcc} &
\textbf{Sens.} &
\textbf{Spec.} &
\textbf{F1} \\
\midrule

\multicolumn{7}{l}{\textbf{MONAI ResNet (MedicalNet)}}\\
\midrule
ResNet10  & $0.706\pm0.041$ & $0.750\pm0.037$ & 0.645 & $0.737\pm0.036$ & $0.554\pm0.052$ & $0.721\pm0.023$\\
ResNet18  & $0.702\pm0.026$ & $0.753\pm0.030$ & 0.617 & $0.801\pm0.025$ & $0.432\pm0.058$ & $0.732\pm0.008$\\
ResNet34  & $0.712\pm0.032$ & $0.758\pm0.031$ & 0.646 & $0.755\pm0.042$ & $0.538\pm0.088$ & $0.729\pm0.017$\\
ResNet50  & $0.711\pm0.047$ & $0.764\pm0.049$ & 0.649 & $0.752\pm0.046$ & $0.547\pm0.079$ & $0.729\pm0.036$\\
ResNet101 & $0.691\pm0.028$ & $0.741\pm0.024$ & 0.644 & $0.762\pm0.049$ & $0.526\pm0.074$ & $0.730\pm0.025$\\
ResNet152 & $0.652\pm0.026$ & $0.706\pm0.033$ & 0.607 & $0.754\pm0.053$ & $0.460\pm0.075$ & $0.710\pm0.022$\\
ResNet200 & $0.693\pm0.018$ & $0.748\pm0.022$ & 0.641 & $0.757\pm0.059$ & $0.526\pm0.128$ & $0.727\pm0.008$\\

\midrule
\multicolumn{7}{l}{\textbf{DUNE}}\\
\midrule
U-AE  & $0.726\pm0.024$ & $\mathbf{0.788\pm0.028}$ & 0.650 & $0.763\pm0.066$ & $0.538\pm0.095$ & $0.733\pm0.020$\\
U-VAE & $0.548\pm0.015$ & $0.633\pm0.012$ & 0.508 & $0.838\pm0.173$ & $0.179\pm0.202$ & $0.691\pm0.061$\\
UNET  & $0.631\pm0.014$ & $0.697\pm0.022$ & 0.591 & $0.779\pm0.061$ & $0.403\pm0.090$ & $0.712\pm0.020$\\

\midrule
\multicolumn{7}{l}{\textbf{HLIP}}\\
\midrule
ViT & $0.701\pm0.026$ & $0.766\pm0.024$ & 0.625 & $0.757\pm0.027$ & $0.493\pm0.067$ & $0.720\pm0.017$\\

\midrule
\multicolumn{7}{l}{\textbf{3D-Swin-DiT}}\\
\midrule
3D-Swin-lv2 & $0.725\pm0.034$ & $0.768\pm0.022$ & 0.672 & $0.746\pm0.053$ & $0.597\pm0.052$ & $0.738\pm0.027$\\
3D-Swin-lv2-t0 & $0.699\pm0.047$ & $0.754\pm0.046$ & 0.639 & $0.782\pm0.064$ & $0.496\pm0.115$ & $0.735\pm0.025$\\
3D-Swin-lv4-t0 & $0.724\pm0.039$ & $0.759\pm0.034$ & \textbf{0.683} & $0.766\pm0.013$ & $0.599\pm0.061$ & $\mathbf{0.751\pm0.019}$\\
3D-Swin-DiT-lv2 & $0.540\pm0.030$ & $0.624\pm0.018$ & 0.534 & $0.720\pm0.149$ & $0.347\pm0.130$ & $0.657\pm0.075$\\
3D-Swin-DiT-lv4-t0 & $0.696\pm0.053$ & $0.757\pm0.053$ & 0.637 & $0.728\pm0.074$ & $0.547\pm0.091$ & $0.713\pm0.036$\\
3D-Swin-DiT-lv2-t0 & $0.726\pm0.038$ & $0.768\pm0.026$ & 0.670 & $0.740\pm0.047$ & $0.601\pm0.032$ & $0.735\pm0.030$\\
EWC-3D-Swin-DiT-lv4 & $0.727\pm0.048$ & $\mathbf{0.789\pm0.037}$ & 0.667 & $0.775\pm0.030$ & $0.559\pm0.077$ & $0.747\pm0.026$\\
EWC-3D-Swin-lv4 & $0.727\pm0.047$ & $0.785\pm0.039$ & 0.654 & $0.771\pm0.054$ & $0.536\pm0.099$ & $0.739\pm0.034$\\
GBP-3D-Swin-DiT-lv4 & $0.725\pm0.045$ & $0.778\pm0.031$ & 0.657 & $0.738\pm0.058$ & $0.576\pm0.088$ & $0.727\pm0.026$\\
GBP-3D-Swin-lv4 & $0.726\pm0.046$ & $0.779\pm0.032$ & 0.662 & $0.739\pm0.062$ & $0.585\pm0.091$ & $0.730\pm0.028$\\
\bottomrule
\end{tabular}}
\end{table*}

\begin{table*}[t]
\centering
\caption{\textbf{Two-year survival classification across all encoder configurations.}
Patients censored before 2 years were excluded. Results are reported as mean $\pm$ SD across five folds ($n=1334$, prevalence $=0.285$).}
\label{tab:survival_task4_2y}

\scriptsize
\setlength{\tabcolsep}{3pt}

\resizebox{\textwidth}{!}{
\begin{tabular}{lcccccc}
\toprule
\textbf{Encoder} &
\textbf{AUROC} &
\textbf{PR-AUC} &
\textbf{BalAcc} &
\textbf{Sens.} &
\textbf{Spec.} &
\textbf{F1} \\
\midrule

\multicolumn{7}{l}{\textbf{MONAI ResNet (MedicalNet)}}\\
\midrule
ResNet10  & $0.702\pm0.030$ & $0.492\pm0.056$ & 0.608 & $0.347\pm0.045$ & $0.869\pm0.018$ & $0.414\pm0.051$\\
ResNet18  & $0.699\pm0.042$ & $0.482\pm0.069$ & 0.623 & $0.390\pm0.077$ & $0.857\pm0.028$ & $0.442\pm0.059$\\
ResNet34  & $0.702\pm0.036$ & $0.490\pm0.058$ & 0.605 & $0.334\pm0.077$ & $0.876\pm0.030$ & $0.403\pm0.079$\\
ResNet50  & $0.705\pm0.032$ & $0.501\pm0.068$ & 0.634 & $0.408\pm0.093$ & $0.861\pm0.042$ & $0.460\pm0.082$\\
ResNet101 & $0.660\pm0.046$ & $0.437\pm0.052$ & 0.575 & $0.268\pm0.064$ & $0.882\pm0.028$ & $0.340\pm0.063$\\
ResNet152 & $0.658\pm0.016$ & $0.433\pm0.032$ & 0.542 & $0.158\pm0.100$ & $0.927\pm0.047$ & $0.218\pm0.121$\\
ResNet200 & $0.688\pm0.007$ & $0.493\pm0.026$ & 0.604 & $0.332\pm0.095$ & $0.877\pm0.046$ & $0.397\pm0.072$\\

\midrule
\multicolumn{7}{l}{\textbf{DUNE}}\\
\midrule
U-AE  & $0.708\pm0.017$ & $0.528\pm0.032$ & 0.639 & $0.405\pm0.114$ & $0.873\pm0.059$ & $0.460\pm0.076$\\
U-VAE & $0.571\pm0.027$ & $0.343\pm0.024$ & 0.528 & $0.297\pm0.264$ & $0.759\pm0.230$ & $0.238\pm0.186$\\
UNET  & $0.629\pm0.025$ & $0.413\pm0.031$ & 0.557 & $0.184\pm0.144$ & $0.930\pm0.059$ & $0.238\pm0.156$\\

\midrule
\multicolumn{7}{l}{\textbf{HLIP}}\\
\midrule
ViT & $0.719\pm0.032$ & $0.514\pm0.028$ & 0.603 & $0.276\pm0.103$ & $0.930\pm0.036$ & $0.368\pm0.103$\\

\midrule
\multicolumn{7}{l}{\textbf{3D-Swin-DiT}}\\
\midrule
3D-Swin-lv2 & $0.720\pm0.038$ & $0.526\pm0.062$ & 0.633 & $0.392\pm0.071$ & $0.873\pm0.030$ & $0.456\pm0.064$\\
3D-Swin-lv2-t0 & $0.693\pm0.034$ & $0.490\pm0.051$ & 0.633 & $0.424\pm0.097$ & $0.842\pm0.024$ & $0.462\pm0.091$\\
3D-Swin-lv4-t0 & $0.719\pm0.039$ & $0.538\pm0.066$ & 0.634 & $0.387\pm0.052$ & $0.882\pm0.022$ & $0.459\pm0.053$\\
3D-Swin-DiT-lv2 & $0.550\pm0.022$ & $0.318\pm0.013$ & 0.514 & $0.208\pm0.120$ & $0.820\pm0.114$ & $0.225\pm0.120$\\
3D-Swin-DiT-lv4-t0 & $0.693\pm0.039$ & $0.492\pm0.067$ & 0.625 & $0.405\pm0.129$ & $0.845\pm0.044$ & $0.441\pm0.099$\\
3D-Swin-DiT-lv2-t0 & $\mathbf{0.720\pm0.040}$ & $\mathbf{0.545\pm0.070}$ & \textbf{0.643} & $0.418\pm0.063$ & $0.868\pm0.014$ & $\mathbf{0.477\pm0.061}$\\
EWC-3D-Swin-DiT-lv4 & $0.716\pm0.037$ & $0.536\pm0.040$ & 0.614 & $0.337\pm0.074$ & $0.891\pm0.038$ & $0.414\pm0.051$\\
EWC-3D-Swin-lv4 & $0.716\pm0.036$ & $0.535\pm0.041$ & 0.613 & $0.329\pm0.062$ & $0.896\pm0.032$ & $0.411\pm0.049$\\
GBP-3D-Swin-DiT-lv4 & $0.719\pm0.036$ & $0.530\pm0.061$ & 0.624 & $0.366\pm0.067$ & $0.883\pm0.010$ & $0.438\pm0.061$\\
GBP-3D-Swin-lv4 & $0.718\pm0.037$ & $0.523\pm0.065$ & 0.626 & $0.366\pm0.067$ & $0.887\pm0.012$ & $0.441\pm0.059$\\

\bottomrule
\end{tabular}}
\end{table*}

\begin{table*}[t]
\centering
\caption{\textbf{Three-year survival classification across all encoder configurations.}
Patients censored before 3 years were excluded. Results are reported as mean $\pm$ SD across five folds ($n=1285$, prevalence $=0.173$).}
\label{tab:survival_task4_3y}

\scriptsize
\setlength{\tabcolsep}{3pt}

\resizebox{\textwidth}{!}{
\begin{tabular}{lcccccc}
\toprule
\textbf{Encoder} &
\textbf{AUROC} &
\textbf{PR-AUC} &
\textbf{BalAcc} &
\textbf{Sens.} &
\textbf{Spec.} &
\textbf{F1} \\
\midrule

\multicolumn{7}{l}{\textbf{MONAI ResNet (MedicalNet)}}\\
\midrule
ResNet10  & $0.735\pm0.023$ & $0.404\pm0.027$ & 0.552 & $0.136\pm0.109$ & $0.968\pm0.034$ & $0.186\pm0.132$\\
ResNet18  & $0.709\pm0.043$ & $0.382\pm0.048$ & 0.564 & $0.177\pm0.112$ & $0.952\pm0.041$ & $0.227\pm0.123$\\
ResNet34  & $0.737\pm0.033$ & $0.392\pm0.064$ & 0.577 & $0.195\pm0.147$ & $0.960\pm0.028$ & $0.251\pm0.158$\\
ResNet50  & $0.717\pm0.037$ & $0.397\pm0.062$ & 0.600 & $0.253\pm0.189$ & $0.947\pm0.036$ & $0.296\pm0.176$\\
ResNet101 & $0.680\pm0.019$ & $0.343\pm0.022$ & 0.540 & $0.113\pm0.062$ & $0.968\pm0.024$ & $0.168\pm0.087$\\
ResNet152 & $0.635\pm0.009$ & $0.272\pm0.025$ & 0.513 & $0.045\pm0.069$ & $0.981\pm0.022$ & $0.066\pm0.092$\\
ResNet200 & $0.690\pm0.020$ & $0.370\pm0.045$ & 0.600 & $0.279\pm0.068$ & $0.921\pm0.029$ & $0.332\pm0.054$\\

\midrule
\multicolumn{7}{l}{\textbf{DUNE}}\\
\midrule
U-AE  & $\mathbf{0.748\pm0.034}$ & $\mathbf{0.444\pm0.031}$ & \textbf{0.634} & $0.343\pm0.045$ & $0.926\pm0.030$ & $\mathbf{0.402\pm0.016}$\\
U-VAE & $0.549\pm0.042$ & $0.207\pm0.022$ & 0.521 & $0.268\pm0.323$ & $0.775\pm0.282$ & $0.139\pm0.132$\\
UNET  & $0.659\pm0.031$ & $0.322\pm0.051$ & 0.535 & $0.099\pm0.072$ & $0.971\pm0.019$ & $0.148\pm0.097$\\

\midrule
\multicolumn{7}{l}{\textbf{HLIP}}\\
\midrule
ViT & $0.728\pm0.061$ & $0.407\pm0.080$ & 0.567 & $0.194\pm0.102$ & $0.939\pm0.038$ & $0.242\pm0.126$\\

\midrule
\multicolumn{7}{l}{\textbf{3D-Swin-DiT}}\\
\midrule
3D-Swin-lv4 & $0.737\pm0.023$ & $0.408\pm0.043$ & 0.603 & $0.257\pm0.118$ & $0.949\pm0.018$ & $0.329\pm0.112$\\
3D-Swin-DiT-lv4 & $0.570\pm0.025$ & $0.253\pm0.055$ & 0.513 & $0.059\pm0.086$ & $0.967\pm0.029$ & $0.085\pm0.122$\\
3D-Swin-DiT-lv2-t0 & $0.722\pm0.024$ & $0.355\pm0.048$ & 0.555 & $0.159\pm0.086$ & $0.951\pm0.023$ & $0.214\pm0.116$\\
3D-Swin-DiT-lv4-t0 & $0.736\pm0.022$ & $0.392\pm0.051$ & 0.602 & $0.275\pm0.091$ & $0.928\pm0.020$ & $0.332\pm0.076$\\
EWC-3D-Swin-DiT-lv4 & $0.729\pm0.013$ & $0.398\pm0.030$ & 0.619 & $0.319\pm0.135$ & $0.919\pm0.064$ & $0.361\pm0.072$\\
GBP-3D-Swin-DiT-lv4 & $0.736\pm0.013$ & $0.403\pm0.046$ & 0.625 & $0.329\pm0.084$ & $0.922\pm0.042$ & $0.381\pm0.047$\\
GBP-3D-Swin-lv4 & $0.735\pm0.013$ & $0.413\pm0.033$ & 0.630 & $0.333\pm0.087$ & $0.926\pm0.040$ & $0.388\pm0.047$\\

\bottomrule
\end{tabular}}
\end{table*} 

\label{sec:survival_validation}

We benchmarked frozen MRI representations from 3D-Swin-DiT against three established encoder families on multicentre glioma overall survival (OS; $n=1{,}551$). Nineteen encoder configurations (seven MONAI ResNet depths, three DUNE variants, HLIP, and fourteen DiT checkpoints) were evaluated under an identical lightweight survival protocol (Methods). Endpoints spanned time-to-event discrimination (Cox models with two head architectures), log-time regression among patients who died during follow-up, and alive/dead classification at 1, 2 and 3 years.

3D-Swin-DiT achieved the highest time-to-event discrimination. Concordance indices from both Cox formulations were largest for 3D-Swin-DiT with elastic-weight consolidation and concatenated T1C+FLAIR diffusion features (MLP Cox: $0.592 \pm 0.014$; DeepSurv-style Cox: $0.601 \pm 0.016$; Table~S61). This configuration exceeded the strongest external comparators in the table (ResNet-34: $0.570 \pm 0.006$ and $0.574 \pm 0.013$; DUNE U-AE: $0.565 \pm 0.021$ and $0.581 \pm 0.022$; HLIP: $0.569 \pm 0.013$ and $0.564 \pm 0.010$). A 3D-Swin-DiT checkpoint remained close ($0.585 \pm 0.015$ and $0.598 \pm 0.011$), indicating that continual-learning strategy modulates but does not solely drive the gain. Across ResNet depths, Cox performance varied modestly (concordance $\approx 0.54$--$0.57$; Table~\ref{tab:survival_cox}). DUNE U-NET lagged other DUNE variants on these endpoints ($0.546 \pm 0.011$ for MLP Cox).

\begin{table}[t]
\centering
\caption{\textbf{Representative encoders for multicentre glioma overall survival}
($n=1551$). Results are reported as mean $\pm$ SD across five folds.}
\label{tab:survival_main_compact2}

\scriptsize
\setlength{\tabcolsep}{2pt}

\begin{tabular}{lcccccc}
\toprule
Encoder & Cox-MLP & DeepSurv & 1Y & 2Y & 3Y & Time Reg.\\
\midrule
ResNet-34 &
$0.570\pm0.006$ &
$0.574\pm0.013$ &
$0.712\pm0.032$ &
$0.702\pm0.036$ &
$0.737\pm0.033$ &
$0.534\pm0.013$ \\
DUNE U-AE &
$0.565\pm0.021$ &
$0.581\pm0.022$ &
$0.726\pm0.024$ &
$0.708\pm0.017$ &
$0.748\pm0.034$ &
$0.574\pm0.023$ \\
HLIP ViT &
$0.569\pm0.013$ &
$0.564\pm0.010$ &
$0.701\pm0.026$ &
$0.719\pm0.032$ &
$0.728\pm0.061$ &
$0.537\pm0.015$ \\
NeurMRI-DiT (SEQ) &
$0.585\pm0.015$ &
$0.598\pm0.011$ &
$0.731\pm0.044$ &
$0.717\pm0.037$ &
$0.729\pm0.010$ &
$0.552\pm0.017$ \\
NeurMRI-DiT (EWC) &
$0.592\pm0.014$ &
$0.601\pm0.016$ &
$0.731\pm0.044$ &
$0.715\pm0.034$ &
$0.727\pm0.011$ &
$0.549\pm0.018$ \\
NeurMRI-DiT (GBP) &
$0.585\pm0.014$ &
$0.601\pm0.020$ &
$0.725\pm0.045$ &
$0.719\pm0.036$ &
$0.736\pm0.013$ &
$0.550\pm0.017$ \\
\bottomrule
\end{tabular}
\end{table}
 
Horizon-specific classification favoured DiT at 1--2 years and DUNE at 3 years. One-year AUROC reached $0.731 \pm 0.044$ for 3D-Swin-DiT (baseline and EWC concatenation models). At two years, the highest AUROC in the full benchmark was $0.720 \pm 0.040$ (3D-Swin), with 3D-Swin-DiT models near $0.715$--$0.719$. At three years, DUNE U-AE achieved the top AUROC ($0.748 \pm 0.034$), marginally above the strongest 3D-Swin-DiT configuration ($0.737 \pm 0.023$; 3D-Swin) and ResNet-34 ($0.737 \pm 0.033$); fold-wise uncertainty overlapped across these leaders. Effective sample size fell at longer horizons ($n=1{,}419$, $1{,}334$ and $1{,}285$ at 1, 2 and 3 years; positive-class prevalence $\approx 59\%$, $29\%$ and $17\%$), widening confidence intervals for late-horizon metrics.

Absolute survival time was estimated most accurately by DUNE, not by the best ranking model. Among patients with observed death, DUNE U-AE and U-VAE yielded the lowest mean absolute error in predicted OS ($510 \pm 32$ and $502 \pm 35$ days), despite weaker Cox concordance than 3D-Swin-DiT. EWC-3D-Swin-DiT-lv2 showed larger errors ($1{,}160 \pm 370$ days) while still ranking risks well, illustrating that prognostic ordering and calendar-time calibration respond differently to representation choice. HLIP and ResNet spanned a broad MAE range across depths ( Table~\ref{tab:survival_task3}).

Within 3D-Swin-DiT, multi-contrast fusion mattered more than any single branch. Concatenating T1C and FLAIR embeddings at diffusion timestep $t{=}0$ and spatial level 4 outperformed single-modality or Swin-only features on Cox and 1-year endpoints. EWC-3D-Swin-DiT outperforms on concordance, whereas the strongest three-year AUROC among 3D-Swin-DiT variants was achieved by Swin-only features ($0.737 \pm 0.023$), closely followed by GBP concatenation ($0.736 \pm 0.013$;  Table~\ref{tab:survival_task4_3y}). No single continual-learning checkpoint dominated every endpoint, supporting task-aware selection when deploying 3D-Swin-DiT for prognosis.

\subsection{Survival Analysis and corresponding UMAP embeddings}

\begin{figure*}
    \centering
    \includegraphics[width=\textwidth]{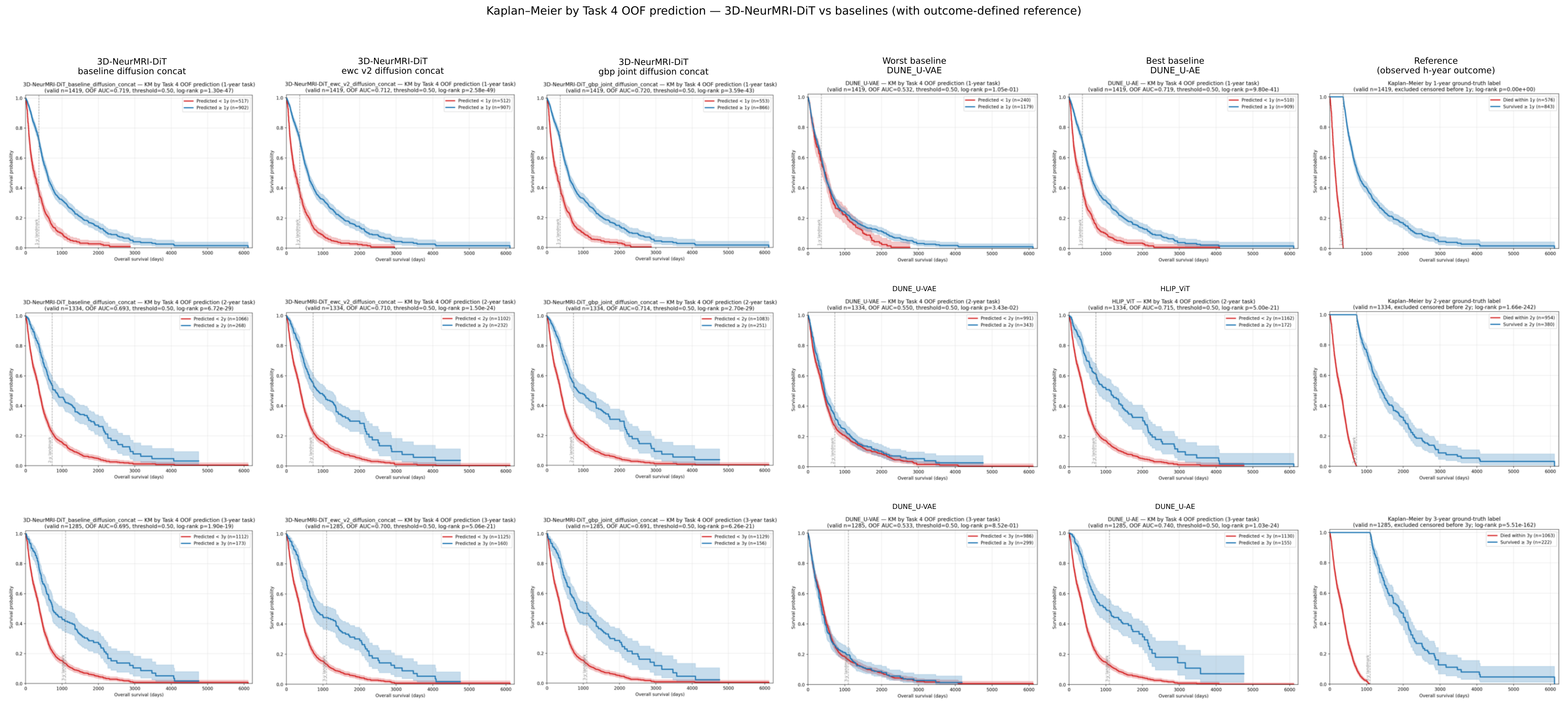}
    \caption{
    \textbf{Kaplan--Meier analysis of Task 4 out-of-fold survival predictions across 3D-Swin-DiT and baseline models.}
    The figure compares predicted survival stratification across multiple model variants, including the 3D-Swin-DiT baseline diffusion model, EWC v2 diffusion model, GBP joint diffusion model, the worst-performing baseline, the best-performing baseline, and an outcome-defined reference. Each panel shows Kaplan--Meier curves for predicted high-risk and low-risk groups, with confidence intervals, enabling visual comparison of survival separation across models. The reference panels based on observed outcome labels provide an upper-bound comparison for evaluating how closely model-derived risk groups reproduce clinically meaningful survival patterns. Overall, the GBP joint diffusion and related 3D-Swin-DiT variants are presented against established baselines to assess whether foundation-model-derived representations improve prognostic stratification in Task 4.
    }
    \label{fig:km_task4_survival}
\end{figure*}

\begin{figure*}
    \centering
    \includegraphics[width=\textwidth]{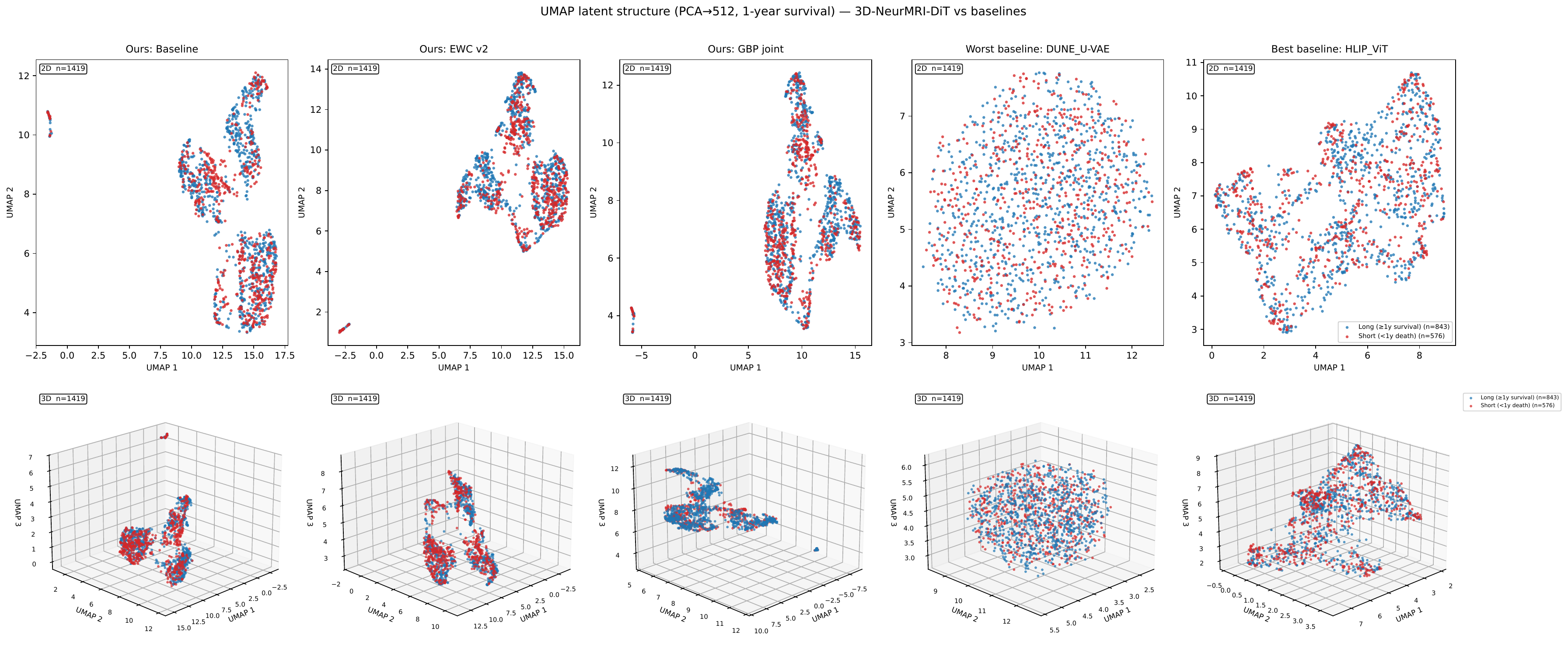}
    \caption{
    \textbf{Latent representation of 1-year survival learned by 3D-Swin-DiT and baseline models.}
    Two-dimensional (top row) and three-dimensional (bottom row) UMAP projections of latent representations obtained after PCA reduction to 512 dimensions are shown for the proposed 3D-Swin-DiT variants (Baseline, EWC v2, and GBP joint) and competing baselines (DUNE\_U-VAE and HLIP\_ViT). Each point corresponds to an individual subject and is colored according to the observed survival outcome: long survival ($\geq$1 year) or short survival (<1 year). Compared with baseline approaches, the proposed foundation model variants exhibit a more structured latent organization with improved separation of survival groups, indicating that the learned representations capture clinically meaningful prognostic information.
    }
    \label{fig:umap_survival_1y}
\end{figure*}

\begin{figure*}
    \centering
    \includegraphics[width=\textwidth]{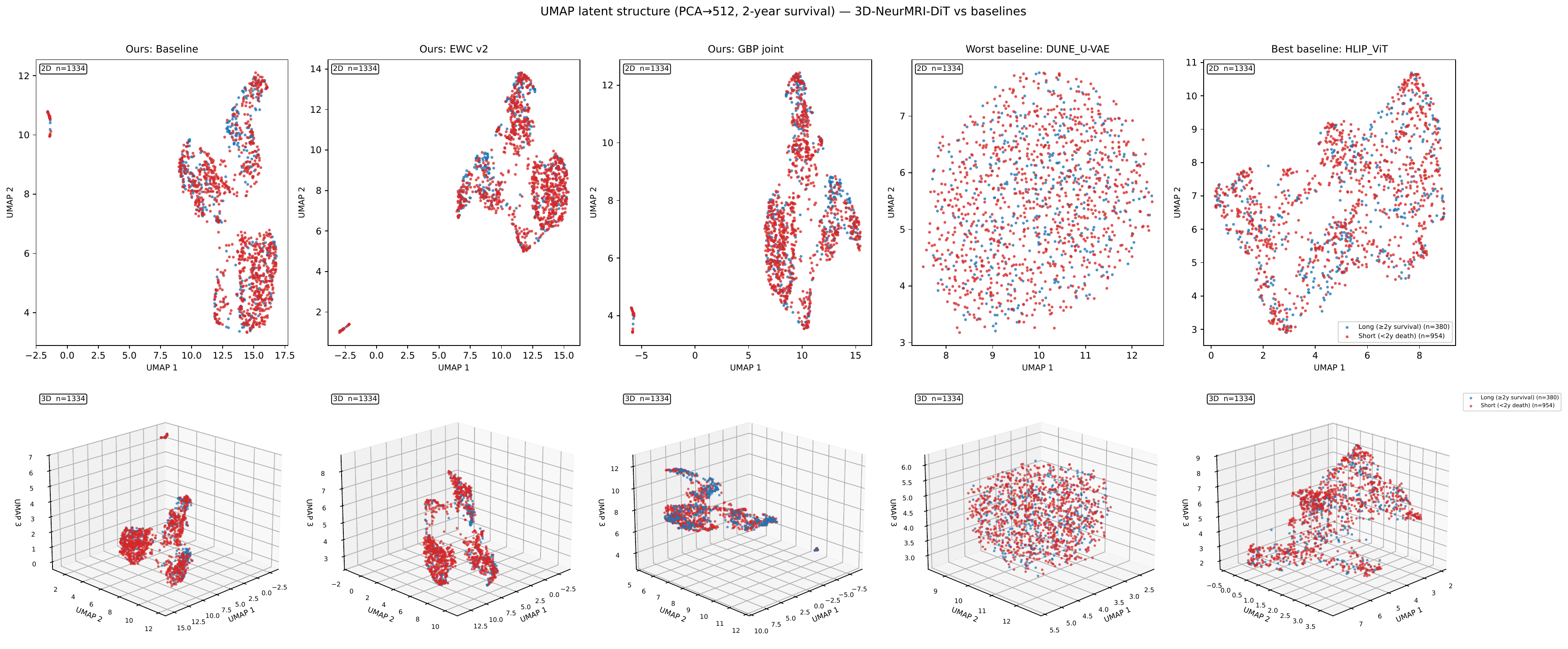}
    \caption{
    \textbf{Latent representation of 2-year survival learned by 3D-Swin-DiT and baseline models.}
    UMAP visualizations of latent embeddings after PCA reduction to 512 dimensions are shown for the proposed 3D-Swin-DiT variants and baseline methods. The top row presents 2D embeddings, whereas the bottom row shows their corresponding 3D representations. Subjects are colored according to observed survival outcome, distinguishing long survival ($\geq$2 years) from short survival (<2 years). The GBP joint model exhibits a more coherent latent topology and improved organization of survival-related patterns compared with competing baselines, suggesting that the learned foundation representations encode clinically relevant prognostic features.
    }
    \label{fig:umap_survival_2y}
\end{figure*}

\begin{figure*}
    \centering
    \includegraphics[width=\textwidth]{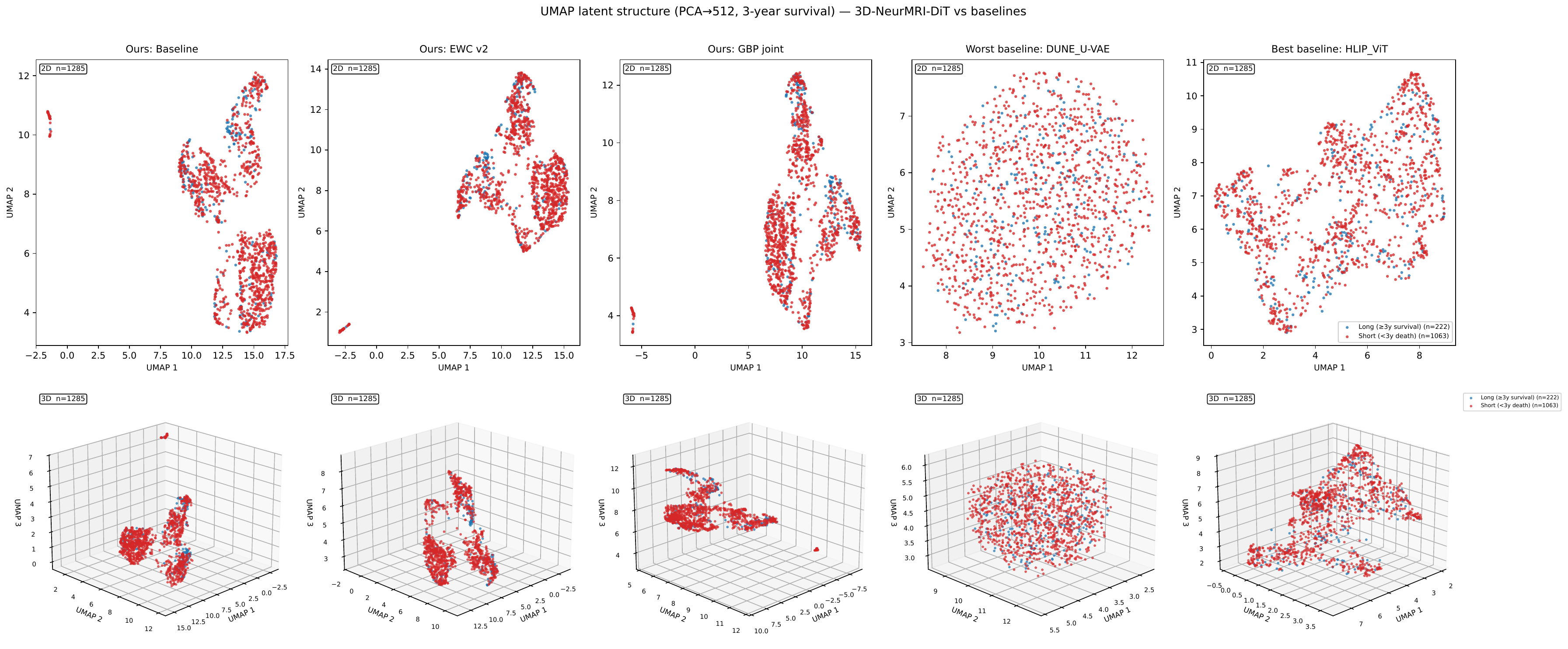}
    \caption{
    \textbf{Latent representation of 3-year survival learned by 3D-Swin-DiT and baseline models.}
    Two-dimensional and three-dimensional UMAP projections of the latent feature space are shown for the proposed 3D-Swin-DiT variants and representative baseline methods after PCA reduction to 512 dimensions. Subjects are stratified according to long survival ($\geq$3 years) and short survival (<3 years). The proposed models, particularly the GBP joint variant, reveal a more organized latent manifold with enhanced discrimination of survival phenotypes compared with baseline approaches, demonstrating the ability of foundation-model-derived representations to capture long-term prognostic information from MRI data.
    }
    \label{fig:umap_survival_3y}
\end{figure*}

\subsubsection{Outline of the analysis}

3D-Swin-DiT embeddings contain multicentre prognostic information for glioma OS that is accessible to simple survival heads. Under a shared downstream protocol, 3D-Swin-DiT achieved the highest concordance in both Cox formulations (up to $0.601 \pm 0.016$), with strong one-year classification (AUROC $0.731 \pm 0.044$). These results support the view that generative--discriminative pretraining at scale can complement, and in this benchmark exceed, established transfer-learning and autoencoder pipelines for time-to-event modelling without OS-specific fine-tuning of the backbone. They do not, however, imply uniform superiority across every prognostic formulation: DUNE led three-year discrimination and absolute time regression.

The dissociation between concordance and mean absolute error is particularly informative. 3D-Swin-DiT ranked risks well yet predicted calendar survival less accurately than DUNE among patients who died, whereas DUNE excelled at time regression while underperforming 3D-Swin-DiT on Cox concordance. Ranking metrics reward monotonic ordering of relative hazard; regression metrics penalise deviation in days on a log-time scale trained only on observed deaths. Encoder inductive biases therefore interact with endpoint choice. DUNE was developed to compress whole-brain MRI into compact embeddings~\cite{barba2025dune}, a representation that may align with magnitude-related survival variation, whereas 3D-Swin-DiT at $t=0$ appears better suited to ordering patients for short and medium follow-up. HLIP and ResNet spanned intermediate performance without leading either primary axis, consistent with differences in pretraining objective, dimensionality and three-dimensional modelling capacity~\cite{zhao2025hlip,chen2019med3d}.

Within DiT, T1C+FLAIR concatenation outperformed single-modality or single-branch features on concordance and early horizons, emphasising that multi-contrast fusion is not optional for this application. Continual-learning checkpoints were not interchangeable: elastic-weight consolidation helped Cox discrimination, whereas three-year AUROC varied only modestly across top-performing DiT variants (best: 3D-Swin $0.737 \pm 0.023$; close: GBP concatenation $0.736 \pm 0.013$). Task-aware deployment, rather than a single biased checkpoint, is therefore prudent until external studies confirm which strategy generalises.

Methodological choices frame how strongly these conclusions transfer. Validation used publicly available, harmonised glioma cohorts only; prospective and in-house testing is needed. Each encoder retained its native preprocessing (Table~\ref{tab:preprocessing_by_model}), reflecting realistic use but coupling representation quality to pipeline differences. Metrics were obtained by five-fold cross-validation without a fully locked hold-out set, and downstream hyperparameters were fixed without nested tuning, so reported intervals describe internal stability rather than definitive generalisation bounds. Late horizons further carried fewer evaluable patients and lower event prevalence, inflating variance for three-year AUROC despite point estimates near 0.75.

For clinical translation, frozen-encoder evaluation is a deliberate staging step. It tests whether a foundation model already encodes prognosis before committing to end-to-end fine-tuning, multimodal fusion with molecular and clinical variables, or deployment workflows. The breadth of endpoints used here reduces the risk of optimising a single metric that does not reflect clinical utility. Next steps should include external cohorts, calibration and decision-curve analysis, and interpretability linking diffusion latents to imaging phenotypes that clinicians recognise. Integrating non-imaging covariates while preserving transparent benchmarking will clarify how much prognostic signal is uniquely carried by 3D-Swin-DiT representations versus routinely collected clinical data.

\subsection{Clinical Task of Adult's Brain Tumor Survival After Surgery}
\subsubsection{Few-shot Adaptation and Continual Learning for Tumour Classification}

Tables~\ref{tab:d3_d4_s8_fewshot}, and \ref{tab:d3_d4_l8_fewshot} summarize the performance of Elastic Weight Consolidation (EWC), Graph-Blueprint Pruning (GBP), and the sequential model without a continual learning approach.(SEQ), and direct tumour fine-tuning under 0-shot, 1-shot, 3-shot, and 5-shot adaptation settings. Performance was evaluated using F1-score, sensitivity, precision, and accuracy.

Across all datasets and architectures, increasing the number of labeled examples consistently improved performance. For the SEQ and S8 architecture on D3 (Table~\ref{tab:d3_d4_s8_fewshot}), F1-score increased from approximately 0.38--0.48 in the zero-shot setting to 0.53--0.58 with five labelled examples. Similar trends were observed for sensitivity, precision, and accuracy, indicating that the improvement was not restricted to a single evaluation metric but reflected a more balanced enhancement of classification performance.

A similar behaviour was observed on the D4 dataset (Table~\ref{tab:d3_d4_s8_fewshot}), although gains were generally smaller and accompanied by larger standard deviations, suggesting that D4 represents a more challenging transfer-learning scenario. Nevertheless, all methods benefited from the introduction of only a few labelled examples, demonstrating the effectiveness of the pretrained foundation representations.

A key observation is that continual-learning approaches consistently outperformed direct tumour fine-tuning in the lowest-shot regimes. For example, on D4 using the S8 architecture (Table~\ref{tab:d3_d4_l8_fewshot}), GBP achieved the highest F1-score at three shots ($0.548 \pm 0.085$), outperforming both EWC ($0.509 \pm 0.053$) and the SEQ baseline ($0.456 \pm 0.028$). This trend suggests that preserving previously acquired representations while selectively adapting model components provides a more robust strategy than unrestricted fine-tuning when supervision is extremely limited.

The relative effectiveness of the continual-learning approaches depended on architectural scale. For the S8 architecture, EWC frequently achieved the strongest overall performance, reaching an F1-score of $0.583 \pm 0.026$ on D3 with five labelled examples (Table~\ref{tab:d3_d4_l8_fewshot}). In contrast, the L8 architecture showed a stronger benefit from graph-based adaptation strategies. As shown in Table~\ref{tab:d3_d4_l8_fewshot}, GBP consistently outperformed EWC, achieving F1-scores of $0.559 \pm 0.038$ and $0.601 \pm 0.050$, respectively.

Among all evaluated methods, GBP demonstrated the most stable and competitive performance across datasets and architectural scales. On D4 with the L8 architecture (Table~\ref{tab:d3_d4_l8_fewshot}), GBP achieved an F1-score of $0.601 \pm 0.050$, sensitivity of $0.639 \pm 0.064$, precision of $0.641 \pm 0.055$, and accuracy of $0.625 \pm 0.047$. In addition to achieving the highest overall performance, GBP generally exhibited lower variance than standard SEQ, indicating improved stability during adaptation.

\begin{table*}[h!]
\centering
\caption{Few-shot classification performance (mean $\pm$ SD) for the D3 and D4 datasets using the S8 configuration and T1/T2 scan MRI modality.}
\label{tab:d3_d4_s8_fewshot}

\scriptsize
\setlength{\tabcolsep}{3pt}

\resizebox{\textwidth}{!}{
\begin{tabular}{ll|ccc|ccc}
\toprule
&& \multicolumn{3}{c|}{\textbf{D3 Dataset}} &
\multicolumn{3}{c}{\textbf{D4 Dataset}}\\
\cmidrule(lr){2-4}
\cmidrule(lr){5-7}

\textbf{Metric} & \textbf{Shot}
& \textbf{EWC}
& \textbf{GBP}
& \textbf{SEQ}
& \textbf{EWC}
& \textbf{GBP}
& \textbf{SEQ}
\\
\midrule

\multirow{4}{*}{F1}
& 0-shot
& 0.4811 & 0.4172 & 0.3933
& $0.4810\pm0.0499$ & $0.4889\pm0.0402$ & $0.3783\pm0.0571$ \\

& 1-shot
& -- & -- & -- 
& $0.4522\pm0.0746$ & $0.4388\pm0.0541$ & $0.4043\pm0.0817$ \\

& 3-shot
& $0.5148\pm0.0584$ & $0.5408\pm0.0315$ & $0.5234\pm0.0338$
& $0.5089\pm0.0531$ & $0.5477\pm0.0851$ & $0.4559\pm0.0276$ \\

& 5-shot
& $0.5828\pm0.0262$ & $0.5475\pm0.0295$ & $0.5661\pm0.0360$
& $0.5563\pm0.0306$ &  $0.5629\pm0.0771$ & $0.5661\pm0.0360$ \\

\midrule

\multirow{4}{*}{Sensitivity}
& 0-shot
& 0.5199 & 0.4994 & 0.4739
& $0.5110\pm0.0528$ & $0.5074\pm0.0456$ & $0.4331\pm0.0542$ \\

& 1-shot
& -- & -- & -- 
& $0.4826\pm0.0748$ & $0.4883\pm0.0212$ & $0.4447\pm0.0798$ \\

& 3-shot
& $0.5641\pm0.0583$ &  $0.5896\pm0.0429$ & $0.5654\pm0.0362$
& $0.5615\pm0.0377$ & $0.5863\pm0.0695$ & $0.5270\pm0.0391$ \\

& 5-shot
& $0.5989\pm0.0331$ & $0.5790\pm0.0351$  & $0.6106\pm0.0377$
& $0.6114\pm0.0318$ & $0.6211\pm0.0671$ & $0.6106\pm0.0377$ \\

\midrule

\multirow{4}{*}{Precision}
& 0-shot
& 0.5238 & 0.4495 & 0.4222
& $0.5058\pm0.0572$ & $0.5005\pm0.0452$  & $0.4270\pm0.0985$ \\

& 1-shot
& -- & -- & -- 
& $0.4768\pm0.0798$ & $0.4581\pm0.0577$ & $0.4319\pm0.0990$ \\

& 3-shot
& $0.5720\pm0.0674$ &  $0.5922\pm0.0625$ & $0.5656\pm0.0405$
& $0.5818\pm0.0580$ & $0.5799\pm0.0779$ & $0.5008\pm0.0284$ \\

& 5-shot
& $0.6187\pm0.0421$ &  $0.5867\pm0.0705$ & $0.6225\pm0.0574$
& $0.6071\pm0.0435$ &  $0.6182\pm0.0828$ & $0.6225\pm0.0574$ \\

\midrule

\multirow{4}{*}{Accuracy}
& 0-shot
& 0.5112 & 0.4936 &  0.4647
& $0.5032\pm0.0397$ & $0.5112\pm0.0410$ & $0.4463\pm0.0558$ \\

& 1-shot
& -- & -- & --
& $0.4786\pm0.0680$ & $0.4869\pm0.0155$ & $0.4598\pm0.0752$ \\

& 3-shot
& $0.5686\pm0.0534$ & $0.5825\pm0.0320$ & $0.5477\pm0.0419$
& $0.5712\pm0.0471$ & $0.5781\pm0.0762$ & $0.5122\pm0.0232$ \\

& 5-shot
& $0.6167\pm0.0236$ &  $0.5917\pm0.0514$ & $0.6167\pm0.0449$
& $0.5792\pm0.0267$ & $0.5917\pm0.0607$ & $0.6167\pm0.0449$ \\

\bottomrule
\end{tabular}}
\end{table*}

\begin{table*}
\centering
\caption{Few-shot classification performance (mean $\pm$ SD) for the D3 and D4 datasets using the L8 configuration and T1/T2 scan MRI modality.}
\label{tab:d3_d4_l8_fewshot}

\scriptsize
\setlength{\tabcolsep}{3pt}

\resizebox{\textwidth}{!}{
\begin{tabular}{ll|ccc|ccc}
\toprule
&& \multicolumn{3}{c|}{\textbf{D3 Dataset}} &
\multicolumn{3}{c}{\textbf{D4 Dataset}}\\
\cmidrule(lr){1-5}
\cmidrule(lr){5-8}

\textbf{Metric} & \textbf{Shot}
& \textbf{EWC}
& \textbf{GBP}
& \textbf{SEQ}
& \textbf{EWC}
& \textbf{GBP}
& \textbf{SEQ}
\\
\midrule

\multirow{4}{*}{F1}
& 0-shot
& 0.3957 & 0.4921 & 0.4776
& $0.3888\pm0.0405$ & $0.4690\pm0.0526$ & $0.4624\pm0.0583$ \\

& 1-shot
& -- & -- & -- 
& $0.4720\pm0.0573$ & $0.4510\pm0.0486$ & $0.4059\pm0.0690$ \\

& 3-shot
& $0.4910\pm0.0182$ & $0.5381\pm0.0312$  & $0.5045\pm0.0625$
& $0.4700\pm0.0573$ & $0.4980\pm0.0572$ & $0.4920\pm0.0248$ \\

& 5-shot
& $0.5124\pm0.0416$ & $0.5593\pm0.0384$  & $0.5238\pm0.0679$
& $0.5431\pm0.0654$ & $0.6009\pm0.0500$ & $0.5452\pm0.0936$ \\

\midrule

\multirow{4}{*}{Sensitivity}
& 0-shot
& 0.4516 &  0.5219 & 0.4998
& $0.4743\pm0.0392$ & $0.4938\pm0.0617$ & $0.4725\pm0.0626$ \\

& 1-shot
& -- & -- & -- 
& $0.5118\pm0.0475$ & $0.4956\pm0.0415$ & $0.5127\pm0.0619$ \\

& 3-shot
& $0.5550\pm0.0116$ & $0.5703\pm0.0324$ & $0.5691\pm0.0587$
& $0.4917\pm0.0577$ & $0.5339\pm0.0615$ & $0.5536\pm0.0459$ \\

& 5-shot
& $0.5712\pm0.0345$ & $0.5869\pm0.0412$  & $0.5904\pm0.0625$
& $0.5684\pm0.0634$ & $0.6390\pm0.0640$ & $0.5975\pm0.0782$ \\

\midrule

\multirow{4}{*}{Precision}
& 0-shot
& 0.4299 & 0.5278 & 0.4903
& $0.4261\pm0.0589$ & $0.5044\pm0.0411$ & $0.4877\pm0.0589$ \\

& 1-shot
& -- & -- & -- 
& $0.5017\pm0.0540$ & $0.5028\pm0.0435$ & $0.4289\pm0.1023$ \\

& 3-shot
& $0.5477\pm0.0448$ & $0.5691\pm0.0231$ & $0.5549\pm0.0591$
& $0.5106\pm0.0726$ & $0.5526\pm0.0469$ & $0.5268\pm0.0419$ \\

& 5-shot
& $0.5596\pm0.0493$  & $0.5794\pm0.0475$ & $0.5667\pm0.0638$
& $0.5717\pm0.0560$  & $0.6411\pm0.0551$ & $0.5946\pm0.0756$ \\

\midrule

\multirow{4}{*}{Accuracy}
& 0-shot
& 0.4679 & 0.5353 & 0.5032
& $0.4904\pm0.0228$ & $0.5080\pm0.0300$ & $0.4936\pm0.0522$ \\

& 1-shot
& -- & -- & -- 
& $0.4994\pm0.0521$ & $0.5030\pm0.0509$ & $0.5092\pm0.0510$ \\

& 3-shot
& $0.5482\pm0.0369$ & $0.5690\pm0.0354$ & $0.5781\pm0.0410$
& $0.5065\pm0.0596$ & $0.5299\pm0.0681$ & $0.5365\pm0.0272$ \\

& 5-shot
& $0.5698\pm0.0442$ & $0.5906\pm0.0391$ & $0.6015\pm0.0472$
& $0.5800\pm0.0481$ &  $0.6250\pm0.0468$ & $0.5900\pm0.1098$ \\

\bottomrule
\end{tabular}}
\end{table*}

Direct comparisons between the S8 and L8 architectures further reveal that model scale alone does not determine transfer performance. On D3 (Table~\ref{tab:chunk_max_results_pm_L8_S8_D3}), GBP achieved the highest precision ($0.719 \pm 0.018$), accuracy ($0.688 \pm 0.045$), and the highest F1-score ($0.667 \pm 0.008$) with the lowest variance among all methods. These findings suggest that the interaction between architecture size and adaptation strategy plays a more important role than model scale alone.

The D4 experiments (Table~\ref{tab:chunk_max_results_S8_L8_D4}) provide additional evidence of robustness under stronger distribution shifts. Although performance variance increased and zero-shot results remained close to random classification, continual-learning approaches maintained a clear advantage over direct fine-tuning. The strongest D4 performance was obtained by EWC ($F1 = 0.648 \pm 0.043$) and GBP ($F1 = 0.625 \pm 0.025$), indicating that both regularisation-based and pruning-based adaptation remain effective under substantial domain mismatch.

\begin{table*}[t]
\centering
\caption{Mean $\pm$ SD of 20-epoch fine-tuning (S2-Level-4) at D3 and D4 evaluation points and T1/T2 scan MRI modality.}
\label{tab:chunk_max_results_compact}

\scriptsize
\setlength{\tabcolsep}{3pt}

\resizebox{\textwidth}{!}{
\begin{tabular}{ll|ccc|ccc}
\toprule
&& \multicolumn{3}{c|}{\textbf{D3}} &
\multicolumn{3}{c}{\textbf{D4}} \\
\cmidrule(lr){1-5}
\cmidrule(lr){5-8}

\textbf{Metric}
& \textbf{Model}
& \textbf{SEQ}
& \textbf{EWC}
& \textbf{GBP}
& \textbf{SEQ}
& \textbf{EWC}
& \textbf{GBP}
\\
\midrule

\multirow{1}{*}{F1}
&
Mean$\pm$SD
&
$0.6556\pm0.0446$
&
$0.6507\pm0.0522$
&
$0.6665\pm0.0076$
&
$0.5801\pm0.0650$
&
$0.6595\pm0.0420$
&
$0.6603\pm0.0248$
\\

\midrule

\multirow{1}{*}{Sensitivity}
&
Mean$\pm$SD
&
$0.6840\pm0.0307$
&
$0.6686\pm0.0337$
&
$0.6922\pm0.0156$
&
$0.6451\pm0.0356$
&
$0.6989\pm0.0555$
&
$0.6886\pm0.0271$
\\

\midrule

\multirow{1}{*}{Precision}
&
Mean$\pm$SD
&
$0.6865\pm0.0398$
&
$0.6997\pm0.0380$
&
$0.6969\pm0.0142$
&
$0.6897\pm0.0105$
&
$0.7185\pm0.0179$
&
$0.6944\pm0.0192$
\\

\midrule

\multirow{1}{*}{Accuracy}
&
Mean$\pm$SD
&
$0.6806\pm0.0314$
&
$0.6649\pm0.0504$
&
$0.6745\pm0.0052$
&
$0.6250\pm0.0454$
&
$0.6875\pm0.0454$
&
$0.6823\pm0.0261$
\\

\bottomrule
\end{tabular}}
\end{table*}

\begin{table*}[t]
\centering
\caption{Mean $\pm$ SD of 20-epoch fine-tuning (S2-Level-4) at D3 and D4 evaluation points and T1/T2 scan MRI modality.}
\label{tab:chunk_max_results_compact_L8}

\scriptsize
\setlength{\tabcolsep}{3pt}

\resizebox{\textwidth}{!}{
\begin{tabular}{l|ccc|ccc}
\toprule
&
\multicolumn{3}{c|}{\textbf{D3}}
&
\multicolumn{3}{c}{\textbf{D4}}
\\
\cmidrule(lr){2-4}
\cmidrule(lr){5-7}

\textbf{Metric}
&
\textbf{SEQ}
&
\textbf{EWC}
&
\textbf{GBP}
&
\textbf{SEQ}
&
\textbf{EWC}
&
\textbf{GBP}
\\
\midrule

F1
&
$0.5419\pm0.0737$
&
$0.5963\pm0.0519$
&
$0.6150\pm0.0176$
&
$0.6094\pm0.0543$
&
$0.6435\pm0.0400$
&
$0.5978\pm0.0530$
\\

\midrule

Sensitivity
&
$0.6458\pm0.0742$
&
$0.6484\pm0.0279$
&
$0.6363\pm0.0159$
&
$0.6753\pm0.0410$
&
$0.6684\pm0.0529$
&
$0.6293\pm0.0388$
\\

\midrule

Precision
&
$0.5503\pm0.0722$
&
$0.6641\pm0.0290$
&
$0.6649\pm0.0232$
&
$0.6493\pm0.0516$
&
$0.6944\pm0.0415$
&
$0.6467\pm0.0403$
\\

\midrule

Accuracy
&
$0.6354\pm0.0675$
&
$0.6615\pm0.0606$
&
$0.6667\pm0.0378$
&
$0.6823\pm0.0415$
&
$0.6927\pm0.0307$
&
$0.6458\pm0.0510$
\\

\bottomrule
\end{tabular}}
\end{table*}

\begin{table*}[t]
\centering
\caption{Mean $\pm$ SD of 20-epoch fine-tuning for the S8 and L8 configurations at the D4 evaluation point and T1/T2 scan MRI modality.}
\label{tab:chunk_max_results_S8_L8_D4}

\scriptsize
\setlength{\tabcolsep}{3pt}

\resizebox{\textwidth}{!}{
\begin{tabular}{l|ccc|ccc}
\toprule
&
\multicolumn{3}{c|}{\textbf{S8}}
&
\multicolumn{3}{c}{\textbf{L8}}
\\
\cmidrule(lr){2-4}
\cmidrule(lr){5-7}

\textbf{Metric}
&
\textbf{SEQ}
&
\textbf{EWC}
&
\textbf{GBP}
&
\textbf{SEQ}
&
\textbf{EWC}
&
\textbf{GBP}
\\
\midrule

F1
&
$0.6302\pm0.0505$
&
$0.6478\pm0.0430$
&
$0.5993\pm0.0238$
&
$0.6271\pm0.0823$
&
$0.6249\pm0.0249$
&
$0.6055\pm0.0465$
\\

\midrule

Sensitivity
&
$0.6607\pm0.0518$
&
$0.6812\pm0.0380$
&
$0.6299\pm0.0096$
&
$0.7028\pm0.0315$
&
$0.6748\pm0.0329$
&
$0.6748\pm0.0386$
\\

\midrule

Precision
&
$0.7092\pm0.0682$
&
$0.6940\pm0.0495$
&
$0.6423\pm0.0201$
&
$0.7488\pm0.0842$
&
$0.6593\pm0.0813$
&
$0.6622\pm0.0466$
\\

\midrule

Accuracy
&
$0.6580\pm0.0568$
&
$0.6632\pm0.0359$
&
$0.6146\pm0.0180$
&
$0.6736\pm0.0535$
&
$0.6493\pm0.0159$
&
$0.6267\pm0.0458$
\\

\bottomrule
\end{tabular}}
\end{table*}

\subsubsection{ODI/NDI Transfer Learning}

The results obtained from ODI/NDI tumour classification are summarized in Table~\ref{tab:odi_ndi_results_compact}. Compared with the T1/T2 experiments, a different performance pattern emerged. The strongest results were achieved by the model of S8 GraphBluePrint-Joint-S8, which reached an F1-score of $0.677 \pm 0.059$, sensitivity of $0.706 \pm 0.062$, precision of $0.739 \pm 0.055$, and accuracy of $0.752 \pm 0.055$.

Compared with direct tumour fine-tuning, GBP-S8 improved F1-score by approximately 17\%, precision by approximately 14\%, and overall accuracy by approximately 27\%. These substantial gains indicate that graph-based continual adaptation is particularly beneficial when transferring between diffusion-derived microstructural representations.

Interestingly, EWC exhibited a much larger performance gap between the S8 and L8 architectures in the ODI/NDI setting than in the T1/T2 experiments. While EWC-L8 achieved competitive results ($F1 = 0.594 \pm 0.055$), EWC-S8 showed substantially lower performance ($F1 = 0.387 \pm 0.055$). This observation suggests that preserving graph-structured representations may be more important than retaining individual feature activations when transferring across microstructural imaging modalities.

Overall, the ODI/NDI results reinforce the conclusions drawn from the T1/T2 experiments. Continual-learning strategies consistently outperform direct fine-tuning, with Graph-Blueprint Joint providing the strongest and most stable performance across metrics. These findings indicate that preserving and selectively reusing graph-based latent representations is a highly effective mechanism for adapting foundation models to downstream tumour classification tasks under both multimodal and microstructural transfer settings.

\begin{table*}[t]
\centering
\caption{Mean $\pm$ SD of ODI/NDI 20-epochs fine-tuning for the S8 and L8 configurations at the D4 evaluation point.}
\label{tab:odi_ndi_results_compact}

\scriptsize
\setlength{\tabcolsep}{3pt}

\resizebox{\textwidth}{!}{
\begin{tabular}{l|ccc|ccc}
\toprule
&
\multicolumn{3}{c|}{\textbf{S8}}
&
\multicolumn{3}{c}{\textbf{L8}}
\\
\cmidrule(lr){2-4}
\cmidrule(lr){5-7}

\textbf{Metric}
&
\textbf{SEQ}
&
\textbf{EWC}
&
\textbf{GBP}
&
\textbf{SEQ}
&
\textbf{EWC}
&
\textbf{GBP}
\\
\midrule

F1
&
$0.5773\pm0.0671$
&
$0.3865\pm0.0551$
&
$0.6770\pm0.0591$
&
$0.5665\pm0.0517$
&
$0.5935\pm0.0545$
&
$0.6424\pm0.0581$
\\

\midrule

Sensitivity
&
$0.6655\pm0.0530$
&
$0.5707\pm0.0081$
&
$0.7057\pm0.0622$
&
$0.6647\pm0.0401$
&
$0.6617\pm0.0756$
&
$0.6646\pm0.0542$
\\

\midrule

Precision
&
$0.6500\pm0.0397$
&
$0.5273\pm0.0443$
&
$0.7388\pm0.0552$
&
$0.6690\pm0.0403$
&
$0.6423\pm0.0582$
&
$0.6799\pm0.0546$
\\

\midrule

Accuracy
&
$0.5938\pm0.0593$
&
$0.4385\pm0.0738$
&
$0.7521\pm0.0553$
&
$0.5875\pm0.0424$
&
$0.6260\pm0.0507$
&
$0.6729\pm0.0552$
\\

\bottomrule
\end{tabular}}
\end{table*}

\subsection{Cross-task analysis and conclusions}

Across the four clinical tasks, a consistent pattern emerges: frozen or lightly adapted foundation representations provide clinically useful signal, but their effectiveness depends strongly on the downstream objective. In the Alzheimer's classification experiments, Swin-UNET latent embeddings achieved performance comparable to conventional CNN baselines, particularly for the binary task, where Level-2 MLP features reached an AUC of $0.845$ and accuracy of $0.757$. Level-4 features remained competitive but showed a modest reduction in performance, suggesting that intermediate latent representations preserve more discriminative disease information than deeper compressed features. In contrast, using the complete diffusion architecture as a direct classifier degraded performance to near chance, indicating that the diffusion backbone alone does not automatically yield clinically separable representations without careful feature selection or task-specific adaptation.

In the brain-tumour generation task, reconstruction and translation performance were not monotonically related. MAISI achieved the strongest overall reconstruction performance, consistent with the benefit of large-scale CT/MRI pretraining. However, the proposed 3D-Swin architecture produced the most transferable latent space for cross-modal generation, outperforming competing encoders in most translation settings, particularly when paired with flow-matching generators. This divergence indicates that high reconstruction fidelity alone is not sufficient for downstream generative utility; the latent geometry must also support semantic alignment across modalities.

In addition, 3D-Swin-DiT achieved either the highest or second-highest performance across most settings compared with the other architectures and diffusion-based latent-space adaptation strategies.

For glioma survival estimation, 3D-Swin-DiT demonstrated the strongest time-to-event discrimination, with EWC-enhanced DiT features achieving the best Cox concordance ($0.592 \pm 0.014$ with MLP Cox and $0.601 \pm 0.016$ with DeepSurv). However, state-of-the-art comparators remained competitive in endpoint-specific tasks: DUNE U-AE achieved the best three-year AUROC and the strongest survival-time regression. Thus, DiT representations appear particularly effective for ranking patient risk, whereas DUNE embeddings better preserve calibrated survival-time information. This dissociation highlights that prognostic ordering and absolute survival-time prediction are related but non-equivalent clinical objectives.

In the post-surgical few-shot tumour adaptation experiments, continual-learning methods were most beneficial when supervision was scarce. GBP and EWC generally outperformed direct tumour fine-tuning in low-shot settings, with GBP providing the most stable performance across D3/D4 and S8/L8 configurations. EWC was competitive in smaller-scale settings, whereas graph-based adaptation became more advantageous at larger scale and under stronger distribution shift. These findings suggest that preserving previously learned representations while selectively adapting task-relevant submodules provides a more robust strategy than unrestricted fine-tuning.

Overall, the results support three main conclusions. First, normal Swin-UNET-style latent models like the proposed (3D-Swin) provide strong and stable discriminative baselines, often matching or exceeding classical CNN/SVM approaches strengthen generalization of the model and avoid overfitting effects. Second, continual-learning mechanisms improve robustness, but their benefit is task-dependent: EWC is strongest for survival-risk ordering, whereas GBP are more reliable for low-shot transfer and adaptation. Third, state-of-the-art external models such as MAISI and DUNE remain highly competitive, particularly for reconstruction and calibrated survival-time prediction, but do not uniformly dominate across tasks. Therefore, no single representation is universally optimal. Instead, the evidence favours a task-aware deployment strategy in which reconstruction, generation, classification, survival ranking, and few-shot adaptation are treated as distinct clinical objectives requiring different latent properties. Our proposed diffusion architecture demonstrates promising results, despite relying solely on a conventional noise-reconstruction fine-tuning paradigm. The model achieves performance comparable to state-of-the-art approaches, including MAISI and DUNE, highlighting the effectiveness of the proposed design. These findings suggest considerable room for further optimization, and future work exploring more sophisticated training, adaptation, and representation-learning strategies may enable substantial performance gains beyond current architectures.

\section*{Data availability}
 \subsection*{Availability of the Training Cohorts}
This study used publicly available, controlled-access and institutional neuroimaging datasets, as detailed in Supplementary Information S5.1. UK-Biobank data were accessed under UK Biobank Application Number 20904. Data from the Parkinson’s Progression Markers Initiative (PPMI) were obtained on 25 August 2025 from the PPMI database (\url{https://www.ppmi-info.org/access-dataspecimens/download-data}; $RRID_006431$) and require registration and a data-use agreement through the LONI Image and Data Archive (\url{https://ida.loni.usc.edu/home/projectPage.jsp?project=PPMI}). Up-to-date information on PPMI is available at \url{https://www.ppmi-info.org}. PPMI is a public--private partnership funded by the Michael J. Fox Foundation for Parkinson’s Research and its funding partners. ADNI data were accessed through the Alzheimer’s Disease Neuroimaging Initiative/LONI portal (\url{https://adni.loni.usc.edu/}) and require registration and approval of the ADNI data-use agreement.

The ABCD diffusion-derived data used in this study were obtained from the Fiber Data Hub ABCD release (\url{https://brain.labsolver.org/abcd.html}). This resource provides derived ABCD FIB files, including baseline ($n=9,713$), 2-year follow-up ($n=7,199$) and 4-year follow-up ($n=2,736$) data, under the Creative Commons Attribution--ShareAlike 4.0 International licence. These files are derived diffusion reconstructions rather than raw MRI data; therefore, no additional NDA request was required for the derived FIB files used here. Raw ABCD MRI data, including T1-weighted, diffusion-weighted and SRC.GZ files, remain subject to the National Data Archive data-use agreement and are not redistributed in this study. Additional neuroimaging resources included the Reproducible Brain Charts dataset, available at \url{https://reprobrainchart.github.io/docs/datasets/}; HABS-HD, available through LONI at \url{https://ida.loni.usc.edu/home/projectPage.jsp?project=HABS_HD}; and PREVENT-AD, available through \url{https://openpreventad.loris.ca/}.

The brain-tumour dataset was assembled from multi-institutional glioma MRI cohorts with public, controlled-access or data-use-agreement provenance. UPenn-GBM (\url{https://doi.org/10.7937/TCIA.709X-DN49}), UCSF-PDGM (\url{https://doi.org/10.7937/TCIA.BDGF-8V37}). The Erasmus Glioma Database was accessed through the BMIA XNAT repository (\url{https://xnat.bmia.nl/data/archive/projects/egd}) under a custom license based on CC BY-NC-SA 4.0.

\subsection*{Availability of the Fine-Tuning Cohorts}
This study used publicly available, controlled-access and institutional neuroimaging datasets, as detailed in Supplementary Information S5.2. Public and repository-based glioma imaging resources were accessed under their original data-use terms and licences. These included BraTS 2025 through Synapse (\url{https://www.synapse.org/Synapse:syn64153130/wiki/630130}), TCGA-LGG (\url{https://www.cancerimagingarchive.net/collection/tcga-lgg/}) and RHUH-GBM (\url{https://www.cancerimagingarchive.net/collection/rhuh-gbm/}) which were used in the generation and translation tasks. Public or repository access links for the adult brain-tumour survival cohorts are: UPenn-GBM (\url{https://doi.org/10.7937/TCIA.709X-DN49}), UCSF-PDGM (\url{https://doi.org/10.7937/TCIA.BDGF-8V37}), TCGA-LGG (\url{https://doi.org/10.7937/K9/TCIA.2016.L4LTD3TK}), TCGA-GBM (\url{https://doi.org/10.7937/K9/TCIA.2016.RNYFUYE9}), LUMIERE (\url{https://doi.org/10.6084/m9.figshare.c.5904905.v1}), REMBRANDT (\url{https://doi.org/10.7937/K9/TCIA.2015.588OZUZB}), RHUH-GBM (\url{https://doi.org/10.7937/4545-C905}), IvyGAP-GBM (\url{https://doi.org/10.7937/K9/TCIA.2016.XLWAN6NL}) and CPTAC-GBM (\url{https://doi.org/10.7937/K9/TCIA.2018.3RJE41Q1}). LUMIERE was obtained from Figshare (\url{https://doi.org/10.6084/m9.figshare.c.5904905.v1}) under CC BY 4.0. Across these resources, available MRI sequences comprised the standard multiparametric glioma protocol: T1-weighted, contrast-enhanced T1-weighted, T2-weighted and FLAIR imaging. The BrainLAT dataset used for Alzheimer's and neurodegenerative-disease analyses is publicly available through Synapse (\url{https://www.synapse.org/Synapse:syn51549340/wiki/624187}). Associated Source Data files for submission should include machine-readable numerical values underlying all main figures, Extended Data figures, retained Supplementary Tables and plotted Supplementary Information results. Aggregated values that can be shared without breaching data-use agreements will be uploaded as separate Source Data files at submission. If any Source Data file cannot be uploaded at initial submission, the same numerical source data will be provided to editors and reviewers during assessment and deposited in a persistent repository before publication.

The institutional SIND dataset cannot be deposited publicly or released through general request because of participant privacy, governance and data-use restrictions. Scientific collaborations may be considered by contacting the corresponding author M.M. (\href{mailto:mm2703@cam.ac.uk}{mm2703@cam.ac.uk}) and S.J.P. (\href{mailto:sjp58@cam.ac.uk}{sjp58@cam.ac.uk}); responses are expected within approximately two weeks. Any access would require independent institutional approvals and formal inter-institutional data-use agreements.

\section*{Ethics and data governance}
All public and controlled-access datasets used in this study consisted of de-identified imaging and clinical data distributed by established repositories under their original ethics approvals, licences and data-use agreements. No attempt was made to re-identify participants. Use of controlled-access resources complied with the access conditions specified by the relevant repositories and data providers. This research has been conducted using the UK-Biobank Resource under Application Number 20904. Access to controlled datasets, including UK-Biobank, PPMI, ADNI and relevant TCIA/GDC-linked resources, complied with the terms specified by the original data providers. Derived ABCD diffusion files were used from the publicly released Fiber Data Hub resource; raw ABCD MRI data were not redistributed and remain governed by NDA access conditions. The study therefore represents a secondary analysis of de-identified public, controlled-access and repository-hosted data, with all restrictions on redistribution preserved.

The institutional SIND dataset was collected as part of the Assessing impact of surgically-induced deficits on patient functioning and quality of life study (SIND study; protocol number SIND-2018), sponsored by Cambridge University Hospitals NHS Foundation Trust and the University of Cambridge. The study received a favourable opinion from the West Midlands - Solihull Research Ethics Committee on 16 May 2019 (REC reference 19/WM/0152; IRAS project ID 256542). Written informed consent was obtained from all participants. SIND data were de-identified before analysis and are not available for public redistribution because of participant privacy, governance and data-use restrictions.

\clearpage


\begin{thebibliography}{92}
\providecommand{\natexlab}[1]{#1}
\providecommand{\url}[1]{\texttt{#1}}
\expandafter\ifx\csname urlstyle\endcsname\relax
  \providecommand{\doi}[1]{doi: #1}\else
  \providecommand{\doi}{doi: \begingroup \urlstyle{rm}\Url}\fi

\bibitem[Wald et~al.(2025)Wald, Ulrich, Suprijadi, Ziegler, Nohel, Peretzke,
  Koehler, and Maier-Hein]{wald2025openmind}
Tassilo Wald, Constantin Ulrich, Jonathan Suprijadi, Sebastian Ziegler, Michal
  Nohel, Robin Peretzke, Gregor Koehler, and Klaus Maier-Hein.
\newblock An openmind for 3d medical vision self-supervised learning.
\newblock In \emph{Proceedings of the IEEE/CVF International Conference on
  Computer Vision (ICCV)}, pages 23839--23879, October 2025.
\newblock \doi{10.48550/arXiv.2412.17041}.
\newblock URL \url{https://arxiv.org/abs/2412.17041}.

\bibitem[Kaczmarek et~al.(2025)Kaczmarek, Szeto, Nichyporuk, and
  Arbel]{kaczmarek2025neurosimclr}
Emily Kaczmarek, Justin Szeto, Brennan Nichyporuk, and Tal Arbel.
\newblock Building a general simclr self-supervised foundation model across
  neurological diseases to advance 3d brain mri diagnoses, 2025.
\newblock URL \url{https://arxiv.org/abs/2509.10620}.

\bibitem[Barba et~al.(2025)Barba, Bagley, Steyaert, Carrillo-Perez, Sad{\'e}e,
  Iv, and Gevaert]{barba2025dune}
Thomas Barba, Bryce~A. Bagley, Sandra Steyaert, Francisco Carrillo-Perez,
  Christoph Sad{\'e}e, Michael Iv, and Olivier Gevaert.
\newblock {DUNE}: a versatile neuroimaging encoder captures brain complexity
  across 3 major diseases: cancer, dementia, and schizophrenia.
\newblock \emph{GigaScience}, 14:\penalty0 giaf116, 2025.
\newblock \doi{10.1093/gigascience/giaf116}.
\newblock URL \url{https://doi.org/10.1093/gigascience/giaf116}.

\bibitem[Tak et~al.(2026)Tak, Garomsa, Zapaishchykova, Chaunzwa, Climent~Pardo,
  Ye, Zielke, Ravipati, Pai, Vajapeyam, Mahootiha, Parker, Pike, Smith,
  Familiar, Liu, Prabhu, Arnaout, Bandopadhayay, Nabavizadeh, Mueller, Aerts,
  Huang, Poussaint, and Kann]{tak2026brainiac}
Divyanshu Tak, Biniam~A. Garomsa, Anna Zapaishchykova, Tafadzwa~L. Chaunzwa,
  Juan~Carlos Climent~Pardo, Zezhong Ye, John Zielke, Yashwanth Ravipati, Suraj
  Pai, Sri Vajapeyam, Maryam Mahootiha, Mitchell Parker, Luke R.~G. Pike,
  Ceilidh Smith, Ariana~M. Familiar, Kevin~X. Liu, Sanjay Prabhu, Omar Arnaout,
  Pratiti Bandopadhayay, Ali Nabavizadeh, Sabine Mueller, Hugo J. W.~L. Aerts,
  Raymond~Y. Huang, Tina~Y. Poussaint, and Benjamin~H. Kann.
\newblock A generalizable foundation model for analysis of human brain {MRI}.
\newblock \emph{Nature Neuroscience}, 29:\penalty0 945--956, 2026.
\newblock \doi{10.1038/s41593-026-02202-6}.

\bibitem[Wu et~al.(2026)Wu, Wang, Li, Safari, Hu, Chang, Veeraraghavan, and
  Yang]{wu2026braindino}
Yizhou Wu, Shansong Wang, Yuheng Li, Mojtaba Safari, Mingzhe Hu, Chih-Wei
  Chang, Harini Veeraraghavan, and Xiaofeng Yang.
\newblock Braindino: A brain mri foundation model for generalizable clinical
  representation learning, 2026.
\newblock URL \url{https://arxiv.org/abs/2604.27277}.

\bibitem[{van de Ven} et~al.(2025){van de Ven}, Soures, and Kudithipudi]{cata}
Gido~M. {van de Ven}, Nicholas Soures, and Dhireesha Kudithipudi.
\newblock 1.09 - continual learning and catastrophic forgetting.
\newblock In John Wixted, editor, \emph{Learning and Memory: A Comprehensive
  Reference (Third Edition)}, pages 153--168. Academic Press, Oxford, third
  edition edition, 2025.
\newblock ISBN 978-0-443-15755-4.
\newblock \doi{10.1016/B978-0-443-15754-7.00073-0}.
\newblock URL
  \url{https://www.sciencedirect.com/science/article/pii/B9780443157547000730}.

\bibitem[Kirkpatrick et~al.(2017)Kirkpatrick, Pascanu, Rabinowitz, Veness,
  Desjardins, Rusu, Milan, Quan, Ramalho, Grabska-Barwinska, Hassabis, Clopath,
  Kumaran, and Hadsell]{kirkpatrick2017}
James Kirkpatrick, Razvan Pascanu, Neil Rabinowitz, Joel Veness, Guillaume
  Desjardins, Andrei~A. Rusu, Kieran Milan, John Quan, Tiago Ramalho, Agnieszka
  Grabska-Barwinska, Demis Hassabis, Claudia Clopath, Dharshan Kumaran, and
  Raia Hadsell.
\newblock Overcoming catastrophic forgetting in neural networks.
\newblock \emph{Proceedings of the National Academy of Sciences of the United
  States of America}, 114\penalty0 (13):\penalty0 3521--3526, 2017.
\newblock \doi{10.1073/pnas.1611835114}.

\bibitem[Lindsey et~al.(2025)Lindsey, Gurnee, Ameisen, Chen, Pearce, Turner,
  Citro, Abrahams, Carter, Hosmer, Marcus, Sklar, Templeton, Bricken,
  McDougall, Cunningham, Henighan, Jermyn, Jones, Persic, Qi, Thompson,
  Zimmerman, Rivoire, Conerly, Olah, and Batson]{anthropic2025graphs}
Jack Lindsey, Wes Gurnee, Emmanuel Ameisen, Brian Chen, Adam Pearce,
  Nicholas~L. Turner, Craig Citro, David Abrahams, Shan Carter, Basil Hosmer,
  Jonathan Marcus, Michael Sklar, Adly Templeton, Trenton Bricken, Callum
  McDougall, Hoagy Cunningham, Thomas Henighan, Adam Jermyn, Andy Jones, Andrew
  Persic, Zhenyi Qi, T.~Ben Thompson, Sam Zimmerman, Kelley Rivoire, Thomas
  Conerly, Chris Olah, and Joshua Batson.
\newblock On the biology of a large language model.
\newblock \emph{Transformer Circuits Thread}, 2025.
\newblock URL
  \url{https://transformer-circuits.pub/2025/attribution-graphs/biology.html}.

\bibitem[Kang et~al.(2022)Kang, Mina, Madjid, Yoon, Hasegawa-Johnson, Hwang,
  and Yoo]{kang2022wsn}
Haeyong Kang, Rusty John~Lloyd Mina, Sultan Rizky~Hikmawan Madjid, Jaehong
  Yoon, Mark Hasegawa-Johnson, Sung~Ju Hwang, and Chang~D. Yoo.
\newblock Forget-free continual learning with winning subnetworks.
\newblock In \emph{Proceedings of the 39th International Conference on Machine
  Learning}, volume 162 of \emph{Proceedings of Machine Learning Research},
  pages 10734--10750. PMLR, 2022.
\newblock URL \url{https://proceedings.mlr.press/v162/kang22b.html}.

\bibitem[Serr{\`a} et~al.(2018)Serr{\`a}, Sur{\'i}s, Miron, and
  Karatzoglou]{serra2018hat}
Joan Serr{\`a}, D{\'i}dac Sur{\'i}s, Marius Miron, and Alexandros Karatzoglou.
\newblock Overcoming catastrophic forgetting with hard attention to the task.
\newblock In \emph{Proceedings of the 35th International Conference on Machine
  Learning}, volume~80 of \emph{Proceedings of Machine Learning Research},
  pages 4548--4557. PMLR, 2018.
\newblock URL \url{https://proceedings.mlr.press/v80/serra18a.html}.

\bibitem[Mirzadeh et~al.(2020)Mirzadeh, Farajtabar, G{\"{o}}r{\"{u}}r, Pascanu,
  and Ghasemzadeh]{reg}
Seyed{-}Iman Mirzadeh, Mehrdad Farajtabar, Dilan G{\"{o}}r{\"{u}}r, Razvan
  Pascanu, and Hassan Ghasemzadeh.
\newblock Linear mode connectivity in multitask and continual learning.
\newblock \emph{CoRR}, abs/2010.04495, 2020.
\newblock URL \url{https://arxiv.org/abs/2010.04495}.

\bibitem[Chen et~al.(2025)Chen, Wuerkaixi, Cui, Li, Li, Zhang, Han, Niu, Liu,
  Yang, Yang, Zhang, and Ren]{chen2025pathwayprotection}
Zhikang Chen, Abudukelimu Wuerkaixi, Sen Cui, Haoxuan Li, Ding Li, Jingfeng
  Zhang, Bo~Han, Gang Niu, Houfang Liu, Yi~Yang, Sifan Yang, Changshui Zhang,
  and Tianling Ren.
\newblock Learning without isolation: Pathway protection for continual
  learning.
\newblock In \emph{Proceedings of the 42nd International Conference on Machine
  Learning}, volume 267 of \emph{Proceedings of Machine Learning Research},
  pages 9377--9399. PMLR, 2025.
\newblock URL \url{https://proceedings.mlr.press/v267/chen25bt.html}.

\bibitem[Liu et~al.(2021)Liu, Lin, Cao, Hu, Wei, Zhang, Lin, and
  Guo]{liu2021swin}
Ze~Liu, Yutong Lin, Yue Cao, Han Hu, Yixuan Wei, Zheng Zhang, Stephen Lin, and
  Baining Guo.
\newblock Swin transformer: Hierarchical vision transformer using shifted
  windows.
\newblock \emph{Proceedings of the IEEE/CVF International Conference on
  Computer Vision (ICCV)}, 2021.
\newblock URL \url{https://arxiv.org/abs/2103.14030}.

\bibitem[Tang et~al.(2022)Tang, Yang, Li, Roth, Landman, Xu, Nath, and
  Hatamizadeh]{tang2022self}
Yucheng Tang, Dong Yang, Wenqi Li, Holger~R. Roth, Bennett~A. Landman, Daguang
  Xu, Vishwesh Nath, and Ali Hatamizadeh.
\newblock Self-supervised pre-training of swin transformers for 3d medical
  image analysis.
\newblock \emph{Proceedings of the IEEE/CVF Conference on Computer Vision and
  Pattern Recognition}, pages 20730--20740, 2022.
\newblock \doi{10.1109/CVPR52688.2022.02007}.
\newblock URL
  \url{https://openaccess.thecvf.com/content/CVPR2022/html/Tang_Self-Supervised_Pre-Training_of_Swin_Transformers_for_3D_Medical_Image_Analysis_CVPR_2022_paper.html}.

\bibitem[{\c{C}}i{\c{c}}ek et~al.(2016){\c{C}}i{\c{c}}ek, Abdulkadir, Lienkamp,
  Brox, and Ronneberger]{cicek2016}
{\"O}zg{\"u}n {\c{C}}i{\c{c}}ek, Ahmed Abdulkadir, Soeren~S. Lienkamp, Thomas
  Brox, and Olaf Ronneberger.
\newblock 3d u-net: Learning dense volumetric segmentation from sparse
  annotation.
\newblock In \emph{Medical Image Computing and Computer-Assisted Intervention
  -- MICCAI 2016}, volume 9901 of \emph{Lecture Notes in Computer Science},
  pages 424--432. Springer, 2016.
\newblock \doi{10.1007/978-3-319-46723-8_49}.

\bibitem[Peebles and Xie(2023)]{peebles2023}
William Peebles and Saining Xie.
\newblock Scalable diffusion models with transformers.
\newblock In \emph{Proceedings of the IEEE/CVF International Conference on
  Computer Vision (ICCV)}, pages 4195--4205, 2023.
\newblock \doi{10.1109/ICCV51070.2023.00387}.

\bibitem[Dosovitskiy et~al.(2021)Dosovitskiy, Beyer, Kolesnikov, Weissenborn,
  Zhai, Unterthiner, Dehghani, Minderer, Heigold, Gelly, Uszkoreit, and
  Houlsby]{dosovitskiy2021}
Alexey Dosovitskiy, Lucas Beyer, Alexander Kolesnikov, Dirk Weissenborn,
  Xiaohua Zhai, Thomas Unterthiner, Mostafa Dehghani, Matthias Minderer, Georg
  Heigold, Sylvain Gelly, Jakob Uszkoreit, and Neil Houlsby.
\newblock An image is worth 16x16 words: Transformers for image recognition at
  scale.
\newblock In \emph{International Conference on Learning Representations
  (ICLR)}, 2021.
\newblock URL \url{https://openreview.net/forum?id=YicbFdNTTy}.

\bibitem[Ho et~al.(2020)Ho, Jain, and Abbeel]{ho2020}
Jonathan Ho, Ajay Jain, and Pieter Abbeel.
\newblock Denoising diffusion probabilistic models.
\newblock In \emph{Advances in Neural Information Processing Systems},
  volume~33, pages 6840--6851, 2020.
\newblock URL
  \url{https://proceedings.neurips.cc/paper/2020/hash/4c5bcfec8584af0d967f1ab10179ca4b-Abstract.html}.

\bibitem[Ronneberger et~al.(2015)Ronneberger, Fischer, and
  Brox]{ronneberger2015}
Olaf Ronneberger, Philipp Fischer, and Thomas Brox.
\newblock U-net: Convolutional networks for biomedical image segmentation.
\newblock In \emph{Medical Image Computing and Computer-Assisted Intervention
  -- MICCAI 2015}, volume 9351 of \emph{Lecture Notes in Computer Science},
  pages 234--241. Springer, 2015.
\newblock \doi{10.1007/978-3-319-24574-4_28}.

\bibitem[Kingma and Welling(2014{\natexlab{a}})]{kingma2014}
Diederik~P. Kingma and Max Welling.
\newblock Auto-encoding variational bayes.
\newblock In \emph{International Conference on Learning Representations
  (ICLR)}, 2014{\natexlab{a}}.

\bibitem[van~den Oord et~al.(2017)van~den Oord, Vinyals, and
  Kavukcuoglu]{van2017neural}
Aaron van~den Oord, Oriol Vinyals, and Koray Kavukcuoglu.
\newblock Neural discrete representation learning.
\newblock In \emph{Advances in Neural Information Processing Systems},
  volume~30, 2017.
\newblock URL
  \url{https://proceedings.neurips.cc/paper/2017/hash/7a98af17e63a0ac09ce2e96d03992fbc-Abstract.html}.

\bibitem[Rombach et~al.(2022)Rombach, Blattmann, Lorenz, Esser, and
  Ommer]{rombach2022high}
Robin Rombach, Andreas Blattmann, Dominik Lorenz, Patrick Esser, and Bj{\"o}rn
  Ommer.
\newblock High-resolution image synthesis with latent diffusion models.
\newblock In \emph{Proceedings of the IEEE/CVF Conference on Computer Vision
  and Pattern Recognition}, pages 10684--10695, 2022.
\newblock \doi{10.1109/CVPR52688.2022.01042}.
\newblock URL
  \url{https://openaccess.thecvf.com/content/CVPR2022/html/Rombach_High-Resolution_Image_Synthesis_With_Latent_Diffusion_Models_CVPR_2022_paper.html}.

\bibitem[Guo et~al.(2025)Guo, Zhao, Yang, Xu, Nath, Tang, Simon, Belue, Harmon,
  Turkbey, and Xu]{guo2025maisi}
Pengfei Guo, Can Zhao, Dong Yang, Ziyue Xu, Vishwesh Nath, Yucheng Tang,
  Benjamin Simon, Mason Belue, Stephanie Harmon, Baris Turkbey, and Daguang Xu.
\newblock Maisi: Medical ai for synthetic imaging.
\newblock In \emph{Proceedings of the Winter Conference on Applications of
  Computer Vision (WACV)}, pages 4430--4441, February 2025.
\newblock \doi{10.1109/WACV61041.2025.00435}.

\bibitem[Yazdani et~al.(2025)Yazdani, Medghalchi, Ashrafian, Hacihaliloglu, and
  Shahriari]{yazdani2025flow}
Milad Yazdani, Yasamin Medghalchi, Pooria Ashrafian, Ilker Hacihaliloglu, and
  Dena Shahriari.
\newblock Flow matching for medical image synthesis: Bridging the gap between
  speed and quality.
\newblock In \emph{Medical Image Computing and Computer Assisted Intervention
  -- MICCAI 2025}, volume 15975 of \emph{Lecture Notes in Computer Science},
  pages 216--226. Springer Nature Switzerland, September 2025.
\newblock URL \url{https://papers.miccai.org/miccai-2025/0343-Paper1056.html}.

\bibitem[Liu et~al.(2022)Liu, Gong, and Liu]{liu2022flow}
Xingchao Liu, Chengyue Gong, and Qiang Liu.
\newblock Flow straight and fast: Learning to generate and transfer data with
  rectified flow.
\newblock \emph{arXiv preprint arXiv:2209.03003}, 2022.

\bibitem[Chen et~al.(2019)Chen, Ma, and Zheng]{chen2019med3d}
Sihong Chen, Kai Ma, and Yefeng Zheng.
\newblock Med3d: Transfer learning for 3d medical image analysis.
\newblock \emph{arXiv preprint arXiv:1904.00625}, 2019.
\newblock URL \url{https://arxiv.org/abs/1904.00625}.

\bibitem[Zhao et~al.(2025)Zhao, Lyu, Chowdury, Harake, Kondepudi, Rao, Hou,
  Lee, and Hollon]{zhao2025hlip}
Chenhui Zhao, Yiwei Lyu, Asadur Chowdury, Edward Harake, Akhil Kondepudi,
  Akshay Rao, Xinhai Hou, Honglak Lee, and Todd Hollon.
\newblock Towards scalable language-image pre-training for 3d medical imaging.
\newblock \emph{arXiv preprint arXiv:2505.21862}, 2025.
\newblock URL \url{https://arxiv.org/abs/2505.21862}.

\bibitem[Sanchez et~al.(2022)Sanchez, Kascenas, Liu, O'Neil, and
  Tsaftaris]{sanchez2022healthy}
Pedro Sanchez, Antanas Kascenas, Xiao Liu, Alison~Q. O'Neil, and Sotirios~A.
  Tsaftaris.
\newblock What is healthy? generative counterfactual diffusion for lesion
  localization.
\newblock In \emph{Deep Generative Models}, volume 13609 of \emph{Lecture Notes
  in Computer Science}, pages 34--44. Springer, 2022.
\newblock \doi{10.1007/978-3-031-18576-2_4}.

\bibitem[Kumar et~al.(2022)Kumar, Hu, Nichyporuk, Falet, Arnold, Tsaftaris, and
  Arbel]{kumar2022counterfactual}
Amar Kumar, Anjun Hu, Brennan Nichyporuk, Jean-Pierre~R. Falet, Douglas~L.
  Arnold, Sotirios~A. Tsaftaris, and Tal Arbel.
\newblock Counterfactual image synthesis for discovery of personalized
  predictive image markers.
\newblock In \emph{Artificial Intelligence over Infrared Images for Medical
  Applications and Medical Image Assisted Biomarker Discovery}, volume 13602 of
  \emph{Lecture Notes in Computer Science}, pages 113--124. Springer, 2022.
\newblock \doi{10.1007/978-3-031-19660-7_11}.

\bibitem[P{\'a}lsson et~al.(2022)P{\'a}lsson, Cerri, Poulsen, Urup, Law, and
  Van~Leemput]{palsson2022survival}
Sveinn P{\'a}lsson, Stefano Cerri, Hans~Skovgaard Poulsen, Thomas Urup, Ian
  Law, and Koen Van~Leemput.
\newblock Predicting survival of glioblastoma from automatic whole-brain and
  tumor segmentation of {MR} images.
\newblock \emph{Scientific Reports}, 12:\penalty0 19744, 2022.
\newblock \doi{10.1038/s41598-022-19223-3}.

\bibitem[Yoon et~al.(2020)Yoon, Cheon, Jeong, Kim, Kim, Nam, Han, and
  Lim]{yoon2020multiparametric}
Han~Gyul Yoon, Wonjoong Cheon, Sang~Woon Jeong, Hye~Seung Kim, Kyunga Kim,
  Heerim Nam, Youngyih Han, and Do~Hoon Lim.
\newblock Multi-parametric deep learning model for prediction of overall
  survival after postoperative concurrent chemoradiotherapy in glioblastoma
  patients.
\newblock \emph{Cancers}, 12\penalty0 (8):\penalty0 2284, 2020.
\newblock \doi{10.3390/cancers12082284}.

\bibitem[Rutherford et~al.(2022)Rutherford, Kia, Wolfers, Fraza, Zabihi, Dinga,
  Berthet, Worker, Verdi, Ruhe, Beckmann, and
  Marquand]{rutherford2022normative}
Saige Rutherford, Seyed~Mostafa Kia, Thomas Wolfers, Charlotte Fraza, Mariam
  Zabihi, Richard Dinga, Pierre Berthet, Amanda Worker, Serena Verdi,
  Henricus~G. Ruhe, Christian~F. Beckmann, and Andre~F. Marquand.
\newblock The normative modeling framework for computational psychiatry.
\newblock \emph{Nature Protocols}, 17\penalty0 (7):\penalty0 1711--1734, 2022.
\newblock \doi{10.1038/s41596-022-00696-5}.

\bibitem[Young et~al.(2018)Young, Marinescu, Oxtoby, et~al.]{young2018sustain}
Alexandra~L. Young, Razvan~V. Marinescu, Neil~P. Oxtoby, et~al.
\newblock Uncovering the heterogeneity and temporal complexity of
  neurodegenerative diseases with subtype and stage inference.
\newblock \emph{Nature Communications}, 9:\penalty0 4273, 2018.
\newblock \doi{10.1038/s41467-018-05892-0}.

\bibitem[{UK Biobank}(2026)]{ukbiobank}
{UK Biobank}.
\newblock Uk biobank.
\newblock \url{https://www.ukbiobank.ac.uk/}, 2026.
\newblock URL \url{https://www.ukbiobank.ac.uk/}.
\newblock Accessed 17 August 2026.

\bibitem[Sudlow et~al.(2015)Sudlow, Gallacher, Allen, et~al.]{ukbiobank_paper}
Cathie Sudlow, John Gallacher, Naomi Allen, et~al.
\newblock Uk biobank: an open access resource for identifying the causes of a
  wide range of complex diseases of middle and old age.
\newblock \emph{PLoS Medicine}, 12\penalty0 (3):\penalty0 e1001779, 2015.
\newblock \doi{10.1371/journal.pmed.1001779}.
\newblock URL \url{https://doi.org/10.1371/journal.pmed.1001779}.
\newblock PMID:25826379; PMCID:PMC4380465.

\bibitem[Marek et~al.(2018)Marek, Chowdhury, Siderowf, Lasch, Coffey,
  Caspell-Garcia, Simuni, Jennings, Tanner, Trojanowski, Shaw, Seibyl, Schuff,
  Singleton, Kieburtz, Toga, Mollenhauer, Galasko, Chahine, Weintraub, Foroud,
  Tosun-Turgut, Poston, Arnedo, Frasier, Sherer, and the Parkinson's
  Progression Markers~Initiative]{ppmi}
Kenneth Marek, Sohini Chowdhury, Andrew Siderowf, Shirley Lasch, Christopher~S.
  Coffey, Chelsea Caspell-Garcia, Tanya Simuni, Danna Jennings, Caroline~M.
  Tanner, John~Q. Trojanowski, Leslie~M. Shaw, John Seibyl, Norbert Schuff,
  Andrew Singleton, Karl Kieburtz, Arthur~W. Toga, Brit Mollenhauer, Doug
  Galasko, Lana~M. Chahine, Daniel Weintraub, Tatiana Foroud, Duygu
  Tosun-Turgut, Kathleen Poston, Vanessa Arnedo, Mark Frasier, Todd Sherer, and
  the Parkinson's Progression Markers~Initiative.
\newblock The parkinson's progression markers initiative (ppmi) –
  establishing a pd biomarker cohort.
\newblock \emph{Annals of Clinical and Translational Neurology}, 5\penalty0
  (12):\penalty0 1460--1477, 2018.
\newblock \doi{10.1002/acn3.644}.
\newblock URL \url{https://onlinelibrary.wiley.com/doi/abs/10.1002/acn3.644}.

\bibitem[Weiner et~al.(2017)Weiner, Veitch, Aisen, Beckett, Cairns, Green,
  Harvey, Jack~Jr., Jagust, Morris, Petersen, Salazar, Saykin, Shaw, Toga,
  Trojanowski, and Initiative]{adni}
Michael~W. Weiner, Dallas~P. Veitch, Paul~S. Aisen, Laurel~A. Beckett, Nigel~J.
  Cairns, Robert~C. Green, Danielle Harvey, Clifford~R. Jack~Jr., William
  Jagust, John~C. Morris, Ronald~C. Petersen, Jennifer Salazar, Andrew~J.
  Saykin, Leslie~M. Shaw, Arthur~W. Toga, John~Q. Trojanowski, and Alzheimer's
  Disease~Neuroimaging Initiative.
\newblock The alzheimer's disease neuroimaging initiative 3: Continued
  innovation for clinical trial improvement.
\newblock \emph{Alzheimer's \& Dementia}, 13\penalty0 (5):\penalty0 561--571,
  2017.
\newblock \doi{10.1016/j.jalz.2016.10.006}.
\newblock URL
  \url{https://alz-journals.onlinelibrary.wiley.com/doi/abs/10.1016/j.jalz.2016.10.006}.

\bibitem[Villeneuve et~al.(2025)Villeneuve, Poirier, Breitner,
  Tremblay-Mercier, Remz, Raoult, Yakoub, Gallego-Rudolf, Qiu, Fajardo~Valdez,
  Mohammediyan, Javanray, Metz, Sanami, Ourry, Wearn, Pastor-Bernier, Edde,
  Gonneaud, Strikwerda-Brown, Tardif, Gauthier, Descoteaux, Dadar,
  Vachon-Presseau, Baril, Ducharme, Montembeault, Geddes, Soucy, Rajah,
  Laforce, Bocti, Davatzikos, Bellec, Rosa-Neto, Baillet, Evans, Collins,
  Chakravarty, Blennow, Zetterberg, Spreng, Pichet~Binette, and the PREVENT-AD
  Research~Group]{preventad}
Sylvia Villeneuve, Judes Poirier, John C.~S. Breitner, Jennifer
  Tremblay-Mercier, Jordana Remz, Jean-Michel Raoult, Yara Yakoub, Jonathan
  Gallego-Rudolf, Ting Qiu, Alfonso Fajardo~Valdez, Bery Mohammediyan,
  Mohammadali Javanray, Amelie Metz, Safa Sanami, Valentin Ourry, Alfie Wearn,
  Alexandre Pastor-Bernier, Manon Edde, Julie Gonneaud, Cherie
  Strikwerda-Brown, Christine~L. Tardif, Claudine~J. Gauthier, Maxime
  Descoteaux, Mahsa Dadar, Étienne Vachon-Presseau, Andrée-Ann Baril, Simon
  Ducharme, Maxime Montembeault, Maiya~R. Geddes, Jean-Paul Soucy, Natasha
  Rajah, Robert Laforce, Christian Bocti, Christos Davatzikos, Lune Bellec,
  Pedro Rosa-Neto, Sylvain Baillet, Alan~C. Evans, D.~Louis Collins, M.~Mallar
  Chakravarty, Kaj Blennow, Henrik Zetterberg, R.~Nathan Spreng, Alexa
  Pichet~Binette, and the PREVENT-AD Research~Group.
\newblock The prevent-ad cohort: Accelerating alzheimer's disease research and
  treatment in canada and beyond.
\newblock \emph{Alzheimer's \& Dementia}, 21\penalty0 (10):\penalty0 e70653,
  2025.
\newblock \doi{10.1002/alz.70653}.
\newblock URL
  \url{https://alz-journals.onlinelibrary.wiley.com/doi/abs/10.1002/alz.70653}.

\bibitem[Casey et~al.(2018)Casey, Cannonier, Conley, Cohen, Barch, Heitzeg,
  Soules, Teslovich, Dellarco, Garavan, Orr, Wager, Banich, Speer, Sutherland,
  Riedel, Dick, Bjork, Thomas, Chaarani, Mejia, Hagler, {Daniela Cornejo},
  Sicat, Harms, Dosenbach, Rosenberg, Earl, Bartsch, Watts, Polimeni, Kuperman,
  Fair, and Dale]{abcd}
B.J. Casey, Tariq Cannonier, May~I. Conley, Alexandra~O. Cohen, Deanna~M.
  Barch, Mary~M. Heitzeg, Mary~E. Soules, Theresa Teslovich, Danielle~V.
  Dellarco, Hugh Garavan, Catherine~A. Orr, Tor~D. Wager, Marie~T. Banich,
  Nicole~K. Speer, Matthew~T. Sutherland, Michael~C. Riedel, Anthony~S. Dick,
  James~M. Bjork, Kathleen~M. Thomas, Bader Chaarani, Margie~H. Mejia,
  Donald~J. Hagler, M.~{Daniela Cornejo}, Chelsea~S. Sicat, Michael~P. Harms,
  Nico~U.F. Dosenbach, Monica Rosenberg, Eric Earl, Hauke Bartsch, Richard
  Watts, Jonathan~R. Polimeni, Joshua~M. Kuperman, Damien~A. Fair, and
  Anders~M. Dale.
\newblock The adolescent brain cognitive development (abcd) study: Imaging
  acquisition across 21 sites.
\newblock \emph{Developmental Cognitive Neuroscience}, 32:\penalty0 43--54,
  2018.
\newblock ISSN 1878-9293.
\newblock \doi{10.1016/j.dcn.2018.03.001}.
\newblock URL
  \url{https://www.sciencedirect.com/science/article/pii/S1878929317301214}.

\bibitem[Shafiei et~al.(2025)Shafiei, Esper, Hoffmann, Ai, Chen, Cluce, Covitz,
  Giavasis, Lane, Mehta, et~al.]{shafiei2025rbc}
Golia Shafiei, Nicholas~B. Esper, Matheus~S. Hoffmann, Lei Ai, Albert~A. Chen,
  Julia Cluce, Sarah Covitz, Sophia Giavasis, Chloe Lane, Kruti Mehta, et~al.
\newblock Reproducible brain charts: An open data resource for mapping brain
  development and its associations with mental health.
\newblock \emph{Neuron}, 113\penalty0 (22):\penalty0 3758--3779.e6, 2025.
\newblock \doi{10.1016/j.neuron.2025.08.026}.

\bibitem[Petersen et~al.(2025)Petersen, Zhou, Hall, Phillips, Meeker, Borzage,
  Braskie, Clark, Shi, Rissman, Zhang, Vintimilla, Casas, Rhodes, Barber,
  Johnson, Yaffe, Toga, O'Bryant, and for~the HABS-HD Study~Team]{habs_hd}
Melissa~E. Petersen, Zhengyang Zhou, James~R. Hall, Nicole Phillips, Karin~L.
  Meeker, Matthew~T. Borzage, Meredith~N. Braskie, Alexandra~L. Clark, Yonggang
  Shi, Robert~A. Rissman, Fan Zhang, Raul Vintimilla, Antonio Casas, Jill
  Rhodes, Robert~C. Barber, Leigh Johnson, Kristine Yaffe, Arthur~W. Toga,
  Sid~E. O'Bryant, and for~the HABS-HD Study~Team.
\newblock Health and aging brain study–health disparities (habs-hd) methods
  and partner characteristics.
\newblock \emph{Alzheimer's \& Dementia: Translational Research \& Clinical
  Interventions}, 11\penalty0 (3):\penalty0 e70140, 2025.
\newblock \doi{10.1002/trc2.70140}.
\newblock URL
  \url{https://alz-journals.onlinelibrary.wiley.com/doi/abs/10.1002/trc2.70140}.

\bibitem[{The Cancer Imaging Archive}(2021)]{upenn_gbm}
{The Cancer Imaging Archive}.
\newblock The university of pennsylvania glioblastoma cohort (upenn-gbm), 2021.
\newblock URL \url{https://doi.org/10.7937/TCIA.709X-DN49}.
\newblock Data set.

\bibitem[{The Cancer Imaging Archive}(2022)]{ucsf_pdgm}
{The Cancer Imaging Archive}.
\newblock The university of california san francisco preoperative diffuse
  glioma mri dataset (ucsf-pdgm), 2022.
\newblock URL \url{https://doi.org/10.7937/TCIA.BDGF-8V37}.
\newblock Data set.

\bibitem[{Erasmus MC} and {Health-RI}(2021)]{egd}
{Erasmus MC} and {Health-RI}.
\newblock The erasmus glioma database (egd), 2021.
\newblock URL \url{https://xnat.bmia.nl/data/archive/projects/egd}.
\newblock Controlled-access data set.

\bibitem[Isensee et~al.(2019)]{isensee2019hdbet}
Fabian Isensee et~al.
\newblock Automated brain extraction of multisequence mri using artificial
  neural networks.
\newblock \emph{Human brain mapping}, 40\penalty0 (17):\penalty0 4952--4964,
  2019.
\newblock \doi{10.1002/hbm.24750}.
\newblock URL \url{https://doi.org/10.1002/hbm.24750}.
\newblock PMID:31403237; PMCID:PMC6865732.

\bibitem[Prado et~al.(2023)Prado, Medel, Gonzalez-Gomez, et~al.]{brainlat}
Pablo Prado, Vicente Medel, Ricardo Gonzalez-Gomez, et~al.
\newblock The brainlat project, a multimodal neuroimaging dataset of
  neurodegeneration from underrepresented backgrounds.
\newblock \emph{Scientific Data}, 10:\penalty0 889, 2023.
\newblock \doi{10.1038/s41597-023-02806-8}.
\newblock URL \url{https://doi.org/10.1038/s41597-023-02806-8}.

\bibitem[Adewole et~al.(2023)Adewole, Rudie, Gbdamosi, Toyobo, Raymond, Zhang,
  Omidiji, Akinola, Suwaid, Emegoakor, et~al.]{adewole2023brain}
M.~Adewole, J.~D. Rudie, A.~Gbdamosi, O.~Toyobo, C.~Raymond, D.~Zhang,
  O.~Omidiji, R.~Akinola, M.~A. Suwaid, A.~Emegoakor, et~al.
\newblock The brain tumor segmentation (brats) challenge 2023: Glioma
  segmentation in sub-saharan africa patient population (brats-africa).
\newblock \emph{arXiv preprint arXiv:2305.19369}, 2023.

\bibitem[Moawad et~al.(2023)Moawad, Janas, Baid, Ramakrishnan, Saluja, Ashraf,
  Maleki, Jekel, Yordanov, Fehringer, et~al.]{moawad2024brain}
A.~W. Moawad, A.~Janas, U.~Baid, D.~Ramakrishnan, R.~Saluja, N.~Ashraf,
  N.~Maleki, L.~Jekel, N.~Yordanov, P.~Fehringer, et~al.
\newblock The brain tumor segmentation-metastases (brats-mets) challenge 2023:
  Brain metastasis segmentation on pre-treatment mri.
\newblock \emph{arXiv preprint arXiv:2306.00838}, 2023.

\bibitem[Kazerooni et~al.(2023)Kazerooni, Khalili, Liu, Haldar, Jiang, Anwar,
  Albrecht, Adewole, Anazodo, Anderson, et~al.]{kazerooni2024brain}
A.~F. Kazerooni, N.~Khalili, X.~Liu, D.~Haldar, Z.~Jiang, S.~M. Anwar,
  J.~Albrecht, M.~Adewole, U.~Anazodo, H.~Anderson, et~al.
\newblock The brain tumor segmentation (brats) challenge 2023: Focus on
  pediatrics (cbtn-connect-dipgr-asnr-miccai brats-peds).
\newblock \emph{arXiv preprint arXiv:2305.17033}, 2023.

\bibitem[Pedano et~al.(2016)Pedano, Flanders, Scarpace, Mikkelsen, Eschbacher,
  Hermes, Sisneros, Barnholtz-Sloan, and Ostrom]{pedano2016tcgalgg}
N.~Pedano, A.~E. Flanders, L.~Scarpace, T.~Mikkelsen, J.~M. Eschbacher,
  B.~Hermes, V.~Sisneros, J.~Barnholtz-Sloan, and Q.~Ostrom.
\newblock The cancer genome atlas low grade glioma collection (tcga-lgg), 2016.
\newblock URL \url{https://doi.org/10.7937/K9/TCIA.2016.L4LTD3TK}.
\newblock Data set.

\bibitem[{Biomedical Image Analysis Group, Imperial College London}(2002)]{ixi}
{Biomedical Image Analysis Group, Imperial College London}.
\newblock Ixi dataset: Information extraction from images, 2002.
\newblock URL \url{https://brain-development.org/ixi-dataset/}.
\newblock EPSRC GR/S21533/02.

\bibitem[Cepeda et~al.(2023)Cepeda, Garc{\'i}a-Garc{\'i}a, Arrese, Herrero,
  Escudero, Zamora, and Sarabia]{cepeda2023rhuh}
S.~Cepeda, S.~Garc{\'i}a-Garc{\'i}a, I.~Arrese, F.~Herrero, T.~Escudero,
  T.~Zamora, and R.~Sarabia.
\newblock The r{\'i}o hortega university hospital glioblastoma dataset: A
  comprehensive collection of preoperative, early postoperative and recurrence
  mri scans (rhuh-gbm), 2023.
\newblock URL \url{https://doi.org/10.7937/4545-c905}.
\newblock Data set.

\bibitem[{The Cancer Imaging Archive}(2016{\natexlab{a}})]{tcga_gbm}
{The Cancer Imaging Archive}.
\newblock The cancer genome atlas glioblastoma multiforme collection
  (tcga-gbm), 2016{\natexlab{a}}.
\newblock URL \url{https://doi.org/10.7937/K9/TCIA.2016.RNYFUYE9}.
\newblock Data set.

\bibitem[Suter et~al.(2022)Suter, Knecht, Valenzuela, Notter, Hewer, Schucht,
  Wiest, and Reyes]{lumiere}
Y.~Suter, U.~Knecht, W.~Valenzuela, M.~Notter, E.~Hewer, P.~Schucht, R.~Wiest,
  and M.~Reyes.
\newblock The lumiere dataset: Longitudinal glioblastoma mri with expert rano
  evaluation, 2022.
\newblock URL \url{https://doi.org/10.6084/m9.figshare.c.5904905.v1}.
\newblock Data set.

\bibitem[Scarpace et~al.(2019)Scarpace, Flanders, Jain, Mikkelsen, and
  Andrews]{rembrandt}
L.~Scarpace, A.~E. Flanders, R.~Jain, T.~Mikkelsen, and D.~W. Andrews.
\newblock Data from {REMBRANDT}, 2019.
\newblock URL \url{https://doi.org/10.7937/K9/TCIA.2015.588OZUZB}.
\newblock Data set.

\bibitem[{The Cancer Imaging Archive}(2016{\natexlab{b}})]{ivygap_gbm}
{The Cancer Imaging Archive}.
\newblock Ivy glioblastoma atlas project collection (ivygap-gbm),
  2016{\natexlab{b}}.
\newblock URL \url{https://doi.org/10.7937/K9/TCIA.2016.XLWAN6NL}.
\newblock Data set.

\bibitem[{National Cancer Institute Clinical Proteomic Tumor Analysis
  Consortium}(2018)]{cptac_gbm}
{National Cancer Institute Clinical Proteomic Tumor Analysis Consortium}.
\newblock The clinical proteomic tumor analysis consortium glioblastoma
  multiforme collection (cptac-gbm), 2018.
\newblock URL \url{https://doi.org/10.7937/K9/TCIA.2018.3RJE41Q1}.
\newblock Data set.

\bibitem[Paszke et~al.(2019)Paszke, Gross, Massa, Lerer, Bradbury, Chanan,
  Killeen, Lin, Gimelshein, Antiga, Desmaison, Kopf, Yang, DeVito, Raison,
  Tejani, Chilamkurthy, Steiner, Fang, Bai, and Chintala]{paszke2019pytorch}
Adam Paszke, Sam Gross, Francisco Massa, Adam Lerer, James Bradbury, Gregory
  Chanan, Trevor Killeen, Zeming Lin, Natalia Gimelshein, Luca Antiga, Alban
  Desmaison, Andreas Kopf, Edward Yang, Zachary DeVito, Martin Raison, Alykhan
  Tejani, Sasank Chilamkurthy, Benoit Steiner, Lu~Fang, Junjie Bai, and Soumith
  Chintala.
\newblock Pytorch: An imperative style, high-performance deep learning library.
\newblock \emph{Advances in Neural Information Processing Systems}, 32, 2019.
\newblock URL
  \url{https://proceedings.neurips.cc/paper/2019/hash/bdbca288fee7f92f2bfa9f7012727740-Abstract.html}.

\bibitem[Cardoso et~al.(2022)]{cardoso2022monai}
M.~Jorge Cardoso et~al.
\newblock Monai: An open-source framework for deep learning in healthcare.
\newblock \emph{arXiv preprint arXiv:2211.02701}, 2022.
\newblock URL \url{https://arxiv.org/abs/2211.02701}.

\bibitem[Kingma and Welling(2014{\natexlab{b}})]{kingma2014vae}
Diederik~P. Kingma and Max Welling.
\newblock Auto-encoding variational {B}ayes.
\newblock In \emph{Proceedings of the 2nd International Conference on Learning
  Representations (ICLR)}, Banff, Canada, 2014{\natexlab{b}}.

\bibitem[Li et~al.(2022)Li, Xu, Lv, Cui, Zhang, and Wei]{dit}
Junlong Li, Yiheng Xu, Tengchao Lv, Lei Cui, Cha Zhang, and Furu Wei.
\newblock Dit: Self-supervised pre-training for document image transformer.
\newblock In \emph{ACM Multimedia 2022}, October 2022.
\newblock URL
  \url{https://www.microsoft.com/en-us/research/publication/dit-self-supervised-pre-training-for-document-image-transformer/}.

\bibitem[Mo et~al.(2023)Mo, Xie, Chu, Yao, Hong, Nießner, and Li]{3dit}
Shentong Mo, Enze Xie, Ruihang Chu, Lewei Yao, Lanqing Hong, Matthias Nießner,
  and Zhenguo Li.
\newblock Dit-3d: Exploring plain diffusion transformers for 3d shape
  generation, 2023.
\newblock URL \url{https://arxiv.org/abs/2307.01831}.

\bibitem[Nguyen et~al.(2018)Nguyen, Li, Bui, and Turner]{nguyen2018variational}
Cuong~V. Nguyen, Yingzhen Li, Thang~D. Bui, and Richard~E. Turner.
\newblock Variational continual learning.
\newblock \emph{International Conference on Learning Representations (ICLR)},
  2018.
\newblock URL \url{https://openreview.net/forum?id=BkQqq0gRb}.

\bibitem[Farquhar and Gal(2018)]{farquhar2019towards}
Sebastian Farquhar and Yarin Gal.
\newblock Towards robust evaluations of continual learning, 2018.
\newblock URL \url{https://arxiv.org/abs/1805.09733}.

\bibitem[Arco et~al.(2025)Arco, Jim{\'e}nez-Mesa, Ortiz, Ram{\'\i}rez, Levin,
  and G{\'o}rriz]{arco2025explainable}
Juan~E Arco, Carmen Jim{\'e}nez-Mesa, Andr{\'e}s Ortiz, Javier Ram{\'\i}rez,
  Johannes Levin, and Juan~M G{\'o}rriz.
\newblock Explainable inter-modality medical information transfer using siamese
  autoencoders.
\newblock \emph{IEEE Transactions on Radiation and Plasma Medical Sciences},
  2025.
\newblock \doi{10.1109/TRPMS.2025.3577309}.

\bibitem[Chen et~al.(2026)Chen, Yin, Chen, Wu, and
  Li]{chen2026contrastxmultimodalcontrastimage}
Yifan Chen, Fei Yin, Hao Chen, Jia Wu, and Chao Li.
\newblock Contrast-x: A multi-modal contrast image synthesis benchmark and
  universal modality flow matching, 2026.
\newblock URL \url{https://arxiv.org/abs/2601.15884}.

\bibitem[He et~al.(2016)He, Zhang, Ren, and Sun]{he2016resnet}
Kaiming He, Xiangyu Zhang, Shaoqing Ren, and Jian Sun.
\newblock Deep residual learning for image recognition.
\newblock \emph{Proceedings of the IEEE/CVF Conference on Computer Vision and
  Pattern Recognition (CVPR)}, 2016.
\newblock URL
  \url{https://openaccess.thecvf.com/content_cvpr_2016/html/He_Deep_Residual_Learning_CVPR_2016_paper.html}.

\bibitem[Baid et~al.(2023)Baid, Ghodasara, Mohan, Bilello, Calabrese, Colak,
  Farahani, Kalpathy-Cramer, Kitamura, Pati, Prevedello, Rudie, Sako,
  Shinohara, Bergquist, Chai, Eddy, Elliott, Reade, Schaffter, Yu, Zheng,
  Davatzikos, Mongan, Hess, Cha, Villanueva-Meyer, Freymann, Kirby, Wiestler,
  Crivellaro, Colen, Kotrotsou, Marcus, Milchenko, Nazeri, Fathallah-Shaykh,
  Wiest, Jakab, Weber, Mahajan, Menze, Flanders, and Bakas]{brats2021}
U.~Baid, S.~Ghodasara, S.~Mohan, M.~Bilello, E.~Calabrese, E.~Colak,
  K.~Farahani, J.~Kalpathy-Cramer, F.~C. Kitamura, S.~Pati, L.~Prevedello,
  J.~Rudie, C.~Sako, R.~Shinohara, T.~Bergquist, R.~Chai, J.~Eddy, J.~Elliott,
  W.~Reade, T.~Schaffter, T.~Yu, J.~Zheng, C.~Davatzikos, J.~Mongan, C.~Hess,
  S.~Cha, J.~Villanueva-Meyer, J.~B. Freymann, J.~S. Kirby, B.~Wiestler,
  P.~Crivellaro, R.~R. Colen, A.~Kotrotsou, D.~Marcus, M.~Milchenko, A.~Nazeri,
  H.~Fathallah-Shaykh, R.~Wiest, A.~Jakab, M.-A. Weber, A.~Mahajan, B.~Menze,
  A.~E. Flanders, and S.~Bakas.
\newblock {RSNA-ASNR-MICCAI-BraTS-2021 Dataset}, 2023.
\newblock Dataset.

\bibitem[Wang et~al.(2004)Wang, Bovik, Sheikh, and Simoncelli]{wang2004image}
Zhou Wang, Alan~C. Bovik, Hamid~R. Sheikh, and Eero~P. Simoncelli.
\newblock Image quality assessment: From error visibility to structural
  similarity.
\newblock \emph{IEEE Transactions on Image Processing}, 13\penalty0
  (4):\penalty0 600--612, 2004.
\newblock \doi{10.1109/TIP.2003.819861}.

\bibitem[Lopez-Paz and Ranzato(2017)]{lopezpaz2017gem}
David Lopez-Paz and Marc'Aurelio Ranzato.
\newblock Gradient episodic memory for continual learning.
\newblock In \emph{Advances in Neural Information Processing Systems
  (NeurIPS)}, volume~30, pages 6467--6476, 2017.
\newblock URL
  \url{https://papers.nips.cc/paper/2017/hash/f87522788a2be2d171666752f97ddebb-Abstract.html}.

\bibitem[van~de Ven et~al.(2020)van~de Ven, Siegelmann, and
  Tolias]{vandeven2020brain}
Gido~M. van~de Ven, Hava~T. Siegelmann, and Andreas~S. Tolias.
\newblock Brain-inspired replay for continual learning with artificial neural
  networks.
\newblock \emph{Nature Communications}, 11\penalty0 (1):\penalty0 4069, 2020.
\newblock \doi{10.1038/s41467-020-17866-2}.

\bibitem[Chaudhry et~al.(2018)Chaudhry, Dokania, Ajanthan, and
  Torr]{chaudhry2018riemannian}
Arslan Chaudhry, Puneet~K. Dokania, Thalaiyasingam Ajanthan, and Philip H.~S.
  Torr.
\newblock Riemannian walk for incremental learning: Understanding forgetting
  and intransigence.
\newblock In \emph{Proceedings of the European Conference on Computer Vision
  (ECCV)}, pages 532--547, 2018.
\newblock \doi{10.1007/978-3-030-01252-6_33}.

\bibitem[Parisi et~al.(2019)Parisi, Kemker, Part, Kanan, and
  Wermter]{parisi2019continual}
German~I. Parisi, Ronald Kemker, Jose~L. Part, Christopher Kanan, and Stefan
  Wermter.
\newblock Continual lifelong learning with neural networks: A review.
\newblock \emph{Neural Networks}, 113:\penalty0 54--71, 2019.
\newblock \doi{10.1016/j.neunet.2019.01.012}.

\bibitem[Wang et~al.(2024)Wang, Zhang, Su, and Zhu]{wang2024comprehensive}
Liyuan Wang, Xingxing Zhang, Hang Su, and Jun Zhu.
\newblock A comprehensive survey of continual learning: Theory, method and
  application.
\newblock \emph{IEEE Transactions on Pattern Analysis and Machine
  Intelligence}, 46\penalty0 (8):\penalty0 5362--5383, 2024.
\newblock \doi{10.1109/TPAMI.2024.3367329}.

\bibitem[Loshchilov and Hutter(2017)]{loshchilov2016sgdr}
Ilya Loshchilov and Frank Hutter.
\newblock Sgdr: Stochastic gradient descent with warm restarts.
\newblock In \emph{International Conference on Learning Representations
  (ICLR)}, 2017.
\newblock URL \url{https://arxiv.org/abs/1608.03983}.

\bibitem[Cox(1972)]{cox1972}
David~R. Cox.
\newblock Regression models and life-tables.
\newblock \emph{Journal of the Royal Statistical Society: Series B},
  34\penalty0 (2):\penalty0 187--220, 1972.
\newblock \doi{10.1111/j.2517-6161.1972.tb00899.x}.
\newblock URL \url{https://doi.org/10.1111/j.2517-6161.1972.tb00899.x}.

\bibitem[Katzman et~al.(2018)Katzman, Shaham, Clisham, Rosset, Jette, Zhang,
  and Komatsoulis]{katzman2018deepsurv}
Jared~L. Katzman, Uri Shaham, Alexander Clisham, Stephen Rosset, Jordan Jette,
  Tianxiao Zhang, and Georgia~A. Komatsoulis.
\newblock Deepsurv: Personalized treatment recommender system using a cox
  proportional hazards deep neural network.
\newblock \emph{BMC Medical Research Methodology}, 18\penalty0 (1):\penalty0
  24, 2018.
\newblock \doi{10.1186/s12874-018-0482-8}.
\newblock URL \url{https://doi.org/10.1186/s12874-018-0482-8}.

\bibitem[HARRELL~Jr. et~al.(1996)HARRELL~Jr., LEE, and MARK]{harrell1982cindex}
FRANK~E. HARRELL~Jr., KERRY~L. LEE, and DANIEL~B. MARK.
\newblock Multivariable prognostic models: Issues in developing models,
  evaluating assumptions and adequacy, and measuring and reducing errors.
\newblock \emph{Statistics in Medicine}, 15\penalty0 (4):\penalty0 361--387,
  1996.
\newblock
  \doi{10.1002/(SICI)1097-0258(19960229)15:4<361::AID-SIM168>3.0.CO;2-4}.
\newblock URL
  \url{https://onlinelibrary.wiley.com/doi/abs/10.1002/\%28SICI\%291097-0258\%2819960229\%2915\%3A4\%3C361\%3A\%3AAID-SIM168\%3E3.0.CO\%3B2-4}.

\bibitem[Hanley and McNeil(1982)]{hanley1982roc}
James~A. Hanley and Barbara~J. McNeil.
\newblock The meaning and use of the area under a receiver operating
  characteristic (roc) curve.
\newblock \emph{Radiology}, 143\penalty0 (1):\penalty0 29--36, 1982.
\newblock \doi{10.1148/radiology.143.1.7063747}.
\newblock URL \url{https://doi.org/10.1148/radiology.143.1.7063747}.

\bibitem[McCandlish et~al.(2018)McCandlish, Kaplan, Amodei, and
  Team]{mccandlish2018empirical}
Sam McCandlish, Jared Kaplan, Dario Amodei, and OpenAI~Dota Team.
\newblock An empirical model of large-batch training.
\newblock \emph{arXiv preprint arXiv:1812.06162}, 2018.
\newblock \doi{10.48550/arXiv.1812.06162}.
\newblock URL \url{https://arxiv.org/abs/1812.06162}.

\bibitem[Smith et~al.(2018)Smith, Kindermans, Ying, and Le]{smith2018dont}
Samuel~L. Smith, Pieter-Jan Kindermans, Chris Ying, and Quoc~V. Le.
\newblock Don't decay the learning rate, increase the batch size.
\newblock \emph{International Conference on Learning Representations (ICLR)},
  2018.
\newblock URL \url{https://openreview.net/forum?id=B1Yy1BxCZ}.

\bibitem[Li et~al.(2024)Li, Zou, Wang, Majumder, Xie, Manmatha, Swaminathan,
  Tu, Ermon, and Soatto]{li2024scalability}
Hao Li, Yang Zou, Ying Wang, Orchid Majumder, Yusheng Xie, R.~Manmatha, Ashwin
  Swaminathan, Zhuowen Tu, Stefano Ermon, and Stefano Soatto.
\newblock On the scalability of diffusion-based text-to-image generation.
\newblock \emph{Amazon Science}, 2024.
\newblock URL
  \url{https://www.amazon.science/publications/on-the-scalability-of-diffusion-based-text-to-image-generation}.

\bibitem[Xu et~al.(2024)Xu, Mi, Wang, and Chen]{xu2024faster}
Tianshuo Xu, Peng Mi, Ruilin Wang, and Yingcong Chen.
\newblock Towards faster training of diffusion models: An inspiration of a
  consistency phenomenon.
\newblock \emph{arXiv preprint arXiv:2404.07946}, 2024.
\newblock \doi{10.48550/arXiv.2404.07946}.
\newblock URL \url{https://arxiv.org/abs/2404.07946}.

\bibitem[Hang et~al.(2023)Hang, Gu, Li, Bao, Chen, Hu, Geng, and
  Guo]{hang2023minsnr}
Tiankai Hang, Shuyang Gu, Chen Li, Jianmin Bao, Dong Chen, Han Hu, Xin Geng,
  and Baining Guo.
\newblock Efficient diffusion training via min-snr weighting strategy.
\newblock In \emph{Proceedings of the IEEE/CVF International Conference on
  Computer Vision (ICCV)}, pages 7441--7451, October 2023.
\newblock \doi{10.1109/ICCV51070.2023.00684}.
\newblock URL
  \url{https://openaccess.thecvf.com/content/ICCV2023/html/Hang_Efficient_Diffusion_Training_via_Min-SNR_Weighting_Strategy_ICCV_2023_paper.html}.

\bibitem[Pinaya et~al.(2022)Pinaya, Tudosiu, Dafflon, Da~Costa, Fernandez,
  Nachev, Ourselin, and Cardoso]{pinaya2022brain}
Walter~HL Pinaya, Petru-Daniel Tudosiu, Jessica Dafflon, Pedro~F Da~Costa,
  Virginia Fernandez, Parashkev Nachev, Sebastien Ourselin, and M~Jorge
  Cardoso.
\newblock Brain imaging generation with latent diffusion models.
\newblock In \emph{MICCAI Workshop on Deep Generative Models (DGM4MICCAI)},
  pages 117--126. Springer, 2022.
\newblock \doi{10.1007/978-3-031-18576-2_12}.
\newblock URL \url{https://arxiv.org/abs/2209.07162}.

\bibitem[Preechakul et~al.(2022)Preechakul, Chatthee, Wizadwongsa, and
  Suwajanakorn]{preechakul2022diffusion}
Konpat Preechakul, Nattanat Chatthee, Suttisak Wizadwongsa, and Supasorn
  Suwajanakorn.
\newblock Diffusion autoencoders: Toward a meaningful and decodable
  representation.
\newblock In \emph{Proceedings of the IEEE/CVF Conference on Computer Vision
  and Pattern Recognition (CVPR)}, pages 10619--10629, 2022.
\newblock \doi{10.1109/CVPR52688.2022.01036}.
\newblock URL
  \url{https://openaccess.thecvf.com/content/CVPR2022/html/Preechakul_Diffusion_Autoencoders_Toward_a_Meaningful_and_Decodable_Representation_CVPR_2022_paper.html}.

\bibitem[Glasser et~al.(2016)Glasser, Coalson, Robinson, et~al.]{Glasser2016}
Matthew~F Glasser, Timothy~S Coalson, Emma~C Robinson, et~al.
\newblock A multi-modal parcellation of human cerebral cortex.
\newblock \emph{Nature}, 536\penalty0 (7615):\penalty0 171--178, 2016.
\newblock \doi{10.1038/nature18933}.

\bibitem[Bedini et~al.(2023)Bedini, Olivetti, Avesani, and
  Baldauf]{bedini_hcpmmp_fsl}
Marco Bedini, Emanuele Olivetti, Paolo Avesani, and Daniel Baldauf.
\newblock Accurate localization and coactivation profiles of the frontal eye
  field and inferior frontal junction: an {ALE} and {MACM} {fMRI}
  meta-analysis.
\newblock \emph{Brain Structure and Function}, 228\penalty0 (3):\penalty0
  997--1017, 2023.
\newblock \doi{10.1007/s00429-023-02641-y}.

\bibitem[Rolls et~al.(2020)Rolls, Huang, Lin, Feng, and Joliot]{rolls2020aal3}
Edmund~T. Rolls, Chu-Chung Huang, Ching-Po Lin, Jianfeng Feng, and Marc Joliot.
\newblock Automated anatomical labelling atlas 3.
\newblock \emph{NeuroImage}, 206:\penalty0 116189, 2020.
\newblock \doi{10.1016/j.neuroimage.2019.116189}.

\bibitem[Schaefer et~al.(2018)Schaefer, Kong, Gordon, Laumann, Zuo, Holmes,
  Eickhoff, and Yeo]{schaefer2018localglobal}
Alexander Schaefer, Ru~Kong, Evan~M. Gordon, Timothy~O. Laumann, Xi-Nian Zuo,
  Avram~J. Holmes, Simon~B. Eickhoff, and B.~T.~Thomas Yeo.
\newblock Local-global parcellation of the human cerebral cortex from intrinsic
  functional connectivity {MRI}.
\newblock \emph{Cerebral Cortex}, 28\penalty0 (9):\penalty0 3095--3114, 2018.
\newblock \doi{10.1093/cercor/bhx179}.

\bibitem[Yeo et~al.(2011)Yeo, Krienen, Sepulcre, Sabuncu, Lashkari,
  Hollinshead, Roffman, Smoller, Z{\"o}llei, Polimeni, Fischl, Liu, and
  Buckner]{yeo2011organization}
B.~T.~Thomas Yeo, Fenna~M. Krienen, Jorge Sepulcre, Mert~R. Sabuncu, Danial
  Lashkari, Marisa Hollinshead, Joshua~L. Roffman, Jordan~W. Smoller, Lilla
  Z{\"o}llei, Jonathan~R. Polimeni, Bruce Fischl, Hesheng Liu, and Randy~L.
  Buckner.
\newblock The organization of the human cerebral cortex estimated by intrinsic
  functional connectivity.
\newblock \emph{Journal of Neurophysiology}, 106\penalty0 (3):\penalty0
  1125--1165, 2011.
\newblock \doi{10.1152/jn.00338.2011}.

\bibitem[Syed et~al.(2024)Syed, Rager, and Conmy]{syed2023attributionpatching}
Aaquib Syed, Can Rager, and Arthur Conmy.
\newblock Attribution patching outperforms automated circuit discovery.
\newblock In Yonatan Belinkov, Najoung Kim, Jaap Jumelet, Hosein Mohebbi, Aaron
  Mueller, and Hanjie Chen, editors, \emph{Proceedings of the 7th BlackboxNLP
  Workshop: Analyzing and Interpreting Neural Networks for NLP}, pages
  407--416, Miami, Florida, US, November 2024. Association for Computational
  Linguistics.
\newblock \doi{10.18653/v1/2024.blackboxnlp-1.25}.
\newblock URL \url{https://aclanthology.org/2024.blackboxnlp-1.25/}.

\end{thebibliography}
\end{document}